\documentclass[12pt]{gatechthesis}
\usepackage{tabularx,booktabs}
\usepackage{multirow}
\usepackage{amssymb}
\usepackage{tikz}
\usepackage{dsfont}
\usepackage{wrapfig}
\usepackage{color}
\usepackage[dvipsnames]{}
\usepackage{xcolor,colortbl}
\usepackage{siunitx}
\usepackage{glossaries}

\usepackage[ruled,linesnumbered]{algorithm2e}
\definecolor{lightblue}{RGB}{173,216,230}
\definecolor{darkblue}{RGB}{8,81,156}
\definecolor{stanfordgrey}{RGB}{46,45,41}
\definecolor{cardinalred}{RGB}{253,141,60}
\definecolor{codegreen}{rgb}{0,0.4,0}
\definecolor{codegray}{rgb}{1.0,0.5,0.5}
\definecolor{codepurple}{rgb}{0.58,0,0}
\definecolor{tealblue}{rgb}{0,0.5,0.5}
\definecolor{codebackcolour}{rgb}{0.95,0.95,0.92}
\definecolor{darkgreen}{RGB}{0,127,0}
\definecolor{yellow}{RGB}{240,164,47}
\definecolor{darkred}{RGB}{200,0,0}
\definecolor{orange}{rgb}{1,0.5,0}
\definecolor{pink}{RGB}{213,126,190}
\definecolor{falured}{rgb}{0.5, 0.09, 0.09}
\definecolor{teal}{rgb}{0.1176, 0.5019, 0.5019}

\usepackage{pifont} %

\def\greencheckmark{\textcolor{darkgreen}{\checkmark}}
\def\redxmark{\textcolor{darkred}{\ding{55}}}  %

\title{Learning 3D Robotics Perception using Inductive Priors}
\author{Muhammad Zubair Irshad}
\approvaldate{December 04, 2023}
\school{Woodruff School}
\department{Department of Mechanical Engineering}
\bibliography{references}
\newcommand{\xmark}{\ding{55}}%
\newcolumntype{C}{>{\centering\arraybackslash}X}

\definecolor{tabfirst}{rgb}{1, 0.85, 0.7} 
\definecolor{tabsecond}{rgb}{1, 1, 0.7} 
\definecolor{tabthird}{rgb}{1,1,1} 

\begin{document}

\makeTitlePage{December}{2023}


\begin{approvalPage}{6}


\committeeMember{Dr. Zsolt Kira}{School of Interactive Computing}{Georgia Institute of Technology}

\committeeMember{Dr. Aaron Young}{Department of Mechanical Engineering}{Georgia Institute of Technology}
\committeeMember{Dr. Nader Sadegh}{Department of Mechanical Engineering}{Georgia Institute of Technology}
\committeeMember{Dr. Shreyas Kousik}{Department of Mechanical Engineering}{Georgia Institute of Technology}
\\ 
\committeeMember{Dr. Adrien Gaidon}{Head of Machine Learning}{Toyota Research Institute}

\end{approvalPage}
\makeEpigraph{Knowledge is love and light and vision.}{Helen Keller}
\makeDedication{I dedicate this thesis to \\
\textbf{My Mother} Najma Irshad, \\
\textbf{My Father} Irshad Ahmed
}

\begin{frontmatter}
    \begin{acknowledgments}


Embarking on a PhD involves significant effort and sacrifice, representing one of the most intellectually demanding challenges I've faced to date. Despite its difficulty, it has been an immensely rewarding journey. The successful completion wouldn't have been possible without the unwavering support of those around me. This section is dedicated to acknowledging everyone who played a crucial role in this chapter of my life, and I strive to do justice in recognizing their contributions. 

I would like to begin by extending my sincere gratitude to Allah~(SWT) for sharing a piece of His eternal knowledge with me. All the knowledge in this world, whether visible or concealed, belongs to You, and I am profoundly thankful to You for guiding me in learning and discovering things beyond my understanding.

I express my heartfelt gratitude to my loving parents, my Ammi and Abbu, for their unwavering love and support during my entire PhD journey and the years leading up to the start of my PhD degree. I am thankful to you for providing me with an excellent education from my early years to undergraduate, even if it meant relocating to Karachi. Your teachings of profound morals and the values instilled in all of us continue to resonate with me to this day. I am forever grateful beyond words for everything else you have done for me.

I want to express my heartfelt appreciation to my wife, Fatima Azfar, for accompanying me to Atlanta during this challenging phase of my life. Your love and support hold immense significance for me. You have brought joy into my life, and I am grateful for the beautiful two and a half years we have shared together leading up to this moment. May we continue to live happily together in this world and in the~\textit{aakhira} inshAllah.

I want to express my gratitude to my siblings, Shifa, Nuwaira, and Hamza, for ensuring that I never felt far from home. Your insightful and cheerful conversations over the years have been a source of comfort and joy. Thank you for the warmth and connection that your presence has brought into my life.

I extend my sincere gratitude to my advisor, Dr. Zsolt Kira, for his unwavering support throughout my PhD journey. You took a chance on me when I approached you from a different school, despite my limited experience in deep learning. I am thankful for teaching me how to do great research while being patient with me as I learned new skills. Your efforts in creating a conducive lab environment and alleviating concerns about funding have been invaluable in achieving my research goals. Thank you for all the great personal and professional advice over these years which shaped me as a researcher.

I express my gratitude to my PhD dissertation committee, namely, Dr. Aaron Young, Dr. Shreyas Kousik, Dr. Nader Sadegh, and Dr. Adrien Gaidon, for your unwavering support and guidance during the final year of my PhD. Your constructive feedback and valuable suggestions significantly enhanced the quality of my thesis writing. I appreciate your time, expertise, and commitment to ensuring the successful completion of my doctoral journey.

I express my gratitude to Dr. Wayne Whiteman and Ms. Camellia Henry for their kindness over the years. They have not only listened to my academic questions but also supported me in meeting the academic requirements of the ME school. I especially thank Dr. Whiteman for offering me a valuable first TA-ship funding opportunity in the ME school, enabling me to come to the U.S. and pursue my ambitions of obtaining a Ph.D. degree.

I extend my gratitude to all my teachers who have shared their knowledge, contributing to my personal growth. I express gratitude to Dr. Seth Hutchinson for accepting me as an M.S. student researcher and for his motivation and support in pursuing a PhD degree. I also appreciate the professors at Ghulam Ishaq Khan University of Science and Technology for sharing their knowledge with me and teaching me how to meta-learn new skills.

I express my gratitude to Toyota Research Institute and SRI International for allowing me to spend four wonderful summers and springs during my Ph.D. program. Special thanks to my mentor and collaborators at Toyota Research Institute, particularly Adrien Gaidon, for instilling in me a bird's-eye view approach of leading research projects and fostering creative thinking. I extend my thanks to Rares Ambrus for being a great manager and mentor, supporting me in everything, from compute resources to pushing me to excel in research and emphasizing attention to detail in all of my projects. I would like to thank Sergey Zakharov for being a valuable mentor and collaborator on my research projects. I appreciate Thomas Kollar for his great advice and mentorship during my first internship project and ongoing support in subsequent projects. I also extend my thanks to mentors at SRI International, including Han-Pang Chiu, Niluhtpol Mithun, Zachary Seymour, and Rakesh Kumar, for their invaluable advice and guidance during my research internship. 

A special thank you to all my other co-authors, namely Vitor Guizilini, Katherine Liu, Dian Chen, Michael Laskey, Kevin Stone, Supun Samarasakera, Chih-Yao Ma~(from Meta Gen AI), Nick Heppert~(University of Freiburg), Jeanette Bohg~(Stanford University), Abhinav Valada~(University of Freiburg), Mayank Lunayach, Vishnu Jaganathan, Amit Raj~(Google Reserch), Varun Jampani~(Google Research) and Hannah Huang for your support and contributions that have played a pivotal role in shaping my research journey.

I appreciate the financial support from Toyota Research Institute and DARPA throughout my Ph.D. journey. A special acknowledgment goes to Toyota Research Institute for providing essential compute resources in the latter part of my doctoral studies. I also extend sincere thanks to the Fulbright program for sponsoring two impactful years of my M.S. degree at Georgia Tech. The Fulbright experience not only enriched my M.S. journey but also served as a motivating force for pursuing and successfully completing my Ph.D.

I express my gratitude to my lab members for fostering an exceptional lab environment and for their valuable peer support, advice, and feedback throughout my journey. Special thanks to Nathan Glaser, James Smith, Mayank Lunayach, Yousef Emam, Jeremiah Coholich, Junjiao Tian, Yen-Cheng Liu, Shanak Halbe, Andrew Szot, Chih-Yao Ma and Ming-Hun Chen for insightful conversation over the years in both personal and professional sphere.

I express my appreciation to my mentees for allowing me to guide and mentor you throughout your M.S., B.S., and pursuit to Ph.D. journeys. A special acknowledgment to Mayank Lunayach, Vishnu Jaganathan, Shiva Gantha, Hanna Huang, and Asawaree Bhide from Georgia Tech. I also extend my gratitude to Ahnaf Munir and Anas Zafar from Fatima Fellowship for their dedication and collaboration. Being your mentor and witnessing your success in projects and careers has added joy and memorable moments to my Ph.D. journey.

Finally, I express my gratitude to my support network of friends and family who have been unwavering in their support throughout my Ph.D. journey. A heartfelt thank you to Saima Aunty and Azfar Uncle for their warm welcome and for making me feel at home in Chicago whenever I visited your beautiful city. I extend my appreciation to my Grandfather, Nanoo, for your boundless love and for consistently coming to Karachi to meet me whenever I visited, along with the countless prayers you sends my way. I also want to thank my support group of friends in Atlanta for the shared dinners, cricket and tennis matches, board game sessions, and adventurous and memorable trips across the U.S. Special thanks to the Pakistani Student Association, Muslim Student Association, and Alfarooq Mosque of Atlanta and all the organizing folks from these communities for contributing to the memorable moments and making me feel at home away from home during my time in Atlanta.

\end{acknowledgments}

    \makeTOC
    \makeListOfTables
    \makeListOfFigures

\newacronym{robovln}{Robo-VLN}{Robotics Vision-and-Language Navigation}
\newacronym{nocs}{NOCS}{Normalized Object Coordinate Space}
\newacronym{fpn}{FPN}{Feature Pyramid Network}
\newacronym{mos}{MOS}{Multi-Object ShapeNet}
\newacronym{iou}{IOU}{Intersection-over-Union}
\newacronym{map}{mAP}{Mean Average Precision}
\newacronym{cd}{CD}{Chamfer Distance}
\newacronym{fps}{FPS}{Frames-per-second}
\newacronym{nerf}{NeRF}{Neural Radiance Fields}
\newacronym{sdf}{SDF}{Signed Distance Function}
\newacronym{tf}{TF}{Texture Fields}
\newacronym{mlp}{MLP}{Multi-Layer Perceptron}
\newacronym{lod}{LoD}{level of detail}
\newacronym{vln}{VLN}{Vision-and-Language Navigation}
\newacronym{sasra}{SASRA}{Semantically-aware Spatio-Temporal Reasoning Agent}
\newacronym{nerds}{NERDS}{NeRF for Reconstruction, Decomposition and Scene Synthesis}
\newacronym{nerfmae}{NeRF-MAE}{Neural Radiance Field Mask Auto-Encoder}
\newacronym{swin}{SwinTransformer}{Shifted Window Transformer}
\newacronym{vit}{ViT}{Vision Transformers}
\newacronym{hcm}{HCM}{Hierarchical Cross-Modal Agent}
\newacronym{cnn}{CNN}{Convolution Neural Network}

\makeListOfAcronyms
    
\begin{summary}

Recent advances in deep learning have led to a 'data-centric intelligence' in the last decade i.e. artificially intelligent models unlocking the potential to ingest a large amount of data and be really good at performing digital tasks such as text-to-image generation, machine-human conversation, and image recognition. This thesis covers the topic of learning with structured inductive bias and priors to design approaches and algorithms unlocking the potential of 'principle-centric intelligence' for the real-world. ~\textit{Prior knowledge}~(priors for short), often available in terms of past experience as well as assumptions of how the world works, helps the autonomous agent generalize better and adapt their behavior based on past experience. In this thesis, I demonstrate the use of prior knowledge in three different robotics perception problems. 1. object-centric 3D reconstruction, 2. vision and language for decision-making, and 3. 3D scene understanding. To solve these challenging problems, I propose various sources of prior knowledge including 1.~\textit{geometry and appearance priors} from synthetic data, 2.~\textit{modularity and semantic map} priors and 3.~\textit{semantic, structural, and contextual} priors. I study these priors for solving robotics 3D perception tasks and propose ways to efficiently encode them in deep learning models. Some priors are used to warm-start the network for transfer learning, others are used as hard constraints to restrict the action space of robotics agents. While classical techniques are brittle and fail to generalize to unseen scenarios and data-centric approaches require a large amount of labeled data, this thesis aims to build intelligent agents which require very-less real-world data or data acquired only from simulation to generalize to highly dynamic and cluttered environments in novel simulations (i.e. sim2sim) or real-world unseen environments (i.e. sim2real) for a holistic scene understanding of the 3D world.


\end{summary}
\addcontentsline{toc}{chapter}{Summary}
\end{frontmatter}

\begin{thesisbody}
    \chapter{Introduction and Background}

\looseness=-1
The promise of personal assistant robots that can seamlessly perform complex tasks such as manipulating the 3D world around the agent and following human instructions to navigate to a goal in real-life environments has
long been sought after. This problem is hard due to the uncertainties of real-world systems such as occlusions, clutter, object diversity, variability, and dynamic nature of environments as well as limited data in the real world to train policies. Existing classical systems are often brittle and design specialized algorithms for these complicated tasks which often fail to generalize to unseen environments, such as a robotic agent in an unseen home or an autonomous driving car on a dense road.

Recent advancements in deep learning to extract meaningful information from raw sensor data have enabled some progress toward this
goal. One way to train these agents is to collect massive amounts of labeled data in the real world. Training such a system in a fully supervised manner has previously been shown to exhibit great generalization capabilities. However, annotations are expensive to obtain, and they are not always readily available or cover all possible edge cases.

\looseness=-1
Moreover, reactive agents~(such as autonomous vehicles) or robots in homes, need to be versatile and adapt their behaviors. This~\textit{adaptability} in dynamic environments can either come from trial-and-error over a vast number of iterations or through learning over a limited subset of data in a~\textit{principled manner}. The process of teaching a robot through data to make it adapt to new surroundings is what is referred to as~\textit{learning}. My thesis is about how robots can~\textit{learn 3D perception tasks} using prior knowledge to make them adapt faster to new surroundings and in cases where a limited amount of real-world data might be available.

\looseness=-1
One attractive alternative solution to the data scarcity problem is to utilize data from simulated environments. This offers~\textit{free} access to annotations covering all edge cases. However, a domain gap still exists between simulation and real-world systems which leads us to ask three important questions:~\textbf{1.}~\textit{How to bridge the gap between simulation and real-world systems}.~\textbf{2.}~\textit{How to design systems that work with limited real-world labeled data or sparsity of data} and~\textbf{3.}~\textit{How to leverage unsupervised 2D data to build strong 3D representations of the real world which can be readily used for a variety of downstream tasks.}

The real world offers many priors~\cite{bengio2013representation}, which the agents do not need to learn twice. They can be either universal truths like laws of physics, widely accepted mathematical models of how the world works or they can be heuristics beneficial for the hypothesis space of the problem we are trying to solve. Examples of~\textit{priors} in the realm of computer vision are the use of convolutional-neural-network~\cite{lecun1995convolutional}~(ConvNets) architecture acting as general priors. ConvNets incorporate the concept of translation equivariance. This prior assumption acknowledges that in image data, objects and features often exhibit translational invariance, meaning that their characteristics remain consistent even when shifted or translated within the image. ConvNets leverage this prior by incorporating convolutional layers, which apply filters (kernels) to small local regions of the input data. These convolutional layers are designed to capture and exploit the local structural patterns within the data, respecting the principle of translation equivariance.

Another~\textit{prior} used in computer vision tasks is multi-view epipoplar geometry. It defines the geometric relationship between two cameras observing the same 3D scene, and it constrains the possible correspondence between points in different views. The epipolar constraint helps in feature matching and can reduce ambiguities in stereo vision and structure-from-motion tasks.

My thesis is that 3D perception in the context of robotics offers many task-specific and general priors that can aid learning efficiently, learning from limited data, and offering better adaption to unique environments and surroundings. 

\section{Thesis Statement}
\looseness=-1
~\textit{Deep Learning agents can address the challenges of data and annotation scarcity in the real-world through utilizing data from simulation and by~\textbf{1.} bridging the gap between simulation and real-world with strong inductive prior knowledge~(priors for short) in the form of hierarchy, geometry, and structural context and~\textbf{2.} enabling systems to perform a variety of 3D tasks in the real-world by creating strong 3D representations from unsupervised 2D data.}

\section{Outline and Contributions}

The structure of this thesis proposal document is as follows: I first discuss in detail a broad overview of the related literature in~(Chapter~\ref{chap:background_and_literature_review}). This is followed by key technical contributions relating to each of the three thrust areas which comprises two chapters each. Finally, I summarize the main contributions and discuss future directions in the conclusion section. The three key technical contributions of this thesis are summarized as follows:

\noindent \textbf{Part~\ref{part1}: Efficient Object-Centric Neural 3D
Representations
}

This part focuses on studying efficient object-centric 3D representations and linking these with 2D detections to propose efficient end-to-end systems for improved real-world 3D object understanding. We study~\emph{geometry and appearance priors} from synthetically simulated data which aids in learning strong 3D representations that transfer to the real world with minimal fine-tuning. 

\begin{itemize}
    \item \textbf{Chapter~\ref{chap:centersnap}: CenterSnap: Single-Shot Multi-Object 3D Shape Reconstruction and Categorical 6D Pose and Size Estimation -} This chapter studies and formalizes a novel~\emph{single-shot setting} for 3D shape reconstruction and 6D pose and size estimation. Departing from conventional methods that rely on multi-stage pipelines and CAD models, this chapter presents a one-stage solution without the need for 2D bounding boxes or separate models for different categories during inference. Through a per-pixel representation, the approach achieves real-time reconstruction and demonstrates substantial performance improvements, making a noteworthy contribution to the field.
    
    \item \textbf{Chapter~\ref{chap:shapo}: ShAPO: Implicit Representations for Multi-Object Shape Appearance and Pose Optimization -} This chapter introduces a disentangled shape and appearance representation coupled with an octree-based iterative optimization, which enables accurate identification of multiple object's appearances along with shapes and 6D poses. We show that utilizing shape and appearance priors from neural networks trained primarily on synthetic data transfers well to the real-world scenario with minimal fine-tuning while also allowing inference-time optimization from single-view observations.
\end{itemize}

\noindent \textbf{Part~\ref{part2}: Hierarchical Vision-and-Language for Action
}

This part investigates the impact of injecting prior knowledge into agent-centric deep learning systems~(i.e. scenarios where the agent takes an action based on raw-visual observations as an input). We specifically show semantic maps~(a top-down representation of the 3D scene) as a strong prior as well as explore the utility of encoding hierarchy as a means to disentangle systems reliant on predicting actions through deep learning frameworks. Our study shows improved generalization capabilities in novel environments by employing both the hierarchy and semantic map prior techniques, as detailed below: 

\begin{itemize}
    \item \textbf{Chapter~\ref{chap:robovln}: Hierarchical Cross-Modal Agent for Robotics Vision-and-Language Navigation -} 
    This chapter introduces the novel Robo-VLN setting, which extends Vision-and-Language Navigation (VLN) to continuous 3D environments with longer trajectories and obstacles; thus making the VLN problem more challenging and closer to the real-world. We also show that existing baselines struggle at this task and propose a~\gls{hcm}. Through comprehensive experiments with baselines, this work sets a new benchmark for Robo-VLN by outperforming existing methods.
    
    \item \textbf{Chapter~\ref{chap:sasra}: SASRA: Semantically-aware Spatio-temporal Reasoning Agent for Vision-and-Language Navigation in Continuous Environments -} identifies a crucial gap in the existing approach to Vision-and-Language Navigation (VLN) in continuous 3D environments, which primarily relies on raw visual data. This chapter demonstrates that combining classical semantic mapping techniques with learning-based methods significantly enhances the agent's ability to adapt to new scenes. By leveraging a Transformer architecture for the fusion of map and language features, along with a hybrid Transformer for action prediction, our approach achieves state-of-the-art results in the challenging continuous VLN task.
\end{itemize}

\noindent \textbf{Part~\ref{part3}: Generalizable Self-Supervised 3D Scene Understanding
}

This part presents studies obtaining robust feature representations for two key objectives: \textbf{1.} Leveraging unsupervised multi-view RGB data for robust 3D representation learning, and \textbf{2.} harnessing sparse multiview data to enhance the efficiency of 3D scene comprehension, encompassing both indoor and outdoor environments. Our investigation highlights the significant contributions of high-quality representations acquired from extensive synthetic data in facilitating challenging tasks such as sparse view novel-view synthesis. Furthermore, we demonstrate that representations acquired through a masked self-supervised learning paradigm can yield substantial improvements across a diverse spectrum of downstream 3D tasks, especially in scenarios where acquiring extensive 3D annotations proves to be a challenging endeavor.

\begin{itemize}
    \item \textbf{Chapter~\ref{chap:neo360}: NeO 360: Neural Fields for Sparse View Synthesis of Outdoor Scenes - }
    This chapter presents an innovative approach for sparse view synthesis of outdoor scenes. While recent implicit neural representations excel at novel view synthesis, they often demand require many dense-view annotations. NeO 360 addresses this by reconstructing 360~$^{\circ}$ scenes from just a single or a few RGB images. We propose a novel hybrid image-conditional triplane representation that is trained on a large-scale synthetic dataset. We show that hybrid local and global features can aid novel view synthesis from very sparse views. We show superior generalization performance on the challenging NeRDS 360 dataset while also offering powerful editing and composition capabilities through our method.
    
    \item \textbf{Chapter~\ref{chap:nerfmae}: NeRF-MAE: Masked AutoEncoders for Self Supervised 3D representations Learning for Neural Radiance Fields -} 
    
    This chapter explores the effectiveness of applying masked autoencoders as a pretraining method for Neural Radiance Fields (NeRFs) using Swin Transformers. Referred to as NeRF-Transformer, the approach pretrains on multi-view posed RGB data and is inspired by the success of similar techniques in 2D and 3D vision tasks. NeRF-MAE, the resulting self-supervised pre-trained model, significantly enhances performance in downstream 3D vision tasks, achieving state-of-the-art results in 3D object detection, voxel-grid super-resolution, and 3D voxel labeling, all while utilizing only posed 2D data for pretraining.
\end{itemize}

    \chapter{Background and Literature Review}
\label{chap:background_and_literature_review}

This chapter provides a comprehensive overview of the key research domains covered in this thesis while pointing out critical research gaps and includes a discussion of how this thesis addresses these gaps. Each section delves into the background and includes relevant literature related to the unique sub-challenges studied in this thesis, especially focusing on the connection of each research domain with prior knowledge or~\textit{prior}-centric learning.

\section{3D Scene Representations}

Lately, there has been a growing trend in the adoption of learning-based methods for 3D reconstruction~\cite{fan2017point, choy20163d, groueix2018, yuan2018pcn}. Unlike traditional multi-view stereo algorithms~\cite{yao2020blendedmvs, aanaes2016large}, these learned models have the capability to encompass extensive prior knowledge about 3D shapes, which proves invaluable in resolving uncertainties in the input data. Existing representations can be broadly categorized into explicit~(Voxel-based~\cite{zhou2018voxelnet}, Point-based~\cite{fan2017point, qi2017pointnet, qi2017pointnet++}, and mesh~\cite{gkioxari2019mesh, remelli2020meshsdf} representation) and implicit representations~\cite{park2019deepsdf, mescheder2019occupancy}~(shape represented implicitly using a neural-network as a scalar-field which is converted to an explicit representation using marching cubes or other similar procedures). Prior works mainly focus on single-object reconstruction or employ complex pipelines for reconstructing multiple objects~\cite{engelmann2021points, runz2020frodo, gkioxari2019mesh}. In contrast, we propose efficient retrieval of shape and appearance priors for multi-object reconstruction in Chapters~\ref{chap:centersnap} and~\ref{chap:shapo}. Next, we discuss common ways to encode priors into each of the two categories~(i.e., explicit and implicit) 3D scene representations.

\subsection{Priors for Object and Scene-Centric 3D Representations}

Dedicated neural network architectures have been designed for encoding prior knowledge of shape and appearance to achieve generalizable 3D reconstruction. These network architectures can be broadly categorized into auto-encoders~\cite{yuan2018pcn, choy20163d} and auto-decoder neural networks~\cite{park2019deepsdf, OechsleICCV2019}. An auto-encoder comprises an encoder neural network that embeds inputs into a low-dimensional latent representation. This latent representation is then decoded to produce the desired output, with the neural network weights learned through gradient descent. This architecture has found success in applications involving point clouds, hybrid deep marching tetrahedra (DMTet)~\cite{shen2021dmtet}, as well as voxels and meshes for shape completion~\cite{wen2020point, mittal2022autosdf}, reconstruction~\cite{liu2021voxel, choy20163d, fan2017point}, and 3D representation learning~\cite{yu2022point, liang2022meshmae}. In contrast, auto-decoder architectures~\cite{park2019deepsdf, heppert2023carto, zakharov2021single} utilize instance-specific latent codes to represent shapes, using a single neural network per category. While these architectures excel in representing individual shapes, they encounter challenges when it comes to representing complete scenes. Consequently, to encode scene-level priors, the approach of local feature conditioning~\cite{peng2020convolutional, chabra2020deep} has been introduced to achieve accurate scene and complex object reconstruction. Furthermore, to enhance the efficiency of encoding priors, various works have incorporated data structures like octrees~\cite{zakharov2022road, takikawa2021nglod}, enabling fast and accurate categorical reconstruction for many instances. Less efforts have been dedicated to efficient retrieval of shape and appearance priors as well as test-time optimization of priors obtained for scenes comprising multiple objects. We show single-shot shape and appearance prior retrieval in Chapter~\ref{chap:centersnap} and fast test-time optimization of 3D priors in Chapter~\ref{chap:shapo} using an octree-based differentiable optimization. 

\section{Neural Fields}
\label{sec:neural_fields}
Neural Fields~\cite{xie2021neural}, also termed as \emph{implicit neural representations} or \emph{coordinate-based MLPs}, take spatial 2D or 3D coordinates as input and produce scalar quantities, such as 1-channel \gls{sdf}~\cite{park2019deepsdf} for shape reconstruction or 4-channel RGB and density for \gls{nerf}~\cite{mildenhall2020nerf} reconstruction. The output of a coordinate-based \gls{mlp} is typically transformed into the desired output through a forward map, which in the case of SDF denotes sphere-tracing~\cite{liu2020dist} or iso-surface extraction~\cite{sdflabel}, among other methods. In the case of \gls{nerf}, it denotes volumetric rendering~\cite{barron2021mip, zhang2020nerf++}. Neural fields have been successfully applied in various domains, including object reconstruction~\cite{jang2021codenerf, mueller2022autorf}, unbounded scene modeling~\cite{fu2022panoptic, zhang2020nerf++}, few-view view synthesis~\cite{chen2021mvsnerf, yu2020pixelnerf}, and robotics applications~\cite{irshad2022implicitnerfroboticsresources}, such as object pose estimation~\cite{yen2020inerf}, navigation~\cite{adamkiewicz2022vision}, manipulation~\cite{simeonov2022neural}, and SLAM~\cite{zhu2022nice, Ortiz:etal:iSDF2022}. Less efforts have been dedicated to studying representation learning from NeRF. Our work in Chapter~\ref{chap:nerfmae} studies semantic and structural similarity prior to learn strong 3D representation from NeRFs while utilizing only posed 2D data as input. Next, we discuss methods to encode prior knowledge into neural field representations.

\subsection{Priors for Neural Fields}

The purpose of encoding priors in neural fields is to enable fast (amortized) inference as well as to offer a generalizable solution, which contrasts with the slow and per-scene optimization methods~\cite{mildenhall2020nerf, wang2021neus, barron2021mip, park2020nerfies} of neural fields, as depicted in Figure~\ref{fig:related_work} later in Section~\ref{chap:neo360,sec:intro}. These priors come in the form of global conditioners~\cite{jang2021codenerf, park2019deepsdf, xie2021fig}, where the conditioning variable ($z_{i}$) is optimized with neural-network weights ($\theta_{i}$) using a reconstruction loss. During inference, a new latent code ($z_{n}$) is optimized using fixed network weights learned during training. Additionally, local conditioning priors either use multi-view geometry constraints to encode local features~\cite{chen2021mvsnerf, yu2020pixelnerf, srt22} as conditioners or introduce multiple latent codes ($z$), where each latent code has control over a local neighborhood of coordinates~\cite{peng2020convolutional, liu2020neural, chabra2020deep, sharma2022seeing} to improve generalization and the efficiency of coordinate-based \gls{mlp}. While generalizable neural fields have shown great progress, they struggle with unbounded scenes. Our study in Chapter~\ref{chap:neo360} demonstrate that both local~(from image level features) and global~(volumetric priors from multi-view geometry) priors aid learning strong representations from sparse-view input data. 

\section{Embodied Aritifical Intelligence}

"Embodied AI"~\cite{duan2022survey} refers to the capability of artificial agents to acquire knowledge by engaging with the world around them. It predominantly entails the incorporation of traditional intelligence principles, including vision, language, and reasoning, into an artificial entity, with the aim of tackling AI-related issues within a simulated setting with efforts on simulation to reality transfer. Particular efforts have been dedicated to creating an infrastructure around bringing in real-world scanned datasets such as Matterport3D~\cite{Matterport3D}, Gibson~\cite{gibsonenv}, and others with emphasis on environments, assets, physics, control of agents and speed~\cite{duan2022survey}. The growing interest in Embodied AI has led to the creation of simulators in both synthetic~\cite{ALFRED20} and realistic domains~\cite{habitat19iccv, gibsonenv}. The development of embodied AI simulators has paved the way for a range of promising research endeavors in the field of embodied AI. These include tasks such as point-goal navigation~\cite{wijmans2020ddppo, chen2018learning}, embodied question-answering~\cite{embodiedqa} and embodied visual exploration~\cite{chaplot2020learning}. We study Vision-and-Language Navigation, which is an important embodied AI task, in Chapters~\ref{chap:robovln} and~\ref{chap:sasra} while incorporating key priors such as modularity and use of semantic maps as input to the neural network architectures to achieve better generalizability. 

\subsection{Vision-and-Language Navigation}

One prominent task in Embodied AI is~\gls{vln}~\cite{mattersim}. VLN is a challenge in which agents acquire the skill of traversing an environment through the guidance of natural language instructions. The difficulty of this endeavor lies in the need to process visual information and language instructions in a sequential manner. VLN continues to be a challenging task since it demands agents to predict sequential actions by drawing from past actions and instructions. Sophisticated architectures such as Recurrent Neural Networks~(RNN)~\cite{sutskever2013training} have been developed for the task of VLN~\cite{mattersim} with focus on multi-modality discriminative learning~\cite{DBLP:journals/corr/abs-1905-13358}, cross-modal matching with self-supervised imitation objective~\cite{DBLP:conf/cvpr/WangHcGSWWZ19}, reinforcement learning objectives~\cite{DBLP:conf/eccv/WangXWW18}, self-supervised auxiliary learning~\cite{ma2019selfmonitoring, ma2019theregretful, Zhu_2020_CVPR} and use of attention and Transformers~\cite{landi2020perceive, vasudevan2020talk2nav}. Existing~\gls{vln}~\cite{mattersim, vasudevan2020talk2nav} works focus on a simpler discrete setting hence making assumptions such as known topology, deterministic navigation, and perfect location. These assumptions make the problem simpler as well as deviate from the real-world both in terms of perception and control.

\subsection{Priors in Embodied AI}

Significant efforts have been dedicated to adding inductive bias and priors in the Embodied AI task. Since traditional deep learning architectures such as~\gls{mlp} usually struggle with reasoning and long-range predictions, one approach to adding priors to aid embodied learning is modularity~\cite{pmlr-v80-le18a}. Modularity has been studied for visual exploration~\cite{chaplot2020learning, chaplot2020semantic} via global and local policy disentanglement and Objet-goal navigation~\cite{chaplot2020object} through semantic-map-based policy disentanglement where the high-level policy predicts an action on the semantic map and the low-level controller takes low-level actions to reach the point-goal specified by the high-level policy. Priors in~\gls{vln} come from the use of auxiliary knowledge where specifically designed losses~\cite{ma2019selfmonitoring, ma2019theregretful} are designed to aid the autonomous agent in tackling the challenging task. One approach to utilizing axillary knowledge in VLN systems is~\cite{Zhu_2020_CVPR}, which introduces four supplementary cognitive tasks, including retelling the trajectory, estimating progress towards the goal, predicting the angle or heading, and alignment of cross-modality. Through dedicated self-supervised losses, the agent acquires the ability to engage in reasoning about past actions and forecast forthcoming information relating to the challenging task of VLN in unstructured environments. Other priors in VLN come from utilizing a graphical planner and a Graph Neural Networks~\cite{deng2020evolving} to learn long-range planning, learning to back-track~\cite{ma2019theregretful, ke2019tactical}, and creating a topological map~\cite{chen2020topological} and an attention map to create a navigation plan on the created map. Structured memory from topologically built scene representation~\cite{wang2021structured} has also been shown to help with long-range planning as well as flexible action spaces due to the modularity of the representation employed. Priors have also been successfully employed in the VLN task through future-view image semantics generation~\cite{li2023improving} as well as mask modelling~\cite{dou2023masked, qiao2022hop}. Particularly,~\cite{qiao2022hop} employs 5 proxy tasks including masked language modeling, trajectory instruction matching, group order modeling, action prediction, and trajectory order modeling. While both modularity and semantic map prior has been studied for embodied AI systems, fewer efforts have been dedicated to studying them for VLN agents. We show their impact on VLN systems in improving generalizability in Chapters~\ref{chap:robovln} and~\ref{chap:sasra}.

\section{3D Self-Supervised Learning}

Self-supervised learning is a popular approach in 2D vision to enable representation learning. Since labeling large amounts of data is difficult, it offers a great alternative for label-free learning on massive amounts of data. One approach is to utilize masked modeling, which has shown to be successful in building representations for language modeling~\cite{devlin2018bert} and 2D vision tasks through image-patch masking~\cite{he2021masked}. This has inspired a number of works in 3D representation learning where the goal is to reconstruct masked 3D patches using auto-encoder architectures. The resulting encoder weights are used for fine-tuning downstream 3D tasks. Masked image modeling has been successfully applied for pointcloud pretraining~\cite{zhang2022point, pang2022masked}, mesh representation learning~\cite{liang2022meshmae}, and voxel pretraining~\cite{min2022voxel}. Other works utilize contrastive point-to-pixel knowledge transfer~\cite{liu2021learning} as well as multi-scale hierarchical features~\cite{zhang2022point} for pre-training 3D representations. A separate line of work creates 3D scene representations in a self-supervised manner from 2D images, as discussed in detail in Section~\ref{sec:neural_fields}. These representations do not require 3D supervision for learning 3D scene representations as common in previous works~\cite{park2019deepsdf, mescheder2019occupancy}. Most notably, DVR~\cite{niemeyer2020differentiable} learns implicit 3D representations without 3D supervision, IDR~\cite{yariv2020multiview} learns multi-view geometry by disentangling shape and appearance, and \gls{nerf}~\cite{mildenhall2020nerf} learns a 3D scene using an implicit neural-network based prediction of radiance and density followed by differentiable volumetric rendering. For more details, please see Section~\ref{sec:neural_fields}. Another line of work utilizes 3D self-supervision in the form of chamfer loss~\cite{lunayach2023fsd} and geometric correspondences~\cite{zhang2022self} to learn 3D object properties, such as 6D pose and shape, using self-supervised learning. Traditional representation learning demands domain-specific data for pretraining. Contrastingly, we show in Chapter~\ref{chap:nerfmae} that only posed 2D data is enough for efficient 3D representation learning utilizing NeRF as a bridge between 2D and 3D domains.

\section{Transformer Architecture in Language and Vision}

Transformer~\cite{NIPS2017_7181} architectures have been a popular choice in deep learning in place of traditional~\gls{mlp}. They have demonstrated great performance in language tasks~\cite{khan2022transformers} such as classification, translation, and question answering. Their success lies in modeling the relationship between sequence elements through the use of an attention mechanism which has been shown to scale well to massive amounts of data~\cite{radford2018improving}. Transformers, in comparison to their convolutional~\cite{heDeepResidualLearning2016} and recurrent counterparts~\cite{sutskever2013training}, operate with little initial understanding of the problem's inherent structure. Therefore, they are frequently pretrained on large, unlabeled datasets with pretext proxy tasks. This pretraining strategy circumvents the need for intensive manual annotations, instead focusing on creating versatile representations that effectively capture intricate relationships among the elements present in a given dataset. The success of transformers in language modeling has carried over to vision.~\cite{dosovitskiy2020image} introduced~\gls{vit} which splits an image into fixed-size patches, and after linearly embedding them and adding positional encoding, feeds them into a standard transformer. Follow-up works have applied Transformers for self-supervised 2D representation~\cite{caron2021emerging, he2021masked} and 3D representation learning~\cite{bachmann2022multimae} as well as several 2D~\cite{liu2021swin, detr} and 3D vision tasks~\cite{hatamizadeh2022unetr, mao2021voxel, zhao2021point, tang2022self, li2022bevformer, stier2021vortx, srt22}. We extend the success in Transformer architectures to show them for cross-modal reasoning for embodied AI tasks in Chapters~\ref{chap:robovln} and~\ref{chap:sasra} and for 3D representation learning from posed 2D data in Chapter~\ref{chap:nerfmae}.

    \part{Efficient Object-Centric Neural 3D Representations}
    \label{part1}

    This part is dedicated to the investigation of efficient object-centric 3D representations, establishing connections with 2D detections to propose end-to-end systems that enhance real-world 3D object understanding. Our study involves the incorporation of \emph{geometry and appearance priors} derived from synthetically simulated data, facilitating the acquisition of robust 3D representations that effectively enable real-world transfer with minimal fine-tuning. In Chapter~\ref{chap:centersnap}, titled "CenterSnap: Single-Shot Multi-Object 3D Shape Reconstruction and Categorical 6D Pose and Size Estimation," we delve into the formalization of an innovative~\emph{single-shot setting} for 3D shape reconstruction and 6D pose and size estimation. Diverging from conventional methods reliant on multi-stage pipelines and CAD models, this chapter introduces a one-stage solution that operates without the need for 2D bounding boxes or separate models for distinct categories during inference. Leveraging a per-pixel representation, our approach achieves real-time reconstruction, showcasing substantial performance enhancements and making a noteworthy contribution to the field. In Chapter~\ref{chap:shapo}, entitled "ShAPO: Implicit Representations for Multi-Object Shape Appearance and Pose Optimization," we present a disentangled shape and appearance representation coupled with an octree-based iterative optimization. This framework enables precise identification of multiple objects' appearances, shapes, and 6D poses. Our study demonstrates that incorporating shape and appearance priors from neural networks trained predominantly on synthetic data seamlessly transfers to real-world scenarios with minimal fine-tuning. Moreover, it allows for inference-time optimization based on single-view observations.
    \clearpage
    
    \chapter{CenterSnap: Single-Shot Multi-Object 3D Shape Reconstruction and Categorical 6D Pose and Size Estimation}
\label{chap:centersnap}
\looseness=-1

In this chapter, I explore the problem setting of~\textbf{object-centric 3D scene reconstruction} with the main aim of making reconstruction and pose estimation~\textit{more efficient} while utilizing~\textit{largre-scale shape priors} in the form of synthetic-data for real-world generalization.

\section{Introduction}
\label{chap:centersnap,sec:intro}

Multi-object 3D shape reconstruction and  6D pose~(i.e. 3D orientation and position) and size estimation from raw visual observations is crucial for robotics manipulation~\cite{cifuentes2016probabilistic, jiang2021synergies, laskey2021simnet}, navigation~\cite{qi2018frustum, chen20153d} and scene understanding~\cite{zhang2021holistic, Nie_2020_CVPR}. The ability to perform pose estimation in real-time leads to fast feedback control~\cite{kappler2018real} and the capability to reconstruct complete 3D shapes~\cite{ kuo2020mask2cad, niemeyer2020differentiable, mescheder2019occupancy} results in fine-grained understanding of local geometry, often helpful in robotic grasping~\cite{jiang2021synergies, ferrari1992planning}. Recent advances in deep learning have enabled great progress in~\textit{instance-level} 6D pose estimation~\cite{kehl2017ssd, rad2017bb8, xiang2018posecnn} where the exact 3D models of objects and their sizes are known a-priori. Unfortunately, these methods~\cite{tekin2018real, peng2019pvnet, wang2019densefusion} do not generalize well to realistic-settings on novel object instances with unknown 3D models in the same category, often referred to as~\textit{category-level settings}. Despite progress in category-level pose estimation, this problem remains challenging even when similar object instances are provided as priors during training, due to a high variance of objects within a category.

\begin{figure}[htb]
\centering
\includegraphics[width=0.95\columnwidth]{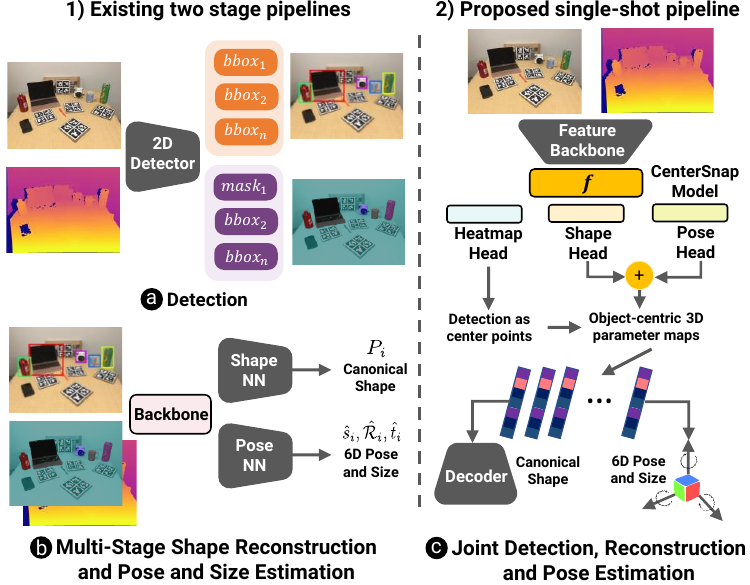}
\centering
  \caption{
  \textbf{Overview:}
    \textbf{(1)} Multi-stage pipelines in comparison to \textbf{(2)} our single-stage approach. The single-stage approach uses object instances as centers to jointly optimize 3D shape, 6D pose, and size.
  }
  \label{ceterpoint_comparison}
\end{figure}

Recent works on shape reconstruction~\cite{gkioxari2019mesh, kato2018neural} and category-level 6D pose and size estimation~\cite{tian2020shape, wang2019normalized, sundermeyer2020augmented} use complex multi-stage pipelines. As shown in Figure~\ref{ceterpoint_comparison}, these approaches independently employ two stages, one for performing 2D detection~\cite{girshick14CVPR, he2017mask, ren2015faster} and another for performing object reconstruction or 6D pose and size estimation. This pipeline is computationally expensive, not scalable, and has low performance on real-world novel object instances, due to the inability to express explicit representation of shape variations within a category. Motivated by above, we propose to reconstruct complete 3D shapes and estimate 6D pose and sizes of novel object instances within a specific category, from a single-view RGB-D in a~\textit{single-shot manner}.\\
To address these challenges, we introduce Center-based Shape reconstruction and 6D pose and size estimation (CenterSnap), a single-shot approach to output complete 3D information (3D shape, 6D pose and sizes of multiple objects) in a bounding-box proposal-free and per-pixel manner. Our approach is inspired by recent success in anchor-free, single-shot 2D key-point estimation and object detection~\cite{nie2019single, zhou2019objects, zhou2020tracking, duan2019centernet}. As shown in Figure~\ref{ceterpoint_comparison}, we propose to learn a spatial per-pixel representation of multiple objects at their center locations using a feature pyramid backbone~\cite{laskey2021simnet, girshick14CVPR}. Our technique directly regresses multiple shape, pose, and size codes, which we denote as \textit{object-centric 3D parameter maps}. At each object's center point in these spatial object-centric 3D parameter maps, we predict vectors denoting the complete 3D information~(i.e. encoding 3D shape, 6D pose and sizes codes). 
A 3D auto-encoder~\cite{fan2017point, park2019deepsdf} is designed to learn canonical shape codes from a large database of shapes. A joint optimization for detection, reconstruction and 6D pose and sizes for each object's spatial center is then carried out using learned shape priors. Hence, we perform complete 3D scene-reconstruction and predict 6D pose and sizes of novel object instances in a single-forward pass, foregoing the need for complex multi-stage pipelines~\cite{girshick14CVPR, gkioxari2019mesh, wang2019normalized}.

Our proposed method leverages a simpler and computationally efficient pipeline for a complete object-centric 3D understanding of multiple objects from a single-view RGB-D observation. We make the following contributions:
\begin{itemize}
\item Present the first work to formulate object-centric~\textit{holistic scene-understanding}~(i.e. 3D shape reconstruction and 6D pose and size estimation) for multiple objects from a single-view RGB-D in a~\textit{single-shot manner}. 
\item Propose a fast~(real-time) joint reconstruction and pose estimation system. Our network runs at 40 FPS on a NVIDIA Quadro RTX 5000 GPU.
\item Our method significantly outperforms all baselines for 6D pose and size estimation on NOCS benchmark, with over $12\%$ absolute improvement in mAP for 6D pose.
\end{itemize}

\section{Related works}
\label{chap:centersnap,sec:related_works}
\textbf{3D shape prediction and completion:} 3D reconstruction from a single-view observation has seen great progress with various input modalities studied. RGB-based shape reconstruction~\cite{fan2017point, choy20163d, groueix2018} has been studied to output either pointclouds, voxels, or meshes~\cite{park2019deepsdf, mescheder2019occupancy, chen2019learning}. Contrarily, learning-based 3D shape completion~\cite{Yang18, varley2017shape, yuan2018pcn} studies the problem of completing partial point clouds obtained from masked depth maps. However, all these works focus on reconstructing a single object. In contrast, our work focuses on multi-object reconstruction from a single RGB-D. Recently, multi-object reconstruction from RGB-D has been studied~\cite{engelmann2021points, runz2020frodo, gkioxari2019mesh}. However, these approaches employ complex multi-stage pipelines employing 2D detections and then predicting canonical shapes. Our approach is a simple, bounding-box proposal-free method that jointly optimizes for detection, shape reconstruction and 6D pose and size. 

\textbf{Instance-Level and Category-Level 6D Pose and Size Estimation:} Works on Instance-level pose estimation use classical techniques such as template matching~\cite{kehl2016deep, sundermeyer2018implicit, tejani2014latent}, direct pose estimation~\cite{kehl2017ssd, wang2019densefusion, xiang2018posecnn} or point correspondences~\cite{tekin2018real, rad2017bb8}. Contrarily, our work closely follows the paradigm of category-level pose and size estimation where CAD models are not available during inference. Previous work has employed complex multi-stage pipelines~\cite{wang2019normalized,tian2020shape, chen2020learning} for category-level pose estimation. Our work optimizes for shape, pose, and sizes jointly, while leveraging the shape priors obtained by training a large dataset of CAD models. CenterSnap is a simpler, more effective, and faster solution.   

\textbf{Per-pixel point-based representation} has been effective for anchor-free object detection and segmentation. These approaches~\cite{zhou2019objects, duan2019centernet, wang2020centermask} represent instances as their centers in a spatial 2D grid. This representation has been further studied for key-point detection~\cite{nie2019single}, segmentation~\cite{tian2020conditional,wang2020solo} and body-mesh recovery~\cite{sun2020centerhmr, ROMP}. Our approach falls in a similar paradigm and further adds a novelty to reconstruct object-centric holistic 3D information in an anchor-free manner. Different from~\cite{engelmann2021points, sun2020centerhmr}, our approach 1) considers pre-trained shape priors on a large collection of CAD models 2) jointly optimizes categorical shape 6D pose and size, instead of 3D-bounding boxes and 3) considers more complicated scenarios~(such as occlusions, a large variety of objects and sim2real transfer with limited real-world supervision).
\section{Method}
\label{chap:centersnap,sec:method}
\begin{figure}[t!]
\centering
\includegraphics[width=1.0\textwidth]{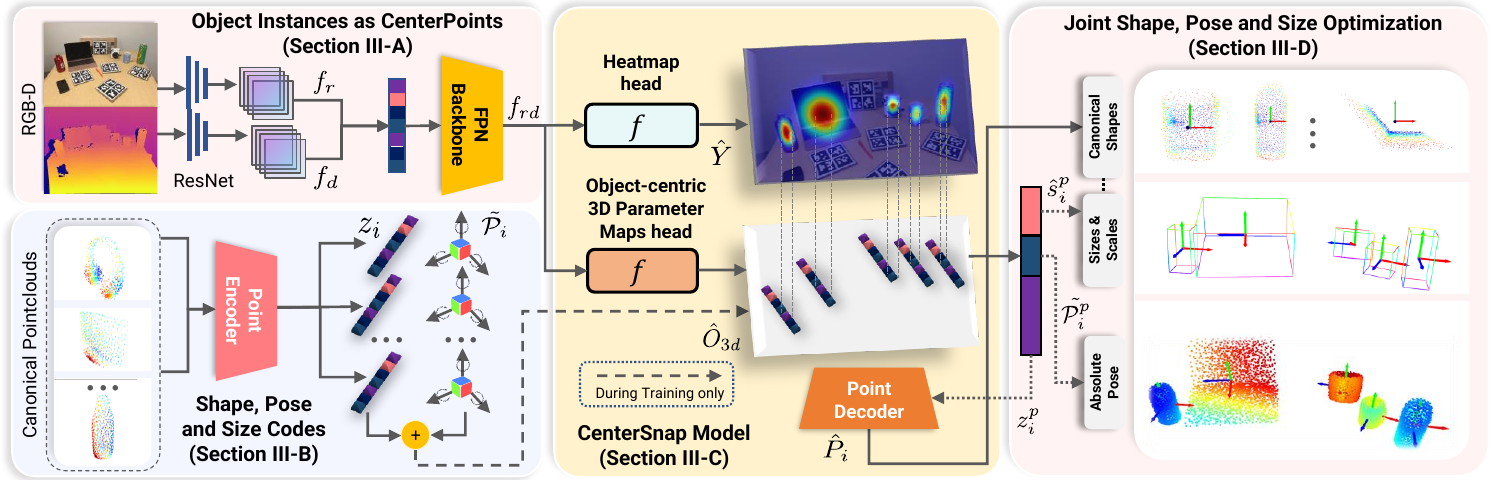}
\captionof{figure}{
\textbf{CenterSnap Method:} Given a single-view RGB-D observation, our proposed approach jointly optimizes for shape, pose, and sizes of each object in a single-shot manner. Our method comprises a joint~\gls{fpn} backbone for feature extraction~(Section \ref{backbone}), a pointcloud auto-encoder to extract shape codes from a large collection of CAD models~(Section~\ref{shapecode}), CenterSnap model which constitutes multiple specialized heads for heatmap and object-centric 3D parameter map prediction~(Section~\ref{centermodel}) and joint optimization for shape, pose, and sizes for each object's spatial center~(Section \ref{optimize}).
}
\label{framework}
\end{figure}

Given an RGB-D image as input, our goal is to simultaneously detect, reconstruct and localize all unknown object instances in the 3D space. In essence, we regard shape reconstruction and pose and size estimation as a point-based representation problem where each object's complete 3D information is represented by its center point in the 2D spatial image. Formally, given an RGB-D single-view observation~($I$ $\in$ $\mathds{R}^{h_{o}\times w_{o}\times 3 }$, $D$ $\in$ $\mathds{R}^{h_{o}\times w_{o}}$) of width $w_{o}$ and height $h_{o}$, our aim is to reconstruct the complete pointclouds~($P$ $\in$ $\mathds{R}^{K\times N\times 3 }$) coupled with 6D pose and scales~($\tilde{\mathbf{\mathcal P}}$ $\in$ $SE(3)$, $\hat{s}$ $\in$ $\mathbb{R}^3$) of all object instances in the 3D scene, where K is the number of arbitrary objects in the scene and N denotes the number of points in the pointcloud. The pose~($\tilde{\mathbf{\mathcal P}}$ $\in$ $SE(3)$) of each object is denoted by a 3D rotation $\hat{\mathcal{R}}$ $\in$ $SO(3)$ and a translation $\hat{t}$ $\in$ $\mathbb{R}^3$. The 6D pose $\tilde{\mathbf{\mathcal P}}$, 3D size (spatial extent obtained from canonical pointclouds~$P$) and 1D scales~$\hat{s}$ completely defines the unknown object instances in 3D space with respect to the camera coordinate frame.
To achieve the above goal, we employ an end-to-end trainable method, as illustrated in Figure~\ref{framework}. First, objects instances are detected as heatmaps in a per-pixel manner~(Section~\ref{backbone}) using a CenterSnap detection backbone based on feature pyramid networks~\cite{laskey2021simnet, lin2017feature}. Second, a joint shape, pose, and size code denoted by object-centric 3D parameter maps is predicted for detected object centers using specialized heads~(Section~\ref{centermodel}). Our pre-training of shape codes is described in Section~\ref{shapecode}. Lastly, 2D heatmaps and our novel object-centric 3D parameter maps are jointly optimized to predict shapes, pose and sizes in a single-forward pass~(Section~\ref{optimize}).

\subsection{Object instances as center points}
\label{backbone}
We represent each object instance by its 2D location in the spatial RGB image following~\cite{zhou2019objects, duan2019centernet}. Given a RGB-D observation~($I$ $\in$ $\mathds{R}^{h_{o}\times w_{o}\times 3 }$, $D$ $\in$ $\mathds{R}^{h_{o}\times w_{o}}$), we generate a low-resolution spatial feature representations $f_{r}$ $\in$ $\mathds{R}^{h_{o}/4\times w_{o}/4\times C_{s}}$ and $f_{d}$ $\in$ $\mathds{R}^{h_{o}/4\times w_{o}/4\times C_{s}}$ by using Resnet~\cite{he2016deep} stems, where $C_{s}=32$. We concatenate computed features $f_{r}$ and $f_{d}$ along the channel dimension before feeding it to Resnet18-FPN backbone~\cite{kirillov2019panoptic} to compute a~\textit{pyramid of features} ($f_{rd}$) with scales ranging from 1/8 to 1/2 resolution, where each pyramid
level has the same channel dimension~(i.e. 64). We use these combined features with a specialized heatmap head to predict object-based heatmaps~$\hat{Y} \in [0,1]^{\frac{h_{o}}{R} \times \frac{w_{o}}{R} \times 1}$ where $R$\,=\,$8$ denotes the heat-map down-sampling factor. Our specialized heatmap head design merges the semantic information from all FPN levels into one output~($\hat{Y}$). We use three upsampling stages followed by element-wise sum and $softmax$ to achieve this. This design allows our network to 1) capture multi-scale information and 2) encode features at higher resolution for effective reasoning at the per-pixel level. We train the network to predict ground-truth heatmaps~($Y$) by minimizing MSE loss, $\mathcal{L}_{\text{inst}}=\sum_{xyg}\left(\hat{Y}-Y\right)^{2}$. The Gaussian kernel~${Y_{xyg} = \exp\left(-\frac{(x-c_x)^2+(y- c_y)^2}{2\sigma^2}\right)}$ of each center in the ground truth heat-maps~($Y$) is relative to the scale-based standard deviation~$\sigma$ of each object, following~\cite{laskey2021simnet, zhou2019objects, duan2019centernet, law2018cornernet}. 

\subsection{Shape, Pose, and Size Codes}
\label{shapecode}
To jointly optimize the object-based heatmaps, 3D shapes and 6D pose and sizes, the complete object-based 3D information (i.e. Pointclouds $P$, 6D pose $\tilde{\mathbf{\mathcal P}}$ and scale $\hat{s}$) are represented as as object-centric 3D parameter maps~($O_{3d}$ $\in$ $\mathds{R}^{h_{o}\times w_{o}\times 141}$). $O_{3d}$ constitutes two parts, shape latent-code and 6D Pose and scales. The pointcloud representation for each object is stored in the object-centric 3D parameter maps as a latent-shape code~($z_{i}$ $\in$ $\mathbb{R}^{128}$). The ground-truth Pose~($\tilde{\mathbf{\mathcal P}}$) represented by a $3\times 3$ rotation $\hat{\mathcal{R}}$ $\in$ $SO(3)$ and translation $\hat{t}$ $\in$ $\mathbb{R}^3$ coupled with 1D scale $\hat{s}$ are vectorized to store in the $O_{3d}$ as 13-D vectors. To learn a shape-code~($z_{i}$) for each object, we design an auto-encoder trained on all 3D shapes from a set of CAD models. Our auto-encoder is representation-invariant and can work with any shape representation. Specifically, we design an encoder-decoder network (Figure~\ref{auto-encoder}), where we utilize a Point-Net encoder~($g_{\phi}$) similar to~\cite{qi2017pointnet}. The decoder network~($d_{\theta}$), which comprises three fully-connected layers, takes the encoded low-dimensional feature vector i.e. the latent shape-code ($z_{i}$) and reconstructs the input pointcloud $\hat{P}_{i} = d_{\theta}(g_{\phi}(P_{i}))$. To train the auto-encoder, we sample 2048 points from the ShapeNet~\cite{chang2015ShapeNet} CAD model repository and use them as ground-truth shapes. Furthermore, we unit canonicalize the input pointclouds by applying a scaling transform to each shape such that the shape is centered at origin and unit normalized. We optimize the encoder and decoder networks jointly using the reconstruction error, denoted by Chamfer Distance, as shown below:
\begin{equation} \label{eq:distance_Chamfer}
     D_{cd}(\mathbf{P}_{i}, \hat{\mathbf{P}}_{i}) = 
     \frac{1}{|\mathbf{P}_{i}|} \sum_{x \in \mathbf{P}_{i}} \min_{y \in \hat{\mathbf{P}}_{i}}\|x-y\|_{2}^{2} + \frac{1}{|\hat{\mathbf{P}}_{i}|}\sum_{\mathbf{y} \in \hat{\mathbf{P}}_{i}} \min_{x \in \mathbf{P}_{i}}\|x-y\|_{2}^{2} \nonumber
\end{equation}

Sample decoder outputs and t-SNE embeddings~\cite{van2008visualizing} for the latent shape-code~($z_{i}$) are shown in Figure~\ref{auto-encoder} and our complete 3D reconstructions on novel real-world object instances are visualized in Figure~\ref{reconstruction_qualitative} as pointclouds, meshes and textures. Our shape-code space provides a compact way to encode 3D shape information from a large number of CAD models. As shown by the t-SNE embeddings (Figure~\ref{auto-encoder}), our shape-code space finds a distinctive 3D space for semantically similar objects and provides an effective way to scale shape prediction to a large number (i.e. 50+) of categories.

\subsection{CenterSnap Model}
\label{centermodel}

\begin{figure}[htb]
\centering
\includegraphics[width=0.90\columnwidth]{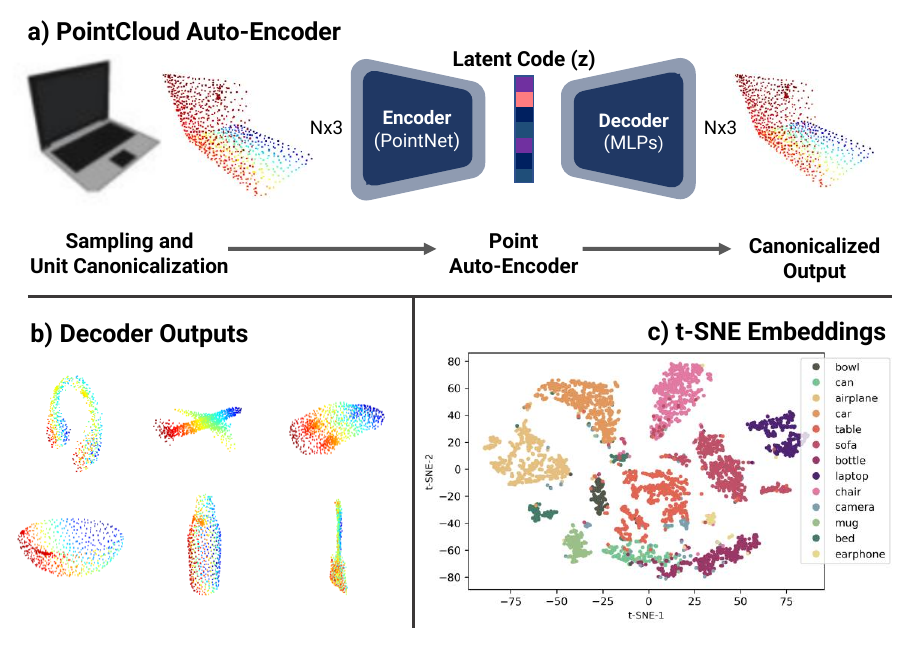}
\centering
  \caption{
  \textbf{Shape Auto-Encoder:} We design a Point Auto-encoder~(\textbf{a}) to find unique shape-code~($z_{i}$) for all the shapes. Unit canonicalized pointcloud outputs from the decoder network are shown in~(\textbf{b}). t-SNE embeddings for shape-code~($z_{i}$) are visualized in~(\textbf{c})
  }
  \label{auto-encoder}
\end{figure}

Given object center heatmaps~(Section~\ref{backbone}), the goal of the CenterSnap model is to infer object-centric 3D parameter maps that define each object instance completely in the 3D space. The CenterSnap model comprises a task-specific head similar to the heatmap head~(Section~\ref{backbone}) with the input being \textit{pyramid of features}~($f_{rd}$). During training, the task-specific head outputs a 3D parameter map $\hat{O}_{3d}$ $\in$ $\mathds{R}^{\frac{h_{o}}{R}\times \frac{w_{o}}{R}\times 141}$ where each pixel in the down-sampled map~($\frac{h_{o}}{R}\times \frac{w_{o}}{R}$) contains the complete object-centric 3D information (i.e. shape-code $z_{i}$, 6D pose $\tilde{\mathbf{\mathcal P}}$ and scale $\hat{s}$) as 141-D vectors, where $R = 8$. Note that, during training, we obtain the ground-truth shape-codes from the pre-trained point-encoder $\hat{z}_{i} = g_{\phi}(P_{i})$. For Pose~($\tilde{\mathbf{\mathcal P}}$), our choice of rotation representation $\hat{\mathcal{R}}$ $\in$ $SO(3)$ is determined by stability during training~\cite{zhou2019continuity}. Furthermore, we project the predicted~$3\times3$ rotation $\hat{\mathcal{R}}$ into $SO(3)$, as follows: $\operatorname{SVD}^{+}(\hat{\mathcal{R}})=U \Sigma^{\prime} V^{T}, \text { where } \Sigma^{\prime}=\operatorname{diag}\left(1, 1, \operatorname{det}\left(U V^{T}\right)\right)$
To handle ambiguities caused by rotational symmetries, we also employ a rotation mapping function defined by~\cite{pitteri2019object}. The mapping function, used only for symmetric objects~\textit{(bottle, bowl, and can)}, maps ambiguous ground-truth rotations to a single canonical rotation by normalizing the pose rotation.
During training, we jointly optimize the predicted object-centric 3D parameter map~($\hat{O}_{3d}$) using a masked Huber loss~(Eq. \ref{huberloss}), where the Huber loss is enforced only where the Gaussian heatmaps ($Y$) have score greater than 0.3 to prevent ambiguity in areas where no objects exist. Similar to the Gaussian distribution of heatmaps in Section \ref{backbone}, the ground-truth \textit{Object 3D-maps}~($O_{3d}$) are calculated using the scale-based Gaussian kernel~$Y_{xyg}$ of each object. 
\begin{equation}
\small
\label{huberloss}
\resizebox{0.91\hsize}{!}{%
$\mathcal{L}_{3D}(O_{3d},\hat{O}_{3d})=\left\{\begin{array}{cc}
\frac{1}{2}(O_{3d}-\hat{O}_{3d})^{2} & if|(O_{3d}-\hat{O}_{3d})|<\delta \\
\delta\left((O_{3d}-\hat{O}_{3d})-\frac{1}{2} \delta\right) & \text {otherwise}
\end{array}\right.$}
\end{equation}

\textbf{Auxiliary Depth-Loss}: We additionally integrate an auxiliary depth reconstruction loss $\mathcal{L}_{D}$ for effective sim2real transfer, where $\mathcal{L}_{D}(D, \hat{D})$ minimizes the Huber loss (Eq. \ref{huberloss}) between target depth~($D$) and the predicted depth~($\hat{D}$) from the output of task-specific head, similar to the one used in Section~\ref{backbone}. The depth auxiliary loss (further investigated empirically using ablation study in Section~\ref{chap:centersnap,sec:exp}) forces the network to learn geometric features by reconstructing artifact-free depth. Since real depth sensors contain artifacts, we enforce this loss by pre-processing the input synthetic depth images to contain noise and random eclipse dropouts~\cite{laskey2021simnet}.
\subsection{Joint Shape, Pose, and Size Optimization}
\label{optimize}
We jointly optimize for detection, reconstruction, and localization. Specifically, we minimize a combination of heatmap instance detection, object-centric 3D map prediction and auxiliary depth losses as $\mathcal{L}= \lambda_{l}\mathcal{L}_{inst} + \lambda_{O_{3d}}\mathcal{L}_{O_{3d}} +  \lambda_{d}\mathcal{L}_{D}
$
where $\lambda$ is a weighting coefficient with values determined empirically as 100, 1.0 and 1.0 respectively.

\textbf{Inference}:

\begin{figure}[htb]
\centering
\includegraphics[width=0.90\columnwidth]{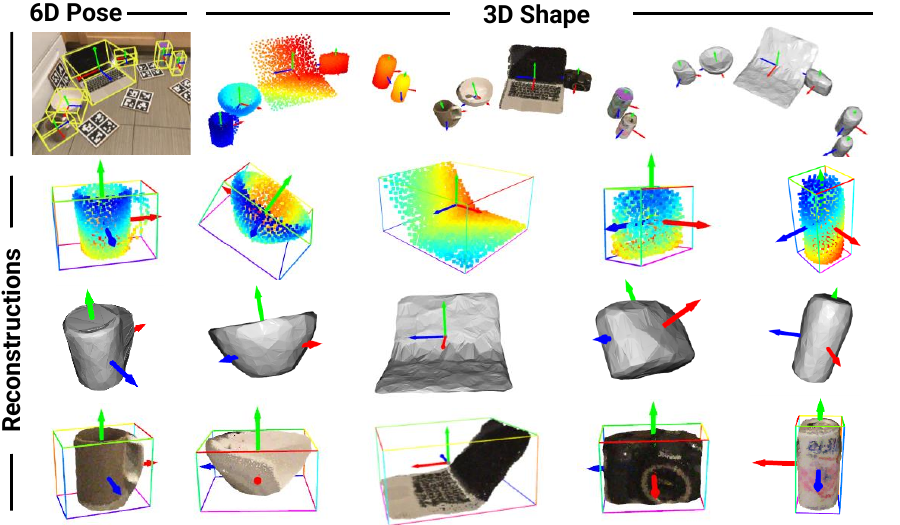}
\centering
  \caption{
  \textbf{Sim2Real Reconstruction:} Single-shot sim2real shape reconstructions on NOCS showing pointclouds, meshes and textures.
  }
  \label{reconstruction_qualitative}
\end{figure}

During inference, we perform peak detection as in~\cite{zhou2019objects} on the heatmap output ($\hat{Y}$) to get detected centerpoints for each object, {$c_{i}$ $in$ $\mathbb{R}^2$} = ($x_{i}, {y_{i}}$) (As shown in Figure~\ref{framework}~\emph{middle}). These centerpoints are the local maximum in heatmap output~($\hat{Y}$). We perform non-maximum suppression on the detected heatmap maximas using a~$3\times 3$ max-pooling, following~\cite{zhou2019objects}. Lastly, we directly sample the object-centric 3D parameter map of each object from~$\hat{O}_{3d}$ at the predicted center location~($c_{i}$) via~$\hat{O}_{3d}(x_{i}, {y_{i}})$. We perform inference on the extracted latent-codes using point-decoder to reconstruct pointclouds~($\hat{P}_{i} = d_{\theta}(z_{i}^{p})$). Finally, we extract~$3\times 3$ rotation $\hat{\mathcal{R}}_{i}^{p}$, 3D translation vector $\hat{t}_{i}^{p}$ and 1D scales $\hat{s}_{i}^{p}$ from~$\hat{O}_{3d}$ to get transformed points in the 3D space $\hat{P}_{i}^{recon} = [\hat{\mathcal{R}}_{i}^{p} | \hat{t}_{i}^{p}]*\hat{s}_{i}^{p}*\hat{P}_{i}$ (As shown in Figure~\ref{framework}~\emph{right-bottom}).
\section{Experiments and Results}
\label{chap:centersnap,sec:exp}

In this section, we aim to answer the following questions: 1) How well does CenterSnap reconstruct multiple objects from a single-view RGB-D observation? 2) Does CenterSnap perform fast pose estimation in real-time for real-world applications? 3) How well does CenterSnap perform in terms of 6D pose and size estimation? 

\textbf{Datasets:} We utilize the \textbf{NOCS}~\cite{wang2019normalized} dataset to evaluate both shape reconstruction and categorical 6D pose and size estimation. We use the CAMERA dataset for training which contains 300K synthetic images, where 25K are held out for evaluation. Our training set comprises 1085 object instances from 6 different categories - \textit{bottle, bowl, camera, can, laptop and mug} whereas the evaluation set contains 184 different instances. The REAL dataset contains
4300 images from 7 different scenes for training, and 2750 real-world images from 6 scenes for evaluation. Further, we evaluate multi-object reconstruction and completion using~\gls{mos}. We generate this dataset using the SimNet~\cite{laskey2021simnet} pipeline. Our datasets contain $640px \times 480px$ renderings of multiple~(3-10) ShapeNet objects~\cite{chang2015ShapeNet} in a table-top scene. Following~\cite{laskey2021simnet}, we randomize over lighting and textures using OpenGL shaders with PyRender~\cite{pyrender}.
Following~\cite{yuan2018pcn}, we utilize 30974 models from 8 different categories for training~(i.e. MOS-train): \textit{airplane, cabinet, car, chair, lamp, sofa, table}. We use the held-out set (MOS-test) of 150 models for testing from a novel set of categories - \textit{bed, bench, bookshelf, and bus}. \\
\textbf{Evaluation Metrics}: Following~\cite{wang2019normalized}, we independently evaluate the performance of 3D object detection and 6D pose estimation using the following key metrics: 1) Average-precision for various~\gls{iou}-overlap thresholds (\textbf{IOU25} and \textbf{IOU50}). 2) Average precision of object instances for which the error is less than $n^{\circ}$ for rotation and $m$ cm for translation (\textbf{5\textdegree \SI{5}{\cm}}, \textbf{\textbf{5\textdegree \SI{10}{\cm}}} and \textbf{\textbf{10\textdegree \SI{10}{\cm}}}). For shape reconstruction, we use Chamfer distance~(CD) following~\cite{yuan2018pcn}.\\
\textbf{Implementation Details:} CenterSnap is trained on the CAMERA training dataset, with fine-tuning on the REAL training set. We use a batch size of 32 and trained the network for 40 epochs with early-stopping based on the performance of the model on the held out validation set. We found data-augmentation~(i.e. color-jitter) on the real-training set to be helpful for stability and training performance. The auto-encoder network comprises a Point-Net encoder~\cite{qi2017pointnet} and three-layered fully-connected decoder, each with output dimension of $512, 1024$ and $1024\times3$. The auto-encoder is frozen after initially training on CAMERA CAD models for 50 epochs. We use Pytorch~\cite{NEURIPS2019_9015} for all our models and training pipeline implementation. For shape-completion experiments, we train only on MOS-train with testing on MOS-test.\\

\begin{table*}[t]
    \centering
    \renewcommand{\arraystretch}{1.3}
    \caption{
    \textbf{Quantitative comparison of 3D object detection and 6D pose estimation on NOCS}~\cite{wang2019normalized}: Comparison with strong baselines. Best results are highlighted in \textbf{bold}. $*$ denotes the method does not evaluate size and scale hence does not report IOU metric. For a fair comparison with other approaches, we report the per-class metrics using nocs-level class predictions. Note that the comparison results are either fair re-evaluations from the author's provided best checkpoints or reported from the original paper.
    }
    \label{comparison_table}
    \vspace{+0.15cm}
    \resizebox{1.0\textwidth}{!}{
    \begin{tabular}{clcccccccccccc}
        \toprule
        & & \multicolumn{6}{c}{\textbf{CAMERA25}} & \multicolumn{6}{c}{\textbf{REAL275}} \\ 
        \cmidrule(r{0.1in}){3-8} \cmidrule(r{0.1in}){9-14}
        & {Method} & \textbf{IOU25} & \textbf{IOU50} & \textbf{5\textdegree \SI{5}{\cm}} & \textbf{5\textdegree \SI{10}{\cm}} & \textbf{10\textdegree \SI{5}{\cm}} & \textbf{10\textdegree \SI{10}{\cm}} & \textbf{IOU25} & \textbf{IOU50} & \textbf{5\textdegree \SI{5}{\cm}} & \textbf{5\textdegree \SI{10}{\cm}}& \textbf{10\textdegree \SI{5}{\cm}} & \textbf{10\textdegree \SI{10}{\cm}}\\
        \cmidrule(r{0.1in}){2-2}
        \cmidrule(r{0.1in}){3-8} \cmidrule(r{0.1in}){9-14}
        1 & {NOCS~\cite{wang2019normalized}}      &  91.1& 83.9 & 40.9  & 38.6 & 64.6 & 65.1 & \textbf{84.8} & 78.0 & 10.0 & 9.8 & 25.2 & 25.8\\
        2 & {Synthesis$^{*}$~\cite{chen2020category}} &  - & - & -  &- & - & - & - & - &  0.9 & 1.4 & 2.4 & 5.5 \\

        3 & {Metric Scale~\cite{lee2021category}}      &  \textbf{93.8}& 90.7 & 20.2  & 28.2 & 55.4 & 58.9 & 81.6 & 68.1 & 5.3 & 5.5 & 24.7 & 26.5 \\
        4 & {ShapePrior~\cite{tian2020shape}} &81.6	&72.4&	59.0&	59.6&  81.0 &  81.3 &	81.2& 77.3	&	21.4	&21.4&	54.1&	54.1\\
        5 & {CASS~\cite{chen2020learning}} & - & - & - & - & - & - & 84.2 & 77.7 &  23.5 & 23.8 & 58.0 & 58.3\\
        \midrule
        6 & {\textbf{CenterSnap~(Ours)}} & 93.2&	92.3&	63.0	& 69.5 &	79.5 & 87.9&	83.5 &	80.2 & 27.2 &	29.2 & 58.8 &	64.4 \\
    7 & {\textbf{CenterSnap-R~(Ours)}} & 93.2&	\textbf{92.5}&	\textbf{66.2}	& \textbf{71.7} &	\textbf{81.3} & \textbf{87.9}&	83.5 &	\textbf{80.2} & \textbf{29.1} &	\textbf{31.6} & \textbf{64.3} &	\textbf{70.9} \\
        \bottomrule
    \end{tabular}
    }
\end{table*}

\begin{table*}[t]
    \centering
    \renewcommand{\arraystretch}{1.3}
    \caption{
    \textbf{Quantitative comparison of 3D shape reconstruction on NOCS}~\cite{wang2019normalized}: Evaluated with \textbf{CD} metric ($10^{-2}).$ Lower is better.
    }
    
    \label{reconstruction_nocs}
    \resizebox{1.0\textwidth}{!}{
    \begin{tabular}{clcccccccccccccc}
        \toprule
        & & \multicolumn{7}{c}{\textbf{CAMERA25}} & \multicolumn{7}{c}{\textbf{REAL275}} \\ \cmidrule(r{0.1in}){3-9} \cmidrule(r{0.1in}){10-16}
        & {Method} & \textbf{Bottle} & \textbf{Bowl} & \textbf{Camera} & \textbf{Can} & \textbf{Laptop} & \textbf{Mug} & \textbf{Mean} & \textbf{Bottle} & \textbf{Bowl} & \textbf{Camera} & \textbf{Can} & \textbf{Laptop} & \textbf{Mug} & \textbf{Mean} \\
        \midrule
        1 & {Reconstruction~\cite{tian2020shape}} & 0.18 & 0.16  & 0.40 & 0.097 & 0.20 & 0.14 & 0.20& 0.34 &0.12 & 0.89 & 0.15 &0.29 &0.10 & 0.32 \\
        2 & {ShapePrior~\cite{tian2020shape}} &0.34 &0.22& 0.90& 0.22& 0.33& 0.21 &0.37 &0.50 &0.12& 0.99& 0.24& 0.71& 0.097& 0.44 \\
        \midrule
        3 & \textbf{CenterSnap~(Ours)} &0.11	&0.10&	0.29&	0.13&	0.07& 0.12	&	0.14 & 0.13 &  0.10 & 0.43 &0.09 & 0.07 & 0.06 & 0.15 \\
        \bottomrule
    \end{tabular}
    }
\end{table*}

\textbf{NOCS Baselines:}
We compare seven model variants to show effectiveness of our method:
(1) \textbf{NOCS}~\cite{wang2019normalized}: Extends Mask-RCNN architecture to predict~\gls{nocs} map and uses similarity transform with depth to predict pose and size. Our results are compared against the best pose-estimation configuration in NOCS~(i.e. 32-bin classification) (2) \textbf{Shape Prior}~\cite{tian2020shape}: Infers 2D bounding-box for each object and predicts a shape-deformation. (3) \textbf{CASS}~\cite{chen2020learning}: Employs a 2-stage approach to first detect 2D bounding-boxes and second regress the pose and size. (4) \textbf{Metric-Scale}~\cite{lee2021category}: Extends NOCS to predict object center and metric shape separately (5) \textbf{CenterSnap:} Our single-shot approach with direct pose and shape regression. (6) \textbf{CenterSnap-R:} Our model with a standard point-to-plane iterative pose refinement~\cite{segal2009generalized, Zhou2018} between the projected canonical pointclouds in the 3D space and the depth-map.  Note that we do not include comparisons to 6D pose tracking baselines such as~\cite{wang20206, wen2021bundletrack} which are not detection-based~(i.e. do not report mAP metrics) and require pose initialization. \\

\begin{figure}[!b]
\centering
\includegraphics[width=0.95\columnwidth]{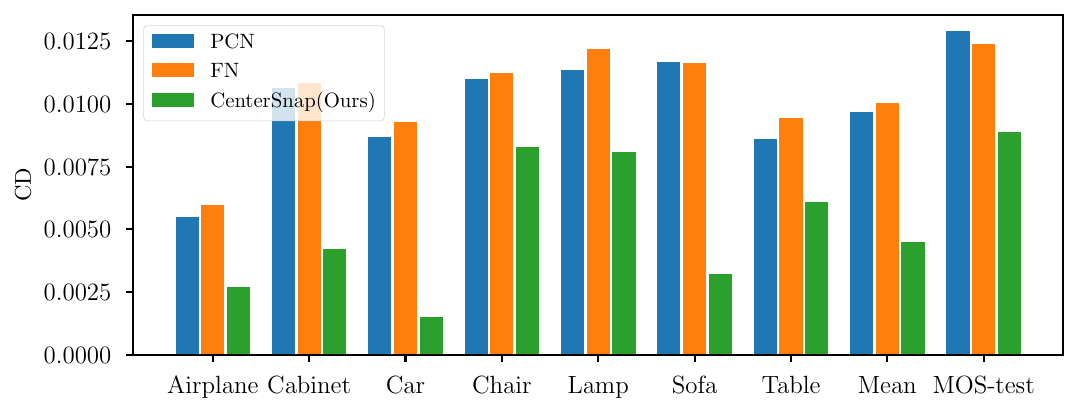}
\centering
  \caption{
  \textbf{Shape Completion:} Chamfer distance (CD reported on y-axis) evaluation on Multi-object ShapeNet dataset.
  }
  \label{shape_completion}
\end{figure}

\begin{figure}[!b]
\centering
\includegraphics[width=0.95\columnwidth]{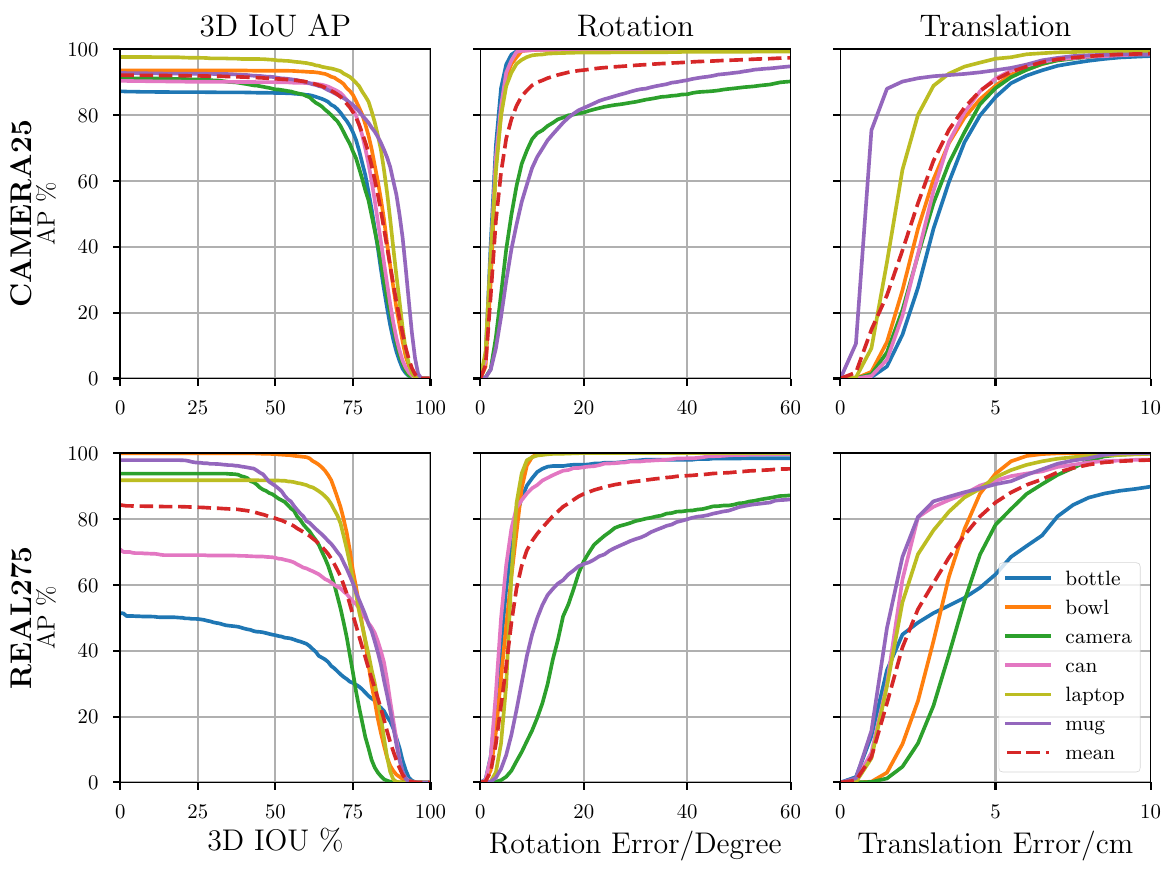}
\centering
  \caption{
  \textbf{\gls{map} on Real-275 test-set:} Our method's mean Average Precision on NOCS for various IOU and pose error thresholds.
  }
  \label{nocs_resut}
\end{figure}

\textbf{Comparison with NOCS baselines:} The results of our proposed CenterSnap method are reported in Table~\ref{comparison_table} and Figure~\ref{nocs_resut}. Our proposed approach consistently outperforms all the baseline methods on both 3D object detection and 6D pose estimation. Among our variants, CenterSnap-R achieves the best performance. Our method~(i.e. CenterSnap) is able to outperform strong baselines (\#1 - \#5 in Table~\ref{comparison_table}) even without iterative refinement. Specifically, CenterSnap-R method shows superior performance on the REAL test-set by achieving a mAP of 80.2\% for 3D IOU at 0.5, 31.6\% for 6D pose at 5\textdegree \SI{10}{\cm} and 70.9\% for 6D pose at 10\textdegree \SI{10}{\cm}, hence demonstrating an absolute improvement of 2.7\%, 10.8\% and 12.6\% over the best-performing baseline on the Real dataset.  Our method also achieves superior test-time performance on CAMERA evaluation never seen during training. We achieve a mAP of 92.5\% for 3D IOU at 0.5, 71.7\% for 6D pose at 5\textdegree \SI{10}{\cm} and 87.9\% for 6D pose at 10\textdegree \SI{10}{\cm}, demonstrating an absolute improvement of 1.8\%, 12.1\% and 6.6\% over the best-performing baseline. \\

\textbf{NOCS Reconstruction:} To quantitatively analyze the reconstruction accuracy, we measure the~\gls{cd} of our reconstructed pointclouds with ground-truth CAD model in NOCS. Our results are reported in Figure~\ref{reconstruction_nocs}. Our results show consistently lower CD metrics for all class categories, which shows superior reconstruction performance on novel object instances. We report a lower mean Chamfer distance of 0.14 on CAMERA25 and 0.15 on REAL275 compared to 0.20 and 0.32 reported by the competitive baseline~\cite{tian2020shape}. \\ 
\textbf{Comparison with Shape Completion Baselines}: We further test our network’s ability to reconstruct complete 3D
shapes by comparing against depth-based shape-completion
baselines, i.e. PCN~\cite{yuan2018pcn} and Folding-Net~\cite{yang2018foldingnet}. The results of
our CenterSnap method are reported in Figure 5. Our consistently lower Chamfer distance (CD) compared to strong
shape-completion baselines show our network’s ability to
reconstruct complete 3D shapes from partial 3D information
such as depth-maps. We report a lower mean CD of 0.089 on
test instances from categories not included during training vs
0.0129 for PCN and 0.0124 for Folding-Net respectively.
\begin{figure}[!t]
\centering
\includegraphics[width=0.95\columnwidth]{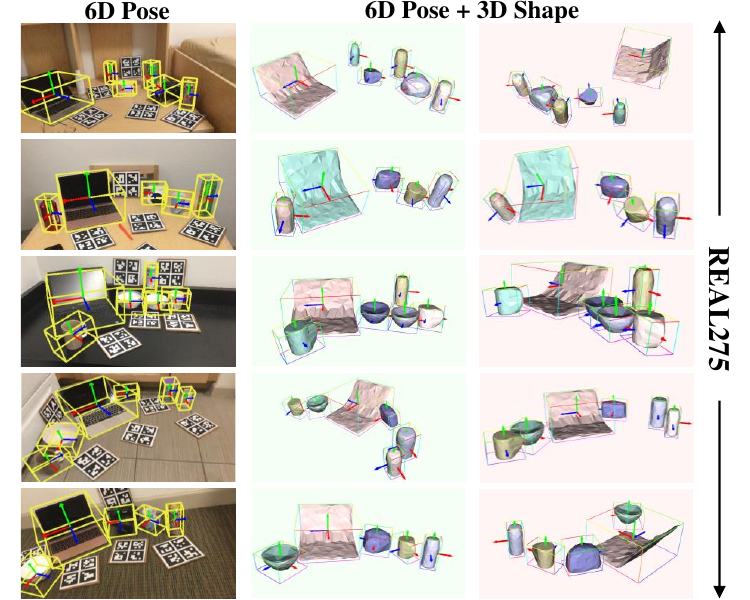}
\centering
  \caption{
  \textbf{Qualitative Results:} We visualize the real-world 3D shape prediction and 6D pose and size estimation of our method~(CenterSnap) from different viewpoints (green and red backgrounds).
  }
  \label{qualitative}
\end{figure} 

\textbf{Inference time:}
Given RGB-D images of size $640 \times 480$, our method performs fast~(real-time) joint reconstruction and pose and size estimation. We achieve an interactive rate of around 40~\gls{fps} for CenterSnap on a desktop with an Intel Xeon W-10855M@2.80GHz CPU and NVIDIA Quadro RTX 5000 GPU, which is fast enough for real-time applications. Specifically, our network takes 22 ms and reconstruction takes around 3 ms. In comparison, on the same machine, competitive multi-stage baselines~\cite{wang2019normalized,tian2020shape} achieve an interactive rate of 4 FPS for pose estimation. \\
\renewcommand{\arraystretch}{1.0}
\begin{table}[t]
\centering
\caption{\textbf{Ablation Study}: Study of the Proposed CenterSnap method on NOCS Real-test set to investigate the impact of different components i.e. Input, Shape, Training Regime~(TR) and Depth-Auxiliary loss~(D-Aux) on performance. C indicates Camera-train, R indicates Real-train and RF indicates Real-train with finetuning. 3D shape reconstruction evaluated with \textbf{CD} ($10^{-2}$). $*$ denotes the method does not evaluate size and scale and so has no IOU metric.}
\label{tab:ablation}
\resizebox{0.95\textwidth}{!}{%
\begin{tabular}{c@{\hskip 0.1in}lccccccccc} \\
\toprule
& & & & \multicolumn{6}{c}{\textbf{Metrics}} \\ 
\cmidrule(r{0.05in}){6-6}
\cmidrule(r{0.05in}){7-10}

& & & & & \textbf{3D Shape} & \multicolumn{4}{c}{\textbf{6D Pose}} \\ 
\cmidrule(r{0.05in}){6-6}
\cmidrule(r{0.05in}){7-10}
 {\textbf{\#}} & \text{Input} &\text{Shape} &\text{TR} & D-Aux&\textbf{CD}~$\downarrow$ & \textbf{IOU25}~$\uparrow$ & \textbf{IOU50}~$\uparrow$ &\textbf{5\textdegree \SI{10}{\cm}}~$\uparrow$ & \textbf{10\textdegree \SI{10}{\cm}}~$\uparrow$  \\

\midrule
 1 &RGB-D& \checkmark & C & \checkmark & 0.19 & 28.4 & 27.0 & 14.2 &48.2 \\
 2 &RGB-D& \checkmark & C+R & \checkmark & 0.19 & 41.5 & 40.1 & 27.1 & 58.2\\
 3 &RGB-D$^{*}$& &C+RF & \checkmark & --- & --- & --- & 13.8 & 50.2 \\
 4 & RGB & \checkmark & C+RF & \checkmark & 0.20 & 63.7 & 31.5 & 8.30 & 30.1\\
 5 &Depth& \checkmark & C+RF & \checkmark & 0.15 & 74.2 & 66.7 & 30.2 & 63.2  \\
 6 &RGB-D&\checkmark  & C+RF&  & 0.17 & 82.3 & 78.3 & 30.8 &68.3 \\
 7 &RGB-D& \checkmark & C+RF & \checkmark & 0.15 & 83.5 & 80.2 & 31.6 &70.9 \\
\bottomrule
\end{tabular}}
\end{table}
\textbf{Ablation Study:} An empirical study to validate the significance of different design choices and modalities in our proposed CenterSnap model was carried out. Our results are summarized in Table~\ref{tab:ablation}. We investigate the performance impact of \textbf{Input-modality}~(i.e. RGB, Depth or RGB-D), \textbf{Shape}, \textbf{Training-regime} and \textbf{Depth-Auxiliary loss} on the held out Real-275 set. Our ablations results show that our network with just mono-RGB sensor performs the worst~(31.5\% IOU50 and 30.1\% 6D pose at 10\textdegree \SI{10}{\cm}) likely because 2D-3D is an ill-posed problem and the task is 3D in nature. The networks with Depth-only~(66.7\% IOU50 and 63.2\% 6D pose at 10\textdegree \SI{10}{\cm}) and RGB-D~(80.2\% IOU50 and 70.9\% 6D pose at 10\textdegree \SI{10}{\cm}) perform much better. Our model without shape prediction under-performs the model with shape~(\#3 vs \#8 in Table~\ref{tab:ablation}), indicating shape understanding is needed to enable robust 6D pose estimation performance. The result without depth auxiliary loss~(0.17 CD, 78.3\% IOU50 and 68.3\% 6D pose at 10\textdegree \SI{10}{\cm}) indicates that adding a depth prediction task improved the performance of our model~(1.9\% absolute for IOU50 and 2.6\% absolute for 6D pose at 10\textdegree \SI{10}{\cm}) on real-world novel object instances. Our model trained on NOCS CAMERA-train with fine-tuning on Real-train~(80.2\% IOU50 and 70.9\% 6D pose at 10\textdegree \SI{10}{\cm}) outperforms all other training-regime ablations such as training only on CAMERA-train or combined CAMERA and REAL train-sets~(\#1 and- \#2 in Table~\ref{tab:ablation}) which indicates that sequential learning in this case leads to more robust sim2real transfer. \\
\textbf{Qualitative Results:} We qualitatively analyze the performance of CenterSnap on NOCS Real-275 test-set never seen during training. As shown in Figure~\ref{qualitative}, our method performs accurate 6D pose estimation and joint shape reconstruction on 5 different real-world scenes containing novel object instances. Our method also reconstructs complete 3D shapes~(visualized with two different camera viewpoints) with accurate aspect ratios and fine-grained geometric details such as mug-handle and can-head.
\section{Summary} 
\label{chap:centersnap,sec:conclusion}
Despite recent progress, existing categorical 6D pose and size estimation approaches suffer from high computational costs and low performance. In this chapter, we proposed an anchor-free and single-shot approach for holistic object-centric 3D scene understanding from a single-view RGB-D. Our approach runs in real-time~(40 FPS) and performs accurate categorical pose and size estimation, achieving significant improvements against strong baselines on the NOCS REAL275 benchmark on novel object instances. Specifically, we utilized large-scale priors from synthetic data and utilized them in an end-to-end learning-based pipeline to show improved generalization to real-world scenes with limited real-world finetuning. 
    \chapter{ShAPO: Implicit Representations for Multi-Object Shape, Appearance, and Pose Optimization}
\label{chap:shapo}

\looseness=-1
In this chapter, I explore the problem setting of~\textbf{3D shape and appearance reconstruction and estimating object poses} with the main aim of improving the accuracy of prior work~(\ref{chap:centersnap}) and while utilizing~\textit{largre-scale shape and appearance priors} in the form of synthetic-data for real-world generalization.

\section{Introduction}
\label{chap:shapo,sec:intro}

\begin{figure}[htb]
   \centering
    \includegraphics[width=1.0\linewidth]{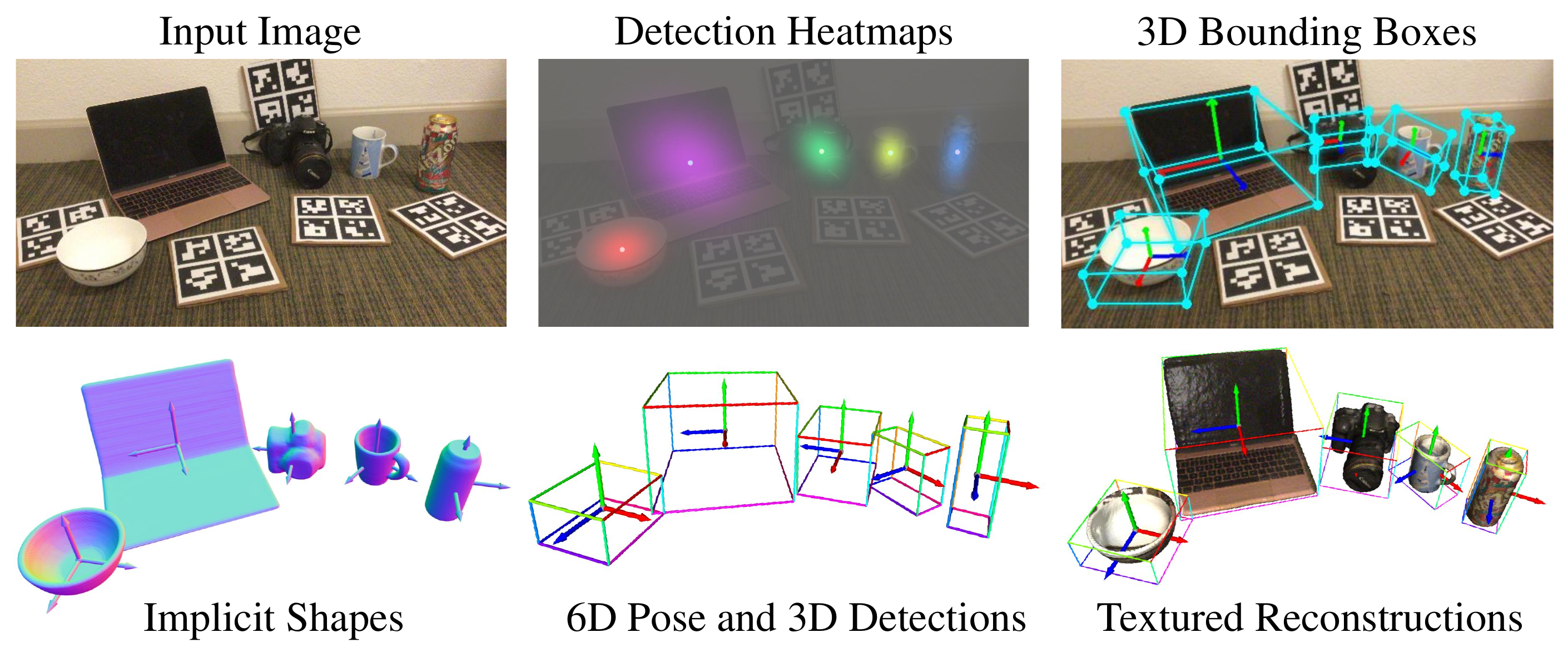}
      \caption{Given a single RGB-D observation, our method, \textbf{ShAPO}, infers 6D pose and size, 3D shapes and appearance of all objects in the scene. Results are shown on novel real-world scene in NOCS~\cite{wang2019normalized}.}
   \label{fig:intro_shapo}
\end{figure}

Holistic 3D object understanding~\cite{zhang2021holistic,Nie_2020_CVPR} from a single RGB-D observation has remained a popular yet challenging problem in computer vision~\cite{He2017,zhou2019objects,gkioxari2019mesh} and robotics~\cite{jiang2021synergies,ferrari1992planning,laskey2021simnet,irshad2022centersnap}. The capability to infer complete object-centric 3D scene information has benefits in robotic manipulation~\cite{cifuentes2016probabilistic,jiang2021synergies, laskey2021simnet, irshad2022centersnap}, navigation~\cite{irshad2022sasra,9561806} and augmented reality. This task demands that the autonomous robot reasons about the 3D geometry of objects from partially observed single-view visual data, infer a 3D shape and 6D pose and size~(i.e. 3D orientation and position), and estimate the appearance of novel object instances. Despite recent progress, this problem remains challenging since inferring 3D shapes from images is an ill-posed problem, and predicting the 6D pose and 3D scale can be extremely ambiguous without having any prior information about the objects of interest. 

Prior works on object-centric scene understanding have attempted to address this challenge in various ways: \textit{object pose understanding} methods obtain 3D bounding box information without shape details. Most prior works in object-pose estimation cast it as an instance-level~\cite{kehl2017ssd,rad2017bb8,xiang2018posecnn} 3D object understanding task, as opposed to category-level. Such methods~\cite{tekin2018real,peng2019pvnet,wang2019densefusion}, while achieving impressive results, rely on provided 3D reconstructions or prior CAD models for successful detection and pose estimation. Category-level approaches~\cite{tian2020shape,wang2019normalized,sundermeyer2020augmented}, on the other hand, rely on learned shape and scale priors during training, making them much more challenging. Despite great progress in category-level pose estimation, the performance of these approaches is limited due to their incapacity to express shape variations explicitly. 
\textit{Object-centric scene-level reconstruction} methods~\cite{wang2018pixel2mesh,niemeyer2020differentiable,kato2018} recover object shapes using 2D or partial 3D information for scene reconstruction. However, most methods are limited in their ability to reconstruct high-quality shapes in a fast manner~(i.e. the studied representation is either voxel-based which is computationally inefficient~\cite{groueix2018} or point-based which results in poor reconstruction quality~\cite{fan2017point,choy20163d}).

For holistic scene-level reconstruction, only predicting shapes in an isolated manner~\cite{groueix2018,kuo2020mask2cad,yuan2018pcn} may not yield good results due to the challenges of aligning objects in the 3D space, reasoning about occlusions and diverse backgrounds. To the best of our knowledge, fewer works have tackled the problem of joint shape reconstruction with appearance and object-centric scene context~(i.e. 3D bounding boxes and object pose and sizes) for a holistic object-centric scene understanding.

Motivated by the above, we present ShAPO, a learnable method unifying accurate shape prediction and alignment with object-centric scene context. As shown in Figure~\ref{fig:intro_shapo}, we infer the complete 3D information of novel object instances~(i.e. 3D shape along with appearance and 6D pose and sizes) from a single-view RGB-D observation; the results shown in Figure~\ref{fig:intro_shapo} are on a novel scene from the NOCS~\cite{wang2019normalized} dataset. In essence, our method represents object instances as center key-points~\cite{irshad2022centersnap,laskey2021simnet,duan2019centernet} in a spatial 2D grid. We regress the complete 3D information i.e. object shape and appearance codes along with the object masks and 6D pose and sizes at each object's spatial center point. A novel joint implicit shape and appearance database of signed distance and texture field priors is utilized, to embed object instances in a unique space and learn from a large collection of CAD models. We further utilize differentiable optimization of implicit shape and appearance representation to iteratively improve shape, pose, size, and appearance jointly in an analysis-by-synthesis fashion. To alleviate the sampling inefficiency inherent in signed distance field shape representations, we propose a novel octree-based point sampling, which leads to significant time and memory improvements as well as increased reconstruction quality. 

Our contributions are summarized as follows:
\begin{itemize}
    \item{\textbf{A generalizable, disentangled shape and appearance space} coupled with \textbf{an efficient octree-based differentiable optimization procedure} which allows us to identify and reconstruct novel object instances without access to their ground truth meshes.}
    \item \textbf{An object-centric scene understanding pipeline} relying on learned joint appearance and implicit differentiable shape priors, which achieves state-of-the-art reconstruction and pose estimation results on benchmark datasets.
    \item Our proposed approach significantly outperforms all baselines for 6D pose and size estimation on NOCS benchmark, showing over 8\% absolute improvement in mAP for 6D pose at 10$^{\circ}$ \SI{10}{\cm}.
\end{itemize}

\section{Related Works}
\label{chap:shapo,sec:related}

In essence, our proposed method infers 3D shape along with predicting the 3D appearance and 6D pose and sizes of multiple object instances to perform object-centric scene reconstruction from a single-view RGB-D, so it relates to multiple areas in 3D scene reconstruction, object understanding, and pose estimation.

\noindent\textbf{Neural Implicit Representations:}
3D shape and appearance reconstructions have recently seen a new prominent direction to use neural nets as scalar field approximators instead of ConvNets. The first works of this class are notably DeepSDF~\cite{park2019deepsdf}, Occ-Net~\cite{mescheder2019occupancy}, and IM-Net~\cite{chen2019learning}. These works use \textit{coordinate-based representation} to output either an occupancy estimate or a continuous SDF value, encoding the object's surface given a 3D coordinate. Improving this direction further, MeshSDF~\cite{remelli2020meshsdf}, NGLOD~\cite{takikawa2021neural} and Texture fields~\cite{OechsleICCV2019} employed end-to-end differentiable mesh representation, efficient octree representation and implicitly representing high-frequency textures respectively. In our pipeline, we build a differentiable database of shape and appearance priors based on the latest advances in neural implicit representations. Our pipeline is end-to-end differentiable and abstains from the expensive Marching Cubes computation at every optimization step. Our database stores not only the geometries of the objects in the form of signed distance fields but also their appearance in the form of texture fields, allowing us to model multiple categories through a single network while also considering test-time optimization of shape, pose, and appearance.

\noindent\textbf{Object-centric Reconstruction:} 3D object reconstruction from a single-view observation has seen great progress to output point clouds, voxels or meshes~\cite{park2019deepsdf,mescheder2019occupancy,chen2019learning}. Similarly, scene representation has been extended to include appearance. SRN~\cite{sitzmann2019scene}, DVR~\cite{niemeyer2020differentiable} learn from multi-view images by employing ray-marching routine and differentiable rendering respectively. Recently,~\gls{nerf}~\cite{mildenhall2020nerf} proposed to regress density and color along a ray and perform volumetric rendering to obtain true color value. Most NeRF-based methods~\cite{li2020neural, Niemeyer2020GIRAFFE, Ost_2021_CVPR} overfit to a single scene, do not promise generalizability, and require dense viewpoint annotations. Our pipeline, on the other hand, is capable of reconstructing shapes, and appearances and inferring the 6D pose and sizes of objects never seen during training from a single-view RGB-D and is also not limited to viewpoints seen during training.

\noindent\textbf{6DoF Object Pose and size estimation} works use direct pose regression~\cite{irshad2022centersnap, kehl2017ssd, wang2019densefusion, xiang2018posecnn}, template matching~\cite{kehl2016deep, sundermeyer2018implicit, tejani2014latent} and establishing correspondences~\cite{wang2019normalized, dpodv2, park2019pix2pose,hodan2020epos, goodwin2022}. However, most works focus only on pose estimation and do not deal with shape and appearance retrieval and their connection to 6D pose. We instead propose a differentiable pipeline to improve the initially regressed pose, along with the shape and appearance, using our novel octree-based differentiable test-time optimization.   

\begin{figure}[htb]
   \centering
       \includegraphics[width=\linewidth]{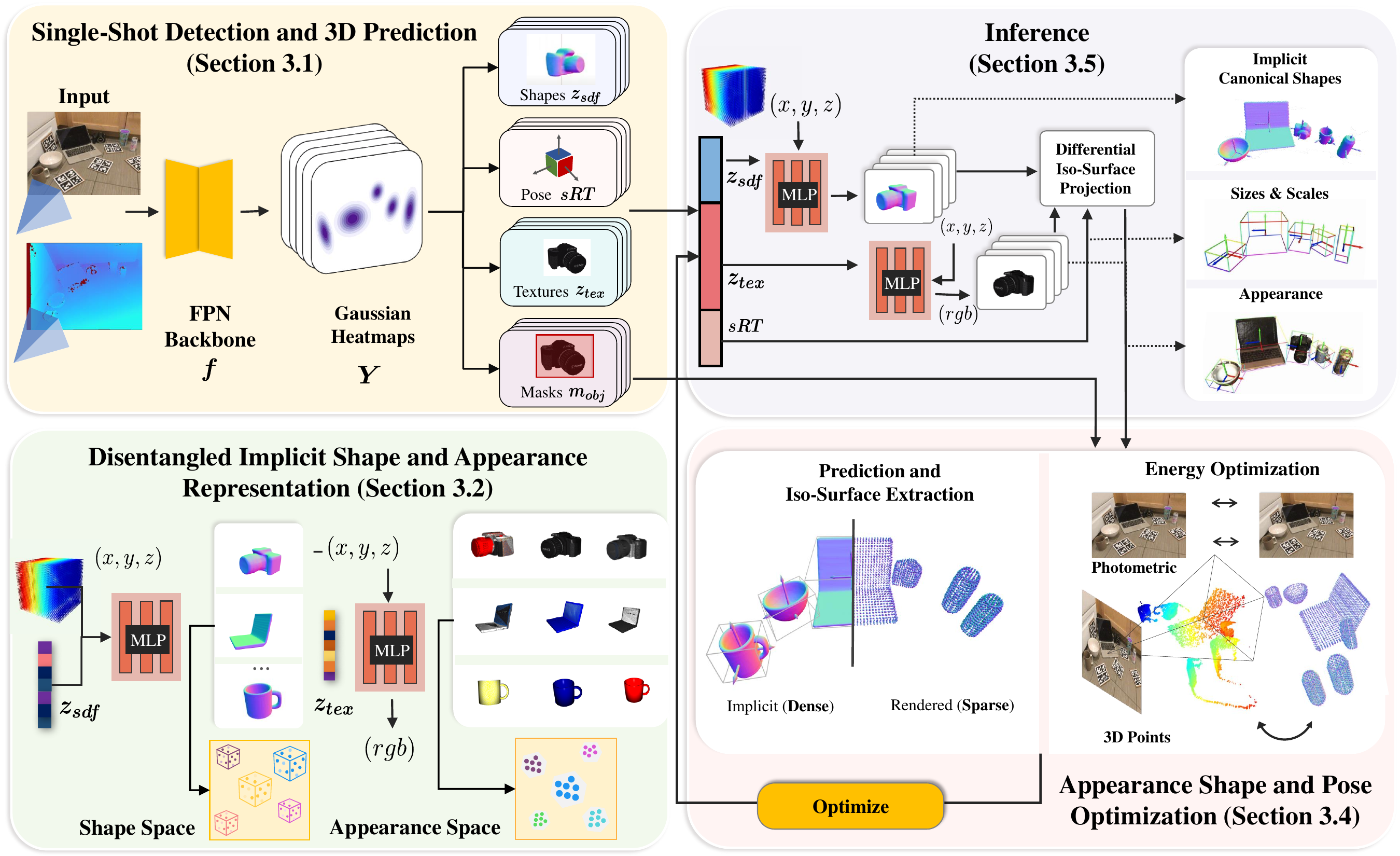}
      \caption{\textbf{ShAPO Method:} Given a single-view RGB-D observation, our method jointly predicts shape, pose, size, and appearance codes along with the masks of each object instance in a single-shot manner. Leveraging a novel octree-based differentiable optimization procedure, we further optimize for shape, pose, size, and appearance jointly in an analysis-by-synthesis manner. 
    }
   \label{fig:method}
\end{figure}
\section{Method}
\label{chap:shapo,sec:method}

ShAPO is a learning-based holistic object-centric scene understanding method that infers 3D shape along with the 6D pose, size and appearance of multiple unknown objects in an RGB-D observation. ShAPO tackles the detection, localization and reconstruction of all unknown object instances in 3D space. Such a goal is made possible by three components: 1) A single-shot detection and 3D prediction module that detects multiple objects based on their center points in the 2D spatial grid and recovers their complete 3D shapes, 6D pose, and sizes along with appearance from partial observations. 2) An implicit joint differentiable database of shape and appearance priors which is used to embed objects in a unique space and represent shapes as signed distance fields~(SDF) and appearance as continuous texture fields~(TF). 3) A 2D/3D refinement method utilizing an octree-based coarse-to-fine differentiable optimization to improve shape, pose, size, and appearance predictions iteratively.

Formally, given a single-view RGB-D observation~($I$ $\in$ $\mathds{R}^{h_{o}\times w_{o}\times 3 }$, $D$ $\in$ $\mathds{R}^{h_{o}\times w_{o}}$) as input, ShAPO infers the complete 3D information of multiple objects including the shape~(represented as a zero-level set of an implicit signed distance field, $\boldsymbol{SDF}$), 6D pose and scales~($\tilde{\mathbf{\mathcal P}}$ $\in$ $SE(3)$, $\hat{s}$ $\in$ $\mathbb{R}^3$) and appearance (represented as a continuous texture field, $\boldsymbol{TF}$). To achieve the above goal, we employ a two-stage approach comprising a single-shot inference
to extract initial pose, shape, and appearance estimates, and subsequently optimize the latent codes and poses keeping network weights fixed. As shown in Figure~\ref{fig:method}, we first formulate object detection as a spatial per-pixel point detection~\cite{irshad2022centersnap,laskey2021simnet,zhou2019objects}. A backbone comprising feature pyramid networks~\cite{girshick14CVPR} is employed with a specialized head to predict object instances as heatmaps in a per-pixel manner. Second, joint shape, pose, size and appearance codes along with instance masks are predicted for detected object centers using specialized heads~(Section~\ref{method:section1}). Our combined shape and appearance implicit differentiable database of priors is described in Section~\ref{method:section2} and the corresponding zero iso-surface based
differentiable rendering is detailed in Section~\ref{method:section3}. Lastly, 3D shapes along with their appearance, coupled with 6D pose and sizes of novel objects are jointly optimized during inference to predict accurate shapes, poses and sizes of novel object instances~(Section~\ref{method:section4}).

\subsection{Single-Shot Detection and 3D prediction}\label{method:section1}

ShAPO represents object instances along with their complete 3D information including shape, pose, appearance and size codes along with corresponding 2D masks through their 2D location in the spatial RGB image, following~\cite{irshad2022centersnap,laskey2021simnet,zhou2019objects,duan2019centernet}. Given an RGB-D observation~($I$ $\in$ $\mathds{R}^{h_{o}\times w_{o}\times 3 }$, $D$ $\in$ $\mathds{R}^{h_{o}\times w_{o}}$), ShAPO predicts object-centric heat maps~$\hat{Y}\in [0,1]^{\frac{h_{o}}{R} \times \frac{w_{o}}{R}\times 1}$ where each detected point~$(\hat{x}_{i}, \hat{y}_{i})$ denotes the local maxima in the heatmap~($\hat{Y}$).
Here $R$ denotes the heatmap down-sampling factor and is set to $8$ in all our experiments. To predict these heatmaps, a feature backbone based on~\gls{fpn}~\cite{kirillov2019panoptic} is utilized along with a specialized heatmap prediction head.
During training, we find the target heatmaps by splatting the ground truth center points~$(x_{i}, y_{i})$ using the Gaussian kernel~$\mathcal{N}(x_{i}, y_{i}, \sigma_{i})$ where $\sigma_{i}$ is relative to the spatial extent of each object (as defined by the corresponding ground truth bounding box annotation).
The network is trained to predict ground-truth heatmaps~($Y$) by minimizing MSE loss over all pixels $(x,y)$ in the heatmap, $\mathcal{L}_{\text{inst}}=\sum_{xyg}\left(\hat{Y}- Y\right)^{2}$. The network also predicts object instance masks~($\hat{M}$) using a specialized head~($f_{\theta_{m}}$) to output $\hat{M} \in \mathds{R}^{h_{o}\times w_{o}}$, similar to the semantic segmentation head described in~\cite{kirillov2019panoptic}. Note that the network must predict masks for accurate downstream optimization~(see Section~\ref{method:section3}).

\subsection{Joint Implicit Shape, Pose and Appearance Prediction}\label{method:section2}
Once the network detects objects, it then predicts their complete 3D information~(i.e. 3D shape, 6D pose, and size along with the 3D appearance) all in a single-forward pass using specialized heads~($f_{\theta_{sdf}}$, $f_{\theta_{P}}$ and $f_{\theta_{tex}}$) with outputs~($Y_{sdf} \in \mathds{R}^{\frac{h_{o}}{R}\times \frac{w_{o}}{R}\times 64}$, $Y_{P}\in \mathds{R}^{\frac{h_{o}}{R}\times \frac{w_{o}}{R}\times 13}$ and $Y_{tex}\in \mathds{R}^{\frac{h_{o}}{R}\times \frac{w_{o}}{R}\times 64}$ respectively).
During training, the task-specific heads output shape code~$z_{sdf}$, 6D pose~$\tilde{\mathbf{\mathcal P}}$, scale~$\hat{s}$ and appearance~$z_{tex}$ information at each pixel in the down-sampled map~($\frac{h_{o}}{R}\times \frac{w_{o}}{R}$).
For each object's Pose~($\tilde{\mathbf{\mathcal P}}$) with respect to the camera coordinate frame, we regress a 3D rotation $\hat{\mathcal{R}}$ $\in$ $SO(3)$, a 3D translation $\hat{t}$ $\in$ $\mathbb{R}^3$ and 1D scales~$\hat{s}$ (totaling thirteen numbers). These parameters are used to transform the object shape from a canonical frame to the 3D world. We select a 9D rotation~$\hat{\mathcal{R}}$ $\in$ $SO(3)$ representation since the neural network can better fit a continuous representation and to avoid discontinuities with lower rotation dimensions, as noted in~\cite{zhou2019continuity}. Furthermore, we employ a rotation mapping function following~\cite{pitteri2019object} to handle ambiguities caused by rotational symmetries. The rotation mapping function is used only for symmetric objects~\textit{(bottle, bowl, and can)} in our database during training, and it maps ambiguous ground-truth rotations to a single canonical rotation by normalizing the pose rotation.
Note that during training, ground-truth shape codes~$z_{sdf}$ and appearance codes~$z_{tex}$ for each object are obtained from our novel implicit textured differentiable representation~(further described in Section~\ref{shapeappearanceemb}).

During training, we jointly optimize for shape, pose, appearance, and mask prediction. Specifically, we minimize the masked $L_{1}$ loss for shape, pose and appearance prediction, denoted as $\mathcal{L}_{\text{sdf}},  \mathcal{L}_{\text{tex}},  \mathcal{L}_{\text{P}}$ and a pixel-wise cross-entropy loss for mask prediction $ \mathcal{L}_{\text {M}}=\sum_{i=1}^{h_{o} \cdot w_{o}}-\log \hat{M}_{i}\left(M_{i}^{\mathrm{gt}}\right)$ where $M_{i}^{\mathrm{gt}}$ denotes the ground truth category label for each pixel.

During training, we minimize a combination of these losses as follows:
\begin{equation}
\mathcal{L}= \lambda_{inst}\mathcal{L}_{inst} + \lambda_{sdf}\mathcal{L}_{sdf} +  \lambda_{tex}\mathcal{L}_{tex} + \lambda_{M}\mathcal{L}_{M} + \lambda_{P}\mathcal{L}_{P}
\end{equation}
where $\lambda$ is a weighting coefficient with values determined empirically as $\lambda_{inst} = 100$ and $\lambda_{sdf} =  \lambda_{tex} =  \lambda_{P}$ = 1.0.
Note that for shape, appearance, and pose predictions, we enforce the $L_{1}$ loss based on the probability estimates of the Gaussian heatmaps ($Y$) i.e. the loss is only applied where Y has a score greater than 0.3
to prevent ambiguity in the space where no object exists. We now describe the shape and appearance representation utilized by our method. 

\begin{figure}[t]
   \centering
       \includegraphics[width=\linewidth]{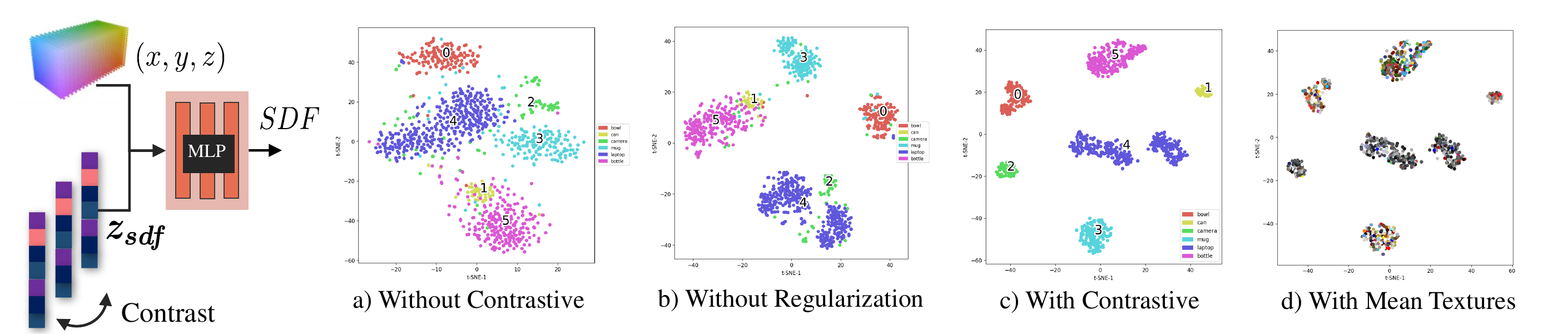}
      \caption{Deep Implicit shape space for NOCS object Categories}
   \label{fig:shape_space}
\end{figure}

\begin{figure}[b]
   \centering
       \includegraphics[width=\linewidth]{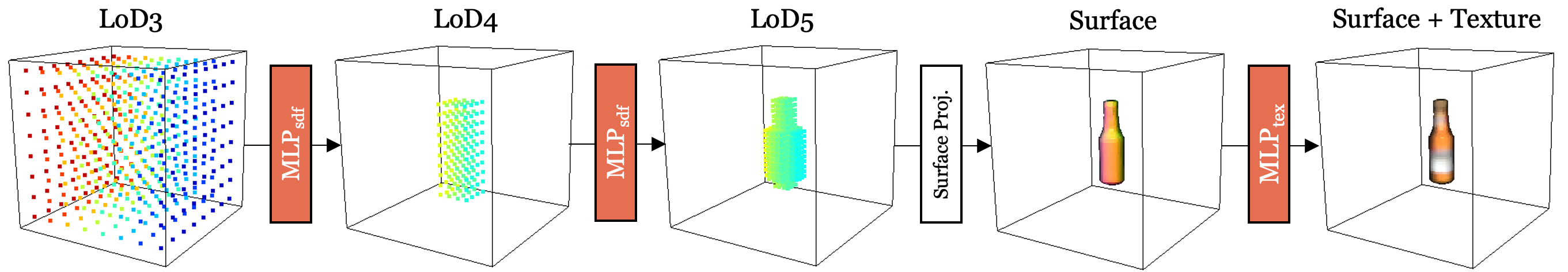}
      \caption{Octree-based object extraction.}
   \label{fig:octree}
\end{figure}

\subsection{Implicit textured differentiable database of priors}\label{shapeappearanceemb}
We propose a novel joint implicit textured representation to learn from a large variety of CAD models and embed objects in a concise space. This novel representation~(As shown in Figure~\ref{fig:method} and Figure~\ref{fig:shape_space}) is also used as a strong inductive prior to efficiently optimize the shape and appearance along with the pose and size of objects in a differentiable manner~(Section~\ref{method:section3}). In our implicit textured database of shape and appearance priors, each object shape is represented as~\gls{sdf} where a neural network learns a signed distance function $G(\boldsymbol{x},\boldsymbol{z}_{sdf})=s: z_{sdf} \in \mathbb{R}^{64}, s\in \mathbb{R}$ for every 3D point $x \in \mathbb{R}^{3}$ and the appearance is represented as~\gls{tf}~($t_{\theta}: \mathbb{R}^{3} \rightarrow \mathbb{R}^{3}$) which maps a  3D point $x \in \mathbb{R}^{3}$ to an RGB value $c \in \mathbb{R}^{3}$. Since the mapping between coordinates and colors is ambiguous without shape information, we propose to learn a texture field only at the predicted shape i.e. $t_{\theta}(\boldsymbol{x},\boldsymbol{z}_{sdf},\boldsymbol{z}_{tex})=c,\boldsymbol{z}_{tex} \in \mathbb{R}^{64}$. The SDF function~($G$) implicitly defines the surface of each object shape by the zero-level set $G(.) = 0$. To learn a shape-code~($\boldsymbol{z}_{sdf}$) and texture code~($\boldsymbol{z}_{tex}$) for each object instance, we design a single~\gls{mlp} each for shape (to reason about the different geometries in the database) and texture~(to predict color information given shape and texture codes). Through conditioning the MLP output on the latent vector, we allow modeling multiple geometries and appearances using a single network each for shape and appearance. Each MLP is trained separately using the supervised ground-truth reconstruction loss $L_{SDF}$ and $L_{RGB}$ as follows:
\begin{gather}
L_{SDF}=\left|\operatorname{clamp}\left(G(\boldsymbol{x},\boldsymbol{z}_{sdf}), \delta\right)-\operatorname{clamp}\left(\boldsymbol{s}_{gt}, \delta\right)\right| + L_{contrastive}(\boldsymbol{z}_{sdf}) \\
L_{RGB} = \sum_{n=1}^{N} \|\boldsymbol{c}_{gt}-t_{\theta}(\boldsymbol{x},\boldsymbol{z}_{sdf},  \boldsymbol{z}_{tex})\|_{2}^{2}
\end{gather}  

where $L_{SDF}$ is a combination of a clipped $L_{1}$ loss between ground-truth signed-distance values $\boldsymbol{s}_{gt}$ and predicted SDF $G(\boldsymbol{x},\boldsymbol{z}_{sdf})$ and a contrastive loss~$L_{\text {contrastive }}=\left[m_{\text {pos  }}-s_{p}\right]_{+}+\left[s_{n}-m_{\text {neg }}\right]_{+}$. As shown by the t-SNE embeddings~\cite{van2008visualizing} (Figure~\ref{fig:shape_space}) for the latent shape-code~($\boldsymbol{z}_{sdf}$), the contrastive loss helps with disentangling the shape space nicely and leads to better downstream regression in the single-shot model (Section~\ref{method:section2}). Once we train the implicit shape auto-decoder, we use the learned shape space $\boldsymbol{z}_{sdf}$ to minimize the color loss, $L_{RGB}$ which is defined as an MSE loss between predicted color at the surface~$t_{\theta}(\boldsymbol{x},\boldsymbol{z}_{sdf},\boldsymbol{z}_{tex})$ and ground-truth color~$\boldsymbol{c}_{gt}$.

We use 3D textured models from the CAD model repository Shapenet~\cite{chang2015ShapeNet} to learn our database of shape and texture priors. Once trained, the MLP networks for both shape and appearance find a disentangled space for color and geometry while keeping semantically similar objects together~(Figure~\ref{fig:shape_space}) and provide us with strong priors to be used for 2D and 3D optimization~(described in Section~\ref{method:section3}).

\looseness=-1
\begin{algorithm}[t!]
	\KwIn{ 
			$\textbf{x} \in \mathbb{R}^3$ grid points,
			$\textbf{l} \in \mathbb{L}$ grid levels,
			$\textbf{z}_{sdf}$ and $\textbf{z}_{tex} \in \mathbb{R}^{64}$ latent vectors 
		}
		\KwOut{ 
			$\textbf{pcd} \in \mathbb{R}^3$ surface points,
			$\textbf{nrm} \in \mathbb{N}^3$ normals,
			$\textbf{col} \in \mathbb{C}^3$ colors
		}
		
		\tcc{Extract object grid (no grad)}
		\For{$ l \in \{1, \ldots, N_{LoD}\} $}{
            $\textbf{sdf} \leftarrow G(\textbf{x}_l, \textbf{z}_{sdf})$ \tcp*{regress $sdf$ values}
            $\textbf{occ} \leftarrow \textbf{sdf} < getCellSize(l)$ \tcp*{estimate cell occupancy}
			$\textbf{x}_{l_{occ}} \leftarrow \textbf{x}_l[occ]$ \tcp*{remove unoccupied cells}
			$\textbf{x}_{l+1} \leftarrow subdivide(\textbf{x}_{l_{occ}})$ \tcp*{subdivide cells to go to next LoD}
			
		}
		\tcc{Extract object shape and appearance}
		$\textbf{nrm} \leftarrow \text{backprop}(\textbf{sdf})$ \tcp*{analytically estimate surface normals}
		$\textbf{pcd} \leftarrow \textbf{x} - \textbf{nrm} * \textbf{sdf}$ \tcp*{project points onto the surface}
		$\textbf{col} \leftarrow t_{\theta}(\textbf{pcd}, \textbf{z}_{sdf}, \textbf{z}_{tex})$ \tcp*{regress surface texture}
		\KwRet{$\textbf{pcd}$, $\textbf{nrm}$, $\textbf{col}$}
		
		\caption{Octree-based implicit surface extraction}
		\label{alg:octree}
	\end{algorithm}
\subsection{Differentiable Optimization}\label{method:section3}
A key component of ShAPO is the optimization scheme which allows refining the initial object predictions with respect to the pose, scale, shape, and appearance. Inspired by sdflabel~\cite{zakharov2020autolabeling}, we develop a new differentiable and fast optimization method. Instead of using mesh-based representations, we rely entirely on implicit surfaces, which not only helps us avoid common connectivity and intersection problems but also provides us full control over sampling density. 

\looseness=-1
\textbf{Surface Projection}
Given input grid points $x_i$ and estimated SDF values $s_i$, we aim to find a differentiable transformation to extract the object surface encoded in $\mathbf{z}_{sdf}$. A trivial solution would be to simply threshold points with SDF values more than a certain threshold. However, this procedure is not differentiable with respect to the input latent vector $\mathbf{z}_{sdf}$. Instead, we utilize the fact that deriving an SDF value $s_i$ with respect to its input coordinate $x_i$ yields a normal at this point, which can be computed in a single backward pass:

\begin{equation}
    n_i = \frac{\partial G(x_i;\mathbf{z}_{sdf})}{\partial x_i}.
\quad\text{and}\quad
    p_i = x_i - \frac{\partial G(x_i;\mathbf{z}_{sdf})}{\partial x_i} G(x_i;\mathbf{z}_{sdf}).
\end{equation}

\textbf{Octree-based Point Sampling}
The brute force solution to recover shapes from a learned SDF representation can be obtained by estimating SDF values for a large collection of grid points similar to the procedure used in \cite{zakharov2020autolabeling}. To obtain clean surface projections, one would then disregard all points $x_i$ outside a narrowband ($|s_i| > 0.03$) of the surface. However, this procedure is extremely inefficient both memory- and compute-wise --- For a grid size of $60^3 = 216000$ points, only around 1600 surface points are extracted (accounting to $0.7\%$ of the total number of points). We propose an Octree-based procedure to efficiently extract points.
We define a coarse voxel grid and estimate SDF values for each of the points using our trained SDF network. We then disregard voxels whose SDF values are larger than the voxel grid size for this resolution level. The remaining voxels are subdivided, each generating eight new voxels. We repeat this procedure until the desired resolution level is reached. In our implementation, we start from~\gls{lod} 3 and traverse up to LoD 6 depending on the desired resolution level. Finally, when points are extracted, we estimate their SDF values and normals and project them onto the object surface. The pseudocode implementation of the Octree-based surface extraction is provided in Alg.~\ref{alg:octree} with the visualization shown in Fig.~\ref{fig:octree}.

\looseness=-1
\subsection{Inference:}\label{method:section4}
During inference, we first perform predictions using our single-shot model. Object detection is performed using peak detection~\cite{zhou2019objects} on the outputs of predicted heatmaps~($\hat{Y}$). Each detected center point~{($\boldsymbol{x}_{i}, {\boldsymbol{y}_{i}}$)} corresponds to maxima in the heatmap output~($\hat{Y}$). Second, we sample shape, pose and appearance codes of each object from the output of task-specific heads at the detected center location~($x_{i}, {y_{i}}$) via~$\boldsymbol{z}_{sdf} = Y_{sdf}(\boldsymbol{x}_{i}, {\boldsymbol{y}_{i}})$, ~$\boldsymbol{z}_{tex} = Y_{tex}(\boldsymbol{x}_{i}, {\boldsymbol{y}_{i}})$ and ~$\tilde{\mathbf{\mathcal P}} = Y_{P}(\boldsymbol{x}_{i}, {\boldsymbol{y}_{i}})$.
We utilize the predicted shape, pose, size and appearance codes as an initial estimate to further refine through our differentiable optimization pipeline. Our optimizer takes as input the predicted implicit shapes in the canonical frame of reference along with the masks predictions~($\hat{M}$),  color codes~($\boldsymbol{z}_{tex}$) and extracted~$3\times 3$ rotation $\hat{\mathcal{R}}_{i}^{p}$, 3D translation vector $\hat{t}_{i}^{p}$ and 1D scales $\hat{s}_{i}^{p}$ from recovered pose $\tilde{\mathbf{\mathcal P}}$. Although a variety of related works consider mean class predictions as initial priors, we mainly utilize the regressed outputs of shape, appearance, and pose for the optimization pipeline since the initial estimates are very robust~(see Table~\ref{comparison_table}). We utilize the predicted SDF to recover the complete surface of each object, in a coarse-to-fine manner, using the proposed differentiable zero-isosurface projection~(Section~\ref{method:section3}). After fixing the decoder~(G) parameters, we optimize the feature vector $\boldsymbol{z}_{sdf}$ by estimating the nearest neighbor between the predicted projected point clouds and masked point clouds obtained from the depth map and predicted masks~($\hat{M}$) of each object. In essence, a shape code $\boldsymbol{z}_{sdf}$ is refined using the Maximum-a-Posterior (MAP) estimation as follows:
\begin{equation}
\boldsymbol{z}_{sdf}=\underset{\boldsymbol{z}}{\arg \min } (\mathcal{L}\left(D(G(\boldsymbol{z}, \boldsymbol{x}\right)), P_{d})
\end{equation}
where D() denotes the differentiable iso-surface projection described in Section~\ref{method:section3} and $P_{d}$ denotes the point clouds obtained from masked depth maps. We further optimize the RGB component similarly by optimizing the difference in colors between the masked image color values~($C_{d}$) and colors obtained using the regressed color codes decoded by the texture field~($t_{\theta}$)  $\boldsymbol{z}_{tex}=\underset{\boldsymbol{z}}{\arg \min } (\mathcal{L}\left(D(t_{\theta}(\boldsymbol{z}, \boldsymbol{x}\right)), C_{d})$. We further allow $t_{\theta}$ weights to change to allow for a finer level of reconstruction.

\setlength{\tabcolsep}{1pt}
\begin{table}[t]
    \small
    \centering
    \renewcommand{\arraystretch}{1.3}
    \caption{
    \textbf{Quantitative comparison of 6D pose estimation and 3D object detection on NOCS}~\cite{wang2019normalized}: Comparison with strong baselines. Best results are highlighted in \textbf{bold}. $*$ denotes the method does not report IOU metrics since size and scale is not evaluated. We report metrics using nocs-level class predictions for a fair comparison with all baselines.
    }
    \label{comparison_table_shapo}
    \resizebox{1.0\textwidth}{!}{
    \begin{tabular}{clcccccccccccc}
        \toprule
        & & \multicolumn{6}{c}{\textbf{CAMERA25}} & \multicolumn{6}{c}{\textbf{REAL275}} \\ 
        \cmidrule(r{0.1in}){3-8} \cmidrule(r{0.1in}){9-14}
        & {Method} & \textbf{IOU25} & \textbf{IOU50} & \textbf{5\textdegree \SI{5}{\cm}} & \textbf{5\textdegree \SI{10}{\cm}} & \textbf{10\textdegree \SI{5}{\cm}} & \textbf{10\textdegree \SI{10}{\cm}} & \textbf{IOU25} & \textbf{IOU50} & \textbf{5\textdegree \SI{5}{\cm}} & \textbf{5\textdegree \SI{10}{\cm}}& \textbf{10\textdegree \SI{5}{\cm}} & \textbf{10\textdegree \SI{10}{\cm}}\\
        \cmidrule(r{0.1in}){2-2}
        \cmidrule(r{0.1in}){3-8} \cmidrule(r{0.1in}){9-14}
        1 & {NOCS~\cite{wang2019normalized}}      &  91.1& 83.9 & 40.9  & 38.6 & 64.6 & 65.1 & 84.8 & 78.0 & 10.0 & 9.8 & 25.2 & 25.8\\
        2 & {Synthesis$^{*}$~\cite{chen2020category}} &  - & - & -  &- & - & - & - & - &  0.9 & 1.4 & 2.4 & 5.5 \\

        3 & {Metric Scale~\cite{lee2021category}}      & 93.8& 90.7 & 20.2  & 28.2 & 55.4 & 58.9 & 81.6 & 68.1 & 5.3 & 5.5 & 24.7 & 26.5 \\
        4 & {ShapePrior~\cite{tian2020shape}} &81.6	&72.4&	59.0&	59.6&  81.0 &  81.3 &	81.2& 77.3	&	21.4	&21.4&	54.1&	54.1\\
        5 & {CASS~\cite{chen2020learning}} & - & - & - & - & - & - & 84.2 & 77.7 &  23.5 & 23.8 & 58.0 & 58.3\\
        6 & {CenterSnap~\cite{irshad2022centersnap}} & 93.2&	92.3&	63.0	& 69.5 &	79.5 & 87.9&	83.5 &	80.2 & 27.2 &	29.2 & 58.8 &	64.4 \\
        7 & {CenterSnap-R~\cite{irshad2022centersnap}} & 93.2&	92.5&	66.2	& 71.7 &81.3 & 87.9&	83.5 &	\textbf{80.2} & 29.1 &	31.6 & 64.3 &	70.9 \\
       \midrule
        8 & {\textbf{ShAPO (Ours)}} & \textbf{94.5}&	\textbf{93.5}&	\textbf{66.6}	& \textbf{75.9} &	\textbf{81.9} & \textbf{89.2}&	\textbf{85.3} &	79.0 & \textbf{48.8} &	\textbf{57.0} & \textbf{66.8} &	\textbf{78.0} \\
      
        \bottomrule
    \end{tabular}
    }
\end{table}

\section{Experiments}
\label{chap:shapo,sec:exp}

In this section, we empirically validate the performance of our method. In essence, our goal is to answer these questions:
1) How well does ShAPO recover the pose and sizes of novel objects?
2) How well does ShAPO perform in terms of reconstructing the geometry and appearance of multiple objects from a single-view RGB-D observation? 3) How well does our differentiable iterative improvement and multi-level optimization impact shape, appearance, pose, and size?\

\begin{figure}[htb]
  \centering
    \includegraphics[width=0.95\textwidth]{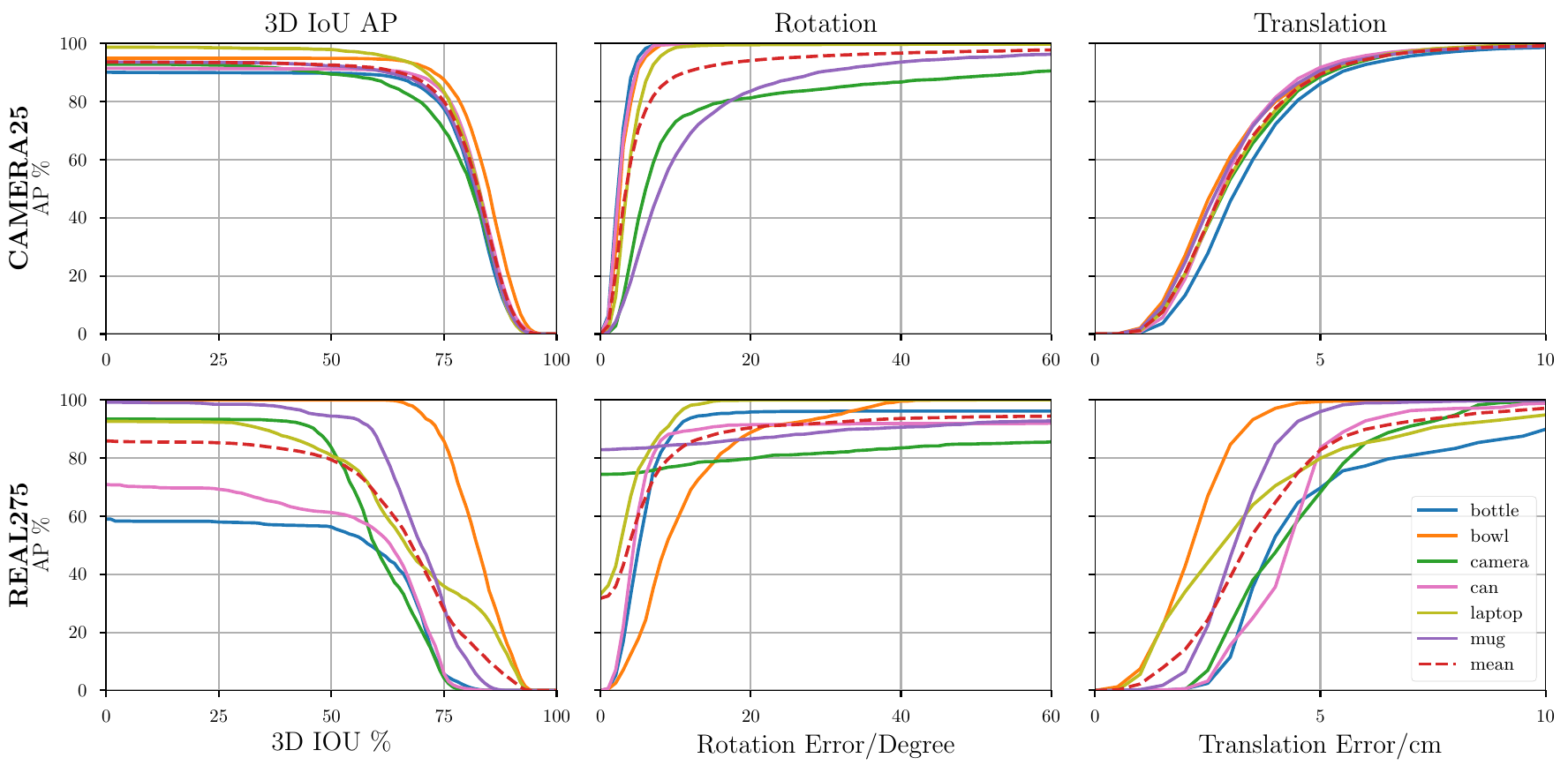}
      \caption{Average precision of ShAPO for various IOU, rotation, and translation thresholds on NOCS CAMERA25 and REAL275 dataset.}
  \label{fig:mAP}
\end{figure}
\setlength{\tabcolsep}{1pt}
\begin{table*}[t]
    \centering
    \small
    \renewcommand{\arraystretch}{1.3}
    \caption{
    \textbf{Quantitative comparison of 3D shape reconstruction on NOCS}~\cite{wang2019normalized}: Evaluated with \textbf{CD} metric ($10^{-2}).$ Lower is better.
    }
    
    \label{reconstruction_nocs_shapo}
    \resizebox{1.0\textwidth}{!}{
    \begin{tabular}{clcccccccccccccc}
        \toprule
        & & \multicolumn{7}{c}{\textbf{CAMERA25}} & \multicolumn{7}{c}{\textbf{REAL275}} \\ \cmidrule(r{0.1in}){3-9} \cmidrule(r{0.1in}){10-16}
        & {Method} & \textbf{Bottle} & \textbf{Bowl} & \textbf{Camera} & \textbf{Can} & \textbf{Laptop} & \textbf{Mug} & \textbf{Mean} & \textbf{Bottle} & \textbf{Bowl} & \textbf{Camera} & \textbf{Can} & \textbf{Laptop} & \textbf{Mug} & \textbf{Mean} \\
        \midrule
        1 & {Reconstruction~\cite{tian2020shape}} & 0.18 & 0.16  & 0.40 & 0.097 & 0.20 & 0.14 & 0.20& 0.34 &0.12 & 0.89 & 0.15 &0.29 &0.10 & 0.32 \\
        2 & {ShapePrior~\cite{tian2020shape}} &0.34 &0.22& 0.90& 0.22& 0.33& 0.21 &0.37 &0.50 &0.12& 0.99& 0.24& 0.71& 0.097& 0.44 \\
        3 & CenterSnap &0.11	&0.10&	0.29&	0.13&	0.07& 0.12	&	0.14 & 0.13 &  0.10 & 0.43 &0.09 & 0.07 & 0.06 & 0.15 \\
        \midrule
        3 & \textbf{ShAPO{}(Ours)} &0.14	&0.08&	0.2&	0.14&	0.07& 0.11	&	0.16 & 0.1 &  0.08 &0.4 &0.07 & 0.08 & 0.06 & 0.13 \\
        \bottomrule
    \end{tabular}
    }
\end{table*}

\textbf{Datasets} We utilize the \textbf{NOCS}~\cite{wang2019normalized} dataset for both shape reconstruction and category-level 6D pose and size estimation evaluation. For training, we utilize the CAMERA dataset which comprises 300K synthetic images, of which 25K are held out for evaluation. The training dataset includes 1085 object models from 6 different categories - \textit{bottle, bowl, camera, can, laptop, and mug} whereas the evaluation dataset includes 184 different models. The REAL dataset train set comprises 7 scenes with 4300 images, and the test-set comprises 6 scenes with 2750 real-world images. \\
\looseness=-1
\textbf{Metrics}
We evaluate the performance of 3D object detection and 6D pose estimation independently following~\cite{wang2019normalized}. We use the following key metrics: 1) Average precision for different IOU-overlap thresholds (\textbf{IOU25} and \textbf{IOU50}). 2) Average precision for which the error is less than $n^{\circ}$ for rotation and $m$ cm for translation (\textbf{5\textdegree \SI{5}{\cm}}, \textbf{\textbf{5\textdegree \SI{10}{\cm}}} and \textbf{\textbf{10\textdegree \SI{10}{\cm}}}). We use Chamfer distance~(CD) for shape reconstruction following~\cite{yuan2018pcn}.

\setlength{\tabcolsep}{6pt}
\begin{table}[t]
  \centering
  \small
  \renewcommand{\arraystretch}{1.3}
  \caption{\textbf{Generalizable Implicit Representation Ablation}: We evaluate the efficiency (point sampling/time(s)/memory(MB)) and generalization (shape(CD) and texture(PSNR) reconstruction) capabilities of our implicit object representation as well as its sampling efficiency for different levels of detail (LoDs) and compare it to the ordinary grid sampling. All ablations were executed on NVIDIA RTX A6000 GPU.}
  \resizebox{1.0\textwidth}{!}{%
    \begin{tabular}{c|c|cc|cc|cc}
    \toprule
    \multirow{2}[4]{*}{Grid type} & \multirow{2}[4]{*}{Resolution} & \multicolumn{2}{c|}{Point Sampling} & \multicolumn{2}{c|}{Efficiency (per object)} & \multicolumn{2}{c}{Reconstruction} \\
\cmidrule{3-8}          &       & Input & Output & Time (s) & Memory (MB) & Shape (CD) & Texture (PSNR) \\
    \midrule
    \multirow{3}[2]{*}{Ordinary} & 40    & 64000 & 412   & 10.96 & 3994  & 0.30  & 10.08 \\
          & 50    & 125000 & 835   & 18.78 & 5570  & 0.19  & 12.83 \\
          & 60    & 216000 & 1400  & 30.51 & 7850  & 0.33  & \textbf{19.52} \\
    \midrule
    \multirow{3}[2]{*}{OctGrid} & LoD5  & 1521  & 704   & \textbf{5.53} & \textbf{2376} & 0.19  & 9.27 \\
          & LoD6  & 5192  & 3228  & 6.88  & 2880  & \textbf{0.18} & 13.63 \\
          & LoD7  & 20246 & 13023 & 12.29 & 5848  & 0.24  & 16.14 \\
    \bottomrule
    \end{tabular}%
    }
  \label{tab:optimization_3d}
\end{table}%
\subsection{Implementation Details}

ShAPO is sequentially trained first on the CAMERA set, with minimal fine-tuning on Real training set. For SDF, we use an MLP with 8 layers and a hidden size of 512. For color, we utilize a Siren MLP~\cite{sitzmann2020implicit} as it can fit higher frequencies better. We train the SDF and Color MLPs on all categories for 2000 epochs. We use Pytorch~\cite{NEURIPS2019_9015} for our models and training pipeline implementation. For optimization, we use an adaptive learning rate that varies with the obtained masks of each object, since we believe masks capture the confidence of heatmap prediction during detection. We optimize each object for 200 iterations. Our backbone is implemented as Feature pyramid network~\cite{kirillov2019panoptic} with takes as input Resnet~\cite{heDeepResidualLearning2016} outputs at various spatial resolutions and adds lateral connections with a top-down pathway. Each of our specialized heads comprises a series of convolution layers (i.e. up-sampling stages), as described in~\cite{Kirillov_2019_CVPR} with the final prediction layer comprising a $1\times1$ convolution and $4x$ bilinear up-sampling. We train the combined backbone and heads network for 30 epochs with early stopping based on the performance on the validation set. We use a learning rate of $6e^{-4}$ and a polynomial weight decay with a coefficient of $1e^{-4}$. Our texture network $t_{\theta}$ is a Siren-based~\cite{sitzmann2020implicit} 6-layer MLP consisting of 512-dimensional hidden layers and with $\omega_0$ set to 128. Siren networks demonstrate superior results at representing fine details when compared to standard ReLU-based MLPs thanks to the use of periodic activation functions. After we train the shape MLP~($G$) and texture MLP~($t_{\theta}$), we freeze the networks for single-shot supervision at the Gaussian center locations. During inference, we use the frozen networks~(G) and~($t_{\theta}$) to optimize for shape, pose, size, and appearance latent codes.

\begin{figure}[b!]
   \centering
       \includegraphics[width=\linewidth]{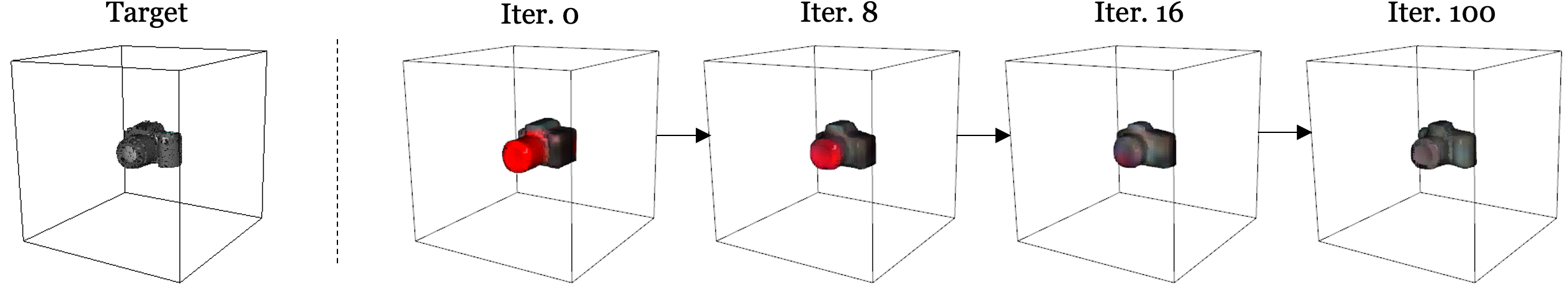}
      \caption{Shape/Appearance optimization.}
   \label{fig:optimization}
\end{figure}

\begin{figure}[htb]
   \centering
       \includegraphics[width=\linewidth]{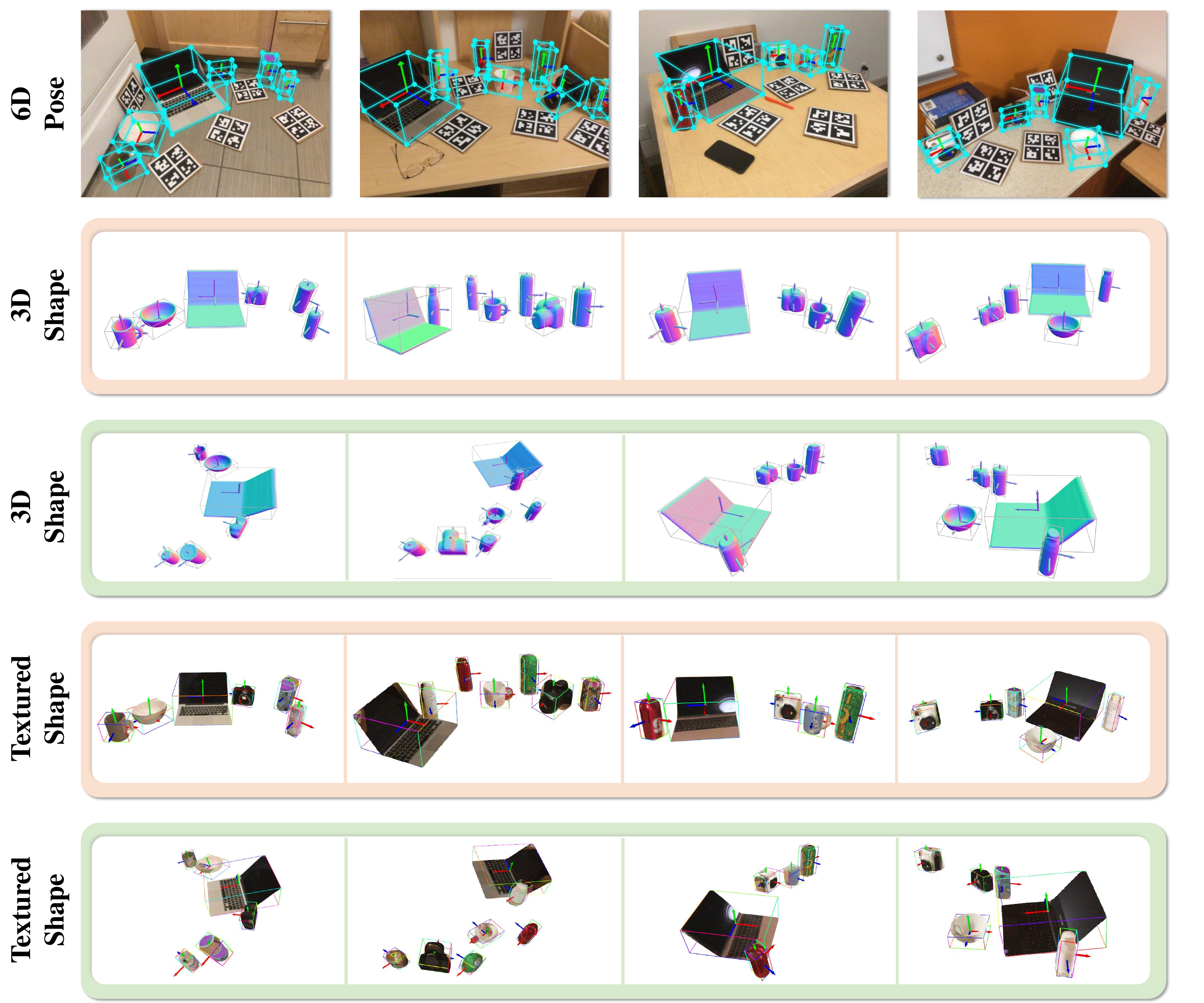}
      \caption{\textbf{ShAPO Qualitative 6D pose estimation and 3D Shape Reconstructions including Appearance:} Given a single-view RGB-D observation, our method reconstructs accurate 3D shapes along with appearances and estimates the 6D pose and sizes of multiple objects in the scene. Here, reconstructions are shown with different camera viewpoints i.e. orange and green backgrounds.} 
   \label{fig:qualitative_reconstructions}
\end{figure}

\subsection{NOCS 6D Pose and Size Estimation Benchmark}
\textbf{NOCS Baselines:} We compare eight model variants to show the effectiveness of our method:
(1) \textbf{NOCS}~\cite{wang2019normalized}: Regresses NOCS map and predicts poses using similarity transform with depth maps. We report the best pose estimation configuration in NOCS~(i.e. 32-bin classification) (2) \textbf{Shape Prior}~\cite{tian2020shape}: Estimates shape deformation for inter-class variation. (3) \textbf{CASS}~\cite{chen2020learning}: Regress the pose and size with first detecting 2D bounding boxes. (4) \textbf{Metric-Scale}~\cite{lee2021category}: Estimates the object center and metric scale (5) \textbf{CenterSnap~\cite{irshad2022centersnap}:} Single-shot approach to predict pose and size (6) \textbf{CenterSnap-R~\cite{irshad2022centersnap}:} ICP optimization done with the predicted point-cloud based shape. Following~\cite{irshad2022centersnap}, we do not compare against 6D pose tracking baselines such as~\cite{wang20206,wen2021bundletrack} which need pose initialization and are not end-to-end detection based~(i.e. they do not report mAP metrics) \\
\textbf{Comparison with strong baselines on NOCS:} Table~\ref{comparison_table_shapo} and Figure~\ref{fig:mAP} shows the result of our proposed ShAPO method. ShAPO consistently outperforms all the baseline methods on 6D pose and size estimation and 3D object detection. Specifically, ShAPO method shows superior generalization on the REAL test-set by achieving a mAP of 85.3$\%$ for 3D IOU at 0.25, 57.0$\%$ for 6D pose at 5\textdegree \SI{10}{\cm} and 78.0$\%$ for 6D pose at 10\textdegree \SI{10}{\cm}, hence demonstrating an absolute improvement of 1.8$\%$, 25.4$\%$ and 7.1$\%$ over the best-performing baseline on the Real dataset. Our method also achieves better generalization on the never-seen CAMERA evaluation set. We achieve a mAP of 94.5$\%$ for 3D IOU at 0.25, 75.9$\%$ for 6D pose at 5\textdegree \SI{10}{\cm} and 89.2$\%$ for 6D pose at 10\textdegree \SI{10}{\cm}, demonstrating an absolute improvement of 1.3$\%$, 4.2$\%$ and 1.3$\%$ over the best-performing baseline.

\begin{figure}[htbp]
   \centering
       \includegraphics[width=\linewidth]{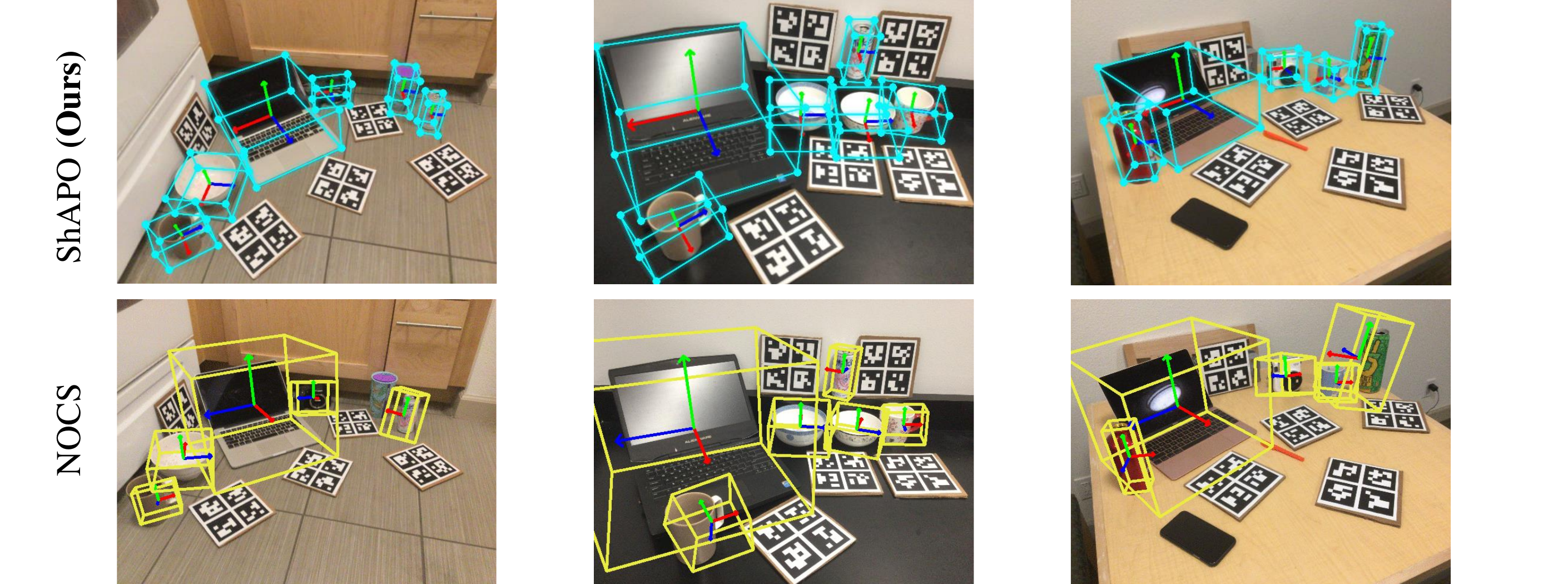}
      \caption{\textbf{ShAPO Qualitative 6D pose estimation comparison with NOCS:} Our method's 6D pose estimation in comparison to the best pose estimation configuration, i.e. 32-bin classification on the NOCS dataset. We show accurate 3D bounding box predictions and 6D pose and size estimation of multiple novel object categories than the strong baseline.} 
   \label{fig:nocs_comparison}
\end{figure}

\begin{figure}[htbp]
   \centering
       \includegraphics[width=\linewidth]{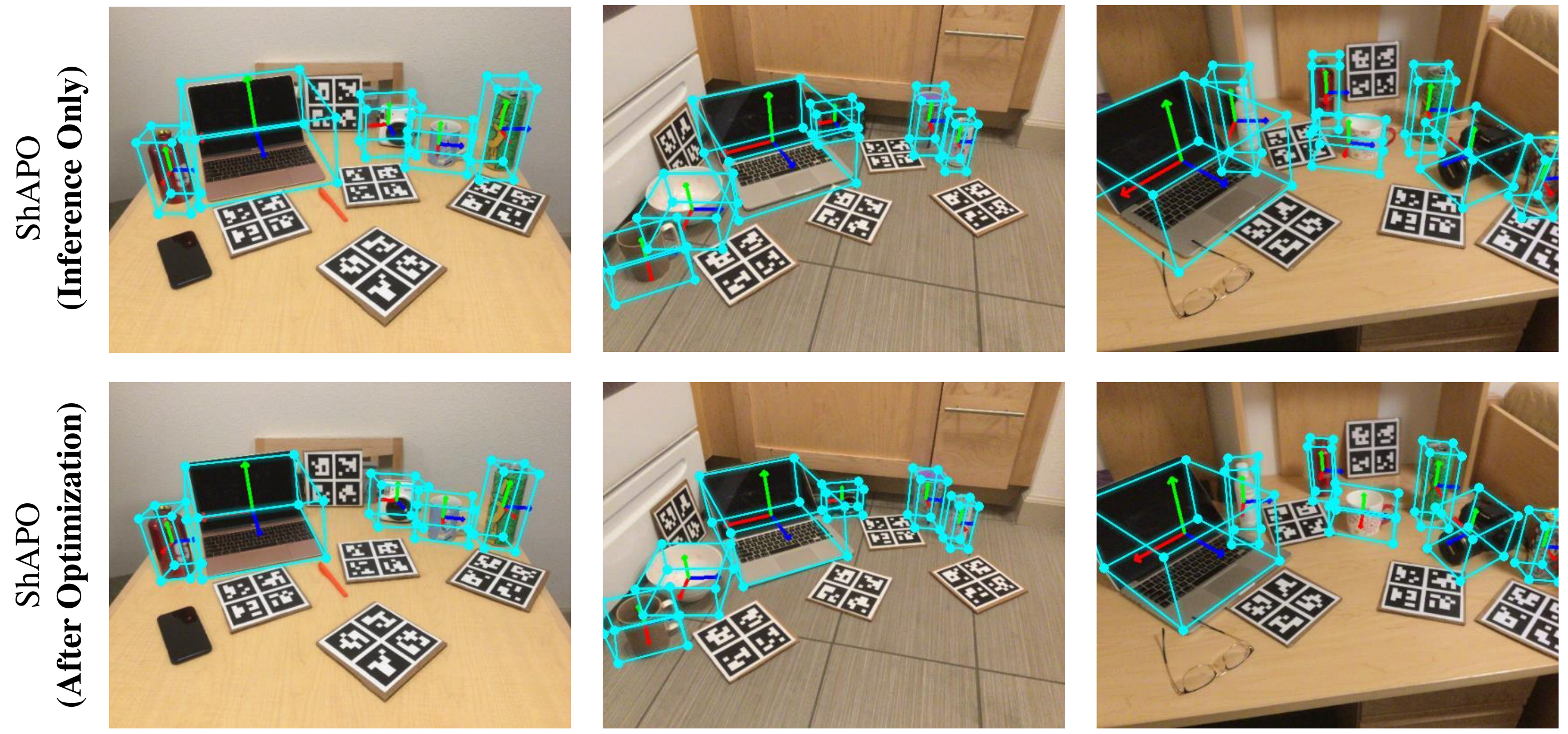}
      \caption{\textbf{ShAPO Qualitative Comparison of 6D pose and size Inference and Optimization:} Our method's 6D pose and size comparison shown on 3 novel scenes in NOCS Real275 test-set. After optimization, our method predicts accurate bounding boxes, as shown by the bottom row in the figure.} 
   \label{fig:before_after_optim_pose}
\end{figure}

\subsection{Generalizable Implicit Object Representation}
In this experiment, we evaluate the effectiveness of our generalizable implicit object representation as well as the efficiency of our octree-based differentiable optimization. To do that, we isolate our implicit representation from the detection pipeline and consider the 3D reconstruction/identification task - given a novel object from NOCS test split, we optimize our representation for 200 iterations, while keeping $f_{sdf}$ and $f_{rgb}$ weights frozen, to find the best fitting model in terms of both shape and texture. We initialize the optimization using the average latent feature per class. The standard Adam solver with a learning rate of 0.001 for both shape and appearance losses (L2 norms) is used with weight factors 1 and 0.3 respectively. We use three different octree resolution levels - from LoD5 to LoD7. Additionally, we show three different resolution levels for the standard ordinary grid sampling~\cite{zakharov2020autolabeling} (40, 50, 60). Table~\ref{tab:optimization_3d} summarizes the results of the ablation by comparing different modalities with respect to the average point sampling (input vs output) and time efficiency, average GPU memory consumption, as well as reconstruction for shape (Chamfer distance) and texture (PSNR). One can see that our representation is significantly more efficient than the ordinary grid representation with respect to all metrics. While LoD7 provides the best overall results, we use LoD6 for our experiments since it results in the optimal speed/memory/reconstruction trade-off. We show an example optimization procedure in Fig.~\ref{fig:optimization}.

\setlength{\tabcolsep}{16pt}
\begin{table}[htbp]
  \centering
  \caption{\textbf{Texture quality ablation.} We compare texture quality using the PSNR metric between three modalities: network prediction, optimization, and fine-tuning of the $t_{\theta}$ network.}
    \resizebox{0.7\textwidth}{!}{%
    \begin{tabular}{l|ccc}
    \toprule
          & \multicolumn{1}{l}{Inference} & \multicolumn{1}{l}{Optimization} & \multicolumn{1}{l}{Fine-tuning} \\
    \midrule
    PSNR  & 11.41 & 20.64 & 24.32 \\
    \bottomrule
    \end{tabular}%
    }
  \label{tab:psnr_shapo}%
\end{table}%

\textbf{Texture Quality Ablation on NOCS REAL 275}: In this section, we provide an ablation on the output texture quality on NOCS Real275 test scenes. In particular, we compare the direct network texture prediction with the result after our differentiable optimization and the result after our differentiable optimization with additional fine-tuning of the $t_{\theta}$ network weights. We use the learning rate of $10^{-5}$ for the weight fine-tuning. Table~\ref{tab:psnr_shapo} demonstrates that our optimization procedure almost doubles the texture quality in terms of PSNR. Additional fine-tuning of the network weights allows us to improve texture reconstruction results even further. For qualitative results, see Figure~\ref{fig:inference_optim_comparison}.

\begin{figure}[htbp]
   \centering
       \includegraphics[width=1.0\linewidth]{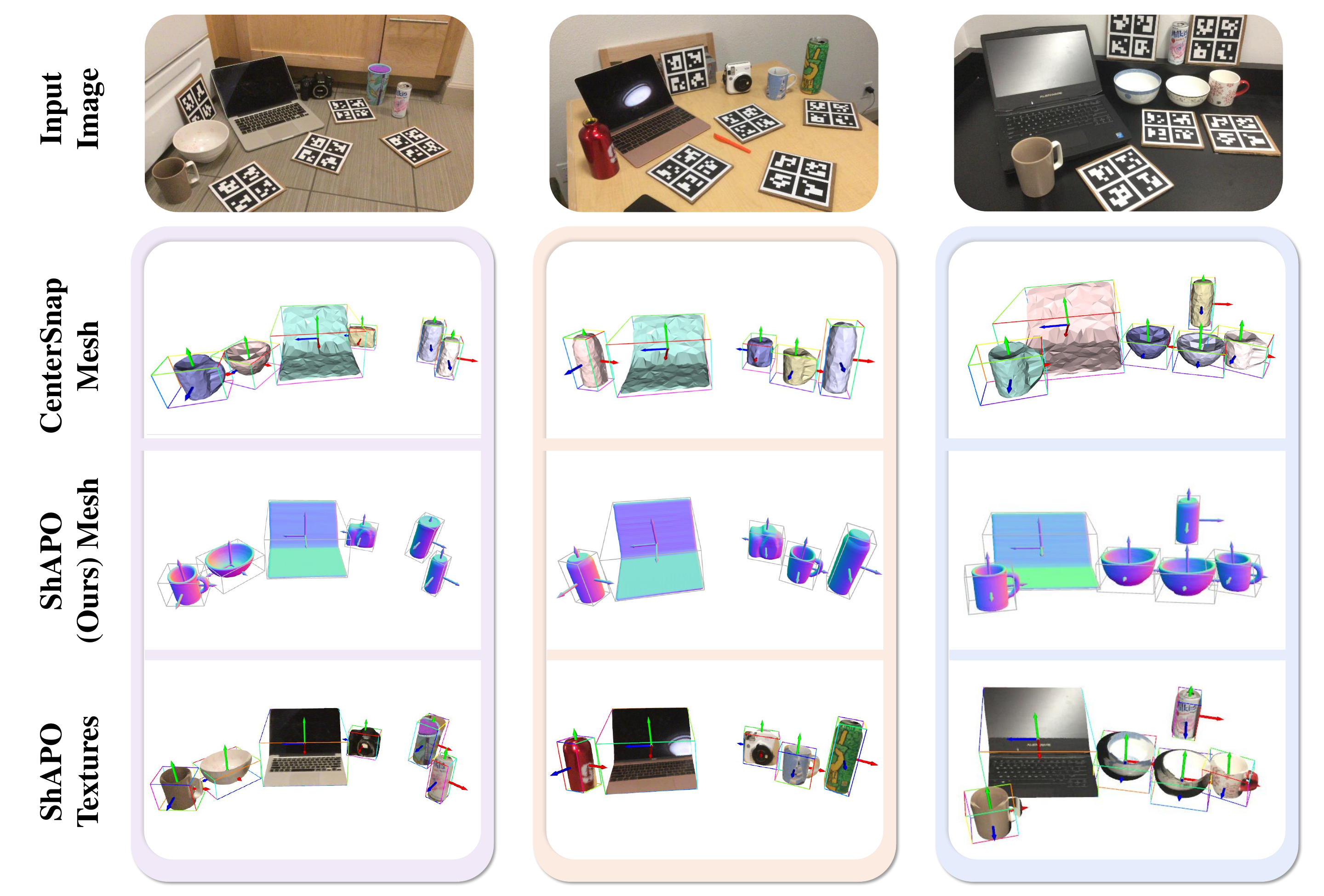}
      \caption{\textbf{ShAPO Qualitative Reconstruction Comparison with CenterSnap~\cite{irshad2022centersnap}:} The figure qualitatively shows the superior reconstruction performance of our method with the strong state-of-the-art i.e. CenterSnap~\cite{irshad2022centersnap} on novel scene in NOCS Real275 test-set. Our method produces finer reconstruction surfaces both in terms of shape accuracy and textures with details such as mug handle and camera lens.} 
   \label{fig:centersnap_comparison}
\end{figure}

\subsection{NOCS Object Reconstruction Benchmark}
To quantitatively analyze the reconstruction accuracy, we measure the Chamfer distance~(CD) between our reconstructed point clouds and the ground-truth CAD model in NOCS. Our results are reported in Table~\ref{reconstruction_nocs_shapo}. Our results show consistently lower CD metrics for all class categories, which shows superior reconstruction performance on novel object instances. We report a lower mean Chamfer distance of 0.14 on CAMERA25 and 0.15 on REAL275 compared to 0.20 and 0.32 reported by the competitive baseline~\cite{tian2020shape}.

\begin{figure}[htbp]
   \centering
       \includegraphics[width=1.0\linewidth]{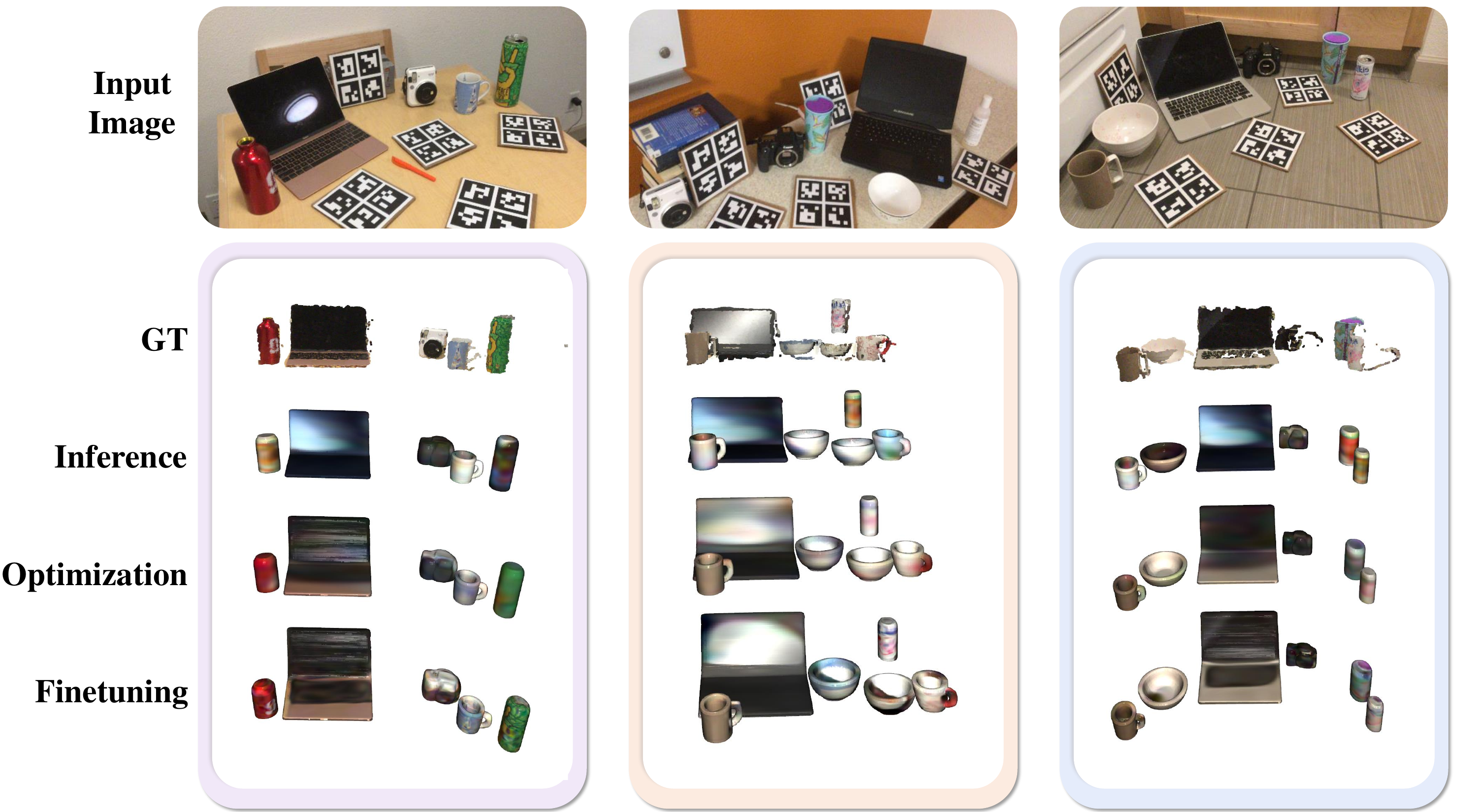}
      \caption{\textbf{ShAPO Qualitative Inference, Optimization and Fine-tuning Comparison:} The figure qualitatively shows the inference, latent-only optimization, and latent with appearance network optimization. Note that, as noted earlier, we let the appearance network weights change to allow for a finer level of reconstruction. } 
   \label{fig:inference_optim_comparison}
\end{figure}

\section{Qualitative Results}
\label{chap:shapo,sec:qualitative}

We qualitatively analyze the performance of our method \textbf{ShAPO} on the NOCS Real275 dataset~\cite{wang2019normalized}. As shown in Figure~\ref{fig:qualitative_reconstructions}, our method reconstructs accurate 3D shapes and appearances of multiple novel objects along with estimating the 6D pose and sizes without requiring 3D CAD models of these novel instances (Results shown on 4 different real-world scenes containing novel
object instances using different camera-view points i.e. orange and green backgrounds).

We show more visual qualitative results of superior single-view multi-object \textit{Shape Reconstruction}, \textit{6D pose and size estimation} and \textit{Appearance Reconstruction} done using our technique, \textbf{ShaPO}. Our method shows very promising results for superior 6D pose and size estimation compared to the strong baseline NOCS~\cite{wang2019normalized}~(Figure~\ref{fig:nocs_comparison}). Our network also performs more accurate shape and texture reconstruction compared to the strong-baseline, CenterSnap~\cite{irshad2022centersnap}~(Figure~\ref{fig:centersnap_comparison}), which only performs shape reconstruction~(i.e. meshes obtained through surface reconstruction of coarse pointcloud predictions i.e. 2048 points). We also visualize the improved pose estimation performance of our method after inference-time optimization~(Figure~\ref{fig:before_after_optim_pose}). Figure~\ref{fig:HSR} also shows zero-shot generalization results on HSR robot, i.e. no re-training was done. 

\begin{figure}[htbp]
\begin{center}
\includegraphics[width=0.75\linewidth]{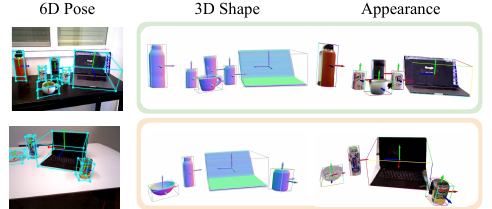}
\captionof{figure}{\textbf{Zero-shot real-world generalization experiments on HSR robot}}
  \label{fig:HSR}
\end{center}
\end{figure}

\section{Summary}
\label{chap:shapo,sec:conclusion}

In this chapter, we proposed ShAPO, an end-to-end method for joint multi-object detection, 3D textured reconstruction, 6D object pose, and size estimation. Our method detects and reconstructs novel objects without having access to their ground truth 3D meshes. To facilitate this, we proposed a novel, generalizable shape and appearance space that yields accurate textured 3D reconstructions of objects in the wild. To alleviate sampling inefficiencies leading to increased time and memory requirements, we proposed a novel octree-based differentiable optimization procedure that is significantly more efficient than alternative grid-based representations. For future work, we will explore how the proposed method can be used to build object databases in new environments, to alleviate the cost and time required to construct high-quality 3D textured assets. A second avenue of future work consists of extensions to multi-view settings and integrating our work with simultaneous location and mapping~(SLAM) pipelines for joint camera motion and object pose, shape, and texture estimation in static and dynamic scenes. Specifically, we made use of large-scale shape and appearance priors from synthetic data to demonstrate greater generalization capability to real-world scenes, requiring very little real-world fine-tuning.  
    
    \part{Hierarchical Vision-and-Language for Action}
    \label{part2}

    This part delves into the exploration of incorporating prior knowledge into agent-centric deep learning systems, specifically where the agent takes actions based on visual observations. We highlight the effectiveness of semantic maps, representing a top-down view of the 3D scene, as a robust prior. Additionally, we investigate the utility of encoding hierarchy to disentangle systems reliant on predicting actions through deep learning frameworks. Our research demonstrates enhanced generalization capabilities in novel environments through the application of both hierarchy and semantic map prior techniques. In Chapter~\ref{chap:robovln}, titled "Hierarchical Cross-Modal Agent for Robotics Vision-and-Language Navigation," we introduce the innovative Robo-VLN setting, extending Vision-and-Language Navigation (VLN) to continuous 3D environments with longer trajectories and obstacles. ~\textit{Continuous} here refers to simulated agents not teleporting from one viewpoint to the next in a sparse graph where the nodes denote navigability but rather being equipped with continuous action spaces such as velocity and steering commands. This extension makes the VLN problem more challenging and closer to real-world, where the problem is not just a simple graph search over nodes, but rather dealing with the challenge of imperfect topology, localization, and stochastic navigation. Our study reveals that existing baselines struggle with this task, prompting the proposal of a Hierarchical Cross-Modal (HCM) approach. Through extensive experiments, this work establishes a new benchmark for Robo-VLN, surpassing the performance of existing methods. In Chapter~\ref{chap:sasra}, titled "SASRA: Semantically-aware Spatio-temporal Reasoning Agent for Vision-and-Language Navigation in Continuous Environments," we identify a critical gap in the existing approach to Vision-and-Language Navigation (VLN) in continuous 3D environments, which heavily relies on raw visual data. This chapter demonstrates that the integration of classical semantic mapping techniques with learning-based methods significantly enhances the agent's adaptability to new scenes. By leveraging a Transformer architecture for the fusion of map and language features, coupled with a hybrid Transformer for action prediction, our approach achieves state-of-the-art results in the challenging continuous VLN task.
    \clearpage

    \chapter{Robo-VLN: Hierarchical Cross-Modal Agent for Robotics Vision and Language Navigation}
\label{chap:robovln}

In the first chapter of this part, I concentrate on studying the impact of injecting prior knowledge into agent-centric deep learning systems. I specifically explore the problem setting of~\textbf{utilizing hierarchy as a means to disentangle systems} which predict actions in an end-to-end manner using deep learning. Our main aim is to propose a realistic Vision-and-Language (VLN) navigation setting and improve the generalizability of VLN agents by introducing~\textit{modularity and disentanglement} priors in an end-to-end manner.

\section{Introduction}
\label{chap:robovln,sec:intro}

\begin{figure}[h]
\centering
\includegraphics[width=1.0\columnwidth]{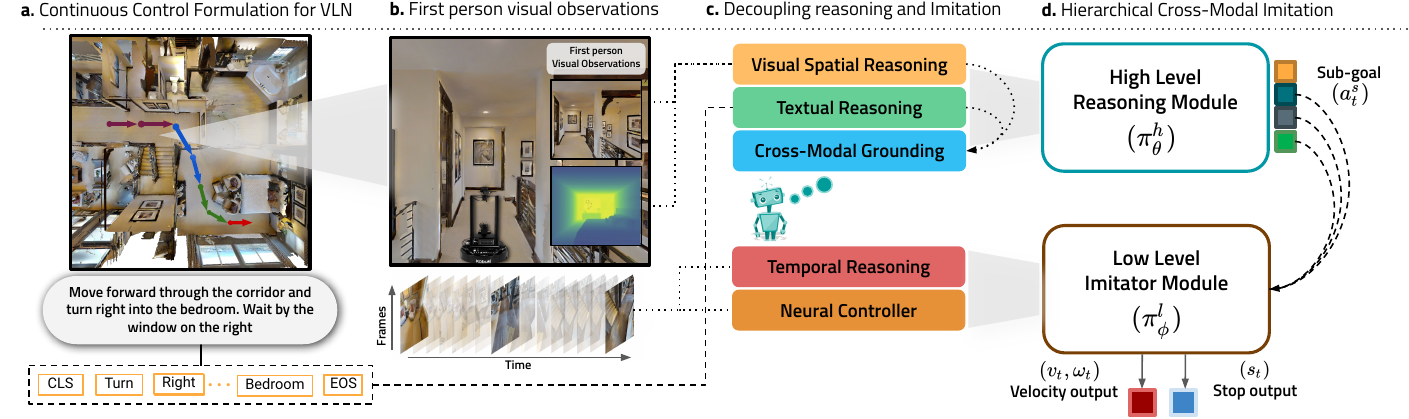}
\centering
  \caption{
  Our proposed method,~\textbf{~\gls{robovln}}~\cite{irshad2021hierarchical} task in continuous environments and our proposed Hierarchical Cross-Modal (HCM) agent. The agent decouples reasoning and imitation through a modularized training regime to solve the complex long-horizon Robo-VLN task}
  \label{overview_robovln}
\end{figure}

The promise of personal assistant robots that can seamlessly follow human instructions in real life environments has long been sought after. Recent advancements in deep learning (to extract meaningful information from raw sensor data) and deep reinforcement learning (to learn effective decision-making policies) have enabled some progress towards this goal~\cite{vasudevan2020talk2nav, majumdar2020improving, hao2020towards}. Due to the difficulty of collecting data in these contexts, a great deal of work has been done using photo-realistic simulations such as those captured through Matterport3D panoramas in homes~\cite{Matterport3D} or point-cloud meshes in Gibson~\cite{gibsonenv}. For example, a number of works have investigated autonomous agents that can follow rich, natural-language instructions in such simulations~\cite{ma2019theregretful, Wang2019ReinforcedCM, DBLP:conf/nips/FriedHCRAMBSKD18}. Precisely defined,~\gls{vln} is a task that requires the agent to navigate to a goal location purely based on visual inputs and provided instructions in the absence of a prior global map~\cite{mattersim}.

\begin{table}[t!]
\centering
\caption{\textbf{Comparison between our proposed Robo-VLN setting and prior environments used for Vision-and-Language Navigation}} \label{tab:a}
\label{rvln_comparison}
\resizebox{1.0\textwidth}{!}{%
\begin{tabular}{c*{8}{>{$}c<{$}}}
\toprule

& \multicolumn{3}{c}{\textbf{---Simulation---}} &\multicolumn{2}{c}{\textbf{---Environment---}}&\multicolumn{2}{c}{\textbf{---Instructions---}} \\
\cline{1-4}\cline{4-6}\cline{6-8} \\ 
\textbf{} & \textbf{Action space}& \textbf{Granularity} & \textbf{Agent} & \textbf{Navigation} & \textbf{Type}  & \textbf{Richness} & \textbf{Generation} \\
\midrule
\textbf{Touchdown~\cite{ai2thor}, R2R~\cite{mattersim}} & $Discrete$ & $High$ & $Virtual$ & $Unconstrained$ & $Photo-realistic$ & $Complex$ & $Human-annotated$\\

\textbf{Follow-net~\cite{46942}}  & $Discrete$ & $High$ & $Virtual$ & $Constrained$ & $Synthetic$ & $Simple$ & $Human-annotated$\\

\textbf{LANI~\cite{misra-etal-2018-mapping}}  & $Discrete$ & $High$ & $Virtual$ & $Constrained$ & $Synthetic$ & $Simple$ & $Template based$\\

\textbf{VLN-CE~\cite{krantz2020navgraph}} & $Discrete$ & $High$ & $Virtual$ & $Unconstrained$ & $Photo-realistic$ & $Complex$ & $Human-annotated$\\

\midrule
\textbf{Robo-VLN (Ours)}  & $Continuous$ & $High/Low$ & $Robotics$ & $Unconstrained$ & $Photo-realistic$ & $Complex$ & $Human-annotated$ \\
\bottomrule
\end{tabular}
}
\end{table}

While increasingly effective neural network architectures have been developed for these tasks, many limitations still exist that prevent their applicability to real-world robotics problems. Specifically, previous works~\cite{Wang2019ReinforcedCM, DBLP:conf/nips/FriedHCRAMBSKD18,46942, Matterport3D, Zhuetal} have focused on a simpler subset of this problem by defining the instruction-guided robot trajectories as either a discrete navigation graph~\cite{Matterport3D, mattersim} or assuming the action space of the autonomous agent comprises discrete values~\cite{ALFRED20, krantz2020navgraph}. These formulations assume known topology, perfect localization, and deterministic navigation from one viewpoint to the next in the absence of any obstacles~\cite{krantz2020navgraph}. Hence, these assumptions significantly deviate from the real world both in terms of control and perception.

As a first contribution, we focus on a richer VLN formulation, which is defined in continuous environments over long-horizon trajectories. Our proposed setting, \textbf{Robo-VLN} (\textbf{Robo}tics \textbf{V}ision-and-\textbf{L}anguage \textbf{N}avigation), is summarized in Figure~\ref{overview_robovln}. We lift the agent off the navigation graph, making the language-guided navigation problem richer, more challenging, and closer to the real world. In this context, the term "continuous" pertains to simulated agents avoiding teleportation between viewpoints in a sparse graph, where nodes represent navigability. Instead, these agents are equipped with continuous action spaces, such as velocity and steering commands. This expansion introduces greater complexity to the VLN problem, bringing it closer to real-world scenarios. In these scenarios, the challenge extends beyond a mere graph search over nodes, encompassing issues like imperfect topology, localization, and stochastic navigation.

In an attempt to solve the language-guided navigation (VLN) problem, recent learning-based approaches~\cite{ma2019theregretful, ma2019selfmonitoring, wang2019reinforced} make use of sequence-to-sequence architectures~\cite{10.5555/2969033.2969173}. However, when tested for generalization performance in unseen environments, these approaches (initially developed for shorter horizon nav-graph problems) translate poorly to more complex settings~\cite{ALFRED20,krantz2020navgraph}, as we also showed for Robo-VLN in Section~\ref{chap:robovln,sec:exp}.
Hence, for our proposed continuous VLN setting over long-horizon trajectories, we present an approach utilizing \textit{hierarchical decomposition}. Our proposed method leverages hierarchy to decouple cross-modal reasoning and imitation, thus equipping the agent with the following key abilities: 

\textbf{1. Decouple Reasoning and Imitation.}
The agent comprises a high-level policy and a corresponding low-level policy. The high-level policy is tasked with aligning the relevant instructions with observed visual cues as well as reasoning over which instructions have been completed, hence producing a sub-goal output through cross-modal grounding. 
The low-level policy imitates the feedback controller based on sub-goal information and observed visual states. A layered decision-making allows spatially different reasoning at different levels in the hierarchy, hence specializing each policy with a dedicated reasoning abstraction level.

\textbf{2. Modularized Training.}
Disentangling reasoning and controls allows fragmenting a complex long-horizon problem into shorter time horizon problems.
Since each policy is tasked with fulfilling a dedicated goal, each module utilizes separate end-to-end training with sparse communication between the hierarchy in terms of sub-goal information.
In summary, we make the following contributions: 
\begin{itemize}

\item To the best of our knowledge, we present the first work on formulating Vision-and-Language Navigation (VLN) as a continuous control problem in photo-realistic simulations, hence lifting the agent of the assumptions enforced by navigation graphs and discrete action spaces.

\item We formulate a novel hierarchical framework for Robo-VLN, referred to as \textbf{H}ierarchical \textbf{C}ross-\textbf{M}odal Agent (\textbf{HCM}) for effective attention between different input modalities through a modularized training regime, hence tackling a long-horizon and cross-modal task using layered decision-making.
\item Provide a suite of baseline models in Robo-VLN inspired by recent state-of-the-art works in VLN and present a comprehensive comparison against our proposed hierarchical approach --- Our work sets a new strong benchmark performance for a long horizon complex task, Robo-VLN, with over 13\% improvement in absolute success rate in unseen validation environments.
\end{itemize} 

\section{Related Work} 
\label{chap:robovln,sec:relatedworks}

\textbf{Vision-and-Language Navigation.} 
Learning based navigation has been explored in both synthetic~\cite{DBLP:journals/corr/KempkaWRTJ16, DBLP:journals/corr/abs-1712-05474, wu2018building} and photo-realistic~\cite{habitat19iccv, Matterport3D, gibsonenv} environments. For a navigation graph-based formulation of the VLN problem (i.e. discrete action space), previous works have utilized hybrid reinforcement learning~\cite{DBLP:conf/eccv/WangXWW18}, behavior cloning~\cite{eqa_modular}, speaker-follower~\cite{NIPS2018_7592} and sequence-to-sequence based approaches~\cite{mattersim}. Subsequent methods have focused on utilizing auxiliary losses~\cite{ma2019selfmonitoring, Zhu_2020_CVPR}, backtracking~\cite{ma2019theregretful} and cross-modal attention techniques~\cite{DBLP:conf/cvpr/WangHcGSWWZ19,DBLP:journals/corr/abs-1905-13358,landi2020perceive} to improve the performance of VLN agents. Our work, in contrast to discrete VLN setting \cite{krantz2020navgraph,mattersim} (see Table~\ref{rvln_comparison}), focuses on a much richer VLN formulation, which is defined for continuous action spaces over long-horizon trajectories. 
We study the new continuous Robo-VLN setting and propose hierarchical cross-modal attention and modularized training regime for such task. 

\textbf{Hierarchical Decomposition.}
Hierarchical structure is most commonly utilized in the context of Reinforcement Learning over long-time horizons to improve sample efficiency~\cite{Sutton:1999, Sutton99betweenmdps, Vezhnevets2017FeUdalNF}. Our work closely relates to the options framework in Reinforcement Learning~\cite{,eqa_modular, pmlr-v80-le18a, Sutton:1999, pmlr-v54-fruit17a} where the top-level policy identifies high-level decisions to be fulfilled by a bottom-level policy. In relation to other works which utilize sub-task decomposition for behavior cloning~\cite{pmlr-v80-le18a, Roh2019Conditional}, we show that decomposing hierarchy based on reasoning and imitation are quite effective for long-horizon multi-modal tasks such as Robo-VLN. 

\section{Robotics Vision-and-Language Navigation Environment (Robo-VLN)}\label{robo-vln}

Different from existing VLN environments, we propose a new continuous environment for VLN that more closely mirrors the challenges of the real world, Robo-VLN --- a continuous control formulation for Vision-and-Language Navigation. 
Compared to navigation graph based~\cite{mattersim} and discrete VLN settings~\cite{krantz2020navgraph}, Robo-VLN provides longer horizon trajectories (4.5x average number of steps), more visual frames ($\sim$3.5M visual frames), and a balanced high-level action distribution (see Figure~\ref{comparison_discrete}). Hence, making the problem more challenging and closer to the real-world.

\begin{figure}[t!]
\centering
\includegraphics[width=0.95\columnwidth]{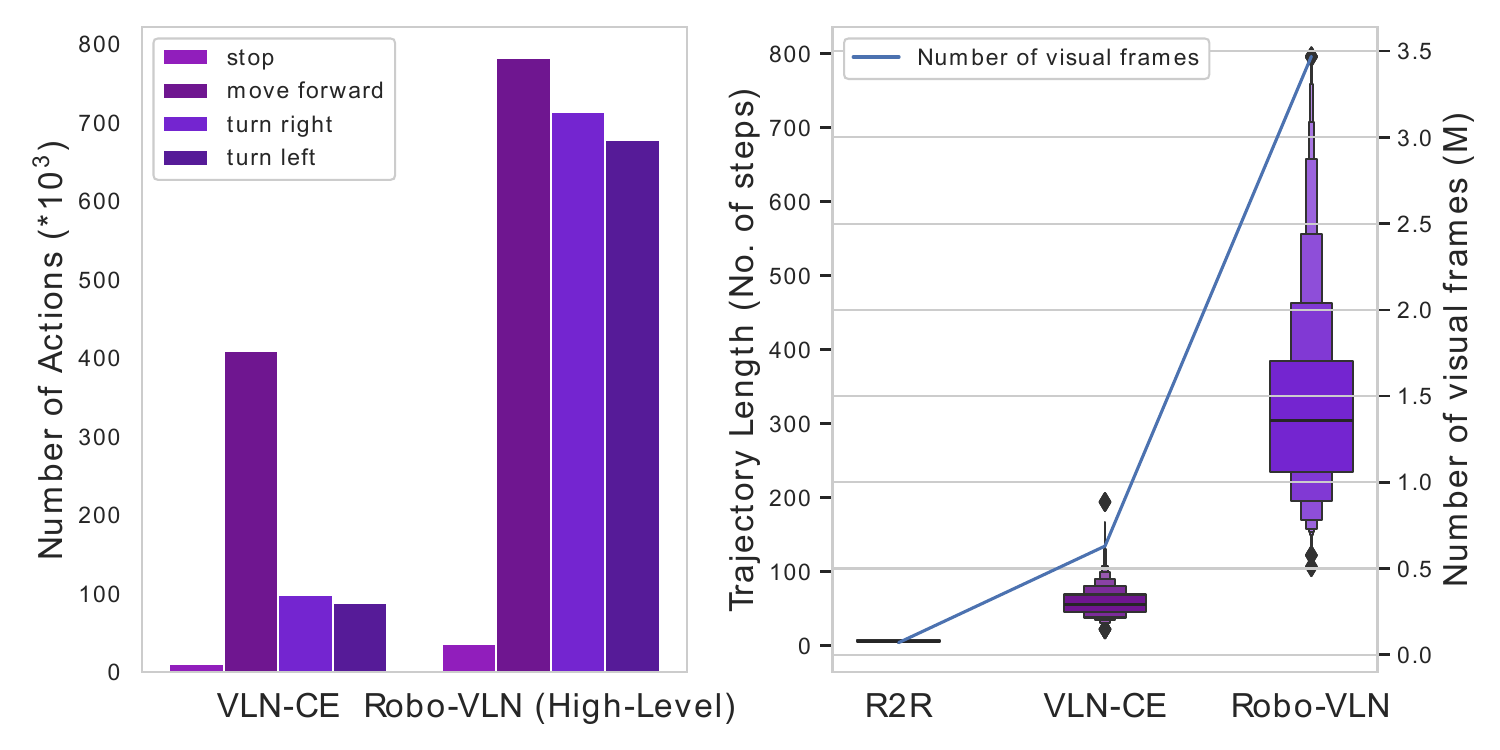}
\centering
  \caption{
  \textbf{Robo-VLN compared with discrete VLN settings:} VLN-CE~\cite{krantz2020navgraph} and R2R~\cite{mattersim}.
  We provide longer horizon trajectories (4.5x average number of steps, over 3M visual frames, and a balanced high-level action distribution.
  }
  \label{comparison_discrete}
\end{figure}
\section{Method}
\label{chap:robovln,sec:method}

Different from existing VLN environments, we propose a new continuous environment for VLN that more closely mirrors the challenges of the real world, Robo-VLN --- a continuous control formulation for Vision-and-Language Navigation. 
Compared to navigation graph based~\cite{mattersim} and discrete VLN settings~\cite{krantz2020navgraph}, Robo-VLN provides longer horizon trajectories (4.5x average number of steps), more visual frames ($\sim$3.5M visual frames), and a balanced high-level action distribution (see Figure~\ref{comparison_discrete}). Hence, making the problem more challenging and closer to the real-world.

\subsection{Problem Definition}
Formally, consider an autonomous agent $\tilde{\mathcal A}$ in an unknown environment $\tilde{\mathcal E}$. 
The goal of a Robo-VLN agent is to learn a policy $a_{t} = \pi(x_{t}, q_{t}, \theta)$ where the agent receives visual observations ($x_{t}$) from the environment $\tilde{\mathcal E}$ at each time-step ($t$) while following a provided instruction ($q$) to navigate to a goal location $\mathcal G$. 
$\theta$ denotes the learnable parameters of the policy $\pi$. 
The action space of the agent consists of continuous linear and angular velocity ($v_{t}, \omega_{t}$) and a discrete stop action ($s_{t}$). 
An episode ($\tau$) is considered successful if the agent's distance to the goal is less than a threshold ($d_{a}<3m$) and the agent comes to a stop by either taking the stop action ($s_{t}$) or decreasing its angular velocity below a certain threshold.

\subsection{Constructing Continuous VLN in 3D Reconstructions}
To make the continuous VLN formulation possible in 3D reconstructed environments, we port over human annotated instructions ($q_{t}$) corresponding to sparse way-points ($z_{t}$) along each instruction-trajectory pair in Room2Room (R2R) dataset \cite{mattersim}, using a continuous control formulation. We do this in 2 stages as follows:

\textbf{Ground-truth oracle feedback controller in 3D reconstructed environments.}
We consider the robotic agent to be a differential drive mobile robot, Locobot~\cite{pyrobot2019}, with a specified radius and height. We develop $A^{\star}$ planner to compute high-level oracle actions ($a^{h}_{t}$) along the shortest path to the goal and use a feedback controller~\cite{9780133496598} to convert the discrete R2R trajectories \cite{mattersim} into continuous ones.
The low-level oracle controller ($u_{t}$) outputs velocity commands $(v_{t}, \omega_{t})$ given sparse way-points ($z_{t}$) along a given language-guided navigation trajectory from the R2R dataset~\cite{Matterport3D}.
The converted continuous actions from the low-level controller will then be used as ground-truth low-level supervisions $a^{l}_{t}$ when training the navigation agents.
We create this continuous control formulation inside Matterport 3D environments~\cite{Matterport3D} by considering the Locobot robot as a 3D mesh inside 3D reconstructed environments (see Figure~\ref{qualitative}).
We use the robot's dynamics~\cite{10.5555/1855026} to predict next state ($\hat{x}_{t+1}$) given current state ($\hat{x}_{t}$) and controller actions ($a^{l}_{t}$). 
Similar to Habitat~\cite{habitat19iccv}, we render the mesh for any arbitrary viewpoint by taking the position generated by the dynamic model inside the 3D reconstruction.

\begin{figure}[t!]
\centering
\includegraphics[width=1.0\textwidth]{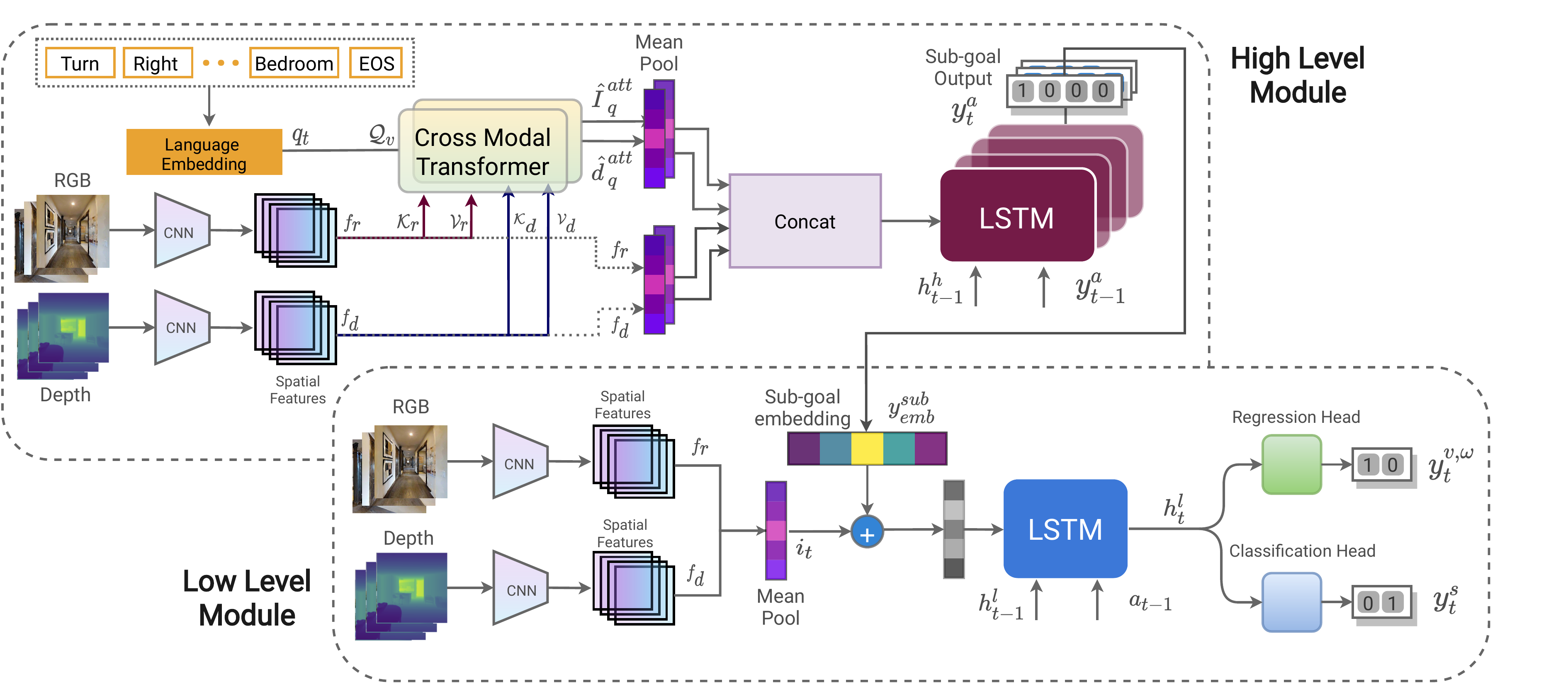}
\captionof{figure}{
\textbf{Hierarchical Cross-Modal Agent (HCM):} Our proposed agent consists of a \textit{high-level module} and a corresponding \textit{low-level module}. High-level module predicts the sub-goal output based on alignment between instructions and visual observations. Low-level module translates the high-level sub-goal output to linear and angular velocities using an imitation learning policy.}
\label{framework_robovln}
\end{figure}

\textbf{Obtaining Navigable Instruction-Trajectory pairs.}
Given a feedback controller of the form $a^{l}_{t} = u_{t}(z_{t})$ and high-level sparse viewpoints ($z_{t} = [z_{1}, \ldots, z_{N}]$ along the language guided navigation trajectory inside a reconstructed mesh, we search for the navigable space $h_{nav}(z_{t})$ using collision detection.
We find navigable space for all the trajectories present in the R2R dataset~\cite{mattersim}.
This procedure ensures the transfer of only the navigable trajectories from R2R dataset to the continuous control formulation in Robo-VLN; hence, we eliminate non-navigable unrealistic paths for a mobile robot, such as climbing up the stairs and moving through obstacles. 
Through this approach, we transferred 71\% of the trajectories from the discrete VLN setting (VLN-CE~\cite{krantz2020navgraph}) while preserving all the environments in the Matterport3D dataset~\cite{Matterport3D}. At the end, Robo-VLN's expert demonstration provide first person RGB-D visual observations ($i_{t}$), human instructions ($q_{t}$), and oracle actions ($a^{h}_{t}, a^{l}_{t}$) for each instruction-trajectory pair.

\section{Hierarchical Cross-Modal Agent}
Learning an effective policy ($\pi$) for a long horizon continuous control problem entails preserving the temporal states as well as spatially reasoning about the surroundings. 
We therefore propose a hierarchical agent to tackle the Robo-VLN task, as it effectively disentangles different dedicated tasks through layered decision-making. 

Given states ($\mathcal{X}=\{x\}$) and instructions ($\mathcal{Q}=\{q\}$), our agent leverages these inputs and learns a high-level policy ($\pi^{h}_{\theta}:  \rightarrow \mathcal{A}_{s,t}$) and a corresponding low-level policy ($\pi^{l}_{\gamma}:\mathcal{X} \times \mathcal{A}_{s,t} \rightarrow \mathcal{A}_{l,t}$). 
The high-level policy consistently reasons about the alignment between input textual and visual modalities to produce a sub-goal output ($\mathcal{A}_{s,t}$). The low-level policy ensures that the high-level sub-goal is translated to low-level actions ($\mathcal{A}_{l,t}$) effectively by imitating the expert controller through an imitation learning policy. Our approach is summarized in Figure~\ref{framework_robovln} and subsequent sections.

\subsection{High-Level Policy}
The high-level policy ($\pi^{h}_{\theta}$) decides a short-term goal ($a^{h}_{t}$) based on the input instructions ($q_{t}$) and observed visual information $x_{t} = \{ r_{t}, d_{t}\}$ from the environment at each time-step, where $r_{t}$, $d_{t}$ denote the RGB and Depth sensor readings respectively. $\pi^{h}_{\theta}$ consists of an encoder-decoder architecture with cross attention between the modules. Subsequent modules of the high-level policy ($\pi^{h}_{\theta}$) are described below.

\textbf{Multi-Modal Cross Attention Encoder.}
Given a natural language instruction comprised of $k$ words, we denote its feature representation as $\{q = q^{1}_{t}, q^{2}_{t}, \ldots, q^{k}_{t} \}$, where $q^{i}_{t}$ is the encoded feature representation of the $i_{th}$ word using BERT embedding~\cite{devlin-etal-2019-bert} to extract meaningful representation of words in the sentence.
To encode the observed RGB-D states ($r_{t}$ $\in$ $\mathds{R}^{h_{o}\times w_{o}\times 3 }$, $d_{t}$ $\in$ $\mathds{R}^{h_{o}\times w_{o}}$), we generate a low-resolution spatial feature representations $f_{r}$ $\in$ $\mathds{R}^{H_{s}\times W_{s}\times C_{s}}$ and $f_{d}$ $\in$ $\mathds{R}^{H_{s}\times W_{s}\times C_{s}}$ by using a pre-trained~\gls{cnn} backbone, where $H_{s}=W_{s} = 7$ and $C_{s}=2048$.
At each time-step $t$, we combine the individual RGB ($f_{r}$) and Depth ($f_{d}$) spatial features with encoded language representation ($q_{t}$) using a Transformer module~\cite{NIPS2017_7181}. Each Transformer module comprises stacked multi-head attention block $(\mathcal{A}_{M})$ followed by a position-wise feed-forward block. We utilize layer normalizations~\cite{ba2016layer} between these blocks along with the residual connection from the previous block such that output of each individual block is $\it{LayerNorm}(z + module(z))$. Each Transformer block is computed as follows:

\begin{equation}\begin{aligned}
\mathcal{A}_{M}(\boldsymbol{Q}, \boldsymbol{K}, \boldsymbol{V}) &=\operatorname {concat}(\boldsymbol{h}_{1},\ldots, \text{ $\boldsymbol{h}_{k}$ }) \boldsymbol{W}^{h}, \\
\text { where $\boldsymbol{h}_{i}$ }&= \mathcal{A}\left(\boldsymbol{Q} W_{i}^{Q}, \boldsymbol{K} W_{i}^{K}, \boldsymbol{V} W_{i}^{V}\right) \\
\mathcal{A}(\boldsymbol{Q}, \boldsymbol{K}, \boldsymbol{V})&=\operatorname{softmax}\left(\frac{\boldsymbol{Q} K^{T}}{\sqrt{d_{k}}}\right) \boldsymbol{V}
\label{transformer}
\end{aligned}\end{equation}
The Attention output ($\mathcal{A}$) is a weighted sum of the values ($V$) calculated using a similarity between projected Query ($Q$) and Key ($K$). $\mathcal{A}_{M}$ represents stacked Attention blocks ($\mathcal{A}$), and $W_{i}^{Q}, W_{i}^{K}, W_{i}^{V}$ and $W^{h}$ are parameters to be learned. 

We utilize Equation \ref{transformer} to perform cross attention between visual spatial representation (RGB $f_{r}$ or Depth $f_{d}$) and language features ($q_{t}$) successively. We do this by utilizing the sum of language features and sinusoidal Positional Encoding (PE~\cite{NIPS2017_7181}) as query ($Q = q_{t}$ + PE$(\boldsymbol{q}_{t})$) and visual representation as Key ($K_r = \boldsymbol{f}_{r}$ or $K_d = \boldsymbol{f}_{d}$) as well as Value ($V_r = \boldsymbol{f}_{r}$ or $V_d = \boldsymbol{f}_{d}$).
The final outputs, which we denote as cross-attended context (from RGB or Depth), are computed using $\mathcal{A}_{M}(\boldsymbol{Q}, \boldsymbol{K}, \boldsymbol{V})$, \textit{e.g.,} $\hat{\boldsymbol{I}}^{att}_{q}$ for RGB input and $\hat{\boldsymbol{d}}^{att}_{q}$ for Depth input.

These cross-attended contexts represent the matching between instructions and corresponding visual features at each time step $t$.
Note that the learnable weights in the Transformer are not shared between the two modalities. 

\textbf{Multi-Modal Attention Decoder.}
To decide on which direction to go next and select the most optimal high-level action ($a^{h}_{t}$) high-level policy preserves a temporal memory of the attended visual-linguistic contexts ($\hat{\boldsymbol{I}}^{att}_{q}$, $\hat{\boldsymbol{d}}^{att}_{q}$), mean-pooled visual features ($\hat{\mathrm{v}}_{t}$) and previous actions (${a}^{h}_{t-1}$). We rely on a Recurrent Neural Network to preserve this temporal information across time.
\begin{equation}
\begin{aligned}
\boldsymbol{h}^{h}_{t}&=\operatorname{LSTM}\left(\left[\hat{\boldsymbol{I}}^{att}_{q}, \hat{\boldsymbol{d}}^{att}_{q}, \hat{\boldsymbol{\mathrm{v}}}_{t}, {a}_{t-1}, \boldsymbol{h}^{h}_{t-1}\right]\right) \\
\hat{\boldsymbol{\mathrm{v}}}_{t}&=\boldsymbol{W}_{i}(\mathrm{g}( [\boldsymbol{f}_{r},\boldsymbol{f}_{d}])^\top+\boldsymbol{b}_{i})
\label{recurrence}
\end{aligned}
\end{equation}

where $\mathrm{g(.)}$ is mean adaptive pooling across the spatial dimensions. $W_{i}$ and $b_{i}$ are learned parameters of a fully-connected layer.

The agent computes a probability ($p^{h}_{a}$) of selecting the most optimal action ($a_{t}$) at each time-step by employing a feed-forward network followed by a $softmax$ as follows:
\begin{equation} \begin{aligned}
p^{h}_{a} &= softmax( \boldsymbol{W}_{a}([ \boldsymbol{h}^{h}_{t}] + \boldsymbol{b}_{a}))
\end{aligned}\end{equation}
where $W_{a}$ and $b_{a}$ are parameters to be learned. High-level action $a_{t}$ comprises the following navigable directions: \texttt{move forward} (0.25m), \texttt{turn-left} or \texttt{turn-right} (15 degrees) and \texttt{stop}.

\subsection{Low-level Policy}
We employ an imitation policy for the low-level module. 
At each time-step $t$, the low-level policy ($\pi^{l}_{\phi}$) selects a low-level action ($a^{l}_{t})$) given the sub-goal ($a^{h}_{t}$), generated by the high-level policy and observed visual states ($r_{t}, d_{t}$) from the environment. Low-level actions are comprised of agent's linear and angular velocity ($v_{t}, \omega_{t}$). Similar to the high-level module, we use mean pooled visual features ($\hat{\mathrm{v}}_{t}$) for the low-level policy and additionally condition the policy on the high-level sub-goal ($a^{h}_{t}$). Furthermore, we utilize stacked LSTM layers with respective fully-connected layers to generate both low-level action and stop probabilities ($p^{l}_{a},p^{s}_{a}$):
\begin{equation}
\boldsymbol{h}^{l}_{t}=\operatorname{LSTM}\left(\left[\hat{\boldsymbol{v}}_{t}, \boldsymbol{a}^{h}_{t}, \boldsymbol{h}^{l}_{t-1}\right]\right)
\end{equation}
\begin{equation}
p^{h}_{a} = tanh(\mathrm{g_{a}} ([ \boldsymbol{h}^{l}_{t}, \boldsymbol{a}^{l}_{t-1} ])), \quad
p^{s}_{a} = \sigma(\mathrm{g_{s}} ([ \boldsymbol{h}^{l}_{t}, \boldsymbol{a}^{l}_{t-1}]))
\end{equation} 
where $\mathrm{g_{a}(.)}$ and $\mathrm{g_{s}(.)}$ are one-layer Multi-Layer Perceptrons (MLP). $\sigma$ and $tanh$ are sigmoid and tanh activation functions, respectively.

\subsection{Training Details}
We train both high- and low-level policies jointly with three different losses. We employ a multi-class cross-entropy loss computed between ground-truth high-level navigable action ($y^{a}_{t}$) and the predicted action probability ($p^{h}_{a}$) for the high-level policy. We employ a mean squared error loss between ground-truth velocity commands ($y^{v,\omega}_{t}$) and predicted low-level action probabilities ($p^{l}_{a}$). Lastly, we use a binary cross-entropy loss between ground-truth stopping actions ($y^{s}_{t}$) and predicted stop probabilities ($p^{s}_{a}$) as follows:

\begin{equation}
\mathcal{L}_{\text {loss}}=\lambda \overbrace{\sum_{t=1}^{T} y_{t}^{a} \log \left(p^{a}_{h}\right)}^{\text {High-Level Action Loss}} + (1-\lambda) (\overbrace{\sum_{t=1}^{T}\left(y_{t}^{v,\omega}-p^{l}_{a}\right)^{2}}^{\text{Low-Level Action Loss }}  + \overbrace{\sum_{t=1}^{T} y^{s}_{t} \log \left(p^{s}_{a}\right)}^{\text{Low-Level Stop Loss }})
\end{equation}

\begin{table*}[t]
    \centering
    \renewcommand{\arraystretch}{1.3}
    \caption{
    \textbf{Quantitative comparison}: Comparison with strong baselines.
    Note that these baselines are reimplementations from VLN-CE~\cite{krantz2020navgraph} with small changes (see~\ref{baselines} for further details). 
    }
    \label{comparison_table_robovln}
    \resizebox{1.0\textwidth}{!}{
    \begin{tabular}{cccccccccccc}
        \toprule
        & & \multicolumn{5}{c}{\textbf{Validation Seen}} & \multicolumn{5}{c}{\textbf{Validation Unseen}} \\ \cline{3-12}
        & {Method} & \textbf{SR}~$\uparrow$ & \textbf{SPL}~$\uparrow$ & \textbf{NDTW}~$\uparrow$ & \textbf{TL}~$\uparrow$ & \textbf{NE}~$\downarrow$ & \textbf{SR}~$\uparrow$ & \textbf{SPL}~$\uparrow$ & \textbf{NDTW}~$\uparrow$ & \textbf{TL}~$\uparrow$ & \textbf{NE}~$\downarrow$ \\
        \midrule
        1 & {Random Agent} &  0.07& 0.07 & 0.14  & 5.26 & 10.25 & 0.08 & 0.08& 0.14  & 5.40 & 9.81 \\
        2 & {Seq2Seq~\cite{Matterport3D}} &0.36	&0.34&	0.32&	11.84&	8.63& 0.33	&	0.30	&0.28&	11.92&	8.97 \\
        3 & {PM~\cite{ma2019selfmonitoring}} & 0.32 & 0.27 & 0.23 & 14.12 & 9.33 & 0.28 & 0.24 & 0.22 &  13.85 & 9.82\\
        4 & {CMA~\cite{wang2019reinforced}} & 0.28&	0.25&	0.22&	11.52&	9.95&	0.28&	0.25&	0.23&	11.57&	9.63 \\
        \midrule
         & {\textbf{HCM (Ours)}} & \textbf{0.49}&	\textbf{0.43}&	\textbf{0.35}	& \textbf{13.53} &	\textbf{7.48} & \textbf{0.46}&	\textbf{0.40} &	\textbf{0.35} & \textbf{14.06} &	\textbf{7.94} \\
        \bottomrule
    \end{tabular}
    }
\end{table*}
\begin{table*}[t]
    \centering
    \renewcommand{\arraystretch}{1.3}
    \caption{
    \textbf{Ablation Study}: Impact of different modules and design choices in our proposed Hierarchical agent.
    }
    \label{ablation}
    \resizebox{1.0\columnwidth}{!}
{
    \begin{tabular}{cccccccccccccccc}
        \toprule
        & & \multicolumn{3}{c}{\textbf{Module}}& \multicolumn{5}{c}{\textbf{Validation Seen}} &\multicolumn{5}{c}{\textbf{Validation Unseen}} \\ \cline{3-15}  
& \# & {Vision} &{Hierarchy} & \begin{tabular}[c]{@{}c@{}}RGB-D\\ Early fusion\end{tabular} & \textbf{SR}~$\uparrow$ & \textbf{SPL}~$\uparrow$ & \textbf{NDTW}~$\uparrow$ & \textbf{TL}~$\uparrow$ & \textbf{NE}~$\downarrow\uparrow$ & \textbf{SR}~$\uparrow$ & \textbf{SPL} & \textbf{NDTW}~$\uparrow$ & \textbf{TL}~$\uparrow$ & \textbf{NE}~$\downarrow$ \\
\midrule
\multirow{4}{*}{\shortstack{Hierarchical\\Agent}} & 1 & &\checkmark &  & 0.07 & 0.07& 0.14& 4.82 & 10.34 & 0.07& 0.07 & 0.14 & 10.2 & 4.81  \\
& 2 &\checkmark& & & 0.44& 0.37 & 0.31& 14.87 & 8.21 & 0.40 & 0.34 & 0.28 &  15.32 & 8.64  \\
& 3 &\checkmark& \checkmark& \checkmark & 0.39 &  0.35 & 0.29 & 13.87 & 9.13 & 0.34 & 0.31 & 0.28 & 12.85 & 8.78  \\
& 4 &\checkmark&\checkmark & & 0.49&	0.43&	0.35	&13.53&	7.48&	0.46&	0.40&	0.35&	14.06&	7.94  \\
\bottomrule
    \end{tabular}
}
\end{table*}

\section{Experiments and Results} 
\label{chap:robovln,sec:exp}

\textbf{Simulation and Dataset.}
We use Habitat simulator~\cite{habitat19iccv} to perform our experiments. Our dataset, Robo-VLN, is built upon Matterport3D dataset~\cite{Matterport3D}, which is a collection of 90 environments captured through around 10k high-definition RGB-D panoramas.
Robo-VLN provides 3177 trajectories, and each trajectory is associated with 3 instructions annotated by humans ported over from the R2R Dataset~\cite{mattersim}. Overall, the dataset comprises 9533 expert instruction-trajectory pairs with an average trajectory length of 326 steps compared to 55.8 in VLN-CE~\cite{krantz2020navgraph} and 5 in R2R~\cite{mattersim}. The corresponding dataset is divided into train, validation seen and validation unseen splits.

\textbf{Evaluation Metrics.} We evaluate our experiments on the following key standard metrics described by Anderson et al.~\cite{DBLP:journals/corr/abs-1807-06757} and Gabriel et al.~\cite{49206}: Success rate ($\textbf{SR}$), Success weighted by path length ($\textbf{SPL}$), Normalized Dynamic Time Warping ($\textbf{NDTW}$), Trajectory Length ($\textbf{TL}$) and Navigation Error ($\textbf{NE}$). We use SPL and NDTW as the primary metrics for comparison. Both of these metrics measure the deviation from ground-truth trajectories; SPL places more emphasis on reaching the goal location, whereas NDTW emphasizes on following the complete ground-truth path. 

\textbf{Implementation Details.}
We use pre-trained ResNet-50 on ImageNet~\cite{7780459} and pre-trained ConvNet on a large scale point-goal navigation task, DDPPO~\cite{wijmans2020ddppo} to extract spatial features for images and depth modalities successively.
For transformer module, we use a hidden size ($H = 256$), number of Transformer heads ($n_{h}=4$), and the size of feed-forward layer $(FF=1024)$. We found that truncated backpropagation through time~\cite{sutskever2013training} was invaluable to train longer sequence recurrent networks in our case. We used a truncation length of 100 to train attention decoders in both policies. We trained the network for 20 epochs and performed early stopping based on the performance of the model on the validation seen dataset.

\begin{figure}[!b]
\centering
\includegraphics[width=0.95\columnwidth]{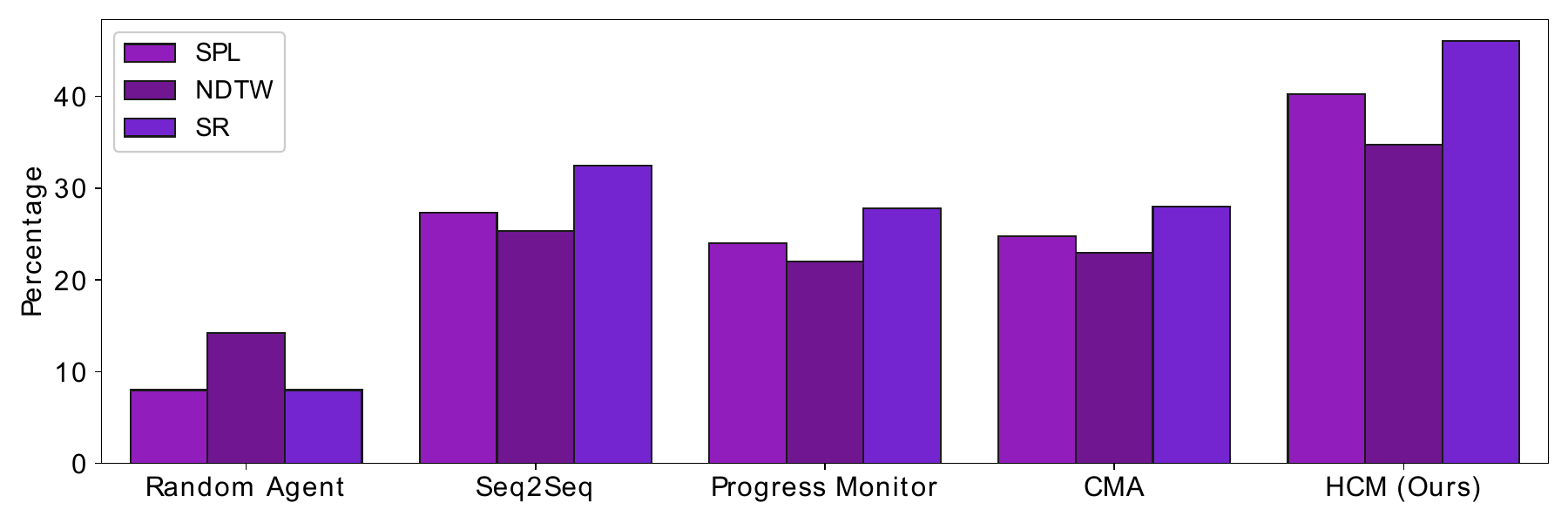}
\centering
  \caption{\textbf{Comparison with strong flat baselines}: Our proposed hierarchical method in comparison with strong flat baselines evaluated on the validation unseen dataset. Our approach shows superior performance and better generalization in unseen settings.
  }
  \label{comparison}
\end{figure}

\begin{figure*}[htp]
\centering
\includegraphics[width=1.0\textwidth]{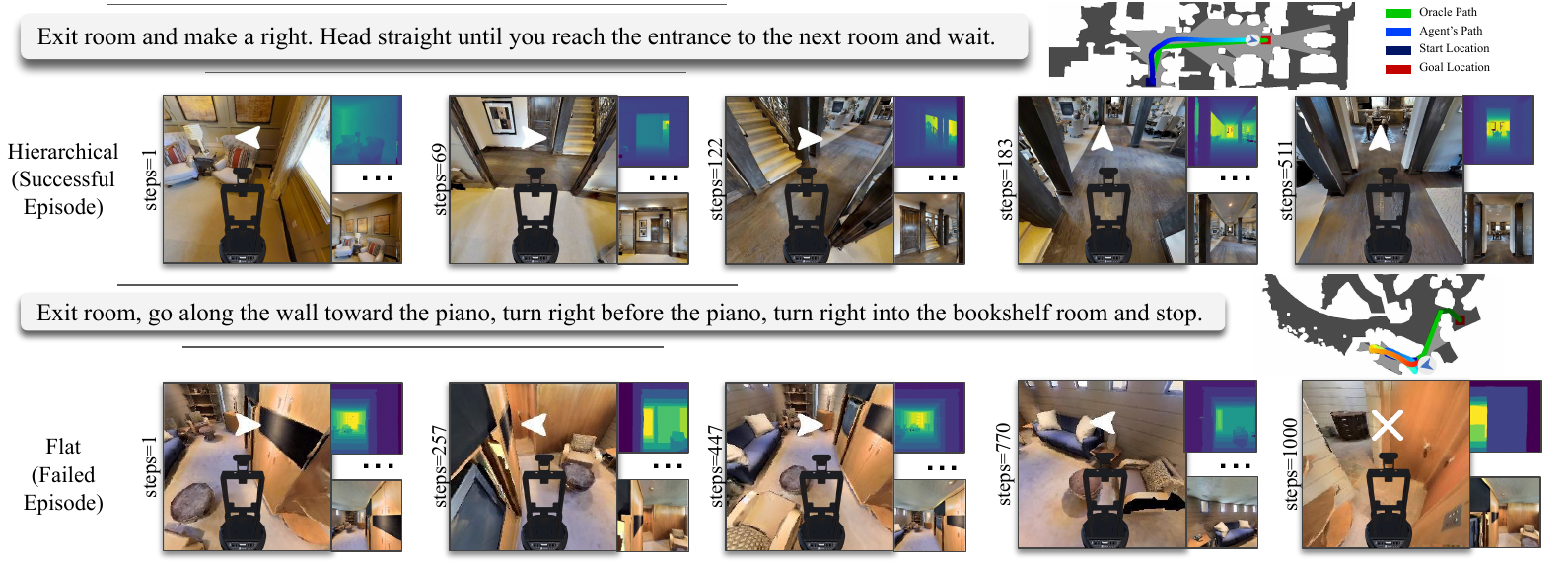}
\captionof{figure}{
\textbf{Qualitative Comparison:} Inference performance of hierarchical and flat model in unseen environments within Robo-VLN. The hierarchical model successfully predicts low-level velocity commands to reach a goal location whereas flat model bumps into obstacles.
}
\label{qualitative_robovln}
\end{figure*}

\section{Experiments \& Results} \label{experiments}
\textbf{Flat Baselines.}\label{baselines}
We introduce a suite of flat\footnote{Flat as in there is no explicit hierarchical design for agent's decision making of high- or low-level actions.} baselines that are similar to the ones used in VLN-CE~\cite{krantz2020navgraph}:
(1) \textbf{Sequence-to-Sequence (Seq2Seq)}: an encoder-decoder architecture trained using teacher-forcing~\cite{mattersim},
(2) \textbf{Progress Monitor (PM)}: an agent based on the Seq2Seq model but with an auxiliary loss for progress monitoring, conceptually similar to \cite{ma2019selfmonitoring}, and 
(3) \textbf{Cross-Modal Attention (CMA)}: an cross-modality attention based agent that is conceptually similar to RCM~\cite{wang2019reinforced}.
We adapt these baselines into our Robo-VLN task but with a single change: the output layers now predict linear and angular velocities as well as the stop action, as opposed to the four actions (forward, turn-left, turn-right, and stop) used in VLN-CE.
Note that baselines are without DAgger~\cite{ross2011reduction} and data augmentation from~\cite{tan2019learning}.

\textbf{Comparison with Flat Baselines.}
The results of our proposed HCM against baselines are summarized in Table~\ref{comparison_table_robovln}. As shown in Table~\ref{comparison_table_robovln} and Table~\ref{comparison}, 
our proposed approach, which uses a hierarchical structure to tackle the long-horizon Robo-VLN problem, consistently outperforms the strong baseline models.  
Specifically, our HCM agent shows superior validation unseen performance by achieving a 40\% SPL and 46\% SR; hence demonstrating an absolute 13\% improvement in SR and 10\% improvement in SPL over the best performing baseline on the validation unseen environments.  

\textbf{Ablation Study.}
In our ablation experiments, we empirically validate the significance of different design choices and modules in our proposed HCM agent. Our results are summarized in Table~\ref{ablation}. First, we ablate \textit{vision} (RGB and Depth) in our model. Our results show that an agent without vision performs as good as a random agent (\textit{i.e.,} 0.07 SPL, 0.07SR). It shows the effectiveness of vision for end-to-end trainable agents in photo-realistic simulations. Second, we consider an architecture with early RGB and Depth fusion before cross attention with language. Our results show that separately aligning RGB and Depth with instructions performs much better than attending to the instructions corresponding to a fused RGB-D representation. We further ablate \textit{hierarchy} to show the importance of hierarchy in our architecture. Our results are summarized as follows.

\textbf{Is the source of improvement from \textit{hierarchy}?}
Our method relies on decomposing the complex task into layered decision-making; the top level predicts a sub-goal whereas the bottom level predicts low-level velocity commands.
To confirm that hierarchy is indeed the source of improvement, we devise an experiment, in which we \textit{flattened} the hierarchical model and provide auxiliary sub-goal supervision to the flattened model in addition to the low-level supervisions.
This model effectively reduced to Seq2Seq baseline model but with high-level action supervision. 
The results are reported in Table~\ref{ablation} (\#2 vs \#4). 
We show that, despite using the same levels of supervisions, the flattened hierarchical model under-performs the hierarchical approach, \textit{e.g.,} 40\% vs 46\% in SR and 34\% vs 40\% in SPL. 
This comparison demonstrates that decoupling reasoning and imitation indeed plays a pivotal role in learning effective individual policies.

\textbf{Qualitative Comparison.}
We qualitatively analyze the performance of hierarchical and flat agents in Robo-VLN. As shown in\ref{qualitative_robovln}, the hierarchical agent (top example) successfully predicts low-level velocity commands while reaching a desired goal location described by the instruction. The agent takes significantly more steps than discrete VLN settings (511 steps) to reach the goal location; hence showing the effectiveness of hierarchical agents to solve long horizon cross-modal trajectory following problem. The flat agent (bottom figure) fails to follow the trajectory and drives into obstacles multiple times. The episode ends after the agent is unsuccessful in reaching the goal at 1000 steps.

\section{Summary} 
\label{chap:robovln,sec:conclusion}

Despite the recent progress, existing VLN environments impose certain unrealistic assumptions such as perfect localization, known topology, and deterministic navigation in the absence of any obstacles. In this section, we first propose the Robo-VLN setting that lifts off the unrealistic assumption of navigation graph and discrete action space and provides a suite of strong baselines inspired by the recent works in discrete VLN setting. We then take the next step to propose a Hierarchical Cross-Modal (HCM) agent that tackles the challenging long-horizon issue in Robo-VLN via a hierarchical model design. Our proposed HCM agent, with trained high- and low-level policies, achieves significant performance improvement against the strong baselines. We believe that our new Robo-VLN setting and strong benchmarks would help build a stronger suite of autonomous agents. Specifically, we utilized hierarchy and modularity as strong prior to disentangling agent-centric deep learning systems that predict actions in an end-to-end manner. We showed that disentangling these systems using their action-space prediction can improve their generalizability to unseen environments. 
    \chapter{SASRA: Semantically-aware Spatio-temporal
Reasoning Agent for Vision-and-Language
Navigation in Continuous Environments}

In this second chapter, my focus shifts to the problem of~\textbf{vision and language navigation within continuous environments}. The primary objective is to incorporate semantic-map priors into agent-centric deep learning systems. The architecture studied in this chapter aims to enhance~\textit{sim2sim generalizability} by seamlessly ~\textit{integrating semantic maps and language modalities} within an end-to-end deep-learning framework for action prediction. This builds upon the prior discussion, emphasizing the importance of prior knowledge integration in our exploration of VLN navigation settings.

\label{chap:sasra}
\section{Introduction}
\label{chap:sasra,sec:intro}

\begin{figure}[h]
\centering
\includegraphics[width=1.0\columnwidth]{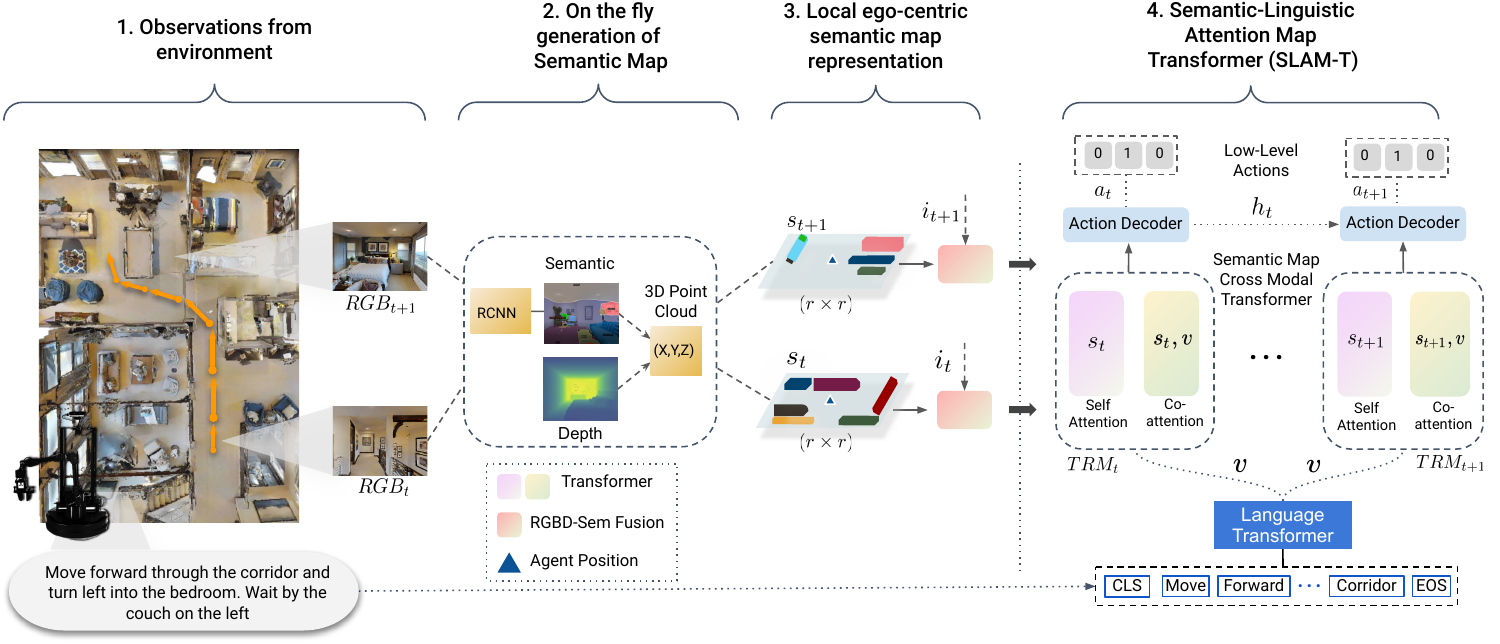}
\centering
  \caption{
\textbf{Overview:} Vision and Language Navigation task and our proposed~\gls{sasra} agent. Our main novelty lies in employing a hybrid transformer-recurrence model for VLN by utilizing a cross-modal Semantic-Linguistic Attention Map Transformer (SLAM-T). At each time-step $t$, the agent generates a local semantic map and consistently reasons with the environment in both spatial (cross-modal attention between semantic map and language) and temporal domains to decode the low-level actions ($a_{t}$) at each time-step.}
  \label{overview_sasra}
\end{figure}

A long-held vision for robotics and artificial intelligence is to create robots that can reason about their environments and perform a task following natural language commands. Recent advances in learning-based approaches~\cite{embodiedqa,johnson2017clevr, Matterport3D} have shown promise in making progress towards solving a complex task referred to as Vision-and-Language Navigation (VLN)~\cite{mattersim, krantz2020navgraph}. VLN task requires an agent to navigate in a 3D environment, following instructions, without prior access to a map. 

Most of the prior VLN systems are trained assuming a simpler navigation-graph~\cite{10.5555/2969033.2969173, Zhuetal,46942}, but the performance of these agents is adversely affected in more complex scenarios such as unconstrained navigation or long-horizon VLN problems~\cite{krantz2020navgraph, ai2thor, irshad2021hierarchical}. In this work, we consider a continuous 3D environment for VLN following~\cite{krantz2020navgraph}. The task is referred to as Vision-and-Language Navigation in Continuous Environments~(VLN-CE). VLN-CE formulation imitates the challenges of real-world much better, and robotic agents trained in such a simulated setting are likely to transfer well. In this context, the term "continuous" specifically denotes that simulated agents do not abruptly teleport between viewpoints in a sparse graph, where nodes signify navigability. Rather, these agents have real-world challenges like imperfect topology, localization uncertainties, and the unpredictable nature of stochastic navigation. These agents also take significantly longer steps to reach the goal than discrete navigation graph-based agents~\cite{krantz2020navgraph}

Prior end-to-end learning-based VLN systems mostly rely on raw visual observations~\cite{ma2019theregretful, Wang2019ReinforcedCM, DBLP:conf/nips/FriedHCRAMBSKD18} and memory components such as LSTM~\cite{Matterport3D}. Structured memory such as local occupancy maps~\cite{chaplot2020learning, Chen_2019_CVPR, gupta2017cognitive, Savinov2019_EC} and scene semantics~\cite{chaplot2020semantic, DBLP:journals/corr/abs-2007-09841, wang2021structured, maast} have shown to previously improve the performance of learning-based navigation. However, their relation to Vision-and-Language navigation~\cite{krantz2020navgraph} has not been effectively investigated~\cite{chen2020topological}. We believe structured scene memory is crucial to operating well in complex long-horizon VLN settings (as we also showed for SASRA experiments in Section~\ref{chap:sasra,sec:exp}) because a longer horizon task demands elaborate spatial reasoning to more effectively learn a VLN Policy.

Motivated by the above, we present a hybrid transformer-recurrence model for VLN in continuous 3D environments, which we refer to as \textbf{S}emantically-\textbf{a}ware \textbf{S}patio-temporal  \textbf{R}easoning  \textbf{A}gent (\textbf{SASRA}). Our approach combines the classical semantic mapping technique with a learning-based method and equips the agent with the following key abilities: 1) Establish a temporal memory by generating local top-down semantic maps from the first-person visual observations (RGB and Depth). 2) Create a spatial relationship between contrasting vision-based mapping and linguistic modalities. Hence, identify the relevant areas in the generated semantic map corresponding to the provided instructions and 3) Preserve the relevant temporal states information through time by combining a transformer-inspired~\cite{NIPS2017_7181} attentive model with a recurrence module; hence make the agent conscious of its environment (spatially scene-aware) as well as its history (temporally conscious).
\noindent The main contributions of the work are summarized below: 
\begin{itemize}

\item To the best of our knowledge, we present the first work on the effective integration of semantic mapping and language in an end-to-end learning methodology for Vision-and-Language Navigation in continuous environments.
\item We introduce a 2-D cross-modal \textbf{S}emantic-\textbf{L}inguistic \textbf{A}ttention \textbf{M}ap \textbf{T}ransformer (SLAM-T) mechanism, based on Transformers, to extract cross-modal map and language features while focusing on the most relevant areas of structured, local top-down spatial information.

\item We show a clear improvement over the state-of-the-art methods and baselines in the Vision-and-Language navigation task in VLN-CE dataset, with over 20\% improvement in success rate in unseen environments.

\end{itemize}
\section{Related Works}
\label{chap:sasra,sec:related_works}

\textbf{Vision-and-Language Navigation:} Prior VLN systems have focused mainly on utilizing raw visual observations with a sequence-to-sequence based approach~\cite{mattersim, deng2020evolving, hao2020towards, ke2019tactical, ma2019theregretful}. Additional progress in VLN has been guided through cross-modal matching techniques~\cite{Wang2019ReinforcedCM, DBLP:journals/corr/abs-1905-13358}, auxiliary losses~\cite{ma2019selfmonitoring, Zhuetal}, data-augmentation~\cite{majumdar2020improving, DBLP:conf/nips/FriedHCRAMBSKD18, tan2019learning} and training-regimes~\cite{krantz2020navgraph}. Our work, in contrast, focuses on incorporating structured memory in the form of scene semantics and its relation to language. Inspired by the recent success of utilizing Transformers for different tasks~\cite{NIPS2017_7181, devlin-etal-2019-bert, DBLP:journals/corr/abs-2005-12872}, our work focuses on end-to-end Transformers for VLN combining visual modalities with semantic mapping and language. 

\textbf{Scene-aware mapping for Navigation:} 
Local map-building during exploration~\cite{chaplot2020neural, gupta2017cognitive,gupta2017unifying, parisotto2017neural,meng2020scaling, chen2020topological} has been an active area of research. Previous works have mainly focused on generating either a local occupancy map~\cite{chaplot2020learning, Chen_2019_CVPR, gupta2017cognitive, Savinov2019_EC} or semantic map~\cite{chaplot2020semantic, DBLP:journals/corr/abs-2007-09841, maast} for better exploration and navigation policies. In contrast, our work
combines classical semantic map generation in an attentive transformer-like architecture~\cite{NIPS2017_7181} to focus on and extract cross-modal features between local semantics maps and input instructions for effective Vision-and-Language Navigation. To the best of our knowledge, this intersection of semantic mapping and language has not been a focus of any prior works.
\section{Method}
\label{chap:sasra,sec:method}

Consider an autonomous agent $\tilde{\mathcal A}$ in an unknown environment. Our goal is to find the most optimal sequence of actions $a_{t} = \pi(x_{t}, \gamma, q)$ which takes the agent from a starting location to a goal location while staying close to the path defined by instructions ($q$). The agent receives visual observations ($x_{t}$) from the environment at each time step ($t$). $\gamma$ denotes the learnable parameters of the policy $\pi$. \\
To learn an effective policy ($\pi$), we propose a hybrid transformer-recurrence model to effectively reason with the environment in both spatial and temporal domains. Our approach, as shown in Figure~\ref{SASRA2}, combines end-to-end transformer blocks with recurrent modules to effectively match the visual cues with the provided instruction. Our methodology comprises three learnable submodules: \textbf{S}emantic-\textbf{L}inguistic \textbf{A}ttention \textbf{M}ap \textbf{T}ransformer (SLAM-T), RGBD-Linguistic Cross-Modal Transformer and Hybrid Action Decoder. The goal of the SLAM-T module is to generate a local semantic map over time and identify the matching between the provided instruction and the semantic map using cross-attention. Hybrid Action Decoder then utilizes the cross-attention features obtained from both RGBD-Linguistic and Semantic Map-Linguistic modules to select an action ($a_{t}$) at each time step. Details of our architecture are summarized as follows:

\begin{figure}[htb]
\begin{center}
\includegraphics[width=0.8\linewidth]{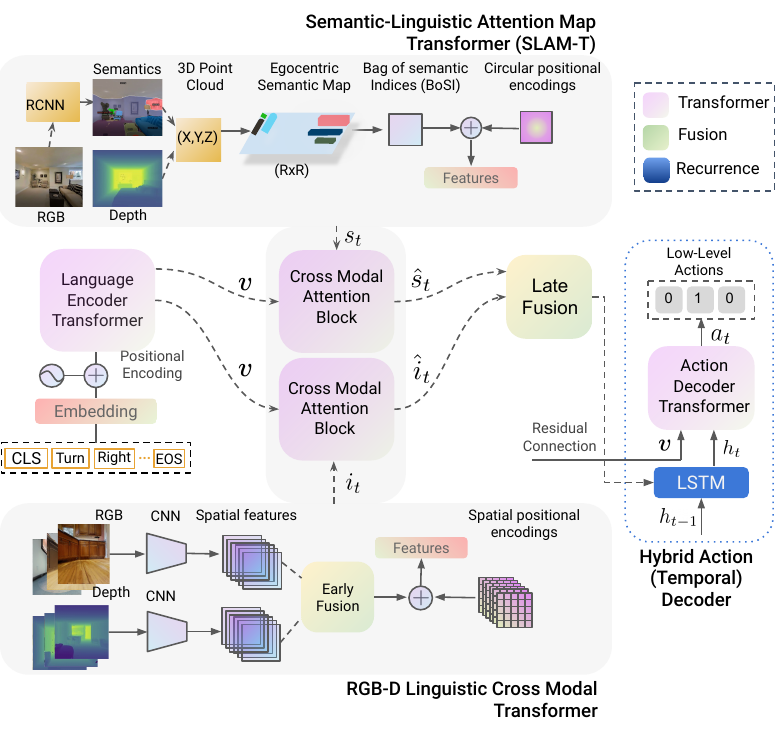}
\captionof{figure}{\textbf{Model Architecture (Detailed model):} Our approach utilises learning-based cross modal attention modules. Semantic-Linguistic Attention Map Transformer (SLAM-T) and RGBD-Linguistic Transformer consistently reason between visual and textual spatial domains. Hybrid Action Decoder captures the temporal dependencies inherent in following a trajectory over time.}
  \label{SASRA2}
\end{center}
\end{figure}

\begin{figure}[htb]
\begin{center}
\includegraphics[width=1.0\textwidth]{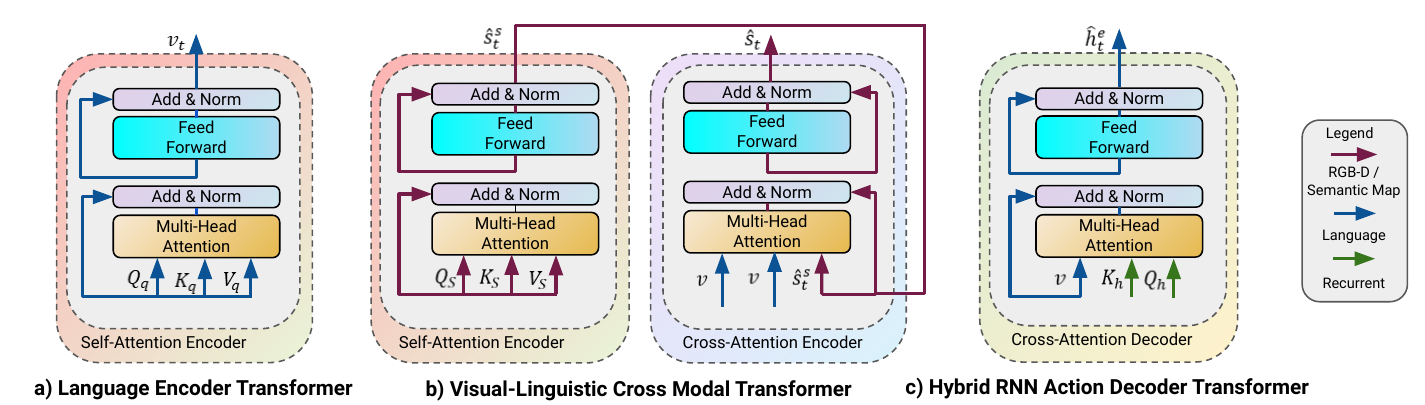}
\captionof{figure}{\textbf{Model Architecture (Individual Blocks):} Our model utilises fully-attentive Transformer blocks at each stage. \textbf{a)} We encode language using a self-attention Transformer. \textbf{b)} Visual-linguistic attention is performed using two-stage Transformer blocks employing both self and cross-attention. \textbf{c)} Action decoder comprises a single cross-modal Transformer module.}
  \label{SASRA3}
\end{center}
\end{figure}

\subsection{Semantic-Linguistic Attention Map Transformer}

To generate a semantic map from visual observations $x_{t} = \{ r_{t}, d_{t}, g_{t} \}$, we employ a three-step approach. Here $r_{t}$, $d_{t}$, $g_{t}$ denote the RGB, Depth, and Semantic sensor readings respectively. 

We first project the pixels in depth observations to a 3D point cloud using camera intrinsic parameters. The agent's current pose estimate is used to find the camera extrinsic parameters, which are used to transform the point cloud to the world frame. The points are projected to a 2D space and stored as either obstacles based on a height threshold or free space. For each point categorized as an obstacle, we store its semantic class, $k$ i.e. the value given by the semantic sensor reading $g_{t}$. In essence, each cell in a discrete $2r \times2r$ map $s_{t}$ is represented as a binary vector $s_{t_{xy}}$ $\in$ $\{0, 1\}$,
where $k = 1$ if a semantic class $k$ is present at that location and 0 otherwise.

To cross-modally attend to the relevant areas of the instructions $q$ corresponding to the generated semantic map, we first encode the instructions $q$ in a fully attentive model as follows: 

\textbf{Language Encoder Transformer:}
Given a natural language instruction comprising $k$ words, its representation is denoted as, $\{ q_{1}, q_{2}, \ldots, q_{k} \}$ where $q_{i}$ is the embedding vector for $i_{th}$ the word in the sentence. We encode the instructions using a transformer module~\cite{NIPS2017_7181} $TRM(q$ + PE$(q))$ to get the instruction encoding vector ($v$) where PE denotes the fixed sinusoidal positional encoding~\cite{NIPS2017_7181} used to encode the relative positions of words appearing in the sentence. The transformer module $TRM$ consists of a stacked multi-head attention block (M.H.A) followed by a position-wise feed-forward block~\cite{NIPS2017_7181} as shown in Figure~\ref{SASRA3}(a) and Equation~\ref{transformer_eq}. Individual transformer blocks are computed as follows:
\begin{equation}\begin{aligned}
\text { M.H.A }(\boldsymbol{Q}, \boldsymbol{K}, \boldsymbol{V}) &=\text{concat( $\boldsymbol{h}_{1}$}, \ldots, \text{ $\boldsymbol{h}_{k}$} ) \boldsymbol{W}^{h} \\
\text {where $\boldsymbol{h}_{i}$ }&=\mathit{A}\left(\boldsymbol{Q} W_{i}^{Q}, \boldsymbol{K} W_{i}^{K}, \boldsymbol{V} W_{i}^{V}\right)
\\
\mathit{A}(\boldsymbol{Q}, \boldsymbol{K}, \boldsymbol{V})&=\operatorname{softmax}\left(\frac{\boldsymbol{Q} K^{T}}{\sqrt{d_{k}}}\right) \boldsymbol{V}
\label{transformer_eq}
\end{aligned}\end{equation}

A basic attention block, as shown in Figure~\ref{SASRA3}, uses a linear projection of the input to find queries ($Q_{q} = q$ + PE$(q) $) and keys ($K_{q} = q$ + PE$(q)$). Based on the similarity between queries ($Q_{q}$) and keys ($K_{q}$), a weighted sum over the values ($V = q$ + PE$(q)$) is computed as the output attention ($\mathit{A})$. Note that the language encoder transformer uses the same language sequences to project the tuple ($Q_{q},K_{q},V_{q}$) and hence the attention output is referred to as \textit{self-attention}. $W_{i}^{Q}, W_{i}^{K}, W_{i}^{V}$ and $W^{h}$ are parameters to be learned. 

\textbf{Semantic-Linguistic Co-Attention Transformer:} The output representation of semantic map ($s_{t}$) is a $2r \times2r$ map centered on the agent. We refer to this map representation ($s_{t}$) as a Bag of Semantic Indices. Each cell in the map ($s_{t_{xy}}$) carries important structural information about the map. To encode the map features in a fully attentive transformer model~(\ref{transformer_eq}), we construct a $2r \times2r$ relative positional encoding matrix ($\mathcal P$) using a Gaussian kernel, centered on the agent, with a scale factor of $\frac{r}{2}$. Gaussian positional encoding ($GPE$) is computed after projecting the Gaussian kernel into an embedding space as follows: 
\begin{equation}\begin{aligned}
\text{GPE}&=\it{embed_{2D}}(F(x,y)) \\
\boldsymbol{F}\left(\boldsymbol{x}, \boldsymbol{y}\right)&=\frac{b^{2}}{\sqrt{2 \pi w^{2}}} \exp\left(-\frac{\left(\boldsymbol{x}-\boldsymbol{y}\right)^{2}}{2 w^{2}}\right)
\label{GPE}
\end{aligned}\end{equation}

where $w$ is the input scale parameter (equivalent to the standard deviation of the Gaussian) and $b$ is the output scale parameter.

We encode the semantic map representation ($s_{t}$) in a 2-step process. First, we compute \textit{self-attention} over the map ($s_{t}$) by employing Equation~\ref{transformer_eq}. We use the sum of Bag of Semantic Indices and Gaussian positional encoding ($\mathcal P$) as Query, Key, and Value ($Q_{s} = K_{s} = V_{s} = s_{t} + GPE(s_{t})$) in Equation~\ref{transformer_eq}. This transformer feature representation $\hat{s}_{t}^{s}$ $\in$ $\mathbb{R}^{2}$ using \textit{self-attention}, as shown in Figure~\ref{SASRA3}(b), is computed as follows:
\begin{equation}
\hat{\boldsymbol{s}}_{t}^{s} = TRM(\boldsymbol{s}_{t} + GPE(\boldsymbol{s}_{t}))    
\end{equation}

$\hat{s}_{t}^{s}$ is $(r\times r)\times H$ matrix, where  $(r\times r)$ is the $2D$ semantic map collapsed into one dimension and $H$ is the hidden size. Typical values we utilized are $r=40$ and $H=512$.

Second, we perform \textit{cross-attention} between computed language features ($v$) and semantic map self-attention features $\hat{s}_{t}^{s}$. We employ~ (\ref{transformer_eq}) by using $\hat{s}_{t}^{s} + GPE(\hat{s}_{t}^{s})$ as Query and instruction encoding ($v$) as Key and Value to get final cross-modal attention features $\hat{s}_{t}$ $\in$ $\mathbb{R}^{(r\times r)\times H}$ as follows:

\begin{equation}
\hat{\boldsymbol{s}}_{t} = TRM(\hat{\boldsymbol{s}}_{t}^{s} +GPE(\hat{\boldsymbol{s}}_{t}^{s}), \boldsymbol{v})   
\end{equation}

\subsection{RGBD-Linguistic Cross Attention Transformer}

Given an observed image ($r_{t}$ $\in$ $\mathbb{R}^{H_{o}\times W_{o}\times 3 }$), we generate a low-resolution spatial representation $f^{r}_{t}$ $\in$ $\mathbb{R}^{H_{r}\times W_{r}\times C_{r}}$  by using a classical CNN. Typical values we utilized are $H_{r}=W_{r} = 7$, $C_{r}=2048$. Furthermore, we process depth modality ($d_{t}$ $\in$ $\mathbb{R}^{H_{o}\times W_{o}}$) using a CNN pre-trained on a large-scale visual navigation task i.e. DDPPO~\cite{wijmans2020ddppo} to generate a spatial representation $f^{d}_{t}$ $\in$ $\mathbb{R}^{H_{d}\times W_{d}\times C_{d}}$. Typical values used in DDPPO training are $H_{d}=W_{d} = 4$, $C_{d}=128$.

\textbf{Early Fusion:} We employ an early fusion of image spatial features $f^{r}_{t}$ and depth spatial features $f^{d}_{t}$. Specifically, we reduce the channel dimension of the spatial image features $f^{r}_{t}$ using a $1\times 1$ convolution and perform average adaptive pooling to reduce the dimensions from $H_{r}\times W_{r}$ to $H_{d}\times W_{d}$. We concatenate the final outputs $f^{r}_{t}$ and $f^{d}_{t}$ along the channel dimension to get a fused RGB-D representation ($i_{t}$).

\textbf{Cross Modal Attention:} We utilize Equation~\ref{transformer_eq} to perform cross modal attention between linguistic features ($v$) and fused RGB-D features ($i_{t}$). We perform this in two steps. First, we use the sum of $i_{t}$ and Learned Positional Encoding ($LPE(i_{t})$) as Query, Key and Value ($Q_{s} = K_{s} = V_{s} = i_{t} + LPE(i_{t})$) in Equation~\ref{transformer_eq} to compute \textit{self-attention} features $\hat{i}_{t}^{s}$ where we learn a spatial $2D$ positional encoding $LPE$ as opposed to utilizing a fixed positional encoding~\cite{NIPS2017_7181}. Second, we perform \textit{cross-attention} by using $i_{t} + LPE(i_{t})$ as Query and instruction encoding ($v$) as Key and Value in Equation~\ref{transformer_eq} to get cross-modal attention features $\hat{i}_{t}$ $\in$ $\mathbb{R}^{(H_{d}\times H_{d})\times H}$.

\textbf{Late Fusion:} We perform a late fusion of cross-modal semantic map features $\hat{s}_{t}$ and cross-modal RGB-D features $\hat{i}_{t}$. Specifically, we utilize average pooling across the spatial dimensions of $\hat{s}_{t}$ before concatenating along the hidden size dimension $H$ to get visual-linguistic embedding ($\hat{V}_{t}^{e}$)

\subsection{Hybrid Action Decoder}

For action selection, the agent preserves a temporal memory of the previous observed visual-linguistic states ($\hat{V}_{t}^{e}$) and previous actions($a_{t-1}$). We utilize a recurrent neural network $\text{RNN}$ to preserve this temporal information across time. 
\begin{equation}\boldsymbol{h}_{t}=\operatorname{RNN}\left(\left[\hat{\boldsymbol{V}}_{t}^{e}, \mathbf{a}_{t-1}\right]\right)\label{recurrence_eq}\end{equation}

\textbf{Temporal Cross-Modal Transformer:} The agent selects an action $a_{t}$ by keeping track of the completed parts of the instructions ($q$) and observed visual-linguistic states ($\hat{V}_{t}^{e}$). We preserve this temporal information regarding instruction completeness using an action decoder transformer ( Figure~\ref{SASRA3}c), which performs \textit{cross-attention} between hidden states of the recurrent network (Equation \ref{recurrence}) and instruction encoding $v$. We compute \textit{cross-attention} $TRM(\boldsymbol{h_{t}} +FPE(\boldsymbol{h_{t}}), \boldsymbol{v})$ utilizing recurrent hidden states $h_{t}$ as Query and Key ($Q_{h} = K_{h} = h_{t} + FPE(h_{t})$) and instruction encoding ($v$) as Value in Equation \ref{transformer}. Finally, we compute the probability ($p_{t}$) of selecting the optimal action ($a_{t}$) at each time-step employing a feed-forward network followed by $softmax$ as follows:
\begin{equation} \begin{aligned}
\hat{\boldsymbol{h}}_{t}^{e} &= TRM(\boldsymbol{h}_{t} +FPE(\boldsymbol{h}_{t}), \boldsymbol{v}) \\
p_{t} &= \text{softmax}(\mathrm{M}(\hat{\boldsymbol{h}}_{t}^{e}))
\end{aligned}\end{equation}
where $\mathrm{M}(.)$ is a multi-layer perceptron and $TRM(.)$ is the Transformer module. \\
Conceptually, our Hybrid Action Decoder performs attention both on temporal sequences (i.e. along time) as well as language sequences. We empirically show (in Section~\ref{ablationsection}) that our hybrid action decoder is indeed crucial for long-horizon and complex tasks such as VLN-CE and this design choice leads to better results than a single-temporal module based on either LSTMs or Transformers alone.   

\subsection{Training}
We train our model using a cross entropy loss, $\mathcal{L}_{loss} = -\sum_{t}^{} y_{t}\log(p_{t})$, computed using ground truth action ($y_{t}$) and log of the predicted action probability ($p_{t}$) at step $t$. 

where $C$ is the number of possible low-level actions.

\begin{equation} \begin{aligned}
\mathcal{L}_{\text {loss}} &= \alpha \mathcal{L}_{CSE} + (1-\alpha) \mathcal{L}_{MSE} \\
\mathcal{L}_{CSE} &= -\sum_{t=1}^{T} y_{a}\log \left(p_{a}\right)\\
\mathcal{L}_{MSE} &= - \sum_{t=1}^{T}\left(d^{*} -p^{d}\right)^{2}\end{aligned}\end{equation}
where $\alpha$ $\in$ $[0,1]$ is the weighting factor and $p^{d}$~\cite{ma2019selfmonitoring} is calculated from the final hidden state output $\hat{\boldsymbol{h}}_{c}^{e}$ of our policy as follows:
\begin{equation}p^{d}=\tanh \left(\mathrm{g}\left( \hat{\boldsymbol{h}}_{c}^{e}\right)\right)\end{equation}

\textbf{Training Regimes:} We explore two popular imitation learning approaches for training, i.e., 1) Teacher-Forcing \cite{williams1989learning}, 2) DAGGER \cite{ross2011reduction}. Teacher-forcing is the most popular training method for RNNs, which minimizes maximum-likelihood loss using ground-truth samples \cite{williams1989learning, goyal2016professor}. However, the teacher-forcing strategy suffers from exposure bias problems due to differences in training and inference \cite{ross2011reduction, krantz2020navgraph}. One popular approach to minimize the exposure bias problem in the teacher-forcing method is DAGGER \cite{ross2011reduction}. In Dagger, the training data set for iteration $N$ is collected with probability $\beta^N$ (where, $0 < \beta < 1$) for ground-truth action and current policy action for remaining. The training for iteration $N$ is done using the aggregated dataset collected till iteration $N$.
\section{Experiments and Results}
\label{chap:sasra,sec:exp}

\renewcommand{\arraystretch}{1.3}
\begin{table*}[!t]
\centering
\caption{\textbf{Quantitative Results on VLN-CE Dataset Comparing Baselines}: Random and Forward-only are non-learning baselines reported to provide context to the performance of learning based baselines. Seq2Seq and CMA are state-of-the-art models on VLN-CE trained following \cite{krantz2020navgraph}. We implemented Seq2Seq-SM and CMA-SM baselines by providing additional semantic map input to Seq2Seq and CMA models. The best results with DAGGER are in \textbf{bold} and with teacher-forcing are \underline{underlined}. Following \cite{Wang2019ReinforcedCM, krantz2020navgraph, DBLP:journals/corr/abs-1807-06757}, \textbf{SPL} and \textbf{SR} are considered as primary evaluation metrics.
Note that, as mentioned by the authors of VLN-CE~\cite{krantz2020navgraph}, different hardware + Habitat builds lead to different results\textsuperscript{$\ddagger$}. Hence, we re-trained all the models using the same build and following the details presented in Sec.~\ref{chap:sasra,sec:exp} for a fair comparison. } \label{tab:quantitative_sasra}
\resizebox{1.0\textwidth}{!}{
\begin{tabular}{cccc@{\hskip 0.1in}cccc@{\hskip 0.2in}ccccc}
\toprule
 & \multicolumn{2}{c}{\textbf{Training Regime}} & \multicolumn{5}{c}{\textbf{Validation Seen}} &\multicolumn{5}{c}{\textbf{Validation Unseen}} \\ 
\cmidrule(r{0.1in}){2-3} \cmidrule(r{0.2in}){4-8} \cmidrule{9-13}
Model & {Teacher-Forcing~\cite{williams1989learning}} &{DAGGER~\cite{ross2011reduction}} &  \textbf{SR}~$\uparrow$ & \textbf{SPL}~$\uparrow$ & \textbf{NDTW}~$\uparrow$ & \textbf{TL}~$\downarrow$ & \textbf{NE}~$\downarrow$ & \textbf{SR}~$\uparrow$ & \textbf{SPL} & \textbf{NDTW}~$\uparrow$ & \textbf{TL}~$\downarrow$ & \textbf{NE}~$\downarrow$ \\
\midrule
Random & & &  0.02 & 0.02 & 0.28 & 3.54 & 10.20 & 0.03& 0.02 & 0.30 & 3.74 & 9.51 \\
Forward-Only & & &  0.04 & 0.04 & 0.33 & 3.83 & 9.56 & 0.03& 0.02 & 0.30 & 3.71 & 10.34 \\
\midrule
\multirow{2}{1.8cm}{\centering Seq2Seq} &\checkmark & &  0.25 & 0.24 & 0.47 & 8.15 & 8.49 & 0.17& 0.16 & 0.43 & 7.55 & 8.91  \\
 & &\checkmark & 0.32 & 0.31 & 0.51 & 8.49 & 7.61 & 0.19 &  0.17 & 0.43 & 7.78 & 8.69  \\
 \midrule
\multirow{2}{1.9cm}{\centering Seq2Seq-SM} &\checkmark& & 0.26& \underline{0.25} & \underline{0.48} & \underline{7.92} & 8.13 &  0.16 &  0.15 &  0.41 &  7.54 &  9.14 \\
 & & \checkmark & 0.30 &  0.28 & 0.49 &  8.58 &  7.98 & 0.19 & 0.17 & 0.43 & 7.69 & 8.98 \\
 \midrule
\multirow{2}{1.8cm}{\centering CMA} &\checkmark&  & 0.24 &  0.23 & 0.47 & 7.94 & 8.69 & 0.19 & 0.18 & \underline{0.44} & \underline{6.96} & \underline{8.52}  \\
 && \checkmark & 0.28 & 0.26 & 0.49 & \textbf{8.43} & 8.12 & 0.20 & 0.18 & 0.44 & \textbf{7.67} & 8.72 \\
\midrule
\multirow{2}{1.8cm}{\centering CMA-SM} &\checkmark& & \underline{0.27} & \underline{0.25} & \underline{0.48} & 8.44 &  8.25 & 0.18	& 0.16 & \underline{0.44} &	7.52 & 8.73 \\
 && \checkmark & 0.29 & 0.28 & 0.49 & 8.49 & 7.79 & 0.19 & 0.18 & 0.43 & 7.70 & 8.92 \\
\midrule
\multirow{2}{1.8cm}{\centering \textbf{SASRA(Ours)~\cite{irshad2022sasra}}} &\checkmark& & \underline{0.27} &	\underline{0.25} &	0.46	&8.84 &	8.48&	\underline{0.22}&	\underline{0.21}&	\underline{0.44} &	8.05&	8.75  \\
 &&\checkmark & \textbf{0.36} &	\textbf{0.34}&	\textbf{0.53}	&8.89 &	\textbf{7.17} & \textbf{0.24}	& \textbf{0.22}	& \textbf{0.47}	& 7.89	& \textbf{8.32} \\
\bottomrule
\end{tabular}}
\end{table*}
\renewcommand{\arraystretch}{1.0}
\begin{table}[htb]
\scriptsize
\centering
\caption{\textbf{Ablation Study} to investigate impact of different components  (i.e., Semantic Map (SM), Hybrid Action Decoder (HAD) and DAGGER (DA)) on VLN-CE validation-seen set.}
\label{tab:ablation_sasra}
\resizebox{1.0\textwidth}{!}{%
\begin{tabular}{c@{\hskip 0.1in}cccccccc} \\
\toprule
& & & & \multicolumn{5}{c}{\textbf{Metrics}} \\ 
\cmidrule(){5-9}
 {\textbf{\#}} & \text{SM} &\text{HAD} &\text{DA} & \textbf{SR}~$\uparrow$ & \textbf{SPL}~$\uparrow$ & \textbf{NDTW}~$\uparrow$ & \textbf{TL}~$\downarrow$  & \textbf{NE}~$\downarrow$  \\
\midrule
  1 & \checkmark & & & 0.22 & 0.20 & 0.41 & 11.07 & 9.09 \\
  2 && \checkmark &  & 0.23 & 0.22 & 0.42 & 8.30 & 8.78  \\
 3 &\checkmark& \checkmark&  & 0.27 & 0.25 & 0.46 & 8.84 & 8.48 \\
 4 &\checkmark&\checkmark &\checkmark & 0.36 & 0.34 & 0.54 & 9.19 & 7.03 \\
\bottomrule
\end{tabular}}
\end{table}
\textbf{Simulator and Dataset:}
We use Habitat simulator~\cite{habitat19iccv} to perform experiments. We use VLN-CE dataset presented by Krantz et al.~\cite{krantz2020navgraph} to evaluate VLN in continuous environments. VLN-CE is built upon Matterport3D dataset~\cite{Matterport3D} which is a collection of $90$ environments captured around 10k RGB-D panoramas. VLN-CE provides $4475$ trajectories followed by an agent inside Matterport3D simulation environment available in Habitat Simulator. Each trajectory is associated with 3 human annotated instructions. The dataset is divided into train, validation seen, and validation unseen splits. 

VLN-CE dataset provides the following low-level actions for each instruction-trajectory pair for navigation inside Habitat Simulator: \texttt{move forward} ($0.25m$), \texttt{turn-left} or \texttt{turn-right} ($15 deg.$) and \texttt{stop}. The trajectories in VLN-CE span $55$ steps on average, making the problem realistic and challenging to solve than previously proposed navigation-graph based R2R dataset~\cite{mattersim} which corresponds to an average trajectory length of $4-6$. Note that our experiments do not include R2R dataset~\cite{Matterport3D} since we focus on a richer and more challenging setting of VLN-CE.

\textbf{Evaluation Metrics:} We use standard visual navigation metrics to evaluate our agents as described in previous works~\cite{DBLP:journals/corr/abs-1807-06757, 49206, mattersim}: Success rate ($\textbf{SR}$), Success weighted by path length ($\textbf{SPL}$), Normalized Dynamic Time Warping ($\textbf{NDTW}$), Trajectory Length ($\textbf{TL}$) and Navigation Error ($\textbf{NE}$). Following prior works \cite{Wang2019ReinforcedCM, krantz2020navgraph, DBLP:journals/corr/abs-1807-06757}, we use $\textbf{SPL}$ and $\textbf{SR}$ as the primary metrics to evaluate the performance of our agent. Please see~\cite{DBLP:journals/corr/abs-1807-06757, 49206, mattersim} for details of the metrics.

\textbf{Implementation Details:} 
For Teacher-Forcing, we train the models for at most $20$ epochs. We perform early stopping based on the performance of the model on the validation seen dataset. For DAGGER, we start from the model pre-trained with teacher-forcing and then fine-tune using DAGGER. We set the number of dataset aggregation rounds $N$ for DAGGER to be $3$. $\beta$ is set to $0.75$. Hence, for the $Nth$ Dagger iteration, the dataset is collected with probability $0.75^N$ for Oracle policy actions blended with actions from the agent's policy obtained after training $(N-1)th$ iteration with a probability $(1-0.75^N)$. For each DAGGER round, we perform $4$ epochs of teacher-forcing training. All the models are trained using three RTX 2080 Ti GPUs and Habitat-API-0.1.5~\cite{habitat19iccv}.

To avoid loss curve plateauing, we use Adam Optimizer~\cite{kingma2014method} with a Cyclical Learning Rate~\cite{smith2015cyclical} in the range [$2^{-6}, 1^{-4}$]. The cyclic learning rate was varied over a range of epochs. The inflection weighting coefficient for imitation learning is set to $3.2$ following \cite{krantz2020navgraph}. For individual transformer blocks, we use a consistent hidden size ($H = 512$), Number of Transformer Heads ($n_{h}=4$), and size of feed-forward layer ($s_{F.F}=1024$). We utilize Imagenet~\cite{7780459} pre-trained Resnet-50 for RGB and Resnet-50 pre-trained on DDPPO~\cite{wijmans2020ddppo} to compute low-resolution features for visual modalities. 

\renewcommand{\thefootnote}{\fnsymbol{footnote}}
\footnotetext[3]{\tiny \url{https://github.com/jacobkrantz/VLN-CE\#baseline-performance}}

\subsection{Quantitative Comparison with Baselines}
We report the quantitative results in Table~\ref{tab:quantitative_sasra}. We implement several baselines to compare with the proposed model: 
\textbf{1) Sequence-to-Sequence (Seq2Seq):} Seq2Seq agent employs a recurrent policy that takes a representation of instructions and visual observation (RGB and Depth) at each step and then predicts an action (\cite{mattersim, krantz2020navgraph}). \textbf{2) Seq2Seq with Semantics (Seq2Seq-SM):} Seq2Seq model which includes additional visual observation, (i.e., semantic map) to aid learning. \textbf{3) Cross-Modal Matching (CMA):} A sequence-to-sequence model with cross-modal attention and spatial reasoning (\cite{Wang2019ReinforcedCM, krantz2020navgraph}). \textbf{4) CMA with Semantics (CMA-SM):} CMA with additional visual observation, (i.e., semantic map) to aid learning. 

The Seq2Seq and CMA are state-of-the-art baselines in VLN-CE implemented following \cite{krantz2020navgraph}. Seq2Seq-SM and CMA-SM are our implemented baselines, where base Seq2Seq and CMA models are enhanced by providing semantic map self-attention representation as additional input. We report results in Table~\ref{tab:quantitative_sasra} with both teacher-forcing and DAGGER training. 

\underline{State-of-the-art Baselines.} It is evident from Table~\ref{tab:quantitative_sasra} that the proposed SASRA agent achieves significant performance improvement over state-of-the-art VLN-CE baseline models, i.e., Seq2Seq and CMA. With DAGGER training, we find that the improvement in SPL is $+5\%$ absolute ($+29\%$ relative) in val-unseen and $+3\%$ absolute ($+9.6\%$ relative) in val-seen compared to Seq2Seq. With teacher-forcing training, the performance improvement in SPL is $+5\%$ absolute ($+31\%$ relative) in val-unseen and +1\% absolute ($+4\%$ relative) in val-seen compared to Seq2Seq. Compared to CMA agent with DAGGER training, the absolute performance improvement in SPL is $+4\%$ in val-unseen and $+8\%$ in val-seen. With teacher-forcing training, the absolute performance improvement in SPL is $+3\%$ in val-unseen and $+2\%$ in val-seen. In terms of SR, we also see similar performance improvement.  

\begin{figure}[t!]
\centering
\includegraphics[width=1.0\textwidth]{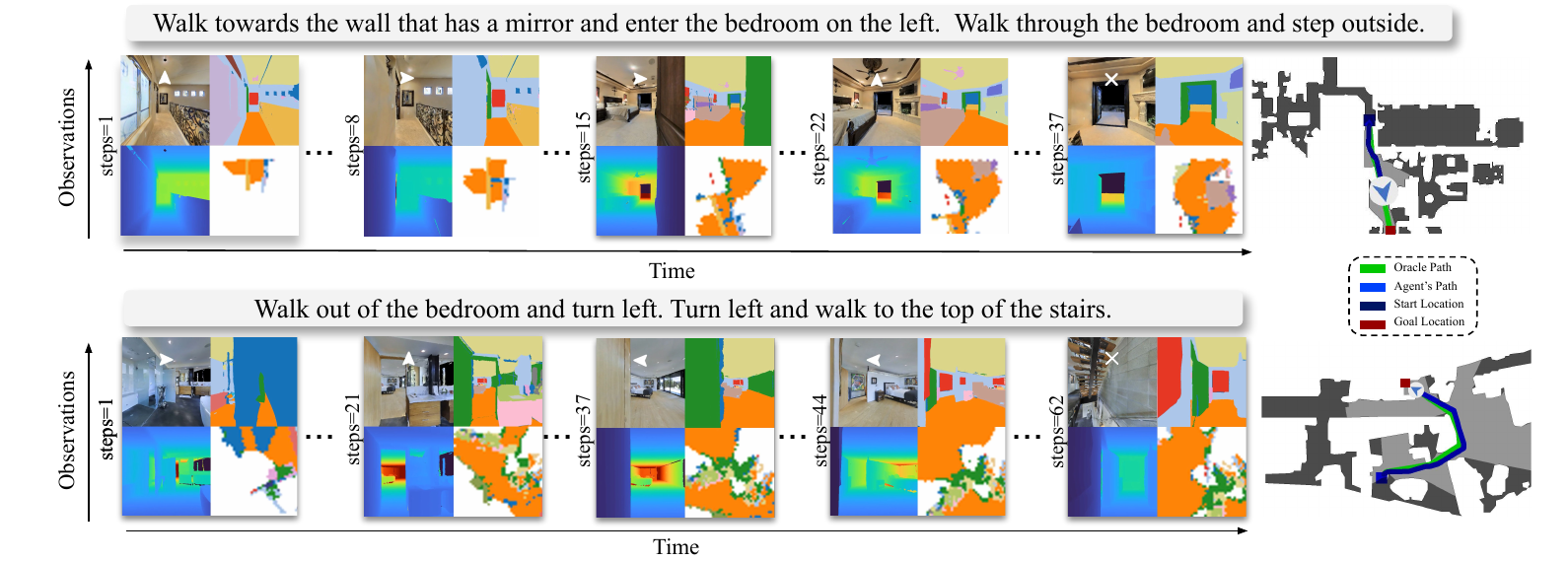}
\captionof{figure}{
\textbf{Qualitative Analysis:} Figure shows instruction-following trajectories of SASRA agent in unseen environments within VLN-CE. Sample observations (Clockwise: RGB, Semantics, Depth, and Semantic Map) seen by the agent and the corresponding actions (overlayed with RGB) are shown at each timestep. The top-down map (shown on the right) is not available to the agent and is only used for evaluation.
}
\label{qualitative_sasra}
\end{figure}

\underline{Baselines with Semantic Map.} We find that the Seq2Seq-SM and CMA-SM are not able to effectively relate the semantic map and language cues information needed for improved
navigation performance.  We find a performance drop for Seq2Seq-SM compared to Seq2Seq with teacher-forcing training
(0.15 SPL vs 0.16 SPL respectively in val-unseen) and with DAGGER training (0.28 SPL vs 0.31 val-seen set). Similarly, we also do not see a consistent performance improvement across different settings with CMA-
SM compared to the CMA model (0.18 SPL vs 0.19 SPL respectively in val-unseen set). From Table 1, we observe that our proposed approach which focuses on the effective integration of semantic mapping and language cues consistently outperforms the strong baseline models i.e. Seq2Seq-SM and CMA-SM; hence demonstrating an absolute improvement of $+6\%$ SPL in val-seen and $+5\%$ SPL in val-unseen over the best-performing baseline.

\underline{Training Regimes.} We observe from Table~\ref{tab:quantitative_sasra} that the models trained with DAGGER in general perform better than the models trained with teacher-forcing which is expected. We find DAGGER training to be most beneficial for SASRA agent and performance improves significantly (e.g., SPL of $0.25$ with teacher-forcing vs. $0.34$ with DAGGER). With teacher-forcing, we find several baselines showing similar performance to the proposed SASRA agent in val-seen, however, the proposed agent performs significantly better in val-unseen. With DAGGER, we see that the proposed model shows consistent large performance improvement over all comparing baselines across val-seen and val-unseen sets. Hence, the proposed model not only performs better in previously explored environments but also generalizes better to unseen environments.

\subsection{Ablation Study}
\label{ablationsection}

We perform an ablation study of SASRA agent in Table~\ref{tab:ablation_sasra}.
First, we observe the effectiveness of our semantic map attention (i.e, SLAM-T) module. We see that the use of semantic map cues leads to significant performance gain (i.e., $+4\%$ absolute in SR and $+3\%$ absolute in SPL) for SASRA comparing row 2 and 3. Second, we evaluate the effect of our Hybrid Action Decoder (HAD). The agent in row 1 utilizes a simple RNN as an action decoder instead of our HAD module. We find that HAD module leads to large performance gains (i.e., $+5\%$ absolute in both SR and SPL) comparing rows 3 and 1. This indicates that an effective combination of semantic map attention and action decoder module is crucial for the performance of SASRA agent. Third, we evaluate the effect of DAGGER training on SASRA agent. We observe that the agent (row 4) with DAGGER achieves a huge improvement compared to the agent with teacher forcing (row 3). The improvement is $+9\%$ absolute in both SR and SPL (Table~\ref{tab:ablation_sasra}). This shows the importance of addressing the exposure bias issue for VLN. 

We qualitatively report the performance of SASRA agent in an unseen photo-realistic simulation environment in Figure~\ref{qualitative_sasra}. 
The agent builds a top-down spatial memory in the form of a semantic map and aligns the language and map features to complete the long-horizon navigation task in 37 and 62 steps.
Episode 1 represents a failure case for our agent. We see that our agent can reach quite close to the goal but stops before reaching the success radius. The agent follows instructions to enter the bedroom and walk. However, it is confused by the phrase ``step outside” which is a bit ambiguous as there are no details to indicate which way, or where it needs to step outside. Episode 2 represents a success case for our agent. Although there are multiple instructions for ``turn left", the agent is able to turn left the correct number of steps to effectively reach its goal. In this environment, there is a staircase going up and down and the goal location is close to that. The agent can immediately understand that it is already at the top of the stairs on the current floor and does not need to walk further up the stairs. We believe semantic cues from the environment help the agent significantly in this case.
\section{Summary} 
\label{chap:sasra,sec:conclusion}

In this section, we proposed an end-to-end learning-based semantically-aware navigation agent to address the VLN task in continuous environments. Our proposed model, SASRA, combines classical semantic mapping in an attentive transformer-like architecture to extract cross-modal features between local semantics maps and provide instructions for effective VLN performance in a new environment. Extensive experiments and ablation studies illustrate the effectiveness of the proposed approach in navigating unseen environments. Specifically, we made use of local semantic mapping in an end-to-end agent-centric deep learning architecture. We showed that local semantic maps as prior knowledge greatly enhance the generalizability of VLN systems.

    \part{Generalizable Self-Supervised 3D Scene Understanding}
    \label{part3}

    This part presents research focusing on obtaining robust feature representations for two critical objectives: \textbf{1.} Utilizing unsupervised multi-view RGB data for robust 3D representation learning, and \textbf{2.} leveraging sparse multiview data to enhance the efficiency of 3D scene comprehension, spanning both indoor and outdoor environments. Our investigation underscores the substantial contributions of high-quality representations derived from extensive synthetic data in tackling challenging tasks, such as sparse view novel-view synthesis. Additionally, we showcase that representations acquired through a masked self-supervised learning paradigm from only posed 2D data yield significant improvements across a diverse spectrum of downstream 3D tasks, particularly in scenarios where obtaining extensive 3D annotations proves challenging. In Chapter~\ref{chap:neo360}, titled "NeO 360: Neural Fields for Sparse View Synthesis of Outdoor Scenes," we introduce an innovative approach for sparse view synthesis of outdoor scenes. While recent implicit neural representations excel at novel view synthesis, they often require numerous dense-view annotations. NeO 360 addresses this challenge by reconstructing 360~$^{\circ}$ scenes from just a single or a few RGB images. We propose a novel hybrid image-conditional triplane representation, trained on a large-scale synthetic dataset. Our approach demonstrates that hybrid local and global features can facilitate novel view synthesis from very sparse views. Moreover, we exhibit superior generalization performance on the challenging NeRDS 360 dataset, providing powerful editing and composition capabilities through our method. In Chapter~\ref{chap:nerfmae}, titled "NeRF-MAE: Masked AutoEncoders for Self Supervised 3D Representations Learning for Neural Radiance Fields," we explore the effectiveness of applying masked autoencoders as a pretraining method for Neural Radiance Fields (NeRFs) using Swin Transformers. Termed NeRF-MAE, this approach pretrains on multi-view posed RGB data, inspired by the success of similar techniques in 2D and 3D vision tasks. The resulting self-supervised pre-trained model, NeRF-MAE, significantly enhances performance in downstream 3D vision tasks, achieving state-of-the-art results in 3D object detection, voxel-grid super-resolution, and 3D voxel labeling, all while utilizing only posed 2D data for pretraining.
    \clearpage

    \chapter{NeO 360: Neural Fields for Sparse View Synthesis of Outdoor Scenes}
\label{chap:neo360}

In this chapter, I explore the problem setting of~\textbf{few-shot novel view synthesis} of outdoor scenes with the main aim of utilizing large-scale synthetic data priors from posed 2D images to make view-synthesis~\textit{more efficient}. The main prior knowledge comes from utilizing~\textit{multi-view 2D geometry principles} such as epipoplar constraints and lifting them to 3D effectively for the complex outdoor few-shot novel-view synthesis task. 

\section{Introduction}
\label{chap:neo360,sec:intro}

\begin{figure}[htb]
\centering
\includegraphics[width=1.0\columnwidth]{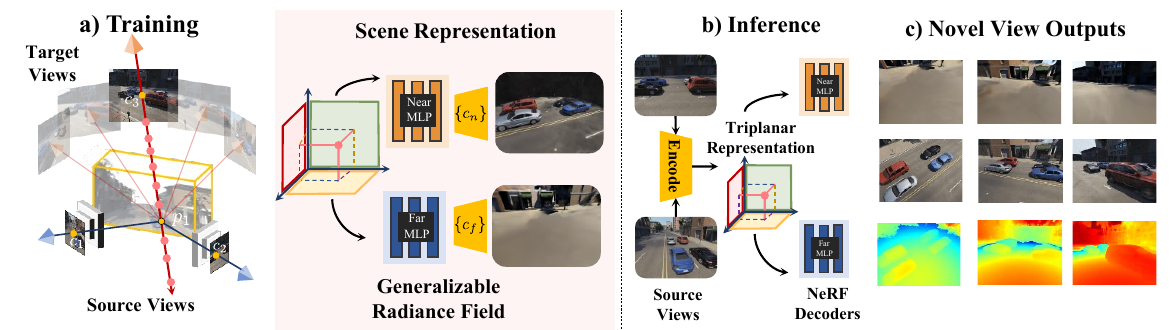}
\centering
  \caption{
  \textbf{Overview:}
    \textbf{(1)} Multi-stage pipelines in comparison to \textbf{(2)} our single-stage approach. The single-stage approach uses object instances as centers to jointly optimize 3D shape, 6D pose and size.
  }
  \label{teaser_neo}
\end{figure}

\looseness=-1
Advances in learning-based implicit neural representations have demonstrated promising results for high-fidelity novel-view synthesis~\cite{xie2021neural}, doing so from multi-view images~\cite{ueda2022neural,wang2021neus}. The capability to infer accurate 3D scene representations has benefits in autonomous driving~\cite{ost2021neural, fu2022panoptic} and robotics~\cite{irshad2022shapo, Ortiz:etal:iSDF2022, irshad2022centersnap}.

\looseness=-1
Despite great progress in neural fields~\cite{tewari2021advances} for indoor novel view synthesis, these techniques are limited in their ability to represent complex urban scenes as well as decompose scenes for reconstruction and editing. Specifically, previous formulations~\cite{boss2020nerd, park2020nerfies, zhang2020nerf++} have focused on per-scene optimization from a large number of views, thus increasing their computational complexity. This requirement limits their application to complex scenarios such as data captured by a moving vehicle where the geometry of interest is observed in just a few views. Another line of work focuses on object reconstructions~\cite{mueller2022autorf, 
irshad2022shapo} from single-view RGB~(\ref{fig:related_work}). However, these approaches require accurate panoptic segmentation and 3D bounding boxes as input which is a strong supervisory signal and consists of multi-stage pipelines that can lead to error-compounding.
\begin{figure}[!t]
  \centering
  \resizebox{0.90\columnwidth}{!}{\begin{tikzpicture}
\draw[ line width=0.2 mm, color=black, opacity=0.4] (3,0) -- (3,4.6);
\draw[ line width=0.2 mm, color=black, opacity=0.4] (0,2.225) -- (6,2.225);

\node [ color=darkblue, rotate=0, anchor=west, fill=white,rounded corners=2pt,inner sep=0.2pt] at (0.1,2.8) {\tiny DM-NeRF~\cite{wang2022dm}};
\node [ color=darkblue, rotate=0, anchor=west, fill=white,rounded corners=2pt,inner sep=0.2pt] at (0.1,2.5) {\tiny Neural Scene Graphs~\cite{ost2020neural}};
\node [ color=darkblue, rotate=0, anchor=west, fill=white,rounded corners=2pt,inner sep=0.2pt] at (0.1,1.9) 
{\tiny Object-NeRF~\cite{yang2021objectnerf}};
\node [ color=darkblue, rotate=0, anchor=west, fill=white,rounded corners=2pt,inner sep=0.2pt] at (0.1,1.6) {\tiny Neuman~\cite{neural-human-radiance-field}};
\node [ color=darkblue, rotate=0, anchor=west, fill=white,rounded corners=2pt,inner sep=0.2pt] at (0.1,1.3) {\tiny Panoptic Neural Fields ~\cite{kundu2022panoptic}};

\node [ color=darkblue, rotate=0, anchor=west, fill=white,rounded corners=2pt,inner sep=0.2pt] at (0.1,0.1) {\tiny NeRF~\cite{mildenhall2020nerf}};
\node [ color=darkblue, rotate=0, anchor=west, fill=white,rounded corners=2pt,inner sep=0.2pt] at (0.1,0.4) {\tiny NeRF-VAE~\cite{kosiorek2021nerf}};

\node [ color=darkblue, rotate=0, anchor=west, fill=white,rounded corners=2pt,inner sep=0.2pt] at (0.1,0.7) {\tiny MipNeRF360~\cite{barron2022mip}};

\node [ color=darkblue, rotate=0, anchor=west, fill=white,rounded corners=2pt,inner sep=0.2pt] at (0.1,4.375) {\tiny DVR~\cite{niemeyer2020differentiable}};
\node [ color=darkblue, rotate=0, anchor=west, fill=white,rounded corners=2pt,inner sep=0.2pt] at (0.1,4.075) {\tiny IDR~\cite{yariv2020multiview}};
\node [ color=darkblue, rotate=0, anchor=west, fill=white,rounded corners=2pt,inner sep=0.2pt] at (0.1,3.775) {\tiny NeUS~\cite{wang2021neus}};
\node [ color=darkblue, rotate=0, anchor=west, fill=white,rounded corners=2pt,inner sep=0.2pt] at (0.1,3.475) {\tiny NeDDF~\cite{ueda2022neural}};

\node [ color=black, rotate=0, anchor=east, fill=white,rounded corners=2pt,inner sep=0.6pt] at (5.8,2.2) {\tiny Generalizable};

\node [ color=black, rotate=0, anchor=east, fill=white,rounded corners=2pt,inner sep=0.6pt] at (1.3,2.2) {\tiny Overfitting};

\node [color=black, rotate=0, anchor=center, fill=white,rounded corners=2pt,inner sep=0.2pt] at (3.0,0.3) {\tiny Scene-centric};

\node [ color=orange, rotate=0, anchor=east, fill=white,rounded corners=2pt,inner sep=0.2pt] at (5.9,2.5) {\tiny \textbf{NeO 360 (Ours)}};
\node [ color=darkblue, rotate=0, anchor=east, fill=white,rounded corners=2pt,inner sep=0.2pt] at (5.9,1.9) {\tiny SSSR~\cite{zakharov2021single}};

\node [ color=darkblue, rotate=0, anchor=east, fill=white,rounded corners=2pt,inner sep=0.2pt] at (5.9,0.1) {\tiny PixelNeRF~\cite{yu2020pixelnerf}};
\node [ color=darkblue, rotate=0, anchor=east, fill=white,rounded corners=2pt,inner sep=0.2pt] at (5.9,0.4) {\tiny DeFiNE~\cite{guizilini2022depth}};
\node [ color=darkblue, rotate=0, anchor=east, fill=white,rounded corners=2pt,inner sep=0.2pt] at (5.9,0.7) {\tiny SRT~\cite{srt22}};
\node [ color=darkblue, rotate=0, anchor=east, fill=white,rounded corners=2pt,inner sep=0.2pt] at (5.9,1.0) {\tiny IBRNet~\cite{wang2021ibrnet}};

\node [ color=black, rotate=0, anchor=center, fill=white,rounded corners=2pt,inner sep=0.6pt] at (3,4.3) {\tiny Object-Centric};

\node [ color=darkblue, rotate=0, anchor=east, fill=white,rounded corners=2pt,inner sep=0.2pt] at (5.9,4.375) {\tiny DeepSDF~\cite{park2019deepsdf}};

\node [ color=darkblue, rotate=0, anchor=east, fill=white,rounded corners=2pt,inner sep=0.2pt] at (5.9,4.075) {\tiny ShAPO~\cite{irshad2022shapo}};

\node [ color=darkblue, rotate=0, anchor=east, fill=white,rounded corners=2pt,inner sep=0.2pt] at (5.9,3.775) {\tiny CodeNeRF~\cite{jang2021codenerf}};
\node [ color=darkblue, rotate=0, anchor=east, fill=white,rounded corners=2pt,inner sep=0.2pt] at (5.9,3.475) {\tiny AutoRF~\cite{mueller2022autorf}};

\node [ color=darkblue, rotate=0, anchor=east, fill=white,rounded corners=2pt,inner sep=0.2pt] at (5.9,3.175) {\tiny CARTO~\cite{heppert2023carto}};

\end{tikzpicture}}
  \caption{\textbf{Taxonomy of implicit representation methods} reconstructing appearance and shapes. The~$x$ and~$y$ correspond to (Generalizable vs Overfitting) and (Object-Centric vs Scene-Centric) dimensions, as discussed in Section~\ref{chap:neo360,sec:intro}.
  } 
  \label{fig:related_work}
\end{figure}
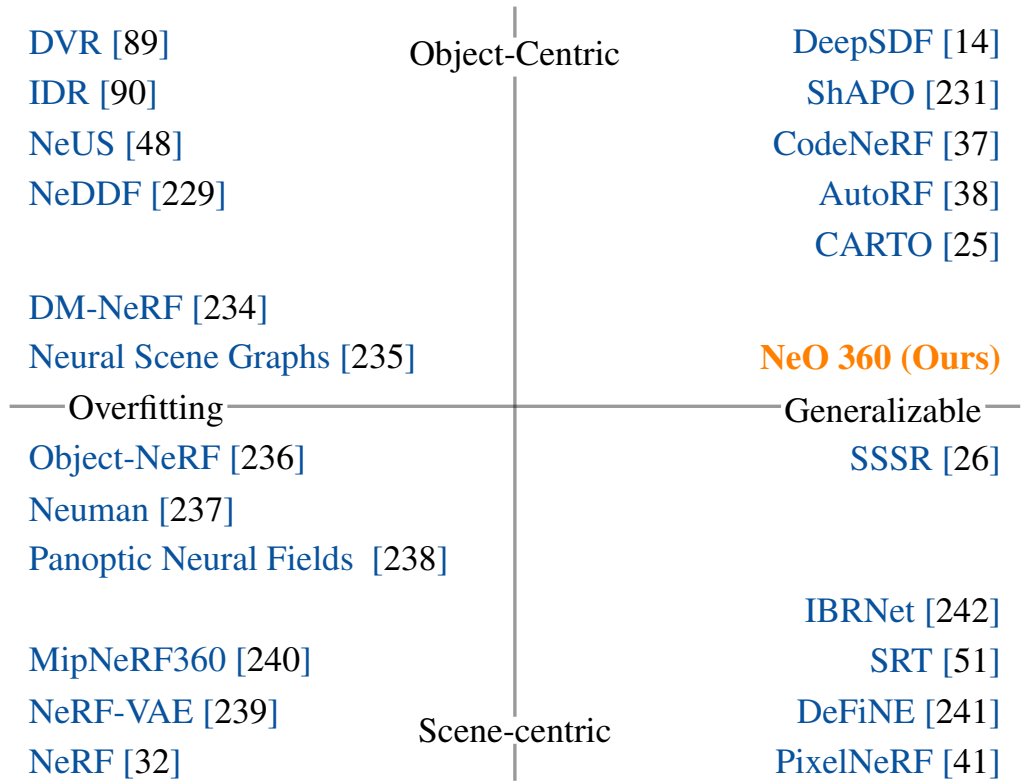

\looseness=-1
To avoid the challenge of acquiring denser input views of a novel scene in order to obtain an accurate 3D representation as well as the computational expense of per-scene optimization from many views, we propose to infer the representation of 360$^{\circ}$~unbounded scenes from just a single or a few posed RGB images of a novel outdoor environment. 

As shown in Figure~\ref{teaser_neo} our approach extends the NeRF++~\cite{zhang2020nerf++} formulation by making it generalizable. At the core of our method are local features represented in the form of triplanes~\cite{chan2022efficient}. This representation is constructed as three perpendicular cross-planes, where each plane
models the 3D surroundings from one perspective and by merging
them, a thorough description of the 3D scene can be achieved. NeO 360's image-conditional tri-planar representation efficiently encodes information from image-level features while offering a compact queryable representation for any world point. We use these features combined with the residual local image-level features to optimize multiple unbounded 3D scenes from a large collection of images. NeO 360's 3D scene representation can build a strong prior for complete 3D scenes, which enables efficient 360$^{\circ}$ novel view synthesis for outdoor scenes from just a few posed RGB images. 

\begin{figure}[t]
    \centering
    \resizebox{0.90\linewidth}{!}{%
\includegraphics{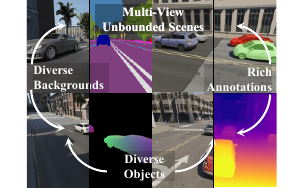}
    }
    \caption{Samples from our large scale \textbf{NeRDS 360}:"\textbf{Ne}RF for \textbf{R}econstruction, \textbf{D}ecomposition and Scene \textbf{S}ynthesis of 360$^{\circ}$ outdoor scenes" dataset comprising 75 unbounded scenes with full multi-view annotations and diverse scenes.
    }
    \label{fig:dataset_fig1}
\end{figure}%

To enable building a strong prior representation of unbounded outdoor scenes and given the scarcity of available multi-view data to train methods like NeRF, we also present a new large scale 360$^{\circ}$ unbounded dataset~(Figure~\ref{fig:dataset_fig1}) comprising of more than 70 scenes across 3 different maps. We demonstrate our approach's effectiveness on this challenging multi-view unbounded dataset in both few-shot novel-view synthesis and prior-based sampling tasks. In addition to learning a strong 3D representation for complete scenes, our method also allows for inference-time pruning of rays using 3D ground truth bounding boxes, thus enabling compositional scene synthesis from a few input views. In summary, we make the following contributions:
\begin{itemize}
   \item \textbf{A generalizable NeRF architecture for outdoor scenes} based on tri-planar representation to extend the NeRF formulation for effective few-shot novel view synthesis of 360$^{\circ}$ unbounded environments.
    \item \textbf{A large scale synthetic 360$^{\circ}$ dataset, called~\gls{nerds} 360, for 3D urban scene understanding} comprising multiple objects, capturing high-fidelity outdoor scenes with dense camera viewpoint annotations.
    \item Our proposed approach significantly outperforms all baselines for few-shot novel view synthesis on the NeRDS 360 dataset, showing 1.89 PNSR and 0.11 SSIM absolute improvement number for the 3-view novel-view synthesis task.
\end{itemize}
\section{Related Works}
\label{chap:neo360,sec:related_works}

\renewcommand{\arraystretch}{1.0}
\begin{table}[t]
\centering
\caption{\textbf{Comparison of NeRDS 360 with prior novel view synthesis datasets} comparing \#~scenes, scale, whether scenes contain multi-objects, are outdoor~i.e. urban setting, supports compositionality, provides 360 cameras~(as opposed to front-facing), and all ground truth information is available~(e.g. depth, 3D bounding boxes, instance masks)}
\resizebox{1.0\textwidth}{!}{%
\begin{tabular}{lcccccccc}
\toprule
Datasets & NeRF\cite{mildenhall2020nerf} & T\&T\cite{knapitsch2017tanks} &  NeRF360\cite{barron2022mip}& BMVS\cite{yao2020blendedmvs} & MP3D\cite{Matterport3D} & DTU\cite{aanaes2016large}&\textbf{Ours}\\
\midrule
\# Scenes & 8 & 15 & 9 & 502 & 90 & 80 & 75 \\
\# Images & 0.8k & 2.8k & 2k & 101k & 190k& 3.9k & 15k \\
Multi-Object & \redxmark & \redxmark & \redxmark & \greencheckmark & \greencheckmark & \redxmark  & \greencheckmark \\
Outdoor scenes & \redxmark & \greencheckmark& \greencheckmark & \greencheckmark &  \redxmark & \redxmark  & \greencheckmark \\
Compositionality & \redxmark & \redxmark & \redxmark & \redxmark & \redxmark & \greencheckmark & \greencheckmark \\
360 Camera & \redxmark & \greencheckmark & \greencheckmark & \redxmark & \greencheckmark &  \redxmark & \greencheckmark          \\
All GT Annotations & \greencheckmark & \redxmark & \redxmark & \redxmark & \redxmark & \redxmark & \greencheckmark \\
\bottomrule
\end{tabular}
}
\label{tab:dataset_tab}
\end{table}

\noindent\textbf{Neural Implicit Representations} use neural networks to map Euclidean or temporal coordinates to target scene properties~\cite{tewari2021advances}. These methods have been successfully used to represent 3D shapes with signed distances~\cite{park2019deepsdf,duan2020curriculum, irshad2022shapo, heppert2023carto, zakharov2022road} or occupancies~\cite{mescheder2019occupancy,takikawa2021nglod}. While earlier methods require ground truth 3D supervision, recent advances in differentiable neural rendering have enabled self-supervised learning of the target signal from only image supervision~\cite{sitzmann2019scene,mildenhall2020nerf}, with Neural Radiance Fields (NeRFs) achieving impressive results, particularly for novel view synthesis. NeRF extensions focus on reducing aliasing effects via multiscale representations~\cite{barron2021mip}, modeling unbounded scenes~\cite{martinbrualla2020nerfw}, disentangled object-background representations and blending~\cite{ost2021neural}, compositional generative models~\cite{Niemeyer2020GIRAFFE} and improving reconstruction and depth estimation accuracy via multi-view consistent features~\cite{stier2021vortx,sun2021neuralrecon,guizilini2022depth}.

\looseness=-1
\noindent\textbf{Generalizable and Feature-conditioned representations:} Neural implicit representations operate in the overfitting context, aiming to accurately encode a single scene/object~\cite{mescheder2019occupancy,mildenhall2020nerf,takikawa2021nglod,muller2022instant}. Other approaches use global conditioning (a single global latent vector) to learn object latent spaces~\cite{park2019deepsdf,jang2021codenerf,mueller2022autorf,rematas2021sharf, irshad2022shapo}, struggling to capture detailed high-frequency information. While usually limited to modeling single scenes, Pixel-NeRF~\cite{yu2020pixelnerf}, Scene Representation Transformers~\cite{srt22}, and GRF~\cite{trevithick2021grf} use local features as conditioners for generalizable neural fields. Recently, grounplanar~\cite{sharma2022seeing} and tri-planar representations gain popularity for efficiently representing scenes using hybrid implicit-explicit representation. EG3D~\cite{chan2022efficient} and GAUDI~\cite{bautista2022gaudi} utilize tri-planar representations in their generative models, employing adversarial training. Notably, GAUDI utilizes GAN inversion for conditional synthesis. Our approach, NeO 360,  does not rely on expensive GAN inversion and directly outputs density and RGB for new views in a feedforward manner.

\noindent\textbf{Novel View Synthesis Datasets:} Existing datasets for novel view synthesis fall under the following major categories: 1.~Synthetic scenes which hemispherical 360-degree views around an object of interest offering dense camera overlap for fine-grained reconstruction (these include SRN rendering~\cite{choy20163d}, RTMV~\cite{tremblay2022rtmv} and Google Scanned Objects~\cite{downs2022google}), 2.~Forward-facing scenes that move the camera in the vicinity of an object without providing full 360-degree coverage (these include LLFF~\cite{mildenhall2019llff}, DTU~\cite{aanaes2016large}, Blender MVS~\cite{yao2020blendedmvs}) and 3.~360-degree real-scenes which provide full surrounding coverage (these include Tanks and Templates~\cite{knapitsch2017tanks}, MipNeRF360 dataset~\cite{barron2021mip} and CO3D~\cite{reizenstein2021common}). These datasets mostly evaluate on indoor scenes and provide little or no compositionality~(i.e.~multi-objects, 3D bounding boxes) for training or evaluation. While MipNeRF360~\cite{barron2021mip} and Tanks and Template~\cite{knapitsch2017tanks} provide 360 coverage, the number of scenes in these datasets is small; hence it is difficult to evaluate the performance of generalizable NeRF methods at scale. Due to these challenges, we collect a large-scale outdoor dataset offering similar camera distributions as NeRF~\cite{mildenhall2020nerf} for 360$^{\circ}$ outdoor scenes. Our dataset, described in detail in~(Section~\ref{chap:neo360,sec:dataset}), offers dense viewpoint annotations for outdoor scenes and is significantly larger than existing outdoor datasets for novel-view synthesis. This allows building effective priors for large-scale scenes which can lead to improved generalizable performance on new scenes with very limited views, as we show in Section~\ref{chap:neo360,sec:exp}.

\section{NeRDS 360 Multi-View Dataset for Outdoor Scenes}
\label{chap:neo360,sec:dataset}

Due to the challenge of obtaining accurate ground-truth 3D and 2D information~(such as denser viewpoint annotations, 3D bounding boxes, and semantic and instance maps), only a handful of outdoor scenes have been available for training and testing.
Specifically, previous formulations~\cite{Ost_2021_CVPR, fu2022panoptic, kundu2022panoptic, rematas2022urban} have focused on reconstructions using existing outdoor scene datasets~\cite{geiger2012we, liao2022kitti, cabon2020vkitti2} offering panoramic-views from the camera mounted on an ego-vehicle. These datasets~\cite{nuscenes2019, sun2020scalability} offer little overlap between adjacent camera views~\cite{xie2023s}, a characteristic known to be useful for training NeRFs and multi-view reconstruction methods. Moreover, the optimization of object-based neural radiance models for these scenes becomes more challenging as the ego car is moving fast and the object of interest is observed in just a few views~(usually less than 5).

\textbf{Dataset:} To address these challenges, we present a large-scale dataset for 3D urban scene understanding. Compared to existing datasets, as demonstrated in Table~\ref{tab:dataset_tab}, our dataset consists of 75 outdoor urban scenes with diverse backgrounds, encompassing over 15,000 images. These scenes offer 360$^{\circ}$ hemispherical views, capturing diverse foreground objects illuminated under various lighting conditions. Additionally, our dataset~(as shown in Figure~\ref{fig:dataset} and Figure~\ref{fig:testcameras}) encompasses scenes that are not limited to forward driving views, addressing the limitations of previous datasets such as limited overlap and coverage between camera views~\cite{geiger2012we, liao2022kitti}). The closest pre-existing dataset for generalizable evaluation is DTU~\cite{aanaes2016large}~(80 scenes) which comprises mostly indoor objects and does not provide multiple foreground objects or background scenes.

\begin{figure}[!t]
    \centering
    \resizebox{1.0\linewidth}{!}{%
      \includegraphics{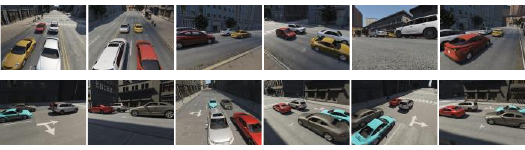}
    }
    \caption{\textbf{Proposed multi-view dataset} RGB renderings for 360$^{\circ}$ novel-view synthesis of outdoor scenes. 
    }
    \label{fig:dataset}
\end{figure}%

We use the Parallel Domain~\cite{parallel_domain} synthetic data generation to render high-fidelity 360$^{\circ}$ scenes. We select 3 different maps i.e. \textit{SF 6thAndMission}, \textit{SF GrantAndCalifornia} and \textit{SF VanNessAveAndTurkSt} and sample 75 different scenes across all 3 maps as our backgrounds~(All 75 scenes across 3 maps are significantly different road scenes from each other, captured at different viewpoints in the city). We select 20 different cars in 50 different textures for training and randomly sample from 1 to 4 cars to render in a scene. We refer to this dataset as~\textbf{NeRDS 360}:~\textbf{Ne}RF for~\textbf{R}econstruction,~\textbf{D}ecomposition and Scene~\textbf{S}ynthesis of 360$^{\circ}$ outdoor scenes. In total, we generate 15k renderings~(Fig.~\ref{fig:dataset}) by sampling 200 cameras in a hemispherical dome at a fixed distance from the center of cars. We held out 5 scenes with 4 different cars and different backgrounds for testing, comprising 100 cameras distributed uniformly sampled in the upper hemisphere, different from the camera distributions used for training. We use the diverse validation camera distribution to test our approach's ability to generalize to unseen viewpoints as well as unseen scenes during training. As shown in Figure~\ref{fig:testcameras}, our dataset and the corresponding task are extremely challenging due to occlusions, diversity of backgrounds, and rendered objects with various lightning and shadows. Our task entails reconstructing 360$^{\circ}$ hemispherical views of complete scenes using a handful of observations i.e. 1 to 5 as shown by red cameras in Figure~\ref{fig:testcameras} whereas evaluating using all 100 hemispherical views, shown as green cameras in Figure~\ref{fig:testcameras}; hence our task requires strong priors for novel view synthesis of outdoor scenes. 

\begin{figure}
    \centering
    \resizebox{\linewidth}{!}{%
      \includegraphics{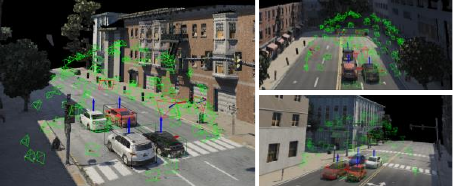}
    }
    \caption{Camera distribution for 1, 3 and 5 source views~(shown in \textcolor{red}{red}) and evaluation views~(shown in \textcolor{green}{green})
    }
    \label{fig:testcameras}
\end{figure}%

We show qualitative examples of our proposed dataset in. Figure~\ref{fig:train_samples} and Figure~\ref{fig:test_samples}. Figure~\ref{fig:train_samples} displays training samples of 3 different scenes from each of the 3 different maps in our dataset. Our dataset is very diverse both in terms of the scenes represented and the foreground car shapes and textures. NeRDS360's scenes also depict high variety in terms of occlusion of foreground objects~(i.e. not all foreground cars are observed from all views and there are various occluders such as trees and lightning poles present in the scene), varied number of objects represented~(i.e. we sample from 1 to 4 foreground cars for each scene with various textures, lightning, and shadows) as well as varied lighting and shadows in a scene~(i.e. lightning and shadows in each scene is not constant). Hence, our dataset and the corresponding task are extremely challenging. We also show different testing samples in Figure~\ref{fig:test_samples}. As shown in the figure, we render completely novel viewpoints not seen during training as well as different textures and shapes of cars that are also not rendered during training. We evaluate all 100 evaluation cameras sampled inside the hemisphere, as shown in Fig.~\ref{fig:testcameras} while giving as input 1, 3, or 5 source views to the network.   

\begin{figure}[htb!]
\begin{center}
\centering
\includegraphics[width=1.0\textwidth]{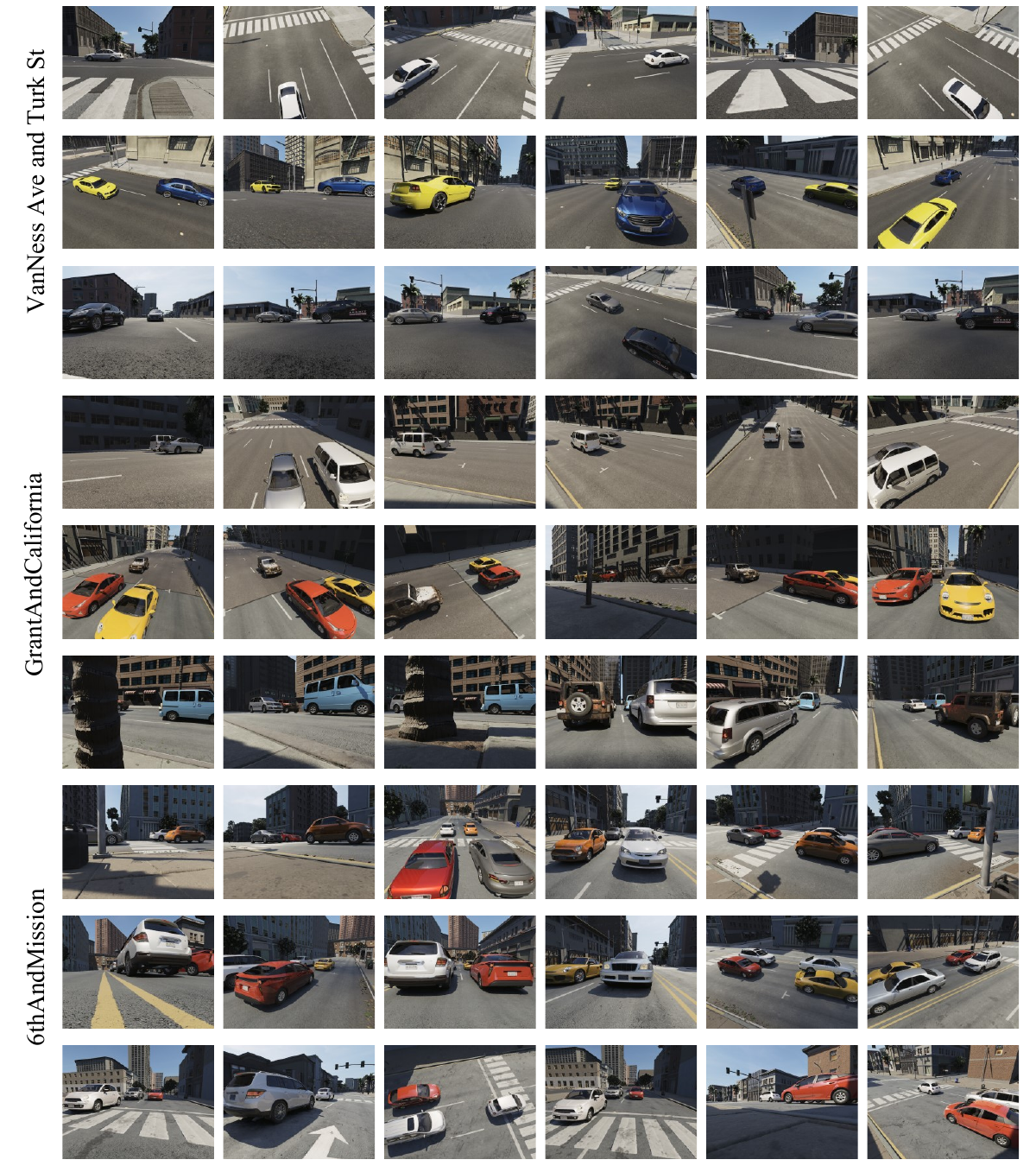}
\captionof{figure}{
\textbf{NeRDS360 training samples:} {We show diverse training samples from our proposed multi-view dataset, showing 3 examples for each map~(shown on the~$y$-axis) and displaying 5 randomly sampled images~(shown on the~$x$-axis) for each scene, from 100 rendered images with cameras placed in a hemisphere at a fixed radius from the center of the scene.
}}
\vspace{1.0cm}
\label{fig:train_samples}
\end{center}
\end{figure}

\begin{figure}[htb!]
\begin{center}
\centering
\includegraphics[width=1.0\textwidth]{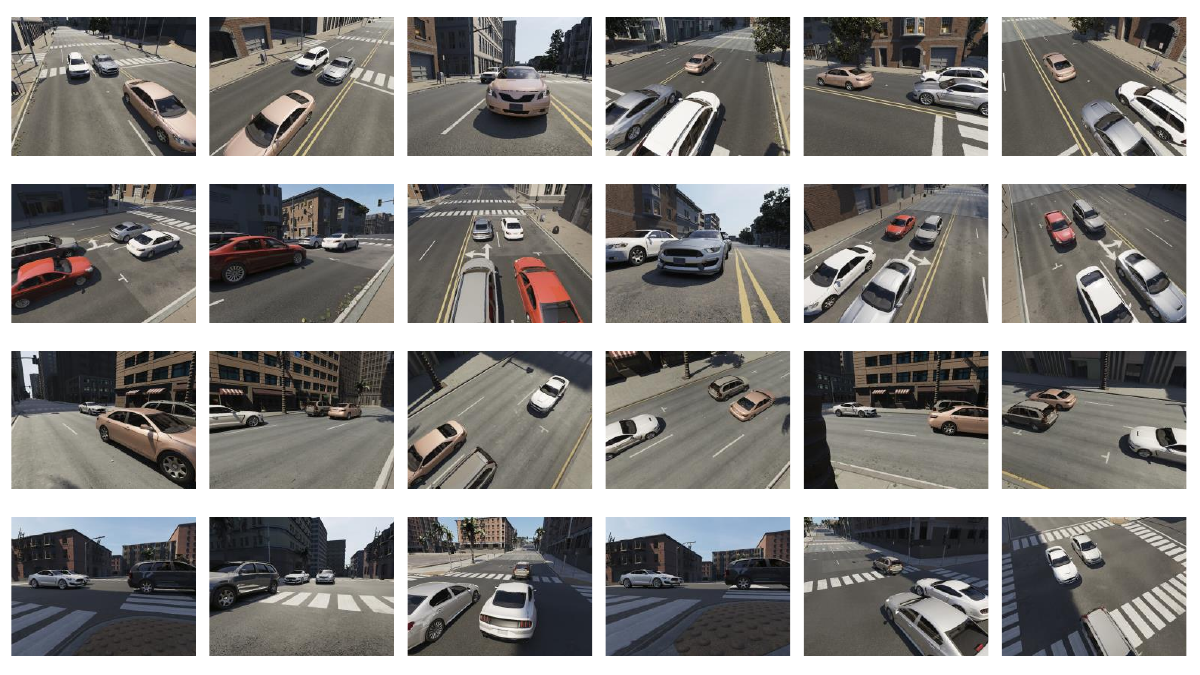}
\captionof{figure}{
\textbf{NeRDS360 test samples:} {We show unseen test samples with completely different backgrounds and objects not seen during training. Test samples include completely different camera viewpoints that are not observed during training which are still sampled in a hemisphere around the foreground objects of interest. Here, we show 4 different scenes from our evaluation dataset, different from the training dataset~(shown on the~$y$-axis) and show 5 randomly sampled images~(shown on the~$x$-axis) for each scene, from 100 rendered images with cameras placed in a hemisphere at a fixed radius from the center of the scene.
}}
\label{fig:test_samples}
\end{center}
\end{figure}

\section{Method}
\label{chap:neo360,sec:method}

Given RGB images of a few views of a novel scene, NeO 360 infers a 3D scene representation capable of performing novel view synthesis and rendering 360$^{\circ}$ scenes. To achieve this goal, we employ a  hybrid local and global feature representation comprised of a triplanar representation that can be queried for any world point. Formally, as shown in Figure~\ref{teaser_neo}, given a few input images,~$I=[I_{1}...I_{n}]$ of a complex scene, where $n =1$ to $5$, and their corresponding camera poses, $\gamma=[\gamma_{1}...\gamma_{n}]$ where $\gamma=[R|T]$, NeO 360 
infers the density and radiance fields for both near and far backgrounds~(similar to NeRF++~\cite{zhang2020nerf++}) with the major difference of using hybrid local and global features for conditioning the radiance field decoders instead of just positions and viewing directions, as employed in the classical NeRF formulation~\cite{mildenhall2020nerf, zhang2020nerf++}. We describe our 3D scene representation in Section~\ref{sec:triplanar}, introduce deep residual local features in Section~\ref{sec:residual_features}, describe how we decode radiance fields conditioned on hybrid local and global features in Section~\ref{sec:decoding}, and discuss performing inference-time scene editing and composition in Section~\ref{sec:editing}.

\subsection{Image-Conditional Triplanar Representation}
\label{sec:triplanar}
\looseness=-1 \textbf{Preliminaries}: NeRF is an implicit 3D scene representation that learns a neural network $f(\mathbf{x}, \theta) \rightarrow(\mathbf{c}, \sigma)$. This end-to-end differentiable function $f$ outputs color $c_{i}$ and density $\sigma_{i}$ for every query 3D position $x_{i}$ and the viewing direction $\theta_{i}$ as input. For each point evaluation, a 4 channel~$(\mathbf{c}, \sigma)$ value is output, which is then alpha composited~(Eq. \ref{eq:alpha_comp} below) to render an image using volume rendering. 
\begin{equation}
\label{eq:alpha_comp}
\mathbf{c}=\sum_i w_i \mathbf{c}_i, \quad \mathbf{acc}=\sum_i w_i
\end{equation}
\begin{equation*}
w_i=\alpha_i \prod_{j<i}\left(1-\alpha_j\right), \quad \alpha_i=1-\exp \left(-\sigma_i\left\|\boldsymbol{x}_i-\boldsymbol{x}_{i+1}\right\|\right)
\end{equation*}

\begin{figure}[t!]
\centering
\includegraphics[width=1.0\textwidth]{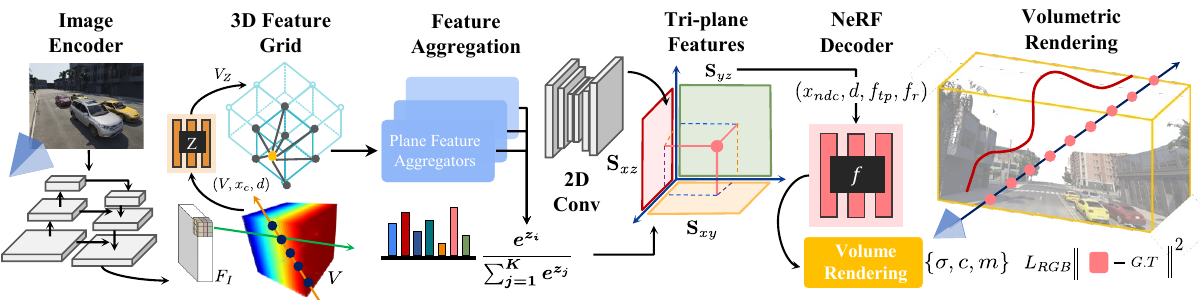}
\captionof{figure}{
\textbf{Method:} {Our method effectively uses local features to infer an image-conditional triplanar representation for both backgrounds and foregrounds. These triplanar features are obtained after orthogonally projecting positions~($x$) into each plane and bilinearly interpolating feature vectors. Dedicated NeRF decoder MLPs are used to regress density and color each for foreground and background.
}}
\label{fig:framework_neo360}
\end{figure}

\textbf{Method:} Although producing high-fidelity scene synthesis, NeRF~\cite{mildenhall2020nerf} is limited in its ability to generalize to novel scenes. In order to effectively use scene-priors and learn from a large collection of unbounded 360$^{\circ}$ scenes, we propose an image-conditional triplanar representation, as shown in Figure~\ref{fig:framework_neo360}. This representation is capable of modeling 3D scenes with full expressiveness at scale without omitting any of its dimensions~(as in 2D or BEV-based representation) and avoiding cubic complexity~(as in voxel-based representations). Our triplanar representation comprises three axis-aligned orthogonal planes~$S=[\textbf{S}_{xy}, \textbf{S}_{xz}, \textbf{S}_{yz}]$, $\in$  $\mathbb{R}^{3\times C \times D \times D}$ where $D \times D$ is the spatial resolution of each plane with feature $C$. 

To construct feature triplanes from input image, we first extract low-resolution spatial feature representations by using an  ImageNet~\cite{deng2009imagenet} pre-trained~\gls{cnn} backbone \textbf{E} which transforms input image $I \in \mathbb{R}^{H_i \times W_i \times 3}$ to 2D feature map $F_{I} \in \mathbb{R}^{H_i/2 \times W_i/2 \times C}$. Similar to prior works involving volumetric reconstruction~\cite{sun2021neuralrecon, murez2020atlas, kar2017learning}, the obtained local features are projected backwards along every ray to the 3D feature volume~($V_{F}$) using camera pose~$\gamma_{i}$ and intrinsic~$K_{i}$. 
While volumetric reconstruction methods~\cite{sun2021neuralrecon, murez2020atlas}, traditionally use the generated volume solely for indoor geometry reconstruction through TSDF, we show that it can also be employed in a computationally efficient way to estimate the entire scene's appearance and enable accurate neural rendering. Since all features along a camera ray are identical in the grid, we further learn depth of individual features by an additional MLP, $V_{Z} = Z(V_{F}, x_{c}, d)$ which takes as input concatenated features in the grid, positions of grid in the camera frame~($x_{c}$) and directions from the positions of grid in the world frame~$x_{w}$ to the camera frame and outputs depth-encoded features $V_{Z}$. Next, we obtain triplane features using learned weights~($w_{i}$) over individual volumetric feature dimensions:
\begin{align}
\textbf{S}_{xy} =\sum_i w_{xy_i} \mathbf{V}_{Z_i},  \quad w_{xy} &= A_{xy}(V_{Z_i}, x_{w_z}) \\
\textbf{S}_{xz} =\sum_i w_{xz_i} \mathbf{V}_{Z_i},\quad   w_{xz} &= A_{xz}(V_{Z_j}, x_{w_y}) \\
\textbf{S}_{yz} =\sum_i w_{yz_i} \mathbf{V}_{Z_i},\quad w_{yz} &= A_{yz}(V_{Z_j}, x_{w_x})
\end{align}

where $A_{xy}, A_{xz}$ and $A_{yz}$ denote feature aggregation MLPs and $w_{xy}, w_{xz}$ and $w_{yz}$ are softmax scores obtained after summing over the $z, y$ and $x$ dimensions, respectively. One motivation to project features into respective planes is to avoid the computationally cubic complexity of 3D CNNs as in~\cite{chen2021mvsnerf, stier2021vortx} and at the same time be more expressive than BEV or 2D feature representations~\cite{li2022bevdepth, li2022bevformer, yu2020pixelnerf} which are computationally more efficient than voxel-based representations but omitting z-axis hurts their expressiveness. We instead rely on 2D convolutions to transform the built image-conditional triplanes into a new G-channel output, where $G=C/4$, while upsampling the spatial dimension of planes from $D\times D$ to image feature space (i.e. $H/2 \times W/2$). The learned convolutions act as inpainting networks to fill in missing features.  
As shown in Figure~\ref{fig:framework_neo360}, our triplanar representation acts as a global feature representation,  as intuitively a complex scene can be better represented when examined from various perspectives. This is because each may offer complementary information that can help understand the scene more effectively.

\subsection{Deep Residual Local Features:}
\label{sec:residual_features}
As noted by~\cite{chen2021mvsnerf, wang2021ibrnet} and inspired by ~\cite{heDeepResidualLearning2016}, for the following radiance field decoding stage, we also use the features $f_{r}$ as a residual connection into the rendering MLP. We obtain $f_{r}$ from $F_{I}$ by projecting the world point $x$ into source view using its camera parameters $\gamma_{i},K_{i}$ and extracting features at the projected pixel locations through bilinear interpolation similar to~\cite{yu2020pixelnerf}. Note that both local and global feature extraction pathways share the same weights~$\theta_{E}$ and encoder~$\textbf{E}$. We find that for complex urban unbounded scenes, using just local features similar to~\cite{yu2020pixelnerf} leads to ineffective performance for occlusions and faraway 360$^{\circ}$ views, as we show quantitatively and qualitatively in Section~\ref{chap:neo360,sec:exp}. Using only global features, on the other hand, leads to hallucinations, as shown in our ablations (Section~\ref{chap:neo360,sec:exp}). Our method combines both local and global feature representations effectively, resulting in a more accurate 360$^{\circ}$ view synthesis from as minimal as a single view of an unbounded scene.
  
\subsection{Decoding Radiance Fields:}
\label{sec:decoding}
The radiance field decoder $D$ is tasked with predicting color $c$ and density $\sigma$ for any arbitrary 3D location $x$ and viewing direction~$d$ from triplanes $S$ and residual features $f_{r}$. We use a modular implementation of rendering MLPs, as proposed by ~\cite{zhang2020nerf++} with the major difference of using our local and global features for conditioning instead of just using positions and viewing directions as an input to the MLPs. The MLP is denoted as: 
\begin{equation}
\label{eq:rendering}
\sigma, c = D(x, d, f_{tp}, f_{r})    
\end{equation}
where we obtain $f_{tp}$ by orthogonally projecting point $x$ into each plane in $S$ and performing bi-linear sampling. We concatenate the three bi-linearly sampled vectors into $f_{tp} = [\textbf{S}_{xy}(i, j),\textbf{S}_{xz}(j, k),\textbf{S}_{yz}(i, k)]$. Note that similar to ~\cite{yu2020pixelnerf}, we establish our coordinate system using the view space of the input image, and then indicate the positions and camera rays within this particular coordinate system. By utilizing the view space, our method can successfully standardize the scales of scenes from various data sources, thereby enhancing its ability to generalize well. Although our method gives reasonable results from single-view observation~(Section~\ref{chap:neo360,sec:exp}), NeO 360 can seamlessly integrate multi-view observations by pooling along the view dimension in the rendering MLPs.

\textbf{Near and Far Decoding MLPs:}
Similar to NeRF++~\cite{zhang2020nerf++}, we define two rendering MLPs for decoding color and density information as follows:
\begin{equation}
D(.) =\begin{cases}
D_{fg}(.) & \text{if } \sum\limits_{i=1}^n (x_i^2+y_i^2+z_i^2) < 1 \\
D_{bg}(.) & \text{if } \sum\limits_{i=1}^n (x_i^2+y_i^2+z_i^2) > 1
\end{cases}
\end{equation}
where we define a coordinate remapping function~($M$) similar to the original NeRF++ formulation~\cite{zhang2020nerf++} to contract the 3D points that lie outside the unit sphere where $M$ maps points~($x,y,z$) outside the unit sphere to the new 4D coordinates$
\left(x^{\prime}, y^{\prime}, z^{\prime}, 1 / r\right)
$, where $(x^{\prime}, y^{\prime}, z^{\prime})$ represents the unit vector in the direction of ($x,y,z$) and $r$ denotes the inverse radius along this dimension. This formulation helps further objects get less resolution in the rendering MLPs. For querying our tri-planar representation, we use the un-contracted coordinates ($x,y,z$) in the actual world coordinates, since our representation is planes instead of spheres. For rendering, we use the respective contracted coordinates~($x^{\prime}, y^{\prime}, z^{\prime}$) for conditioning the MLPs.  
\begin{table*}
\centering
\renewcommand{\arraystretch}{1.0}
\footnotesize
\caption{\small{\textbf{Quantitative novel view synthesis results}: Conditional prior-based sampling and novel-scene rendering. \textcolor{pink}{Pink} \textcolor{pink} {\Large{$\blacktriangle$}}  denotes zero-shot evaluation whereas \textcolor{orange}{Orange}\textcolor{orange} {\Large{$\blacktriangle$}} denotes finetuning only the triplanar network i.e. freezing the encoder \textbf{E} with learning rate $5 \times 10^{-6}$
from 1,3 or 5 source views. The orange triangle for PixelNeRF denotes that we finetune their encoder network similar to how we finetune our triplanes. \underline{Underline} shows best results when trained and evaluated within the same single map~(easier setting), \textbf{Bold} denotes best results with challenging evaluation setting where the evaluation dataset is sampled across 3 different maps with diverse illumination.}
}
\resizebox{\textwidth}{!}{%
\begin{tabular}{cccccccccccccc}
\toprule
\multirow{3}{*}{Method}  & \multirow{3}{*}{\# Views}   & \multicolumn{4}{c}{Single Map (Prior Sampling)} &\multicolumn{4}{c}{Single Map (Novel Scenes)}&\multicolumn{4}{c}{Multi-Map (Novel Scenes)}\\
\cmidrule(lr){3-6}\cmidrule(lr){7-10} \cmidrule(lr){11-14}

& & \multicolumn{3}{c}{\textbf{Scenes}} & \textbf{Objects} &\multicolumn{3}{c}{\textbf{Scenes }} & \textbf{Objects} &\multicolumn{3}{c}{\textbf{Scenes}} & \textbf{Objects}\\
\cmidrule(lr){3-5} \cmidrule(lr){6-6} \cmidrule(lr){7-9} \cmidrule(lr){10-10}
\cmidrule(lr){11-13} \cmidrule(lr){14-14}
& & \textbf{PSNR} $\uparrow$ & \textbf{SSIM} $\uparrow$ & \textbf{LPIPS} $\downarrow$ & \textbf{PSNR} $\uparrow$ & \textbf{PSNR} $\uparrow$ & \textbf{SSIM} $\uparrow$ & \textbf{LPIPS} $\downarrow$ & \textbf{PSNR} $\uparrow$& \textbf{PSNR} $\uparrow$ & \textbf{SSIM} $\uparrow$ & \textbf{LPIPS} $\downarrow$ & \textbf{PSNR} $\uparrow$\\ 

\midrule

\multirow{3}{1.8cm}{\textcolor{pink}{\Large{$\blacktriangle$}} \centering NeRF~\cite{mildenhall2020nerf}}
& 1 & 12.41 & 0.09 & 0.71 & 10.55 & 12.82 & 0.13 & 0.69 & 10.45 & 12.72& 0.12 & 0.69 & 10.39  \\
& 3 & 14.14 & 0.25 & 0.59& 11.67 & 13.76 & 0.18 & 0.62 & 11.81 & 13.82& 0.19& 0.61&  11.85\\
& 5 & 15.37 & 0.38 & 0.50& 12.89 & 16.16 & 0.39 & 0.49 & 14.83 & 16.14& 0.38& 0.48& 14.73\\
\midrule
\multirow{3}{1.8cm}{\textcolor{pink}{\Large{$\blacktriangle$}} \centering MVSNeRF~\cite{chen2021mvsnerf}}
& 1 &  14.40 & 0.40 & 0.65 & 12.42& 13.87 & 0.33 & 0.65 & 11.08 & 13.90&  0.31&  0.65& 11.10\\
& 3 & 13.93 & 0.34 & 0.63 & 11.33 & 14.50 & 0.40 & 0.64 & 12.79 &14.40 & 0.38&  0.63& 12.70\\
& 5 & 14.78 & 0.39 & 0.62 & 12.21& 15.43 & 0.41 & 0.61 & 14.13 & 15.40 & 0.42& 0.62&14.10\\
\midrule

\multirow{3}{1.8cm}{\textcolor{pink}{\Large{$\blacktriangle$}} \centering Pixel-NeRF~\cite{yu2020pixelnerf}}
& 1 & 15.89 & 0.44 & 0.64 & 13.57 & 14.93 & 0.40 & 0.65 & 12.93 & 15.01& 0.47 & 0.65&12.65\\
& 3 & 17.15 & 0.50 & 0.62 & 14.47 & 17.46 & 0.48 & 0.63 & 15.70 & 16.20 & 0.52 & 0.64& 13.00\\\
& 5 & 17.50 & 0.51 & 0.62 & 14.75 & 17.80 & 0.49 & 0.62 & 15.92 & 16.91 & 0.52& 0.62& 14.22\\
\midrule

\multirow{3}{1.8cm}{\textcolor{orange}{\Large{$\blacktriangle$}} \centering Pixel-NeRF\textsubscript{\textit{ft}} ~\cite{yu2020pixelnerf}}
& 1 & -& - & - & -& 15.78 & 0.43 & 0.66 & 14.07 & 14.65& 0.42 & 0.66&11.81 \\
& 3 & - & - &-& -& 17.90 & 0.51 & 0.59 & 17.12 & 16.69 & 0.54 & 0.62& 13.49\\
& 5 & - & - & - & - & 19.26 & 0.55 & 0.57 & 18.54& 17.22 & 0.55& 0.61& 15.21 \\
\midrule

\multirow{3}{1.8cm}{\textcolor{pink}{\Large{$\blacktriangle$}} \centering \textbf{NeO 360} (Ours)}
&1 & 16.91 & 0.51 & 0.56 & 14.11 & 17.60 & 0.56& 0.51&15.80 & 16.30 &0.52 & 0.57& 13.04\\
& 3 & 18.94 & 0.58 & 0.48 & 16.66 & 19.35 & 0.59& 0.50 & 17.60 & 18.59&0.61 &0.52 & 15.93\\

& 5 & \underline{19.64} & \underline{0.62} & \underline{0.47} & \underline{17.34} & 20.10 & 0.62& 0.48& 18.20 & 19.27& 0.64 & 0.49& 16.60\\
\midrule
\multirow{3}{1.8cm}{\textcolor{orange}{\Large{$\blacktriangle$}} \centering 
 \textbf{NeO 360}\textsubscript{\textit{ft}} (Ours)}
&1 & - & - & - & - &{17.93} & {0.58}& {0.49}&{15.95} & 16.42& 0.55& 0.54& 13.80 \\
& 3 & - & - & - & - & 19.56 & {0.61} & 0.46& 18.30 & 18.94 & 0.63 & 0.49& 16.81\\
& 5 & - & -& -& - & \underline{20.56} & \underline{0.64}& \underline{0.45}& \underline{18.62} &\textbf{19.59}&\textbf{0.67} &\textbf{0.46}&\textbf{17.70}\\
\bottomrule
\end{tabular}
}
\label{table:overfitting_objects}
\end{table*}

\textbf{Optimizing radiance fields for few-shot novel-view synthesis:}
Given local and global features constructed from source views, we decode color $c^{i}_{p}$ and density $\sigma^{i}_{p}$ for backgrounds using dedicated near and far background MLPs $D_{near}(.)$ and $D_{far}(.)$~(Eq.~\ref{eq:rendering}) after volumetrically rendering and compositing the near and far backgrounds and enforcing the loss as follows:
\begin{equation}
    L=\left\|c_p-\tilde{c}_t\right\|_2^2 +\lambda_{reg} L_{reg} + \lambda_{LPIPS} L_{LPIPS}
\end{equation}

where $\tilde{c}_t$ is the sampled pixel locations from the target image and $c^{p}$ is the composited color obtained from the rendering output of near and far MLPs as $c^{p}_{i} =  c^{nb}_{i} + \prod_{j<i}\left(1-\alpha^{nb}_j\right) c^{b}_{i}$. We also encourage the weights of near and far background MLPs to be sparse for efficient rendering by enforcing an additional distortion regularization loss~\cite{barron2021mip} and further use ${L}_{LPIPS}$ loss to encourage perceptual similarity b/w patches of rendered color, $c^{p}$ and ground color~$\tilde{c}_t$, where we only enforce it after 30 training epochs to improve background modeling. 

\subsection{Scene Editing and Decomposition:}
\label{sec:editing}

Given 3D bounding boxes obtained from a detector, we can obtain the individual object and background radiance fields for each object by simply sampling rays inside the 3D bounding boxes of the objects and bilinearly interpolating the features at those specific~($x,y,z$) locations in our tri-planar feature grid (\textbf{S}), making it straightforward to edit out and re-render individual objects. As
illustrated in Figure~\ref{fig:qualitative_main}, we perform accurate object re-rendering by considering the features inside the 3D bounding boxes of objects to render the foreground MLP. In essence, we divide the combined editable scene rendering formulation as rendering objects, near backgrounds, and far backgrounds. For far backgrounds, we retrieve the scene color~$c^{b}_{i}$ and density~$\sigma^{b}_{i}$ which is unchanged from the original rendering formulation. For near backgrounds, we obtain color~$c^{nb}_{i}$ and density~$\sigma^{nb}_{i}$ after pruning rays inside the 3D bounding boxes of objects~(i.e. setting $\sigma^{nb}_{i}$ to a negative high value, $-1 \times 10^{-5}$ before volumetrically rendering). For objects, we only consider rays inside the bounding boxes of each object and sampling inside foreground MLP to retrieve~$c^{o}_{i}$ and density~$\sigma^{o}_{i}$. We aggregate the individual opacities and colors along the ray to render composited color using the following equation:
\begin{equation}
\label{eq:alpha_comp_combined}
\mathbf{c}=\sum_i w^{b}_i \mathbf{c}^{b}_i + \sum_i w^{nb}_i \mathbf{c}^{nb}_i + \sum_i w^{o}_i\mathbf{c}^{o}_i
\end{equation}

\subsection{Network Architecture Details:}
\label{sec:architecture}

\begin{figure}[htb]
\centering
\includegraphics[width=1.0\columnwidth]{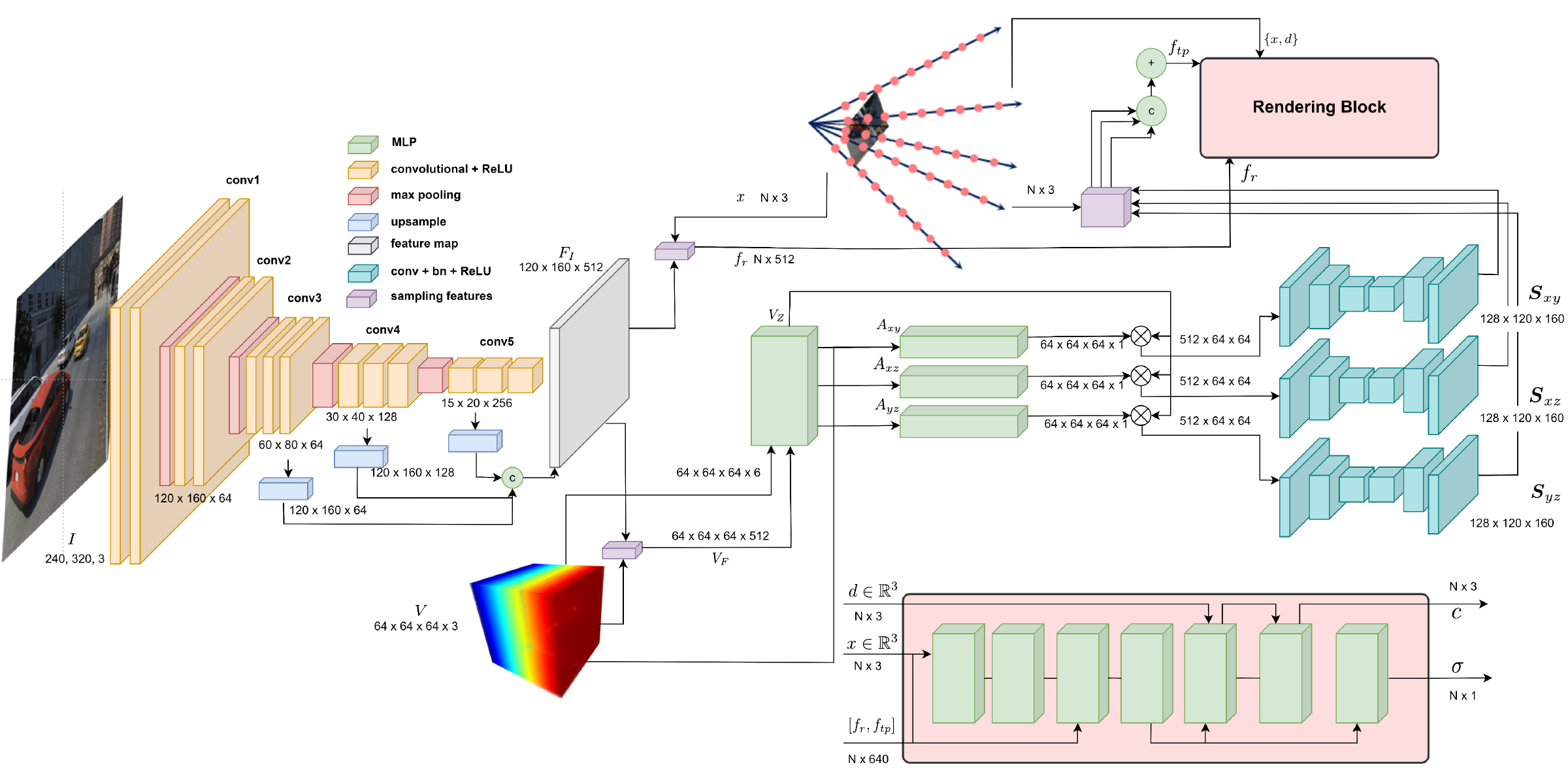}
\centering
  \caption{
  \textbf{Overview:}
    \textbf{(1)} Multi-stage pipelines in comparison to \textbf{(2)} our single-stage approach. The single-stage approach uses object instances as centers to jointly optimize 3D shape, 6D pose, and size.
  }
  \label{fig:supp_architecture}
\end{figure}

In this section, we provide more details about our architectural design, specifically our image-conditional tri-planar representation and our rendering MLPs as shown in Fig.~\ref{fig:framework_neo360}. Our detailed architecture is presented in Fig.~\ref{fig:supp_architecture} and described in the following sections. We first describe the details of our encoder in Sec.~\ref{sec:encoder}, next we describe the details of our tri-planar features and residual features in Sec.~\ref{sec:features}. Finally, we provide details of our rendering MLPs in Sec.~\ref{sec:rendering_mlps}.
\subsection{Encoder:}
\label{sec:encoder}
The Encoder network~$E$ comprises a pre-trained Resnet32~\cite{heDeepResidualLearning2016} backbone. We extract features from the initial convolutional layer and subsequent layer 1 to layer 3 and upsample all features to the same spatial resolution~(i.e. $H/2 \times W/2$) where $W,H$ are image dimensions respectively, before concatenating along the feature dimension to output feature map $F_{I}$ with dimension $512 \times H/2 \times W/2$ as shown in Fig.~\ref{fig:supp_architecture}.

\subsection{Image-conditional tri-planar features and residual features:}

In this section, we delve into the image-conditional tri-planar features and residual features. These elements, formed through volumetric projection, depth encoding, feature aggregation, and 2D convolutions, are essential for our approach's effectiveness in enhancing scene understanding.

\label{sec:features}
\textbf{Volumetric local features:} We back project local feature map~$F_{I}$ along every ray to the world grid~($V$) to get 3D feature volume feature ($V_{F}$) with dimensions $K \times\ K  \times K \times 512$ where K is the resolution of feature grid and we use $K=64$. Note that there is a tradeoff between the size of the feature grid, i.e. expressiveness and computational cost. We found $K=64$ to give a reasonable performance while avoiding any OOM (out-of-memory issues) due to a larger grid size in our network training. \\
\textbf{Feature depth-encoding:} As detailed in Sec.~\ref{sec:triplanar}, we learn the depth of each feature in the feature grid using an additional 2-layer MLP with hidden dimension 512 to output depth-encoded features~$V_{Z}$ of dimensions~$K \times\ K  \times K \times 512$. Our feature aggregation module comprises three 2-layer MLPs with hidden dimension 512 and outputs learned weights $w_{i}$ over individual volumetric feature dimensions to outputs weights~$w_{xy}, w_{xz}$ and $w_{yz}$ each with dimensions $K \times  K \times K \times 1$. After performing $softmax$ and summing over the $z, y$, and $x$ dimensions respectively, we obtain 2D feature maps for each of the three planes with dimension~$K \times K \times 512$. \\
\textbf{2D Convolutions}: We further use a series of 2D convolutions with upsampling layers to transform the planar features to dimension~$ H/2 \times W/2 \times 128$. The convolutional layers comprise three convolutional layers with input channels 512, 256, and 128 and output channels 256, 128, and 128 respectively, with a kernel size of 3, a stride of 2, and a padding of 1, followed by an upsampling layer with a scale factor of 2 and another convolutional layer with an input channel and output channel of $128$. Finally, an upsampling layer with an output dimension of $H/2 \times W/2$ is employed before outputting the features with a final convolution layer with input and output channels $128$. All convolutional layers are followed by the BatchNorm and ReLU layers. The output of each convolutional block becomes our tri-planar features~$S$, each with dimension~$ 128 \times 120 \times 160$. We sample into each plane in~$S$ by projecting $x$ into each plane~i.e. by getting the absolute $xy, xz$ and $yz$ coordinates of $x$ before concatenating and summing over the channel dimension to retrieve feature $f_{tp}$ with dimension $N \times128$ where N denotes the number of sampled points and 128 is the feature dimension. The residual local feature~$f_{r}$ after sampling into $F_{I}$ has dimensions $N \times 512$.

\subsection{Rendering MLPs:}
\label{sec:rendering_mlps}
The rendering MLPs for both foreground and background rendering comprises 7 fully connected layers with hidden dimension of 128 and ReLU activation. We apply positional encoding~\cite{mildenhall2020nerf} to the input positions $x$ and viewing direction~$d$. We concatenate positions~$x$ with triplanar features~$f_{tp}$ and residual features~$f_{r}$ as an input to the first layer of the MLP. We also supply the conditioning feature as a skip connection to the third layer in the MLP and mean pool the features along the viewing dimension in the fourth MLP layer, if there is more than one image in the input. We found this pooling strategy to work better than pooling before the rendering stage~i.e. earlier on in the tri-planar construction stage~(\ref{sec:features}). In total, we use the first 4 layers to output features of dimension~$N \times 128$, before utilizing a final density MLP to output 1 channel value for every sampled point $N$. We further use two additional dedicated MLP layers with a hidden dimension of 128 to output a 3-channel color value for every sampled point~$N$, conditioned on the positionally-encoded viewing direction and the output of the fourth MLP layer.

\section{Experiments and Results}
\label{chap:neo360,sec:exp}
We evaluate our proposed method against various state-of-the-art baselines, focusing on few-shot novel view synthesis including~\textbf{a.} Conditional prior-based sampling and~\textbf{b.} Novel scene rendering tasks. We compare full scenes on the following baselines: 1) \textbf{NeRF}~\cite{mildenhall2020nerf}: Vanilla NeRF formulation which overfits to a scene given posed RGB images 2) \textbf{PixelNeRF}~\cite{yu2020pixelnerf} A generalizable NeRF variant which utilizes local image features for few-shot novel view synthesis 3) \textbf{MVSNeRF}~\cite{chen2021mvsnerf}: Extends NeRF for few-view synthesis using local features obtained by building a cost-volume from source images and 4)~\textbf{NeO 360}: Our proposed architecture which combines local and global features for generalizable scene representation learning. 

\textbf{Metrics:}
We use standard PSNR, SSIM, and LPIPS metrics to evaluate novel-view synthesis and L1 and RMSE to measure depth reconstruction quality.
\renewcommand{\arraystretch}{0.8}
\begin{table}[t]
\centering
\caption{\label{tab:object_discovery} \small{Effect of scene priors on 3-view novel view synthesis on NeRDS 360 dataset~(Single Map)}}
\resizebox{0.85\textwidth}{!}{%
\begin{tabular}{@{}lccc@{}}
\toprule
\multirow{2}{*}{Method}& \multicolumn{2}{c}{\textbf{Backgrounds}} & \textbf{Objects} \\
\cmidrule(lr){2-3} \cmidrule(lr){4-4}
 & \textbf{PSNR} $\!\uparrow$ & \textbf{SSIM} $\!\uparrow$ & \textbf{PSNR} $\!\uparrow$ \\ \midrule
NeRF~\cite{mildenhall2019llff} (No Priors) & 16.16 & 0.34 & 15.42\\
Ours (No Pretraining) & 16.40 & 0.39 & 15.70\\
\midrule
Ours & \textbf{20.48} & \textbf{0.67} & \textbf{19.03}\\ 
\bottomrule
\end{tabular}
}
\end{table}

\textbf{Comparison with strong baselines for novel-view synthesis:} We aim to answer the following key questions: \textbf{1.} Does our generalizable tri-planar representation perform better than other generalizable NeRF variants given access to prior data and a few views for optimization on novel scenes? \textbf{2.} Do scene priors help with zero-shot generalization? and \textbf{3.} Does scaling the data help our network generalize better? We summarize the results in Table~\ref{table:overfitting_objects} and note that NeO 360 achieves superior results compared to state-of-the-art generalizable NeRF variants i.e. PixelNeRF~\cite{yu2020pixelnerf} and MVSNeRF~\cite{chen2021mvsnerf} in both zero-shot testing and fine-tuning given a limited number of source views. Specifically, NeO 360 achieves a PSNR of 19.35, SSIM of 0.59, and LPIPS of 0.50 for complete scenes and 17.60 PSNR for objects with zero-shot evaluation, hence demonstrating an absolute improvement of 1.89 PSNR, 0.11 SSIM, 0.13 LPIPS for complete scenes and 1.90 PSNR for novel objects against the best-performing baseline for a single-map scenario~(i.e. all methods trained on 25 scenes and evaluated on a single novel scene within the same map). Our approach also outperforms the best baselines on the challenging multi-map dataset with zero-shot evaluation, where 5 novel scenes were held out with difficult illuminations and shadows. NeO 360 achieves an absolute PSNR improvement of 2.39 on complete scenes and 2.93 on objects, showing that NeO 360 learns better priors for unbounded scenes. The table also shows both NeO 360 and PixelNeRF perform better than the original NeRF since NeRF has not seen any scene level prior and is optimized per scene from only a few images.

\renewcommand{\arraystretch}{0.8}
\begin{table}[t]
\centering
\caption{\label{tab:ablation_architecture} \small{Effect of different design choices in our architecture for 3-view novel view synthesis on NeRDS 360 dataset~(Single Map)}}
\resizebox{0.80\textwidth}{!}{%
\begin{tabular}{@{}lccc@{}}
\toprule
 Method & \textbf{PSNR} $\!\uparrow$ & \textbf{SSIM}
 $\!\uparrow$ & \textbf{LPIPS} $\!\downarrow$ \\
\cmidrule(r){1-1} \cmidrule(lr){2-4}
with Colors & 16.29 & 0.44 & 0.62\\
w/o Feature Grid  & 17.50 & 0.47 & 0.59\\
w/o Near/Far & 19.02 & 0.57  & 0.52  \\
\midrule
Ours (global + local) & \textbf{19.35} & \textbf{0.59} & \textbf{0.50}\\ 
\bottomrule
\end{tabular}
}
\end{table}
\textbf{Ablation Analysis:} We further show the effect of our design choices in Table~\ref{tab:ablation_architecture}. We show that directly using colors from source views results in a much worse performance of our model due to over-reliance on source view pixels which hurts generalization to out-of-distribution camera views.  Additionally, we ablate the feature grid as well as the near/far MLPs. The results confirm that performance degrades without the use of the 3D feature grid i.e. global features~(row 2) which is one of our major contributions. The near/far decomposition has a less significant but still positive effect. Overall, our model with combined local and global features in the form of triplanes performs the best among all variants.

\begin{figure*}[ht!]
\centering
\includegraphics[width=1.0\textwidth]{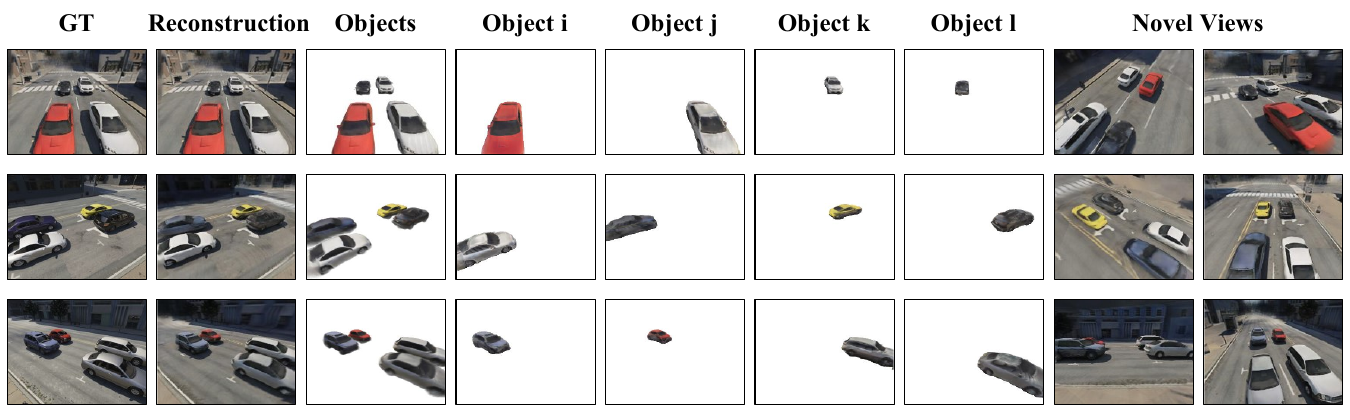}
\captionof{figure}{
\textbf{Scene decomposition qualitative results} showing 3-view scene decomposed individual objects along with novel views on the NeRDS 360 dataset. Our approach performs accurate decomposition by sampling inside the 3D bounding boxes of the objects; hence giving full control over object editability from very few input views.
}
\label{fig:qualitative_main}
\end{figure*}

\textbf{Effect of Scene Priors:}
Table~\ref{table:overfitting_objects} shows NeO 360's ability to overfit to a large number of scenes, hence achieving better PSNR than PixelNeRF on sampling from prior distributions for novel trajectories. To further validate the effect of scene priors, we omit scene priors from our approach's training, and the results are summarized in Table~\ref{tab:object_discovery}. The results further confirm that scene priors actually help our network, resulting in an absolute PSNR improvement of 4.08 comparing the network learned on 25 scenes with the network which has seen no prior scenes during training and is only overfit on 3 unseen views from scratch for a novel scene.  The results show that the architecture design of our network allows us to learn from a large collection of scenes while extending the learned prior to novel scenes with effective zero-shot generalizability from a few views.

\begin{table}[t]
\begin{minipage}{.48\columnwidth}
\centering
\captionof{table}{Quantitative comparison of 3-view view synthesis on NeRDS 360}
\resizebox{0.99\columnwidth}{!}{
\begin{tabular}{@{}lcc@{}}
\toprule
Comparison       & \textbf{PSNR} ↑ & \textbf{SSIM} ↑  \\ 
\midrule
mipNeRF360~\cite{barron2022mip}      & 13.25 & 0.31  \\
SRT~\cite{srt22} & 14.61 & 0.40\\
EG3D*/GAUDI*~\cite{chan2022efficient}~\cite{bautista2022gaudi}   & 12.84 & 0.30 \\

\midrule
\textbf{NeO 360 (Ours)}   & \textbf{19.35} & \textbf{0.59}\\
\bottomrule
\end{tabular}
}
\label{tab:add_comparison}
\end{minipage}
\centering
\hfill\begin{minipage}{.48\columnwidth}
\centering
\captionof{table}{Eval views depth prediction from 3 source views on NeRDS 360}
\resizebox{0.99\columnwidth}{!}{
\begin{tabular}{@{}lcc@{}}
\toprule
Object Depth & \textbf{L1} ↓ & \textbf{RMSE} ↓  \\ 
\midrule
PixelNeRF~\cite{yu2021pixelnerf} & 0.83 & 1.07  \\
NeO 360~(no ft.) & 0.59 & 0.74\\
\midrule
\textbf{NeO 360~(Ours)}   & \textbf{0.20} & \textbf{0.61}\\
\bottomrule
\end{tabular}

}
\label{tab:depth_metrics}

\end{minipage}
\vspace{-16pt}
\end{table}

\begin{figure}[!b]
    \centering
    \resizebox{0.95\linewidth}{!}{%
\includegraphics{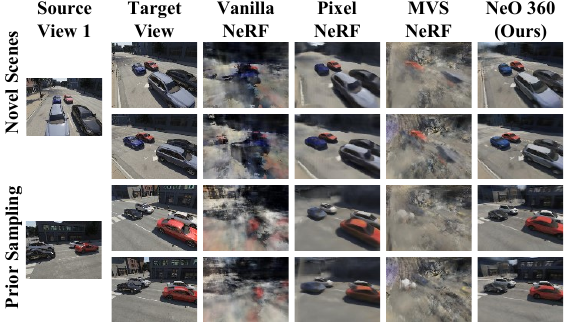}
    }
    \caption{\textbf{Qualitative 3-view view synthesis results}: Comparisons with baselines.
    }
    \label{fig:qualitative_all_baselines}
\end{figure}%

\textbf{Additional baseline comparisons:} We include additional comparisons with novel-view synthesis baselines~(mipNeRF360~\cite{barron2022mip}, EG3D~\cite{chan2022efficient} and SRT~\cite{srt22}) in Tab.~\ref{tab:add_comparison}. One could clearly observe that naively using triplanes~(Tab.~\ref{tab:add_comparison} row 3, *denotes that we take the triplanar representation without generative losses or training) or local features as in PixelNeRF~(Tab.~\ref{tab:ablation_architecture})  hurt the performance. Our method relies on an effective combination of local and global features which serves as a strong baseline for the challenging task of 360$^{\circ}$ view synthesis of outdoor scenes. We also include depth reconstruction metrics~(Tab.~\ref{tab:depth_metrics}) and show our techniques' superior results compared to Pixel-NeRF.

\subsection{Implementation details:}
\label{sec:implementation_details_neo}
\textbf{Sampling rays}: We scale all samples in the dataset so that cameras lie inside a unit hemisphere and use near and far values of 0.02 and 3.0 respectively. We use 64 coarse and 64 fine samples to sample each ray. 

\textbf{Training procedure}: To optimize NeO 360, we first sample 3 source images from one of the 75 scenes in the training dataset. For our initial training phase, we sample 20 random destination views different from the source images used for encoding the NeO 360's network. We sample 1000 rays from all 20 destination views. We use these randomly sampled rays to decode the color and density for each of the 1000 rays. This training strategy helps the network simultaneously decode from a variety of camera distributions and helps with network convergence. We do this by sampling two different sets of points~i.e. one for each near and far background MLP, as employed in~\cite{zhang2020nerf++}. These points samples differ based on the intersection between the origin of rays and the unit sphere. 

\textbf{Loss function and optimizer:} For the first training phase, we employ a mean squared error loss on predicted color and target pixels at the sampled point locations in the ground-truth images, as discussed in Section~\ref{sec:decoding}. We also add a regularization penalty~(Eq.~\ref{eq:regularization}) to encourage the weights to be sparse for both near and far background MLPs, as proposed in~\cite{barron2021mip}. For our second training phase, we select a single destination view and sample~$40 \times 40$ patches of target RGB for training the network using an additional perceptual similarity loss, as described in Section~\ref{sec:decoding} with a~$\lambda$ value set to 0.3.${L}_{LPIPS}$ loss encourages perceptual similarity between patches of rendered color, $c^{p}$ and ground color~$\tilde{c}_t$, where we only enforce it after 30 training epochs to improve background modeling. We optimize the network for 100 epochs in total and employ early stopping based on the validation dataset, which is a subset of the training dataset with different viewpoints than the training camera distribution. We use an Adam optimizer with an initial learning rate of~$5.0e^{-4}$ and a learning-rate ramp-up strategy to increase the learning rate from~$5.0e^{-5}$ to the value~$5.0e^{-4}$ and then decrease it exponentially to a final learning rate~$5.0e^{-6}$. 

\textbf{Compute:} We train the model end-to-end on 8 A-100 Nvidia GPUs for approximately 1 day for network convergence. 

\textbf{Parameters:} Since NeO 360 has the ability to overfit to a large number of scenes, unlike NeRF~\cite{mildenhall2020nerf}, we use a larger model size of 17M parameters. Both ours and NeRF~\cite{mildenhall2020nerf}'s rendering MLP size is the same~(i.e. 1.2M parameters), although our larger model size is attributed to employing ResNet feature block for local features~($\sim$10M parameters) and additional convolutional blocks for tri-planar feature. 

\textbf{Optional fine-tuning:} Although our network gives reasonable zero-shot performance, we also employ an additional finetuning stage using the same few views~(e.g. 1, 3 and 5 source views) to further improve the performance of our network. Note that we employ the same finetuning strategy for the comparing baselines~(Section~\ref{chap:neo360,sec:exp}) and show that the additional finetuning stage improves the performance of both our proposed method and competing baseline, while our approach, NeO 360, still achieves superior overall performance. For our finetuning experiments, we freeze the rest of the network and only the optimize tri-planar network i.e. freezing the encoder~$E$. We employ a lower learning rate of~$5^{10^{-6}}$ to finetune the network from 1,3 or 5 source views.

\begin{equation}
\label{eq:regularization}
\begin{array}{r}
\mathcal{L}_{\mathrm{reg}}(s, w)=\sum_{i=0}^{N-1} \sum_{j=0}^{N-1} w_i w_j|\frac{s_i+s_{i+1}}{2}-\frac{s_j+s_{j+1}}{2}| \\
+\frac{1}{3} \sum_{i=0}^{N-1} w_i^2 (s_{i+1}-s_i)
\end{array}
\end{equation}

\subsection{Experimental Setting Details:}
\label{sec:exp_details}
In this section, we detail our experimental setting to evaluate the effectiveness of our proposed method against the state-of-the-art baselines on the NeRDS360 dataset. We mainly evaluate for~\textbf{a.} Prior-based sampling and~\textbf{b.} Novel-scene rendering. Note that unlike~\cite{bautista2022gaudi} which performs both unconditional and conditional prior-based sampling, our task only considers image-conditional prior-based sampling for~\textbf{a}, since we do not optimize a latent code for each scene and our method does not rely on inference-time GAN-inversion like~\cite{bautista2022gaudi} to find a latent code for a new scene. Rather, our method works in a zero-shot manner reasonably well without any inference time finetuning or inversion, since it takes as input one or few images or a novel scene and is trained as such. We now describe more details about each experimental setting. ~\textbf{a. Prior-based sampling} tests for our network's ability to overfit the training distribution of a large number of scenes. In essence, we keep the evaluation scenes fixed to one of the scenes seen during training and use 1,3, and 5 source camera views as input while decoding from novel camera viewpoints not seen during training. While vanilla NeRF~\cite{mildenhall2020nerf} can do this with many different networks, each optimized from scratch from 100s of views for a new scene, our proposed approach, thanks to its generalizability can overfit to a large number of scenes with just a single network without optimizing a different latent code or vector per-scene, hence demonstrating our network's ability to memorize the training distribution for a large number of scenes seen during training.~\textbf{b. Novel-scene rendering} considers evaluating our approach on a completely new set of scenes and objects never seen during training. We test for our model's ability to generalize well in this scenario which is a core aspect of our approach. This is a more challenging evaluation setup than prior-based sampling since the network has not seen any scenes or objects, neither it has seen these viewpoints during training. Rather, it only relies on the priors learned during training and the few views available during testing~(1, 3, or 5 views in our evaluation setup) to infer the complete 360$^{\circ}$ surroundings of novel scenes.

\section{Qualitative Results:}
\label{chap:neo360,sec:qualitative}

\textbf{Comparison with generalizable NeRF baselines:} As seen in Figure~\ref{fig:qualitative_all_baselines}, our method excels in novel-view synthesis from 3 source views, outperforming strong generalizable NeRF baselines. Vanilla NeRF struggles due to overfitting on these 3 views. MVSNeRF, although generalizable, is limited to nearby views as stated in the original paper, and thus struggles with distant views in this more challenging task, whereas PixelNeRF's renderings also produce artifacts for far backgrounds. 

\begin{figure}[htb!]
    \centering
    \resizebox{0.95\linewidth}{!}{%
\includegraphics{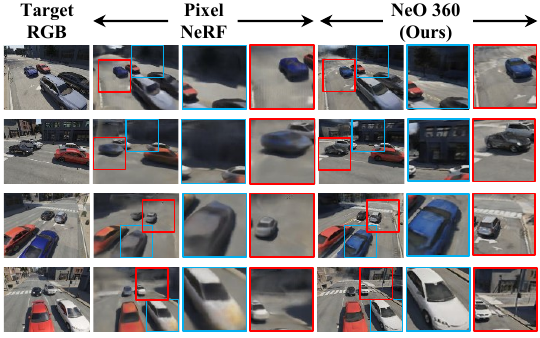}
    }
    \caption{\textbf{Qualitative 3-view view synthesis results}: Close-up comparison with PixelNeRF~\cite{yu2020pixelnerf}.
    }
    \label{fig:qualitative_pixelnerf}
\end{figure}%

\begin{figure}[t]
\centering
\includegraphics[width=1.0\textwidth]{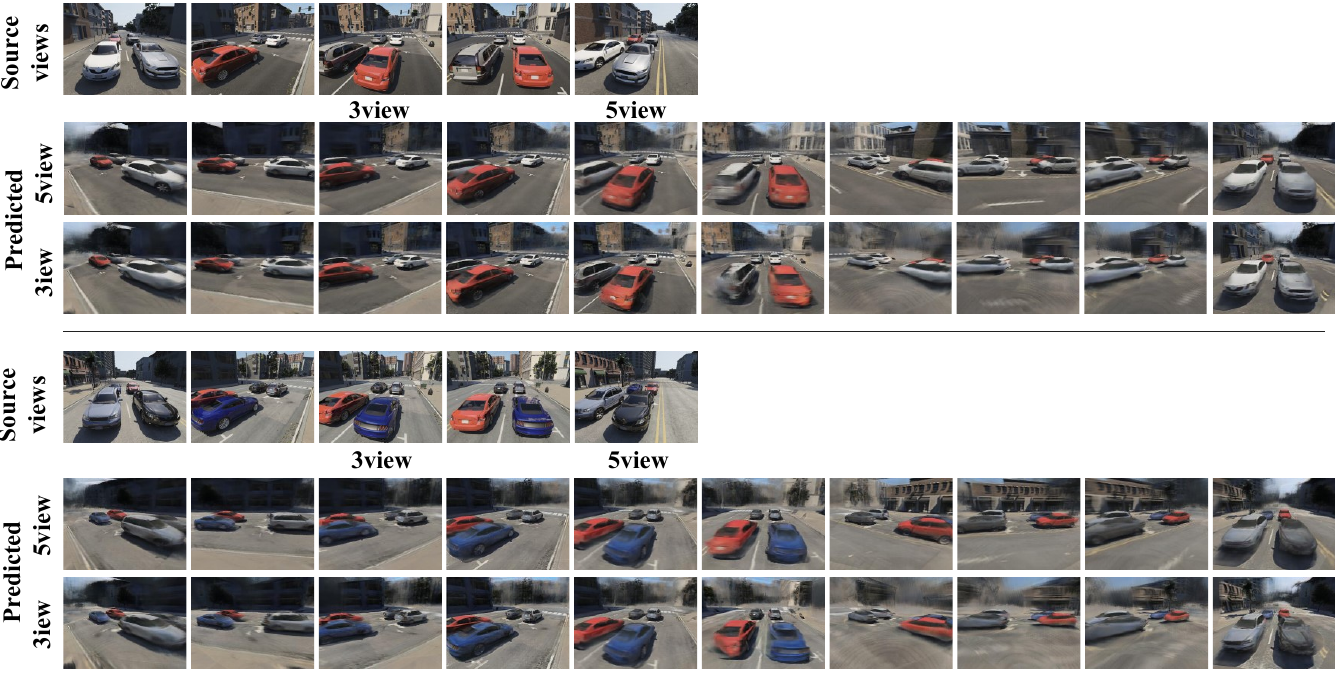}
\captionof{figure}{
\textbf{NeO 360's zero-shot qualitative results:} {We show 360$^{\circ}$ predictions for 3-view and 5-view novel view synthesis. Note that although our network has some shape artifacts for 3-view novel-view synthesis, these are effectively resolved by adding a few more sparse views, showing our network's ability to effectively use learned priors for sparse novel-view synthesis in a zero-shot manner. We show 10 predicted samples with indices 11, 20, 32, 38, 43, 48, 65, 76, 84, and 98 rendered from a circular trajectory generated at a consistent radius around the scene.
}}
\label{fig:360_qualitative}
\end{figure}

\textbf{Detailed comparison with PixelNeRF:} Figure~\ref{fig:qualitative_pixelnerf} presents our method's novel-view synthesis results compared with PixelNeRF. The red and blue boxes focus on close-ups. The visual comparison emphasizes our method's ability to produce crisper and clearer object and background renderings, while PixelNeRF generates blurrier outputs with noticeable artifacts in both foreground and background. 

\textbf{Scene Decomposition:} We further show our network's scene decomposition performance in Figure~\ref{fig:qualitative_main}. The figure demonstrates precise object recovery from the near background MLP output through sampling within each object's GT 3D bounding box (as emphasized in Section~\ref{sec:editing}).  This formulation allows us to easily re-render objects thanks to our feature-based representation which can be queried individually for objects and backgrounds. Note that we do not enforce any objectness prior during training to get this behavior, it is purely learned from multi-view image-based rendering. 

\textbf{360$^{\circ}$ qualitative results:} We further show our network's predicted few-view 360$^{\circ}$ novel view synthesis output in a zero-shot manner on unseen scenes and objects, not observed during training. As shown by Figure~\ref{fig:360_qualitative}, our method performs plausible novel-view synthesis of complete scenes including far-away backgrounds and objects from very few sparse views of an outdoor scene; hence demonstrating 360's ability to use learned priors effectively. We also show that 3-view synthesis introduces some artifacts in parts of the scene where there is no overlap between source views~i.e. where the scene is entirely unobserved. \ref{fig:360_qualitative} shows that by adding a few sparse sets of views in those areas~(i.e.~5-view case), those artifacts can be effectively removed, and a smooth scene representation could be obtained. This shows our network's ability to interpolate smoothly across given source views and also complete the scene in an effective manner.

\subsection{Additional Qualitative Results:}
\label{sec:add_qualitative_results}

\textbf{Results on Kitti-360:} To demonstrate its applicability to real-world data, it's important to capture a comparable dataset to NeRDS360 in a real-world context. While our approach is tailored for a 360$^\circ$ environment, we have successfully adapted NeO 360 for the KITTI-360~\cite{Liao2022PAMI} dataset. This adaptation involves removing the distinction between near and far and employing a single MLP for rendering. Our method employs a source view window of the last 3 frames to render the subsequent frame. Examining overfitting outcomes~(Figure~\ref{fig:KITTI-360}), we observe that our representation achieves significantly improved SSIM and comparable PSNR in contrast to NeRF, when dense views are available for real-world unbounded scenes.

\begin{figure}[t]
\centering
\includegraphics[width=0.80\linewidth]{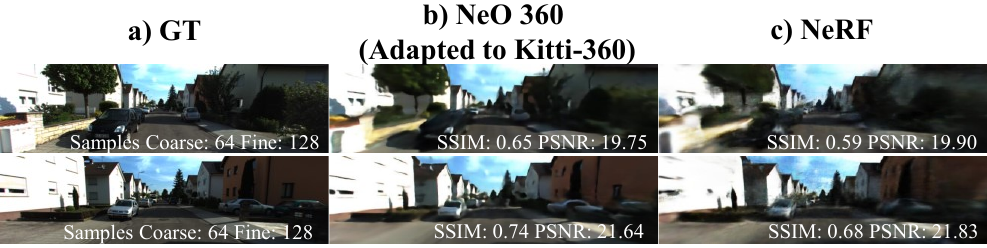}
\captionof{figure}{\textbf{Real-world results:} on KITTI-360~\cite{Liao2022PAMI}, Panoptic NeRF~\cite{fu2022panoptic} test split}
\label{fig:KITTI-360}
\end{figure}
\section{Summary}
\label{chap:neo360,sec:conclusion}

In this section, we proposed NeO 360, a generalizable extension to the NeRF approach for unbounded 360$^{\circ}$ scenes. Our method relies on image-conditional tri-planar representations for few-shot novel view synthesis. In order to build strong priors for unbounded scenes, we propose a large-scale dataset, NERDS 360 to study view synthesis, reconstruction and decomposition in a 360-degree setting. Our method performs significantly better than other generalizable NeRF variants and achieves higher performance when tested on novel scenes. For future work, we will explore how the proposed method can be used to build priors that rely less on labelled data, such as 3D bounding boxes during inference and instead rely on motion cues for effective scene decomposition without labelled data. A second avenue of future work consists of sim2real extensions of this work to alleviate the data and annotations requirement in the real world by using only labelled data in simulation. Specifically, we utilized large-scale priors from synthetic data and utilize them in an end-to-end learning-based pipeline to show improved generalization to real-world scenes with limited real-world fine-tuning. Specifically, we utilized large-scale priors from synthetic data and utilize them in an end-to-end learning-based pipeline to show improved generalization to real-world scenes with limited real-world fine-tuning. Specifically, we utilized~\textit{multi-view 2D geometry constraints} as prior knowledge and effectively lifted them to the 3D domain to show improved few-shot novel view synthesis performance in a challenging outdoor scene task.  
    \chapter{NeRF-MAE: Masked AutoEncoders for Self Supervised 3D representations Learning for Neural Radiance Fields}
\label{chap:nerfmae}

In this chapter, I explore the problem setting of~\textbf{self-supervised 3D representation learning from posed 2D data} with the main aim of improving downstream 3D tasks such as 3D object detection~(as previously discussed in~\ref{chap:centersnap} and~\ref{chap:shapo}), 3D semantic segmentation and voxel super-resolution. We will primarily use the 3D representation discussed in~\ref{chap:neo360} using a self-supervised mask pretraining objective of learning a strong 3D representation from posed 2D images. We will study how learning this 3D representation allows us to improve various downstream 3D tasks such as 3D object detection, voxel super-resolution, and 3D semantic labeling. The main prior knowledge comes from~\textit{contextual and structural similarity} inherent in 3D data. We capture this efficiently by utilizing a training strategy relying on masked auto-encoders to show improved performance on challenging downstream 3D tasks. 

\section{Introduction}
\label{chap:nerfmae,sec:intro}

\begin{figure}[htb]
\centering
\includegraphics[width=1.0\columnwidth]{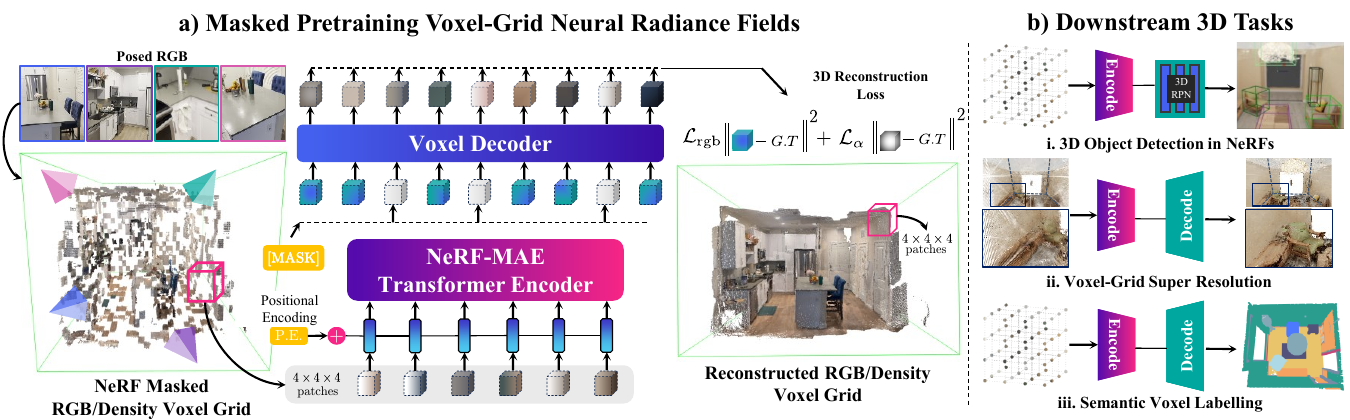}
\centering
  \caption{
    \textbf{Overview:} a) We present~\textbf{NeRF-MAE}, the first large-scale fully self-supervised pretraining for Neural Radiance Fields~(NeRFs). Our approach utilizes a 3D Swin Transformer-based encoder and a voxel decoder to learn a powerful representation in~(\textbf{a}) masked self-supervised learning objective directly in 3D using only posed RGB images as input.~(\textbf{b}) Our representation, when fine-tuned on a small subset of data, improves many 3D downstream tasks such as 3D object detection in NeRFs, voxel-grid super-resolution, and voxel-labelling.
  }
  \label{teaser_nema}
\end{figure}

\begin{table}[!htb]
\small
\centering
\caption{Specifications of supervised or self-supervised approaches to 3D representation learning, highlighting our novel self-supervised 3D pre-training strategy for NeRFs which only requires posed RGB images as input. * denotes multi-view posed RGB images. }
\resizebox{\textwidth}{!}{
\begin{tabular}{cccccccc}
\toprule
Method & PixelNeRF~\cite{yu2020pixelnerf} & NeRFRPN~\cite{hu2023nerf} & PerFception-Scannet~\cite{jeong2022perfception}& Voxel-MAE~\cite{hess2022masked}& PiMAE~\cite{chen2023pimae} & SUNetR~\cite{tang2022self}&\cellcolor{yellow!40}~\textbf{NeRF-MAE}  \\
\midrule
\# Scenes & 40 & 515 & 1500 & 850 & -&5000&\cellcolor{yellow!40}3637 \\
\# Frames & 4k & 80k & 2.5M & - & 10k& - & \cellcolor{yellow!40}1.65M \\
Self-Supervised Pretraining & \textcolor{falured}{\xmark} & \textcolor{falured}{\xmark} & \textcolor{falured}{\xmark} & \textcolor{teal}{\checkmark} & \textcolor{teal}{\checkmark} & \textcolor{teal}{\checkmark} &\cellcolor{yellow!40}\textcolor{teal}{\checkmark} \\
Input Data & RGB* & RGB* & RGB* & Pointcloud & RGB-D& CT Scan & \cellcolor{yellow!40}RGB* \\
3D Representation & NeRF & NeRF & NeRF & Voxel & Pointclouds&Voxel &\cellcolor{yellow!40}NeRF \\
Data domain(s) & Single~(\textcolor{falured}{1}) & Single~(1) & Single~(\textcolor{falured}{1}) & Single~(\textcolor{falured}{1}) & Single~(\textcolor{falured}{1}) & Multiple~(\textcolor{teal}{5}) &\cellcolor{yellow!40}Multiple~(\textcolor{teal}{4})  \\

\bottomrule
\end{tabular}
}

\label{tab:prior_approaches_comparison}
\end{table}

\looseness=-1
Vision Transformers~(\gls{vit})~\cite{Dosovitskiy17} have transformed computer vision in recent years, excelling in pretext task learning~\cite{he2021masked, xie2022simmim, baevski2022data2vec}. Their unique approach, encoding visual representations from patches and leveraging self-attention blocks for long-range global information~\cite{raghu2021vision, zhai2022scaling}, sets them apart from conventional CNNs. ViTs, especially when combined with recent advances in pretraining strategies like masked autoencoders (MAEs)~\cite{he2021masked}, outperform CNNs in robust feature representation learning, demonstrating state-of-the-art results on benchmarks like ImageNet~\cite{deng2009imagenet}. This pre-training advantage translates into significant performance gains during fine-tuning for downstream tasks~\cite{raghu2021vision}.

Unfortunately, these advances in 2D vision have not translated to 3D vision equally well due to a significant domain gap between 3D representations~\cite{pang2022masked, tang2022self, liang2022meshmae} such as meshes and raw RGB data, coupled with the scarcity of high-quality 3D data at scale~\cite{dai2017scannet, deitke2023objaverse}. To address these limitations, we propose a novel method that utilizes RGB images and, by leveraging Neural Radiance Fields (NeRFs), trains a 3D encoder directly through self-supervised learning.

\looseness=-1
Notably, NeRFs~\cite{mildenhall2020nerf, barron2021mip} have exhibited impressive capability in synthesizing novel views by solely utilizing posed RGB images as input. Their ability to capture scenes~\cite{wang2021neus, sucar2021imap, ost2021neural} with high fidelity has led to a surge in NeRF-related works beyond reconstruction, effectively using them for AR/VR~\cite{kuang2022neroic}, robotics~\cite{irshad2022shapo, yen2022nerf}, and driving~\cite{tancik2022block, irshad2023neo360} applications. NeRFs have also emerged as a promising 3D data storage medium~\cite{jeong2022perfception}, and have proven to outperform traditional supervised approaches~\cite{jeong2022perfception, hu2023nerf} for challenging 3D tasks like object detection~\cite{hu2023nerf} and instance segmentation~\cite{liu2023instance}.

Driven by this analysis, we propose a new framework termed~\textbf{NeRF-MAE}~(\textbf{NeRF} \textbf{M}asked \textbf{A}uto \textbf{E}ncoders) for self-supervised learning directly within Neural Radiance Fields. Operating solely on RGB images as input, NeRF-MAE incorporates carefully designed masked autoencoders tailored to NeRF's unique formulation. In comparison to prior generalizable NeRF or 3D representation learning methods~(Table~\ref{tab:prior_approaches_comparison}), NeRF-MAE offers the following distinct advantages:~\textbf{1)} The masked self-supervised learning objective is directly enforced in 3D, i.e.~on pre-trained NeRF grids, eliminating the need for expensive 3D data such as meshes; thus enabling straightforward self-supervised 3D pretraining using 2D data.~\textbf{2)} By disentangling representation learning and NeRF training, our model can effectively utilize a large amount of data from diverse sources. This is in contrast to most previous generalizable NeRF methods that only train on a single type of data source for priors.

\looseness=-1
As shown in Figure~\ref{teaser_nema}, NeRF-MAE consists of a 3D Transformer encoder, and a voxel decoder module. Taking a pre-trained NeRF of a scene as input, we derive a grid of radiance and density, yielding 4D color and density voxels. Subsequently, we partition the 4D voxel grid into smaller patches, randomly masking a substantial ratio to minimize data redundancy. The autoencoder's reconstruction objective is enforced to learn high-level features from the unmasked RGB and density patches and reconstruct the masked color and opacity patches in the canonical world frame. Notably, our encoder's backbone consists of Swin Transformer blocks, extended to the 3D domain, while a voxel decoder is employed for reconstruction.

\looseness=-1
For training our model, we curate a large-scale dataset for NeRF pre-training, encompassing multi-view RGB data from 4 diverse sources in synthetic and real domains. Our dataset comprises multi-view posed images, corresponding pre-trained NeRFs, and sampled color/density voxel grids from Front3D, ScanNet, ARKit Scenes, and Hypersim, accumulating over 1.6M images and 3500+ scenes~(Table.~\ref{tab:prior_approaches_comparison} and Figure~\ref{fig:dataset_fig1}). In essence, our method is the first large-scale fully self-supervised approach for 3D representation learning on Neural Radiance Fields. To summarize, we make the following contributions:

\begin{itemize}
   \item To our knowledge, we introduce the first~\textbf{fully self-supervised transformer-based 3D architecture} for pre-training of Neural Radiance Fields.
   \item A~\textbf{large scale pre-training of Neural Radiance Fields} on 4 different datasets, over 1.6M images, and 3500+ indoor scenes, using~\textbf{a single model}.
   \item Our proposed approach~\textbf{significantly outperforms baselines on multiple downstream 3D tasks}, showing over 20\% AP50 and 12\% mAcc on 3D object detection and semantic voxel labelling respectively on Front3D dataset.
\end{itemize}
\section{Related Work} 
\label{chap:nerfmae,sec:relatedworks}

\textbf{3D Representation Learning:}
Supervised Learning for 3D data has achieved great success recently with sophisticated architectures developed for point-cloud analysis~\cite{Guo_2021,qi2017pointnet, qi2017pointnet++, zhang2023nearest, irshad2022centersnap}, implicit-representation based methods~\cite{mescheder2019occupancy, park2019deepsdf, irshad2022shapo, mittal2022autosdf, heppert2023carto}, learning on voxels for dense 3D prediction tasks~\cite{dai2017scannet, liu2021voxel, zhou2018voxelnet} as well as learning representations for 3D mesh analysis~\cite{verma2018feastnet, liu2023meshdiffusion}. Although powerful, these techniques assume known 3D data as well as annotated labeling in 3D to achieve great performance for a variety of 3D tasks. Contrastively, Masked Auto-Encoders have emerged as a way to learn strong representations for computer vision tasks without needing labeled data. Starting with initial works like~\cite{he2021masked} for 2D representation learning, follow-up works extended the idea of reconstructing 2D masked patches to the 3D domain by directly reconstructing
masked 3D coordinates with point clouds as representation~\cite{yu2022point, chen2023pimae, pang2022masked, zhang2022point}. Other works reconstruct mesh patches~\cite{liang2022meshmae} or study auto-encoding point clouds with redundant spatial information by transforming the point clouds into voxel representations and classifying whether the voxel contains point clouds~\cite{hess2022masked}. Fewer efforts have been dedicated to studying building 3D representations directly from 2D data or 2D priors. Some recent notable works~\cite{zhang2023learning, liu2021learning, sautier2022image} utilize 2D to 3D learning paradigm by either learning from pairwise 2D-3D data for contrastive learning or learning 2D knowledge to guide 3D mask auto encoding. These works inevitably still require 3D data for pre-training. In contrast, our proposed NeRF-MAE utilizes only 2D data for pre-training 3D representations and does not require any 3D data in synthetic simulation or real-world for pre-training our 3D representations.

\textbf{Vision Transformers:}
Recently, computer vision has undergone a transformative shift with the emergence of vision transformers, garnering significant interest. Unlike conventional CNNs that process images using a sliding window approach, Vision Transformers (ViT) break down the image into 16x16 pixel patches~\cite{dosovitskiy2020image, raghu2021vision}. The Swin Transformer~\cite{liu2021swin} has achieved remarkable results by adopting a hierarchical transformer architecture on prominent vision benchmarks. The remarkable performance of visual transformers in various vision tasks has sparked interest in exploring methods to pre-train these backbone models. Two studies, namely MoCoV3~\cite{chen2021mocov3} and MAE~\cite{he2021masked}, delved into different aspects of self-supervised Vision Transformers (ViT). Particularly, MAE~\cite{he2021masked} achieved state-of-the-art outcomes by drawing inspiration from BERT~\cite{devlin2018bert}, which involves randomly masking words in sentences and using masked image reconstruction for self-supervised pre-training. Subsequent research expanded on MAE's contributions by introducing depth priors~\cite{bachmann2022multimae} for MAE pre-training, extended the concept of mask auto-encoding to the 3D domain~\cite{pang2022masked, tang2022self, liang2022meshmae, hess2022masked}, leading to impressive results in downstream 3D vision tasks.

\textbf{NeRF as 3D Scene Representation:} Previous implicit representation methods, such as DeepSDF~\cite{park2019deepsdf}, ROAD~\cite{zakharov2022road}, ShaPO~\cite{irshad2022shapo}, and Convolutional Occupancy Networks~\cite{peng2020convolutional}, have achieved impressive results in 3D shape representation and scene completion. However, these methods heavily rely on ground truth supervision in the form of 3D meshes or point clouds. More recently, advances in differentiable neural rendering have allowed for learning 3D representations using only image supervision~\cite{sitzmann2019scene, mildenhall2020nerf}. Neural Radiance Fields (NeRFs) have emerged as a particularly powerful approach, excelling in novel view synthesis. Several extensions to NeRFs have been proposed to address specific challenges. These include reducing aliasing effects through multiscale representations~\cite{barron2021mip}, modeling unbounded scenes~\cite{martinbrualla2020nerfw}, creating disentangled object-background representations~\cite{ost2021neural}, employing compositional generative models~\cite{Niemeyer2020GIRAFFE}, and improving reconstruction and depth estimation accuracy through multi-view consistent features~\cite{stier2021vortx, sun2021neuralrecon, guizilini2022depth}. However, little attention has been given to pretraining on NeRFs and exploring self-supervised learning on Neural Radiance Fields. This work aims to fill this gap by focusing on self-supervised pretraining for neural radiance fields. The main goal is to leverage only posed 2D data to train 3D representations and investigate its impact on various downstream 3D vision tasks.

\textbf{Meta Learning:} 

Meta-learning~\cite{karunaratne2021robust, finn2017model} has demonstrated its effectiveness in achieving robust generalization to unfamiliar data by employing sample-efficient architectures within the few-shot paradigm. Although highly successful for few-shot learning and sharing some similarities with the transfer learning evaluation strategy of masked autoencoders, there are key distinctions with our approach. Firstly, meta-learning approaches involve training a hypernetwork to adjust the weights of task-specific neural networks in a hierarchical manner, whereas our approach relies on masked autoencoders and utilizes a single neural network for pretraining. During the fine-tuning stage, we discard the decoder and update the weights of the encoder for downstream tasks transfer learning. Secondly, meta-learning approaches like MAML evaluate in a few-shot manner, whereas we provide all the information for downstream tasks in transfer learning.

\section{NeRF-MAE: Self-Supervised Pretraining for Neural Radiance Fields}
\label{chap:nerfmae,sec:method}

\begin{figure}[t]
    \centering
    \resizebox{0.90\linewidth}{!}{%
\includegraphics{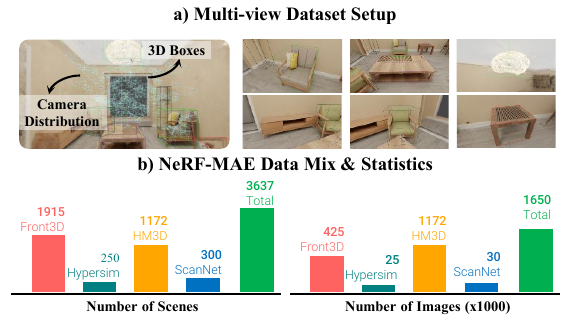}
    }
    \caption{~\textbf{NeRF-MAE~\cite{irshad2024nerfmae} dataset mix for pretraining:} showing~\textbf{a.} Multi-view dataset setup and the corresponding camera distribution and 3D boxes used for pre-training NeRF and~\textbf{b.} Diverse scenes from 4 different data sources i.e. Front3D~\cite{fu20213d}, Hypersim~\cite{roberts2021hypersim}, Habitat-Matterport3D~\cite{ramakrishnan2021hm3d} and ScanNet~\cite{dai2017scannet} totaling over 3500 scenes and over 1.7M images used for pre-training neural radiance fields using our single model i.e. NeRF-MAE.
    }
    \label{fig:nerfmaedataset_fig1}
\end{figure}%

\begin{figure}[t]
    \centering
    \resizebox{0.95\linewidth}{!}{%
\includegraphics{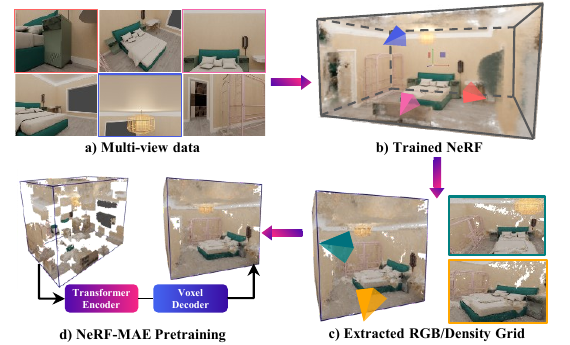}
    }
    \caption{\textbf{NeRF-MAE~\cite{irshad2024nerfmae} data processing flow for pre-training:} showing~\textbf{a.} multi-view training data,~\textbf{b.} trained NeRF representation,~\textbf{c.} extracted RGB/Density grid from the trained NeRF and~\textbf{d.} masked pre-training of the voxel-grid neural radiance field.
    }
    \label{fig:nerfmae_dataset_fig2}
\end{figure}%

\begin{figure}[t]
\centering
\includegraphics[width=1.0\textwidth]{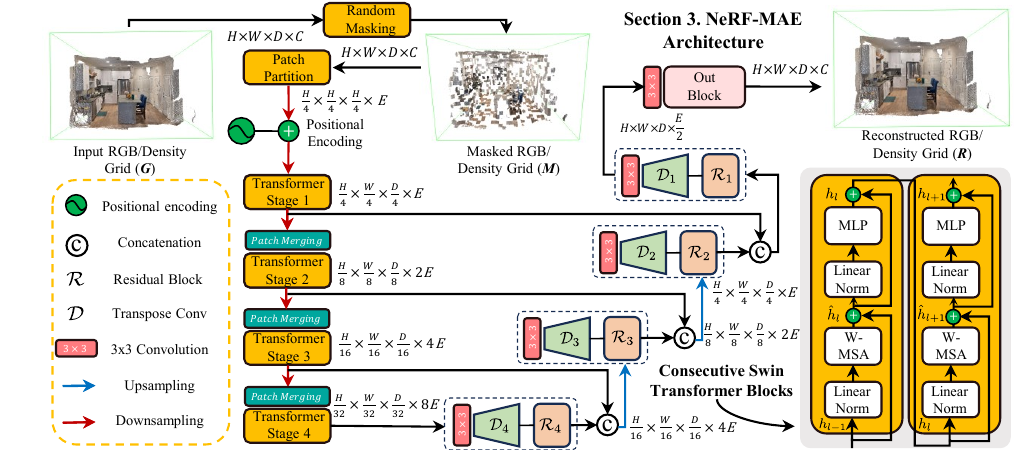}
\captionof{figure}{
\textbf{NeRF-MAE~\cite{irshad2024nerfmae} Architecture:} {Our method utilizes a U-Net~\cite{ronneberger2015u} style architecture employing Swin Transformers~\cite{liu2021swin} as the encoder to encode the RGB/density grid into meaningful multi-resolution low-level features and transposed convolution layers at each stage with skip connections using residual blocks from the features of the encoder.
}}
\label{fig:nerfmae_architecture}
\end{figure}

NeRF-MAE introduces a self-supervised framework aimed at enhancing 3D representation learning in the context of Neural Radiance Fields~\cite{mildenhall2020nerf} (NeRFs). Our approach allows learning robust 3D representations solely from posed RGB images, leading to substantial improvements in various downstream 3D applications, as we show in Section~\ref{chap:nerfmae,sec:exp}. As shown in Figures~\ref{teaser_nema} and~\ref{fig:nerfmae_dataset_fig2}, such a goal is made possible by two major components:~\textbf{(i)} An explicit 4D RGB/Density extraction module in the canonical world frame, from a fully-trained implicit NeRF model~(described in Section~\ref{subsec:nerf_4d_grid}) and~\textbf{(ii)} A novel, masked self-supervised pretraining module, operating directly on the explicit NeRF's 4D radiance and density grid, to train a SwinTransformer~\cite{liu2021swin} encoder and a voxel decoder using a masked reconstruction-based objective in 3D~(discussed in Section~\ref{subsec:masked_pretraining_nerf}). Our NeRF pretraining dataset is described in Section~\ref{subsec:nerf_pt_dataset}.

\begin{figure}[t]
    \centering
    \resizebox{0.75\linewidth}{!}{%
\includegraphics{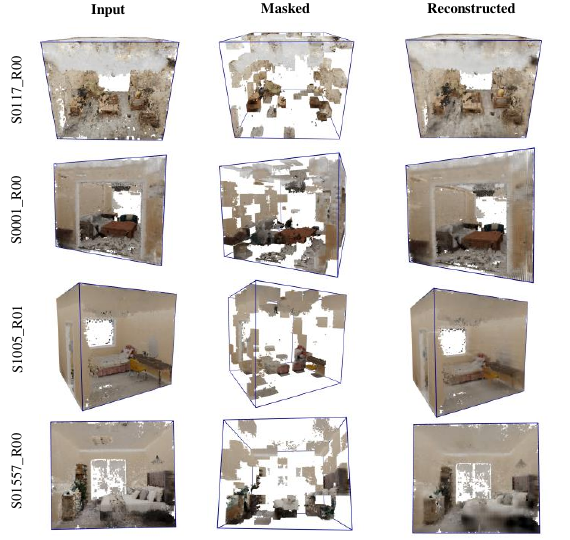}
    }
    \caption{~\textbf{Qualitative NeRF-MAE Reconstruction results:} Examples reconstruction results from Front3D dataset. For each triplet, we show ground truth (left), masked RGB-Density Voxel Grid (middle) and our NeRF-MAE Reconstruction overlayed with unmasked GT Voxel Grid (Right). The masking ratio is 75\%, leaving only 250 patches out of 1000 patches.
    }
    \label{fig:nerfmaequalitative_recon}
\end{figure}%

\subsection{Neural Radiance Field Grid Representation:}
\label{subsec:nerf_4d_grid}

\textbf{Preliminaries - NeRF and InstantNGP:} Given posed 2D images, NeRF represents 
a 3D scene implicitly. It uses a neural network, $f(\mathbf{x}, \theta)$, to predict color ($\mathbf{c}_i$) and density ($\sigma_i$) at any given 3D query position~($x_i$) and viewing direction~($\theta_i$) as input. The 4D color and density outputs are used in an alpha compositing process to generate rendered images through volume rendering with near and far bounds $t_{n}$ and $t_{f}$, as highlighted in the equations below:

\begin{equation}
C(\mathbf{r})=\int_{t_n}^{t_f} T(t) \sigma(\mathbf{r}(t)) \mathbf{c}(\mathbf{r}(t), \mathbf{d}) d t
\end{equation}

\looseness=-1
\noindent where $T(t)=\exp \left(-\int_{t_n}^t \sigma(\mathbf{r}(s)) d s\right)$ and $\boldsymbol{r}$ denotes the camera ray. Although our work is suited to any input NeRF formulation, we choose Instant-NGP~\cite{muller2022instant} to model the radiance and density of a 3D scene using a sparse occupancy grid. A fast ray marching routine is utilized by ray compacting based on occupancy grid values. We also utilize a multi-resolution hash encoding from~\cite{muller2022instant} for faster NeRF training. An example trained scene is shown in Figure~\ref{fig:nerfmae_dataset_fig2}~(b).

\textbf{Radiance and Density Grid Sampling from Instant-NGP NeRF:} NeRF-MAE reconstructs input masked NeRF volumes to pretrain 3D Swin Transformers. The first step in this process is to uniformly sample a radiance and density feature volume from a trained NeRF model~(Figure~\ref{fig:nerfmae_dataset_fig2}), in our case, Instant-NGP NeRF~\cite{muller2022instant}. Querying a pre-trained NeRF for radiance and density information on a regular grid allows~\textbf{1.} extracting an explicit representation that captures the original 3D scene in a compact manner,~\textbf{2.} is invariant to the NeRF formulation used, and~\textbf{3.} opens up the possibility of utilizing existing deep-learning architectures developed for 3D tasks, such as 3D RPN~(Section~\ref{subsec:downstream_tasks}) for detection and segmentation~\cite{hu2023nerf, liu2023instance}. To sample radiance and density information, we query each trained NeRF model from all the training cameras in the traceable scene volume and average the resulting output. 

Formally, let $\mathcal{G} \in \mathbb{R}^{H \times W \times D \times 4}$ be a 4D grid representing the scene volume~(Figure~\ref{fig:nerfmae_dataset_fig2}~c), where, $H,W,D$ denote the height, width, and depth dimension. For a grid point $(i, j, k)$ on a spatial 3D grid, the 4 channel values for each grid point $(i, j, k)$ are the mean of the values obtained by this function $f(x, \theta)$ for all viewing directions. Mathematically, we can write this as:

\begin{equation}
(r_{i, j, k}, g_{i, j, k}, b_{i, j, k}, \alpha_{i, j, k}) = \frac{1}{N} \sum_{\theta=1}^{N} f(x_{i, j, k}, \theta)
\end{equation}
\looseness=-1

\noindent where $\alpha$ can be obtained from volume density~($\sigma_{x}$) as $\alpha=1-\exp (-\sigma_i\|\boldsymbol{x}_i-\boldsymbol{x}_{i+1}\|)$ and ~$\boldsymbol{x}_i-\boldsymbol{x}_{i+1}\ =0.01$, is a small preset distance~\cite{hu2023nerf}. Similar to~\cite{hu2023nerf, liu2023instance}, we determine the traceable volume of the scene by enlarging the axis-aligned bounding box encapsulating all cameras and objects~(where available) in the scene.

\subsection{NeRF Pretraining Datasets}
\label{subsec:nerf_pt_dataset}
A key component of our approach is employing a single model to pre-train 3D representations from NeRFs on a large number of scenes with varying complexity and realism. As shown in Figure~\ref{fig:nerfmaedataset_fig1}, we train our representation in a fully self-supervised manner on the following dataset mix:
\begin{itemize}
    \item \textbf{Front3D}~\cite{fu20213d}: A large-scale synthetic dataset containing 18k rooms with 3D objects. We extend~\cite{fu20213d} to render 5x as many scenes, totaling 1.2k scenes comprising 425k images for our pretraining. The rendering time of this dataset is 6 days on 8 Nvidia 100 GPUs.
    \item \textbf{HM3D}~\cite{ramakrishnan2021hm3d}: A large-scale real2sim dataset of 1000 scans. We render approximately 1.1k scenes comprising 1.2M images using Habitat~\cite{habitat19iccv}. Following~\cite{hong20233d}, we collect the data by randomly selecting 80 navigable points within each room's bounding box. At each navigable point, we rotate the agent 360$^{\circ}$ to render 12 frames with a $[10^{\circ},-10^{\circ}]$ elevation change. The rendering time of this dataset is around 2 days on 8 Nvidia A100 GPUs.
    \item \textbf{Hypersim}~\cite{roberts2021hypersim}: A synthetic dataset with real-world realism and 3D annotations. We use the rendered images~(250 scenes, 25k images) from~\cite{roberts2021hypersim} for our NeRF pretraining as they provide high-quality poses for reconstruction.
    \item \textbf{ScanNet}~\cite{dai2017scannet}: A real-world dataset for indoor 3D scene understanding. We keep this dataset for our cross-dataset transfer experiments and do not use this for NeRF pretraining. 90 out of the 1500 scenes are chosen, similar to~\cite{hu2023nerf} and we use the feature grid extracted from dense-depth-prior NeRF~\cite{roessle2022dense} for our downstream task experiments, which further confirms the generalizability of our approach to various NeRF variants.
\end{itemize}
\looseness=-1
For all datasets, extensive cleaning of the bounding box as well as semantic annotations was performed based on manual filtering, class filtering, and size filtering. This cleaning was only carried out for downstream task fine-tuning and not carried out for our pretraining. We will release the complete dataset, upon acceptance, for reproducibility. 

\begin{figure}[t]
    \centering
    \resizebox{0.85\linewidth}{!}{%
\includegraphics{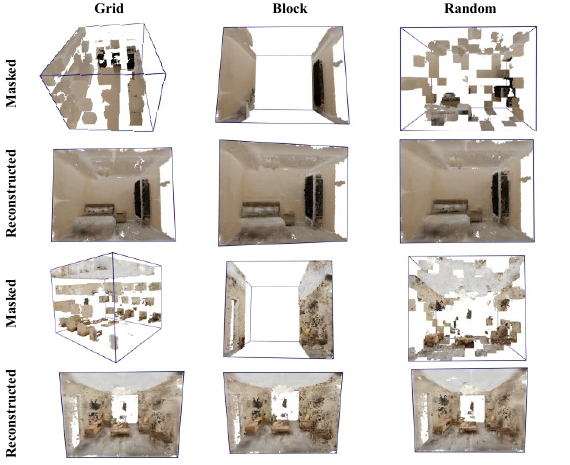}
    }
    \caption{~\textbf{Qualitative NeRF-MAE Reconstruction results:} showing different masking strategies along with the reconstructed output.
    }
    \label{fig:nerfmaequalitative_recon_2}
\end{figure}%

\subsection{Masked Pretraining NeRFs}
\label{subsec:masked_pretraining_nerf}
NeRF-MAE comprises a Swin Transformer encoder~($\mathbb{E}$) and transposed convolution decoders connected with skip connection from the encoded output at multiple feature resolutions of the transformer. Our architecture is summarized in Figure~\ref{fig:nerfmae_architecture}. The aim of the U-Net style architecture is to reconstruct masked 3D radiance and opacity patches. In doing so, the network learns to understand the semantic and spatial structure of the 3D scene solely from posed 2D images which were used to obtain the 3D radiance and density grid~(Section~\ref{subsec:nerf_4d_grid}). Our U-Net~\cite{hatamizadeh2022unetr} style architecture for NeRF-pretraining is described below.

\subsubsection{NeRF-MAE Transformer Encoder:}
The goal of the encoder is to encode the input 3D radiance and density grid~($\mathcal{G}$) into meaningful features at multiple resolutions which can be decoded and used for downstream tasks~(Section~\ref{subsec:downstream_tasks}). We employ the Swin Transformer~\cite{liu2021swin} and extend it to the 3D domain by replacing all the 2D operations of the SwinTransfomer with their 3D equivalent~\cite{hu2023nerf}; for example, 3D convolutions and 3D patch merging. Considering the input to the SwinTransformer be the radiance and density grid~($\mathcal{G} \in \mathbb{R}^{H \times W \times D \times 4}$), we first mask out random patches of size~(P = $p\times p \times p$), where~$p=4$, with a masking ratio of~$m=0.75$ to get the masked radiance and density grid~$\mathcal{M}$. Subsequently, we use~$P$ to divide the masked radiance and density grid into patches to create a series of 3D tokens of size~$\frac{H}{p}\times \frac{W}{p} \times \frac{D}{p}\times E$ using a 3D convolutional operation, where we merge patch partitioning and linear embedding of each patch into one single step. Next, we add 3D positional embedding to each patch, without a classification token, to assign each token to a unique representation. These tokens are further processed by four different stages of shifted window attention and patch merging, where the input patches are subdivided into non-overlapping windows of size~$W \times W \times W$, where~$W=4$, with local multi-head self-attention~(W-MSA-3D) and shifted multi-head self-attention~(SW-MSA-3D) performed for initial and subsequent layers in each block respectively. Each Swin Transformer 3D block also contains a 2-layer~$MLP$ with GeLU activation as non-linearity in between following shifted window-based self-attention modules. A linear norm~($LN$) layer follows every~$\text{W-MSA-3D}$ and~$MLP$ with residual connection applied after each module. The computation of self-attention within each~$\text{W-MSA-3D}$ and~$\text{SW-MSA-3D}$ window is computed as follows:

\begin{equation}
\mathcal{A}(\boldsymbol{Q}, \boldsymbol{K}, \boldsymbol{V}) = \operatorname{softmax}\left(\frac{\boldsymbol{Q} K^T}{\sqrt{d_k}} + B\right) \boldsymbol{V},
\end{equation}

\noindent where the Attention output~($\mathcal{A}$) is a weighted sum of the values~($V$) calculated using a similarity between projected Query~($Q$) and Key~($K$) with a bias~$B \in \mathbb{R}^{M^{3} \times M^{3}}$, where $Q,K,V \in \mathbb{R}^{M^{3} \times d}$ and $d =E/3$. Mathematically, we can write the resulting output of two consecutive layers in each Swin Transformer 3D block as:

\begin{equation}
\begin{aligned}
& \boldsymbol{h}_{t_1}=\boldsymbol{h}_{l-1}+\text{W-MSA-3D}\left(L N\left(\boldsymbol{h}_{l-1}\right)\right) \\
& \boldsymbol{h}_{t}=\boldsymbol{h}_{t_1}+M L P\left(LN\left(\boldsymbol{h}^{t_1}\right)\right) \\
& \boldsymbol{h}_{t_2}=LN\left(\boldsymbol{h}_{t}\right)+\text{SW-MSA-3D}\left(LN\left(\boldsymbol{h}_{t}\right)\right) \\
& \boldsymbol{h}_{l}=LN\left(\boldsymbol{h}_{t_2}\right)+M L P\left(LN\left(\boldsymbol{h}_{t_2}\right)\right)
\end{aligned}
\end{equation}

\noindent Due to the quadratic complexity in computing global self-attention for standard vision transformers, we employ shifted window attention~\cite{liu2021swin} by displacing the windows by $[ \frac{W}{2}, \frac{W}{2}, \frac{W}{2}]$ volume every layer. A patch merging layer follows every transform block, except for the last one, which merges four volumetric patches, so a reduction by a factor of 2 in height, width, and depth of the input volume occurs. A Linear Norm~($LN$) and an MLP layer follow patch merging to increase the feature dimension by a factor of 2. We use an embedding dimension~$E=96$, Swin block depths of~$d=[2,2,18,2]$ and number of heads $[3,6,12,24]$ for each~$\text{W-MSA-3D}$ and~$\text{SW-MSA-3D}$ blocks. 

\subsubsection{Reconstruction with Lightweight Decoders:}
\label{subsec:convolution_decoders}

We utilize the features pyramid~$\mathcal{F} = [f_{i} ... f_{n}]$ obtained after concatenating the resulting output of each SwinTransformer 3D Stage to obtain the reconstructed radiance and density grid~($\mathcal{R}$), where~$i=0$ and $n=3$. The reconstructed grid is obtained by attaching lightweight transposed 3D convolution decoders~($\mathcal{D}_{i}$) to the feature outputs with a kernel size of 3 and adding a residual connection from the previous decoder, where~$d_{i} = Conv(d_{i-1} + \mathcal{D}(f_{i}))$ and~$\mathcal{D}_{0} = f_{3}$. The output of the final decoder block is fed into the residual block with~$3 \times 3 \times 3$ convolutional and a sigmoid activation function to predict a 4-channel reconstructed volume~$\mathcal{R} \in \mathbb{R}^{H \times W \times D \times 4}$. We use the same feature pyramid~$\mathcal{F}$ to output 3D object bounding boxes, voxel semantic labels, and super-resolution voxels with task-specific decoders, as described in Section~\ref{subsec:downstream_tasks}.

\paragraph{Masked NeRF Pretraining Objective:}
The goal of masked voxel-grid pretraining as applied to Neural Radiance Field~(NeRF) is to encode semantic and spatial regions of interest into the network. We use mask volume region reconstruction, similar to~\cite{devlin2018bert} for language and~\cite{he2021masked} for images. We enforce a faithful and accurate reconstruction of masked patches with a custom loss function suited to NeRF's unique formulation. Specifically, we employ a combination of opacity and photometric reconstruction loss, both enforced at the volumetric level, where loss~$L_{recon} = \mathcal{L}_{\text{rgb}} + \mathcal{L}_{\alpha}$ is defined as:

\begin{equation*} \mathcal{L}_{\boldsymbol{rgb}} = \frac{1}{N} \sum_{i=1}^{N} (\hat{\boldsymbol{y}}_{\text{rgb}_i} - \boldsymbol{y}_{\text{rgb}_i})^2 \quad \mathcal{L}_{\boldsymbol{\alpha}} = \frac{1}{M} \sum_{j=1}^{M} (\hat{\boldsymbol{y}}_{\alpha_j} - \boldsymbol{y}_{\alpha_j})^2 \end{equation*}

\noindent where~$N$ is the number of voxels in the mask corresponding to patches where~$\alpha > \boldsymbol{x}_i-\boldsymbol{x}_{i+1}$. We use a small preset distance, ~$\boldsymbol{x}_i-\boldsymbol{x}_{i+1}\ = 0.01$.~$\hat{y}_{\text{rgb}_i}$ is the predicted RGB color at voxel~$i$ and~$y_{\text{rgb}_i}$ is the target RGB color at voxel~$i$.~$M$ denotes the number of voxels in the mask corresponding to removed 3D patches, ~$\hat{y}_{\alpha_j}$ is the predicted opacity at voxel~$j$ and~$y_{\alpha_j}$ is the target opacity at voxel~$j$.

\section{Network Architecture Details}
\label{sec:architecture_details}

In this section, we provide more details about our architectural design. We described in detail the encoder-decoder U-Net style architecture for pretraining in Section~\ref{subsec:masked_pretraining_nerf}. We use a constant W, D, H dimension of 160 for all of our datasets, a patch size $p=4$, and mask out randomly 75\% of the patches. We use 4 Swin Transformer blocks and the spatial feature size of the last block is $5\times5\times5$ with the channel dimension being 768. Consequently, we upsample features with skip connection decoders and use 4 decoders and a final convolution block to restore the original dimension, i.e. $B\times4\times160\times160\times160$. Below, we specify the specifications and architectures of our task-specific decoders.

\subsection{Task-specific Downstream 3D Heads}

Our task-specific downstream heads are attached to the Swin-Transformer encoder. The goal of task-specific heads is to output dedicated meaningful outputs to regress the useful 3D information such as 3D OBBs and semantic labels for each voxel.

\noindent \textbf{3D OBB Prediction:} Our 3D OBB prediction head is similar to NeRF-RPN~\cite{hu2023nerf} except that for all of our transfer learning experiments and comparisons, we start with warm-started encoder weights via our large-scale self-supervised pretraining, whereas NeRF-RPN weights are started from scratch. Starting from scratch here denotes that the network weights were not pretrained on any other dataset, but rather NeRF-RPN initializes the weights of linear layers with truncated normal distribution with a standard deviation of 0.02. We use the author's~\cite {hu2023nerf} official implementation for all the comparisons. Specifically, we utilize an anchor-free RPN head from~\cite{hu2023nerf} which is based on the FCOS detector lifted to the 3D domain via replacing 2D convolutions with their 3D counterparts. We employ a combination of focal loss, IOU loss, and binary cross-entropy loss for downstream transfer learning task similar to~\cite{hu2023nerf}.

\begin{figure*}[t!]
\centering
\includegraphics[width=1.0\textwidth]{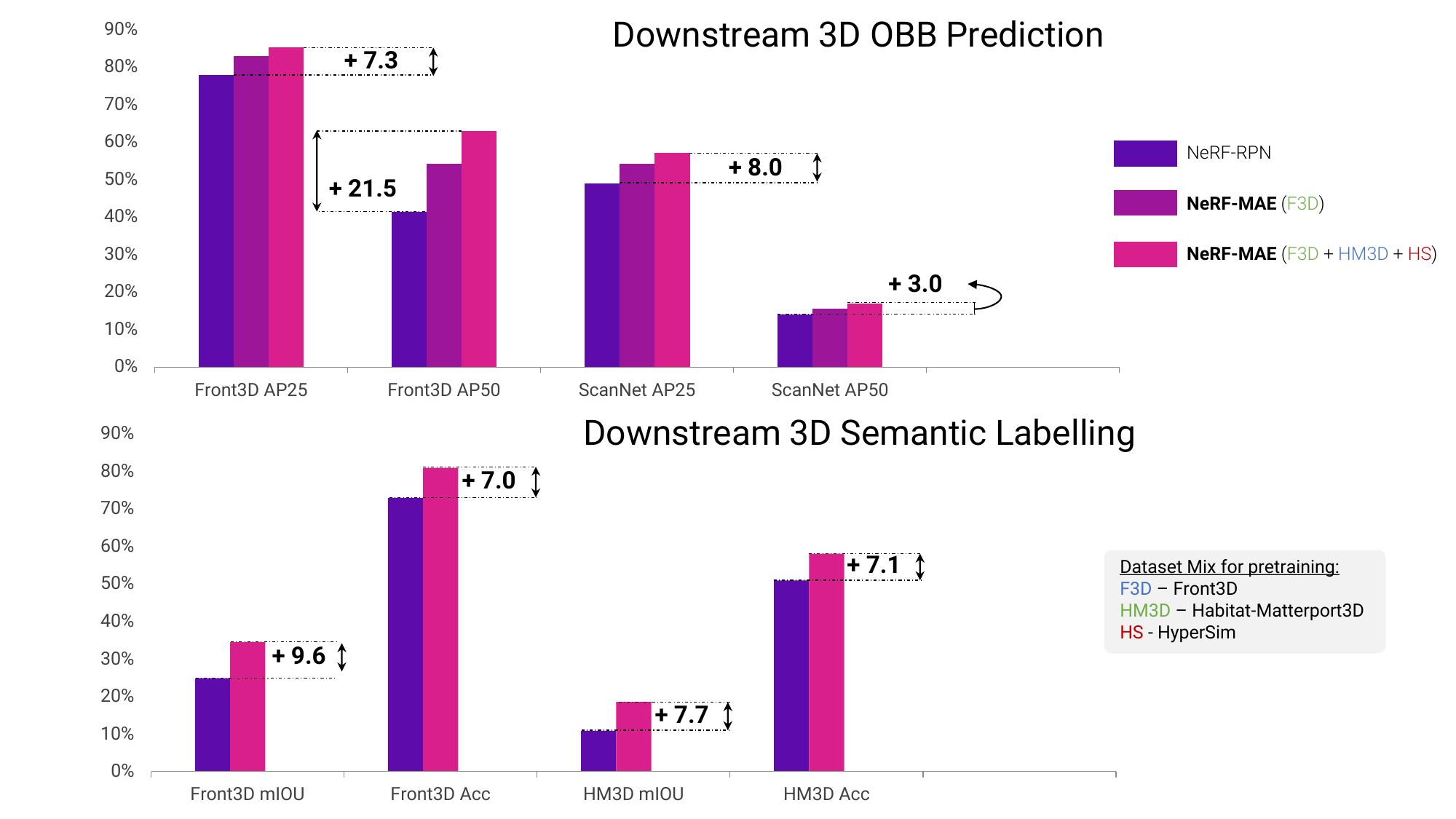}
\captionof{figure}{
\textbf{Quantitative Comparison:} We show further quantitative analysis of results reported in Table~\ref{tab:downstream_3d_obb} and Table~\ref{tab:voxel_labelling}. Specifically, we highlight improvement numbers for AP25 and AP50 on ScanNet~\cite{dai2017scannet} and Front3D~\cite{fu20213d} and mIOU and Acc for HM3D~\cite{ramakrishnan2021hm3d} and Front3D~\cite{fu20213d}. Our results clearly show that our numbers improve over NeRF-RPN~\cite{hu2023nerf} for all cases. Furthermore, we also show that increasing unlabelled posed 2D data from different sources improves performance on 3D OBB prediction downstream task further confirming the benefits of our pre-training strategy. 
}
\label{fig:comparison_chart2}
\end{figure*}

\noindent \textbf{3D Semantic Voxel-Labelling:} The task of semantic voxel labeling is described in detail in Section~\ref{subsec:downstream_tasks}. Specifically, we use the same decoders with skip connections, i.e.~$\mathcal{D}_{i}$, as described in Section~\ref{subsec:convolution_decoders} where~$i$ ranges from 2 to 4. We replace~$\mathcal{D}_{1}$ by incorporating an additional skip connection with the encoded output from the input grid~($\mathcal{G}$). This is followed by a final convolutional output block to output semantic volume~($\mathcal{G}_{S} \in \mathbb{R}^{H \times W \times D \times C}$) of spatial dimension $W,H,D$ where $W=H=D=160$ and $C=18$ for the Front3D dataset and 20 for the Habitat-Matterport3D dataset. We use a weighted masked cross-entropy loss, balanced by the inverse log propensity of each class where propensity is defined to be the frequency of each class in the training dataset. 

\noindent \textbf{Voxel Super-Resolution:} The task of voxel super-resolution is described in detail in Section~\ref{subsec:downstream_tasks}. Specifically, we replace convolution decoders with four 3D convolution layers with instance norm and ReLU activation followed by an upsampling by a factor of 2. This is followed by a final upsampling of 1.6 and 2.4 depending on the output resolution size of~$256^3$ and~$384^{3}$ and a final 3D convolution. Similar to the original reconstruction loss~(Section~\ref{subsec:masked_pretraining_nerf}), we employ a masked color reconstruction loss enforced with the ground-truth upsampled grid. 

\subsection{Parameters}
NeRF-MAE efficiently trains on diverse scenes using a 70 million parameter network, highlighting its capability to handle complex and varied training data. Our Swin-S Transformer encoder architecture as well as the anchor-free RPN head remains the same as NeRF-RPN~\cite{hu2023nerf} for a fair comparison on the downstream 3D OBB prediction task. For reconstruction, we employ lightweight decoders~(Section~\ref{subsec:convolution_decoders}) since for our transfer learning experiments, we eventually discard the decoders and utilize the feature pyramids for downstream 3D tasks as described above. 

\section{Compute Resources}
\label{sec:compute_resources}

Our model was trained on 7 Nvidia A100 GPUs for 1200 epochs for 2 days and 17 hours. The cost for a pretraining run is around 19 GPU days with a GPU capacity of 82GB and an average utilization of 60\%. To pre-train our representation, each NeRF was trained for 8 minutes on a single GPU for 100,000 steps. The total training time for training 3500 NeRF was 58 hours and finished in just over 2 days on 8 Nvidia A100 GPUs. Note that our pretraining strategy is agnostic to the NeRF technique used and can be replaced with any of the faster NeRF techniques developed. We show that our pretraining strategy equally applies to both instant-NGP~\cite{muller2022instant} trained NeRF and dense depth prior NeRF~\cite{roessle2022dense} and does not require any modification to our network architecture or training strategy.

\section{Experiments and Results} 
\label{chap:nerfmae,sec:exp}

\begin{table*}[t!]
\centering
\caption{\textbf{Effect of our NeRF pretraining on downstream 3D OBB prediction:} showing our approach's superior performance compared to the state-of-the-art baseline, NeRF-RPN~\cite{hu2023nerf}, on 3D OBB prediction. We also show that adding unlabelled posed 2D data from different data sources improves 3D performance on unseen data domains, shown here for 3D OBB in Scannet~\cite{dai2017scannet}. Colored text denotes datasets used for pre-training i.e. \textcolor{darkgreen}{Front3D} \cite{fu20213d}, \textcolor{darkblue}{HM3D}\cite{ramakrishnan2021hm3d} and \textcolor{darkred}{Hypersim}~\cite{roberts2021hypersim}. \textcolor{orange}{Orange} and \textcolor{yellow!80!black}{Yellow} denotes best and second best results. \textcolor{gray}{\underline{Underline}} denotes the technique (\textcolor{gray}{FCAF3D*~\cite{rukhovich2022fcaf3d}}) uses additional ground-truth depth information, where all others use multi-view RGB only.}
\normalsize
\resizebox{\textwidth}{!}{
\begin{tabular}{@{}lccccccccccc@{}}
\toprule
& & \multicolumn{4}{c}{Front3D~\cite{fu20213d} 3D OBB} & \multicolumn{4}{c}{ScanNet~\cite{dai2017scannet} 3D OBB} \\
\cmidrule(lr){3-6} \cmidrule(lr){7-10}
 Approach & Aug.(Pretrain)& \textbf{Recall@25}$\uparrow$ & \textbf{Recall@50}$\uparrow$ & \textbf{AP@25}$\uparrow$ & \textbf{AP@50}$\uparrow$ & \textbf{Recall@25}$\uparrow$ & \textbf{Recall@50}$\uparrow$ & \textbf{AP@25}$\uparrow$ & \textbf{AP@50}$\uparrow$ \\
\midrule


NeRF-RPN~\cite{hu2023nerf}              &  - &                    0.961 &                      0.622 &                      0.780 &                      0.415 &                      0.891 &                      0.323 &                      0.491 &                      0.140 \\
ImVoxelNet~\cite{rukhovich2022imvoxelnet}            &  - &                    0.883 & \cellcolor{tabsecond}0.715 &  \cellcolor{tabfirst}0.861 &  \cellcolor{tabfirst}0.664 &                      0.517 &                      0.202 &                      0.373 &                      0.098 \\
\textcolor{gray}{FCAF3D*~\cite{rukhovich2022fcaf3d}}                &  - &                    0.891 &                      0.569 &                      0.731 &                      0.352 &  \cellcolor{tabthird}\textcolor{gray}{\underline{0.902}} &  \cellcolor{tabfirst}\textcolor{gray}{\underline{0.424}}  &  \cellcolor{tabfirst}\textcolor{gray}{\underline{0.637}} &  \cellcolor{tabfirst}\textcolor{gray}{\underline{0.185}} \\
\midrule
NeRF-MAE (\textcolor{darkgreen}{F3D})  &  \textcolor{falured}{\xmark} & \cellcolor{tabthird}0.962 &  \cellcolor{tabthird}0.675 &                      0.780 &                      0.543 &                      0.897 &                      0.361 &                      0.510 &                      0.145 \\
NeRF-MAE (\textcolor{darkgreen}{F3D})         &  \textcolor{teal}{\checkmark}& \cellcolor{tabsecond}0.963 &  \cellcolor{tabfirst}0.743 &  \cellcolor{tabthird}0.830 &  \cellcolor{tabthird}0.591 & \cellcolor{tabsecond}0.905 &  \cellcolor{tabthird}0.391 &  \cellcolor{tabthird}0.543 &  \cellcolor{tabthird}0.155 \\
\midrule
NeRF-MAE (\textcolor{darkgreen}{F3D}+\textcolor{darkblue}{HM3D}+\textcolor{darkred}{HS}) & \textcolor{teal}{\checkmark}& \cellcolor{tabfirst}0.972 &  \cellcolor{tabfirst}0.743 & \cellcolor{tabsecond}0.853 & \cellcolor{tabsecond}0.630 &  \cellcolor{tabfirst}0.920 & \cellcolor{tabsecond}0.395 & \cellcolor{tabsecond}0.571 & \cellcolor{tabsecond}0.170 \\

\bottomrule
\end{tabular}
}

\label{tab:downstream_3d_obb}
\end{table*}




\begin{table*}[t!]
\centering
\caption{\textbf{Effect of amount of 3D labeled scenes on downstream 3D object detection performance:} showing our approach's superior results compared to strong baseline. Note that we use 1515 scenes from FRONT3D~\cite{fu20213d} data to pre-train our NeRF-MAE encoder framework. The FPN network weights in both cases are initialized from scratch.}
\scriptsize
\resizebox{\textwidth}{!}{
\begin{tabular}{@{}ccccccccccc@{}}
\toprule
&  & \multicolumn{4}{c}{MAE pre-trained encoder (\textbf{Ours})} & \multicolumn{4}{c}{NeRF-RPN~\cite{hu2023nerf} Start from scratch} \\
\cmidrule(lr){3-6} \cmidrule(lr){7-10}
 \# Labelled Scenes & \#Scenes & \textbf{Recall@25}$\uparrow$ & \textbf{Recall@50}$\uparrow$ & \textbf{AP@25}$\uparrow$ & \textbf{AP@50}$\uparrow$ & \textbf{Recall@25}$\uparrow$ & \textbf{Recall@50}$\uparrow$ & \textbf{AP@25}$\uparrow$ & \textbf{AP@50}$\uparrow$ \\
\midrule
10\% & 12& 0.93 & 0.41 & 0.517 & 0.175 & 0.91 & 0.352 & 0.499 & 0.152 \\
25\% & 30 & 0.94 & 0.51 & 0.67 & 0.36 & 0.94 & 0.49 & 0.67 & 0.29 \\
50\% & 61 & 0.95 & 0.60 & 0.74 & 0.42 & 0.95 & 0.58 & 0.71 & 0.303\\
100\% & 122& \cellcolor{tabfirst}0.96 & \cellcolor{tabfirst}0.67 & \cellcolor{tabfirst}0.79 & \cellcolor{tabfirst}0.54 & \cellcolor{tabfirst}0.96 & \cellcolor{tabfirst}0.62 & \cellcolor{tabfirst}0.78 & \cellcolor{tabfirst}0.41 \\
\bottomrule
\end{tabular}
}
\vspace{-10pt}
\label{tab:amount_of_labelled_scenes}
\end{table*}

\subsection{Settings}

In this section, we evaluate our proposed method on various 3D downstream tasks against the state-of-the-art baseline, NeRF-RPN~\cite{hu2023nerf}. NeRF-RPN is the first to introduce 3D object detection directly in the Neural Radiance Field formulation. We directly compare with NeRF-RPN's SwinTransformer~(Swin-S) encoder implementation. Our architecture for downstream tasks fine-tuning remains the same as~\cite{hu2023nerf}, whereas the only difference is our encoder weights are pre-trained on unlabeled posed RGB data, and the weights for NeRF-RPN's encoder are started from scratch. We first study the impact of our self-supervised pretraining on various downstream 3D tasks~(Section~\ref{subsec:downstream_tasks}). We then study scaling laws of our self-supervised pretraining as well as ablate the most important factors in our pretraining in Section~\ref{subsec:scaling_laws_and_ablation}. For 3D detection, we also include comparisons to ImVoxelNet~\cite{rukhovich2022imvoxelnet} and FCAF3D~\cite{rukhovich2022fcaf3d}.

\noindent \textbf{Datasets:}
We pre-train our 3D representation in a self-supervised manner on Front3D~\cite{fu20213d}, HM3D~\cite{ramakrishnan2021hm3d} and Hypersim~\cite{roberts2021hypersim}.
For 3D object detection, we additionally test on hold-out Scannet~\cite{dai2017scannet} scenes~(as discussed in Section~\ref{subsec:nerf_pt_dataset}) where we use the train and test split provided by~\cite{hu2023nerf}. These scenes are different from the scenes used for pretraining.  For semantic voxel labelling, a careful curation results in semantic voxel prediction of 18 classes for Front3D and 20 classes for HM3D.

\noindent \textbf{Metrics:} We report 3D PSNR and 3D MSE to evaluate radiance and density grid reconstruction quality. For the 3D OBB downstream task, we report Average Precision at thresholds 25 and 50 along with Recall at 25 and 50 thresholds. For semantic voxel labeling downstream task, we report mean intersection over union~(IOU), mean accuracy~(mAcc), and total accuracy~(Acc) respectively.

\begin{figure}[t]
    \centering
    \resizebox{0.85\linewidth}{!}{%
\includegraphics{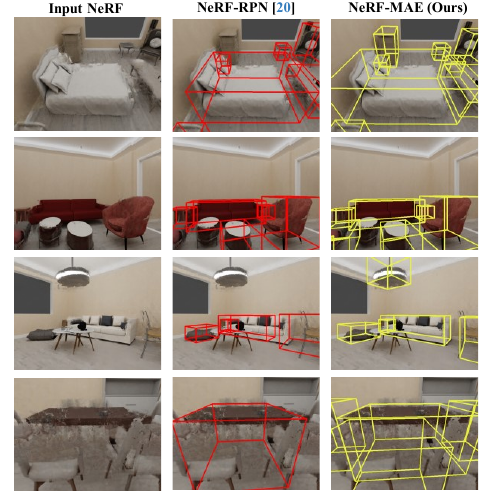}
    }
    \caption{\textbf{Qualitative downstream task generalization comparison:} showing our approach's superior results compared with NeRF-RPN~\cite{hu2023nerf}.
    }
    \label{fig:qualitative_recon_3}
\end{figure}%

\begin{figure}[t]
    \centering
    \resizebox{0.85\linewidth}{!}{%
\includegraphics{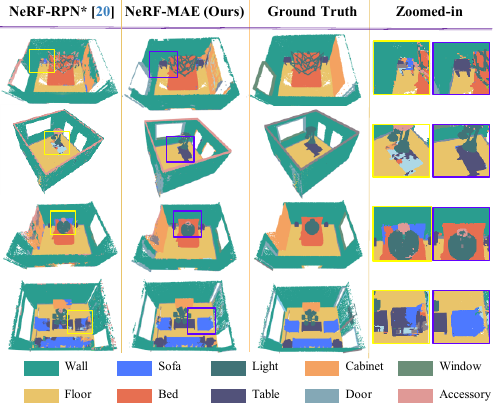}
    }
    \caption{\textbf{Qualitative downstream task generalization comparison:} showing our approach's superior results compared with NeRF-RPN~\cite{hu2023nerf}.
    }
    \label{fig:qualitative_recon_5}
\end{figure}%

\begin{table}[t!]
\centering
\caption{\textbf{Semantic voxel-labelling results:} showing superior performance of our method vs NeRF-RPN~\cite{hu2023nerf}. * denotes we add a task-specific head to NeRF-RPN's Swin-S encoder.}
  \setlength{\tabcolsep}{1.5pt}
 \scriptsize
\begin{tabularx}{\columnwidth}{c|CC|CC}
\toprule

& \multicolumn{2}{c}{Front3D~\cite{fu20213d}} & \multicolumn{2}{c}{HM3D~\cite{ramakrishnan2021hm3d}} \\
\cmidrule(lr){2-3} \cmidrule(lr){4-5}
Metrics & NeRF-RPN*~\cite{hu2023nerf} & NeRF-MAE& NeRF-RPN*~\cite{hu2023nerf} & NeRF-MAE \\
\hline
\textbf{mIOU}$\uparrow$  & 0.249 & \cellcolor{tabfirst}0.345 & 0.109 & \cellcolor{tabfirst}0.186 \\
\textbf{mAcc}$\uparrow$ & 0.338 & \cellcolor{tabfirst}0.450 & 0.160 & \cellcolor{tabfirst}0.264 \\
\textbf{Acc}$\uparrow$ & 0.730 & \cellcolor{tabfirst}0.810 & 0.510 & \cellcolor{tabfirst}0.581 \\
\bottomrule
\end{tabularx}
\vspace{-14pt}
\label{tab:voxel_labelling}
\end{table}




\begin{table}[t!]
\centering
\caption{\textbf{Voxel super-resolution quantitative results:} showing superior performance of our method vs NeRF-RPN~\cite{hu2023nerf}. * denotes we add a task-specific head to NeRF-RPN's Swin-S encoder.}
\setlength{\tabcolsep}{2.5pt}
\scriptsize
\begin{tabularx}{\columnwidth}{c|CC|CC}

\toprule

& \multicolumn{2}{c}{$256^{3}$} & \multicolumn{2}{c}{$384^{3}$}\\
\cmidrule(lr){1-2}
\cmidrule(lr){2-3} \cmidrule(lr){4-5}
 Method & \textbf{PSNR}$\uparrow$& \textbf{MSE} $\downarrow$&  \textbf{PSNR}$\uparrow$& \textbf{MSE} $\downarrow$ \\ 
\hline 

NeRF-RPN*~\cite{hu2023nerf} & 16.25 &0.024 & 16.08 &0.025 \\
NeRF-MAE(Ours) &\cellcolor{tabfirst}17.27 &\cellcolor{tabfirst}0.019 &\cellcolor{tabfirst}17.34 &\cellcolor{tabfirst}0.019\\
\bottomrule
\end{tabularx}
\vspace{-12pt}
\label{tab:voxel_super_resolution}
\end{table}

\begin{table}[t!]
\centering
\caption{\textbf{Voxel super-resolution quantitative results:} for HM3D~\cite{ramakrishnan2021hm3d} test-set showing superior performance of our method vs NeRF-RPN~\cite{hu2023nerf}. * denotes that we add a task-specific head to NeRF-RPN's Swin-S encoder.}
\setlength{\tabcolsep}{2.5pt}
\scriptsize
\begin{tabularx}{\columnwidth}{c|CC|CC}

\toprule

& \multicolumn{2}{c}{$256^{3}$} & \multicolumn{2}{c}{$384^{3}$}\\
\cmidrule(lr){1-2}
\cmidrule(lr){2-3} \cmidrule(lr){4-5}
 Method & \textbf{PSNR}$\uparrow$& \textbf{MSE} $\downarrow$&  \textbf{PSNR}$\uparrow$& \textbf{MSE} $\downarrow$ \\ 
\midrule

NeRF-RPN*~\cite{hu2023nerf} & 14.70 &0.034 & 14.65 &0.035 \\
NeRF-MAE(Ours) &\cellcolor{tabfirst}15.01 &\cellcolor{tabfirst}0.032 &\cellcolor{tabfirst}15.20 &\cellcolor{tabfirst}0.030\\
\bottomrule
\end{tabularx}
\label{tab:voxel_super_resolution_hm3d}
\end{table}
\begin{table}[t!]
\setlength{\tabcolsep}{2.5pt}
\centering
\caption{\textbf{3D OBB prediction quantitative result} for HM3D~\cite{ramakrishnan2021hm3d} test-set showing our approach's superior performance compared with NeRF-RPN~\cite{hu2023nerf}}.
\footnotesize

\begin{tabularx}{\columnwidth}{c|CC}
\toprule
Method & \textbf{AP@25}  & \textbf{Recall@50} \\
\midrule
NeRF-RPN~\cite{hu2023nerf} & 0.30  & 0.31\\
NeRF-MAE (Ours) & \cellcolor{tabfirst}0.37 & \cellcolor{tabfirst}0.40 \\
\bottomrule
\end{tabularx}
\vspace{-10pt}
\label{tab:hm3d_obb}
\end{table}
\begin{table}[t!]
\centering
\caption{\textbf{Effect of percent of pretraining scenes on voxel-grid reconstruction} showing our pretraining on larger datasets helps to learn better representations.}
\scriptsize
\begin{tabularx}{\columnwidth}{cccCC}
\toprule
\multicolumn{3}{c}{Pretraining Scenes} & \multicolumn{2}{c}{Metrics}\\
\cmidrule(lr){1-3}\cmidrule(lr){4-5}
\textbf{\% Scenes} & \textbf{\# Scenes} & ~\textbf{Augmentations} &\textbf{3D PSNR} $\uparrow$ & \textbf{MSE} $\downarrow$ \\ 
\midrule
10\% & 151 & \textcolor{falured}{\xmark}& 14.20 & 0.038 \\
25\% & 378 &\textcolor{falured}{\xmark} & 14.98 & 0.032 \\
50\% & 757 & \textcolor{falured}{\xmark}& 15.42 & 0.028 \\
100\% & 1515 &\textcolor{falured}{\xmark} & 16.00 & 0.025 \\
100\% & 1515 & \textcolor{teal}{\checkmark}&  \cellcolor{tabfirst}19.09 & \cellcolor{tabfirst}0.012 \\
\bottomrule
\end{tabularx}
\label{tab:scaling_law_reconstruction}
\end{table}

\begin{table}[t!]
\centering
\caption{\textbf{Effect of percent of pretraining scenes on downstream 3D object detection performance} showing more unlabelled data helps learn better representation for downstream 3D tasks.}
\scriptsize
\begin{tabularx}{\columnwidth}{ccCCCC}
\toprule
\multicolumn{2}{c}{Pretraining Scenes} & \multicolumn{4}{c}{Metrics}\\
\cmidrule(lr){1-2}\cmidrule(lr){3-6}
\% Scenes& \# Scenes& \textbf{Recall@25} & \textbf{Recall@50} & \textbf{AP@25} & \textbf{AP@50} \\
\midrule
10\% & 151 & 0.94 & 0.61 & 0.74 & 0.45 \\
25\% & 378 & 0.95 & 0.62 & 0.76 & 0.46 \\
50\% & 757 & 0.96 & 0.66 & 0.77 & 0.51 \\
100\% & 1515 & \cellcolor{tabfirst}0.96 & \cellcolor{tabfirst}0.67 & \cellcolor{tabfirst}0.79 & \cellcolor{tabfirst}0.54 \\
\bottomrule
\end{tabularx}
\vspace{-5pt}
\label{tab:effect_pretraining}
\end{table}
\begin{table}[b!]
\centering
\caption{\textbf{Effect of skip connections in the decoder} during pretraining showing NeRF-MAE with skip connection helps learn better representations for downstream tasks}
\scriptsize
\begin{tabularx}{\columnwidth}{cCCCC}
\toprule
& \multicolumn{2}{c}{Front3D~\cite{hu2023nerf}} & \multicolumn{2}{c}{ScanNet~\cite{dai2017scannet}}\\
\cmidrule(lr){2-5}
NeRF-MAE& \textbf{Recall@50} & \textbf{AP@25} & \textbf{Recall@50} & \textbf{AP@25} \\

\midrule
w/o Skip & 0.73 & 0.83 & 0.39 & 0.54 \\
w/ Skip & \cellcolor{tabfirst}0.74 & \cellcolor{tabfirst}0.85 & \cellcolor{tabfirst}0.40 & \cellcolor{tabfirst}0.57 \\
\bottomrule
\end{tabularx}
\label{tab:skip_ablation}
\end{table}

\subsection{Downstream tasks:}
\label{subsec:downstream_tasks}
    \noindent\textbf{3D Object Detection in NeRF}~\cite{hu2023nerf}: The task is to predict 3D OBBs in an anchor-free manner. It entails regressing bounding box-offsets~$\boldsymbol{t}=\left(x_0, y_0, z_0, x_1, y_1, z_1, \Delta \alpha, \Delta \beta\right)$, objectness score~$c$ and a single objectness~$p$ for each voxel. The task is similar to anchor-free RPN from~\cite{hu2023nerf}.
    
    \noindent\textbf{Voxel-grid Super-Resolution}: Given an input voxel grid~($V$) of resolution $160^3$, we predict an upsampled voxel-grid of higher resolutions, such as $384^3$. This can be denoted as~$V_{\text{upsample}}=g(V_{\text{input}}, \text{resolution})$, where~$g$ is a U-Net style encoder-decoder network. It is an important task for fast grid upsampling, since querying an implicit MLP for higher resolution is slow.
    
    \noindent\textbf{Semantic Voxel Labelling}:
    Given an input grid $G$ with dimensions $W \times  H\times D \times 4$, the task is to predict the class labels for each voxel. Each voxel in the grid is assigned a class label, $S_{i,j,k}$ which are integers ranging from 1 to $n$, indicating the semantic category or class to which each voxel belongs. Mathematically, this can be represented as: $S_{i,j,k} \in \{1, 2, \ldots, n\}, \quad \text{for} \quad 1 \leq i \leq W, 1 \leq j \leq H, 1 \leq k \leq D$. Due to the complexity and diversity of objects in a scene, this task requires strong priors for generalization.    

\subsection{Comparative Study}

\textbf{3D Object Detection Downstream Task:} The results of our proposed method are summarized in Tables~\ref{tab:downstream_3d_obb} and~\ref{tab:amount_of_labelled_scenes}. We show the impact of~\textit{dataset mix},~\textit{augmentation during pretraining} and~\textit{amount of labeled scenes for transfer learning}.  Our proposed approach consistently outperforms the state-of-the-art baseline method on 3D object detection in NeRFs on both Front3D and ScanNet scenes. Among our variants, our method pre-trained on multiple data sources~(Table~\ref{tab:downstream_3d_obb},~row 6) and with augmentations enabled during pretraining performs the best. Specifically, NeRF-MAE shows superior performance on the unseen test-set by achieving an AP50 of 0.63 and a Recall50 of 0.743, hence demonstrating an absolute improvement of 21.5\% in AP50 and 12.1\% in Recall50 on the Front-3D dataset. Our method also achieves superior performance on the ScanNet dataset, which is never seen during pretraining. We achieved an AP50 of 0.17 and a Recall50 of 0.395, hence demonstrating an absolute performance improvement of 3\% AP50 and 7.2\% on Recall50 on the ScanNet dataset respectively. Furthermore, we also demonstrate adding random rotation, flip, and scaling augmentation during pretraining further improves performance by 5\% AP50~(row 4 vs 5).

\noindent \textbf{Amount of unlabeled scenes on downstream 3D OBB task:} We further test our method's ability on the 3D OBB prediction task by restricting the amount of scenes available for transfer learning. The results are summarized in Table~\ref{tab:amount_of_labelled_scenes}. Our method outperforms the strong baseline in all scenarios, showing our approach's ability to consistently improve performance in a limited data setting.

\noindent \textbf{Semantic Voxel-labelling and Voxel-Super Resolution:} We further test our approach's ability on two challenging downstream tasks~i.e. semantic voxel-labelling and voxel super-resolution. Our results are summarized in Tables~\ref{tab:voxel_labelling} and~\ref{tab:voxel_super_resolution}. Our higher mIOU, mAcc, and Acc metrics compared to strong baseline show our networks' ability to learn good representation for the challenging downstream 3D voxel-labelling task. Specifically, we achieve 0.81 Acc, 0.45 mAcc, and 0.345 mIOU which is equal to an absolute performance improvement of 6.8\% Acc, 12.9\% mAcc, and 9.8\% mIOU metric against NeRF-RPN~\cite{hu2023nerf} on Front3D. For voxel-super resolution, our high 3D PSNR and low MSE metrics~(absolute improvement of 1.02 for~$256^3$ and 1.259 for~$384^3$ grid show) show our network learns general priors useful for a variety of downstream 3D tasks.

\begin{figure}[t]
    \centering
    \resizebox{1.0\linewidth}{!}{%
\includegraphics{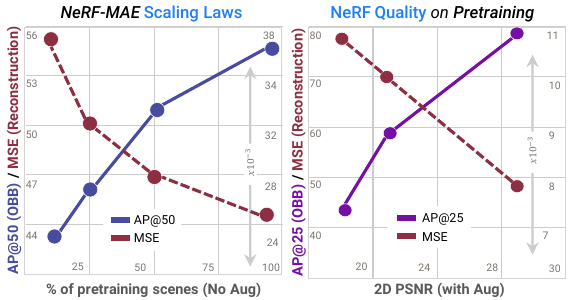}
    }
\caption{\textbf{Quantitative results} showing scaling laws and NeRF quality ablation for NeRF-MAE pretraining and transfer learning.}
    \label{fig:quantitative_results_scaling_laws}
\end{figure}%

\subsection{Scaling Laws and Ablation Analysis}
\label{subsec:scaling_laws_and_ablation}

\looseness=-1
We further analyze our network's ability to learn useful 3D representations by studying how the amount and quality of pretraining data affects downstream performance. We summarize our results in Figure~\ref{fig:quantitative_results_scaling_laws}. To study scaling laws, we select Front3D dataset and use 10\%, 25\%, 50\%, and 100\% of the scenes for pretraining. Our test-set 3D PSNR and MSE metrics clearly show our network's ability to improve with more unlabeled data on the reconstruction and transfer learning tasks. Specifically, we achieved an absolute AP50 improvement of 10\% when adding 10x the number of scenes~(1515 vs 151 scenes), emphasizing our network is able to learn better on more unlabeled data. To understand how the quality of NeRFs during the pretraining stage affects downstream performance, we train the input NeRFs to varying levels of accuracy, providing all 100\% of the scenes at various PSNR values. Our results show that we can learn better representations, achieving downstream AP25 improvement of 36\% when adding higher quality NeRFs to our pretraining~(18 vs 28 2D PSNR).

\section{Implementation Details}
\label{sec:implementation_details}

To pre-train NeRF-MAE, we utilize the Swin-S backbone of the SwinTransformer. As detailed in Section~\ref{subsec:masked_pretraining_nerf}, we utilize four swin blocks of depth 2, 2, 18, and 2 with the number of heads in each block equalling 3, 6, 12, and 24. We pre-train both encoder and decoder networks for 1200 epochs with the Adam optimizer and a maximum learning rate of~$3e^{-4}$, a weight decay of~$1e^{-3}$ with a one-cycle learning rate scheduling. We employ a batch size of 32 for our pretraining. We additionally utilize online data augmentation during both pretraining and transfer learning stages. We employ random flip, rotation, and scaling augmentation with a probability of 50\%. Data augmentation is shown to improve our pretraining, as described in Section~\ref{chap:nerfmae,sec:exp}. For transfer learning, we train both our network and the baseline NeRF-RPN~\cite{hu2023nerf} for 1000 epochs for 3D OBB prediction and 500 epochs for voxel super-resolution with the same learning rate and strategy as our pretraining stage since we empirically determined that changing the learning rate and strategy results in decreased performance for both our method and the baseline. We additionally utilize a maximum normal gradient clip value of 0.1 and fine-tune the network until convergence. We determine transfer learning convergence on the hold-out validation set and employ early stopping based on AP50 metrics for 3D OBB prediction, accuracy metric for 3D semantic voxel labeling, and 3D PSNR metric for voxel super-resolution experiments. For transfer learning experiments, we use a batch size of 8. 

\section{Pretraining Datasets}
\label{sec:pretraining_datasets}

\begin{figure*}[t!]
\centering
\includegraphics[width=1.0\textwidth]{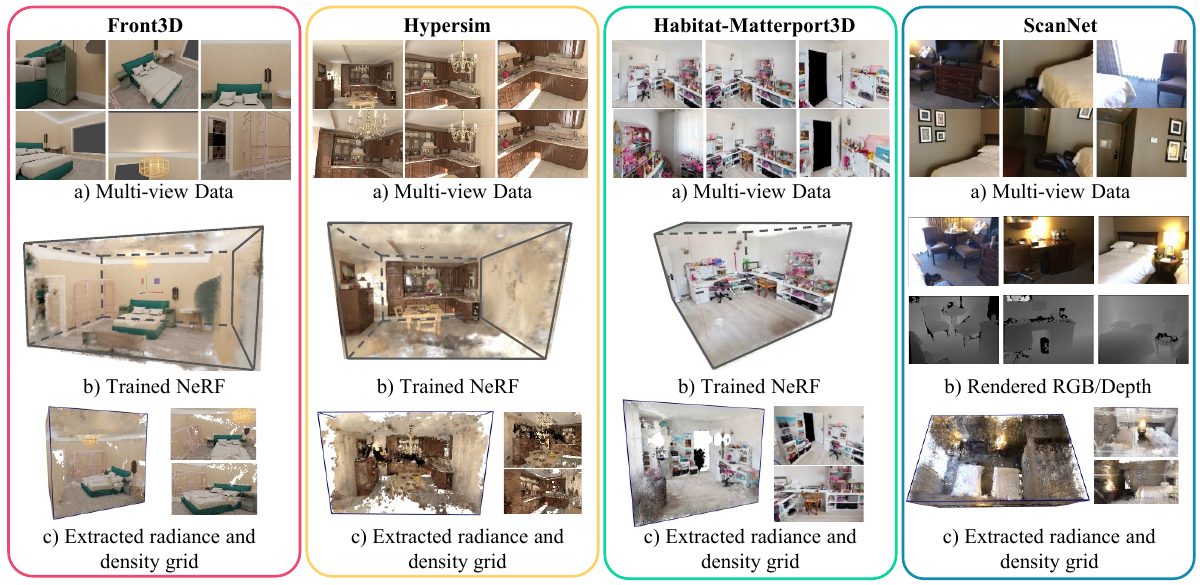}

\captionof{figure}{
\textbf{Data preparation for pretraining:} {showing multi-view images, trained NeRF and extracted radiance and density grid for Instant-NGP~\cite{muller2022instant} trained NeRFs for Front3D~\cite{fu20213d}, HyperSim~\cite{roberts2021hypersim} and Habitat-Matterport3D~\cite{ramakrishnan2021hm3d}. The right column shows multi-view images, rendered RGB/depth images, and extracted radiance and density grid for ScanNet~\cite{dai2017scannet} dataset, which is trained using dense depth prior NeRF~\cite{roessle2022dense}. Note that we do not pre-train our representation on ScanNet, rather use this dataset for our cross-dataset transfer experiment for the downstream 3D OBB prediction task.
}}
\label{fig:pretraining_datasets}
\end{figure*}

We show our pretraining datasets in detail in Figure~\ref{fig:pretraining_datasets} and Figure~\ref{fig:nerf_data}. First, we show the data processing flow for each of the Front3D~\cite{fu20213d}, Scannet~\cite{dai2017scannet}, HM3D~\cite{ramakrishnan2021hm3d} and Hypersim~\cite{roberts2021hypersim} dataset in Figure~\ref{fig:pretraining_datasets} starting with multi-view images to get a NeRF trained for each scene and finally getting an explicit radiance and density grid for all scenes. We train each scene for 100,000 steps to ensure high-quality NeRF are obtained as input to our pretraining stage as further confirmed by our experiments in Figure~\ref{tab:scaling_law_reconstruction}, which quantitatively shows that better NeRF quality results in better pre-training and leads to improved downstream 3D task performance. We further show rendered RGB vs ground-truth RGB for selected scenes from all four datasets in Figure~\ref{fig:nerf_data}. The figure confirms that our NeRF quality for validation images is faithful to the ground-truth images. Secondly, it shows that our pretraining dataset for training good 3D representations from posed 2D data is very diverse in realism, lightning, shadows, number of objects in the scene as well as the setting of each scene i.e. ranging from bedroom scenes to office and living room scenes. This makes our 3D representation learning more robust to the varying input data and helps it improve performance for data outside the training data-distribution, as further confirmed in Section~\ref{chap:nerfmae,sec:exp}. 

\section{Additional Experiment Analysis}
\label{sec:add_experiments}

\begin{figure*}[t!]
\centering
\includegraphics[width=1.0\textwidth]{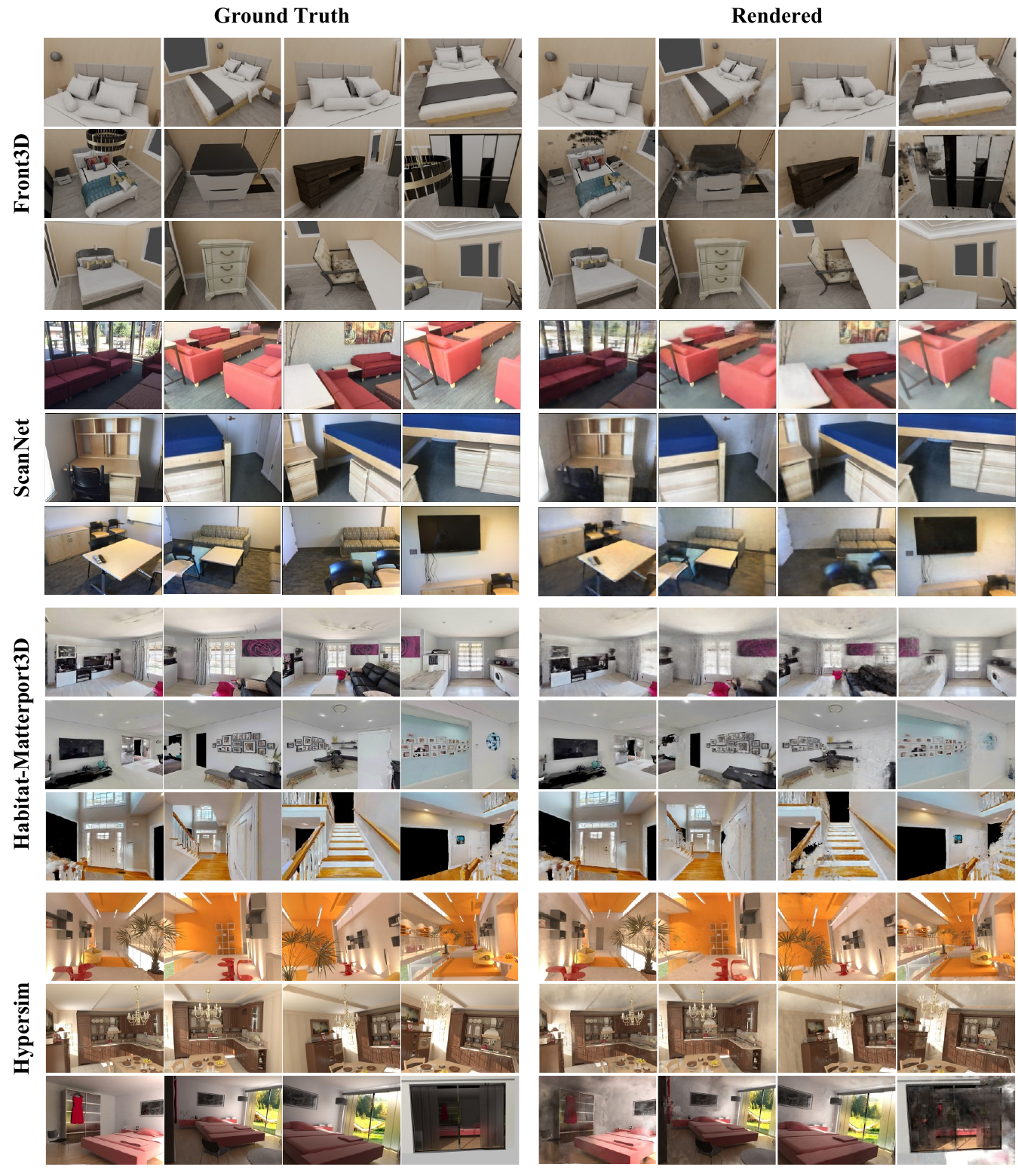}
\captionof{figure}{
\textbf{Multi-view Rendered vs Ground-truth posed RGB} images from four distinct datasets used in this chapter i.e. Front3D~\cite{fu20213d}, ScanNet~\cite{dai2017scannet}, Habitat-Matterport3D~\cite{ramakrishnan2021hm3d} and Hypersim~\cite{roberts2021hypersim}. We use Instant-NGP to train neural radiance fields for Front3D, Hypersim and HM3D, whereas Scannet scenes are trained using dense depth prior nerf~\cite{roessle2022dense}, following~\cite{hu2023nerf}
}
\label{fig:nerf_data}
\end{figure*}

In this section, we present additional experimental analysis. First, we will see a detailed analysis of downstream 3D object detection, semantic voxel-labeling, and voxel super-resolution. We will then move to an additional analysis of scaling laws and ablation studies. 

\noindent \textbf{Comparison on Downstream 3D Tasks:} We analyze all of the improvement numbers mentioned in Table~\ref{tab:downstream_3d_obb} and Table~\ref{tab:voxel_labelling} by highlighting them in Figure~\ref{fig:comparison_chart2}. One can clearly see our pretraining consistently improves all metrics across different datasets, especially on ScanNet which is omitted from our pretraining and used as a hold-out dataset for our cross-dataset transfer experiments. The results also clearly show that downstream 3D task numbers for 3D OBB prediction improve with adding unlabeled data from diverse sources, in this case, adding unlabeled HM3D scenes to our pretraining improves ScanNet numbers showing the efficacy of our pretraining strategy. Note that we only utilized posed 2D data for pre-training, and no 3D information was used in our pretraining pipeline. Next, we present results for the task of voxel grid super-resolution on HM3D dataset hold-out scenes never seen by the model during pretraining or transfer learning. We summarize these results in Table~\ref{tab:voxel_super_resolution_hm3d} and Table~\ref{tab:hm3d_obb}. One can clearly observe from Table~\ref{tab:hm3d_obb}, that our technique achieves a higher Recall@50 and AP@50 compared to strong baseline, specifically showing improvement of 9\% in Recall@50 and 7\% in AP@50, further confirming the efficacy of our pretraining strategy on a challenging dataset and 3D task. The results in Table~\ref{tab:voxel_super_resolution_hm3d} further confirm the finding of Table~\ref{tab:voxel_super_resolution}, achieving consistently higher 3D PSNR and lower MSE metrics for hold-out test scenes on HM3D for the challenging voxel grid super-resolution task.  Moreover, we complement the comparison results to other strong baselines i.e. FCAF3D~\cite{rukhovich2022fcaf3d} and ImVoxelNet~\cite{rukhovich2022imvoxelnet}, included in Table~\ref{tab:downstream_3d_obb}, with additional discussion. Our technique outperforms ImVoxelNet on both Front3D and ScanNet datasets respectively. Despite FCAF3D using ground-truth depth information, our NeRF-MAE approach still outperforms it on Front3D, while achieving very close numbers to the upper bound numbers on ScanNet, which shows our approach can be as strong as the state-of-the-art baseline even when given partial information i.e. only posed RGB images as input.    

\noindent \textbf{Additional Ablation Analysis:} We further report complete metrics for our scaling laws experiments shown in Figure~\ref{fig:quantitative_results_scaling_laws}. We report these metrics in Table~\ref{tab:effect_pretraining} and Table~\ref{tab:scaling_law_reconstruction}. Both results further confirm our findings in the chapter that our results consistently improve on reconstruction task and during transfer learning by utilizing more unlabeled data during pretraining. We additionally include ablations to our NeRF-MAE pretraining architecture in Table~\ref{tab:skip_ablation}. We show that adding skip connections to the light-weight decoders from the feature pyramids increases the accuracy of downstream tasks, specifically increasing the AP@25 by 2\% and 3\% on FRONT3D and ScanNet datasets respectively. 
\section{Summary}
\label{chap:nerfmae,sec:conclusion}

Despite great progress in neural fields, they primarily overfit to single scenes. In this work, we propose to learn strong 3D representations for NeRFs from large-scale posed 2D data only. In essence, we use a single SwinTransformer model for self-supervised pretraining on large-scale indoor neural radiance fields. To pre-train our representation, we curate large-scale indoor scene data totaling 1.6M+ images and 3,500+ scenes from 4 different sources. Our approach, NeRF-MAE, learns strong semantic and spatial structures of scenes using a mask pre-training objectives. Hence, our method excels in transfer learning on three challenging 3D tasks, utilizing only posed 2D data for pretraining. Specifically, we utilized prior knowledge in the form of~\textit{contextual and structural similarity} of the 3D data. Using our proposed strategy, we achieve improved downstream performance on various challenging 3D tasks. 

    \chapter{Conclusion}
\label{conclusion}

In this thesis, we have introduced innovative methodologies aimed at addressing the challenge of training autonomous agents when faced with constraints on real-world data availability. We have delved into several pivotal technical contributions designed to introduce strong inductive biases and priors into the training of deep learning models. These contributions have been carefully designed to achieve three primary objectives~\textbf{1.} bridge the gap between simulated data and real-world deployment using geometry and context priors, ~\textbf{2.} propose generalizable and efficient systems that work with very limited and sparse real-world labeled data and~\textbf{3.} leverage 2D unlabelled data to create 3D representations that show improved performance in downstream 3D tasks. Traditional methods often struggle to adapt to new situations, while data-centric approaches require a lot of labeled data. Our aim has been to develop autonomous agents that need very little real-world data and can learn from simulations. This way, they can handle real-world environments, which are often complex and unpredictable, and improve their ability to understand 3D scenes.

Concretely, we first studied efficient object-centric representations and proposed a fast and efficient end-to-end 3D reconstruction system for improved real-world 3D understanding in Chapter~\ref{chap:centersnap} and Chapter~\ref{chap:shapo}. We established that leveraging geometric and appearance priors obtained from synthetic data significantly aids in the acquisition of 3D representations that can be seamlessly transferred to real-world scenarios with minimal fine-tuning. Next, we studied the utilization of hierarchical structures and semantic maps as a priori knowledge for deep learning systems centered around agents, which is elaborated upon in Chapters~\ref{chap:robovln} and~\ref{chap:sasra}. Our investigations yielded compelling evidence that both semantic map representations and the incorporation of hierarchy to disentangle agent-centric systems substantially improve the generalizability of deep learning models. Lastly, we investigated the potential of leveraging unlabeled posed 2D data for 3D representation learning, in addition to harnessing large-scale synthetic data for the acquisition of 3D representations suitable for sparse multi-view view synthesis, as described in Chapter~\ref{chap:neo360} and Chapter~\ref{chap:nerfmae}. We showed that through an unsupervised approach, high-quality 3D representations can be derived from posed 2D data, employing a NeRF-based masked auto-encoder formulation, leading to substantial improvements in downstream 3D tasks. Furthermore, our findings highlighted the importance of hybrid representations derived from large-scale synthetic data, which facilitates challenging tasks such as few-view novel-view synthesis.

My research spanning object-centric 3D representations, vision-and-language navigation, and 3D scene understanding holds profound implications for the practical application of deep learning models within the domains of real-world robotics and autonomous driving. For instance, the methodologies introduced in Chapters~\ref{chap:centersnap} and~\ref{chap:shapo} offer the capability of performing~\emph{fast and real-time} 6D pose estimation along with 3D shape reconstruction, with a broad spectrum of downstream applications such as robotics grasping and augmented/virtual reality. What sets this approach apart is its minimal dependence on extensive collections of labeled real-world data; instead, it leverages large-scale 3D data obtained from simulations. The strategies outlined in~\ref{chap:robovln} and~\ref{chap:sasra} enable the deployment of models at scale, characterized by a considerably~\emph{reduced sample complexity}. This, in turn, unlocks the potential for navigational proficiency through the use of supervised learning data with priors such as hierarchy and semantic map, which exhibits superior sample complexity compared to agents trained via reinforcement learning from scratch and utilizing only raw visual observations. The algorithms examined in Chapters~\ref{chap:neo360} introduce~\emph{sparse-view novel view synthesis}, eliminating the need for a multitude of densely captured views of novel scenes to achieve precise novel view synthesis and foreground-background decomposition within vast outdoor settings. Consequently, it enables the few-shot understanding of the 3D outdoor world at a large scale. Finally, the algorithm presented in~\ref{chap:nerfmae} makes substantial strides in improving 3D tasks exclusively utilizing posed RGB data, thus eliminating the need to collect large-scale annotated 3D data such as 3D bounding boxes, 3D segmentation masks, and mesh scans of real-world indoor scenes. Instead, our findings show potential for enhancing~\emph{the accuracy of downstream 3D tasks with minimal reliance on real-world annotated data}, and by harnessing multi-view posed 2D data originating from synthetic scenes or casually captured real-world scenes.

\section{Future Research}

This thesis focused on task-specific and general priors for 3D perception relating to the robotics domain. Future work could explore the utility of~\textit{2D priors} for 3D representation learning or learning 3D in a zero-shot manner. In this paradigm, I raise these important questions for future work. 1.~\textit{What forms of 2D priors help 3D representation learning}. 2.~\textit{How high-fidelity real-world simulation can help learning policies for the real-world} and 3.~\textit{How to efficiently distill 2D priors to 3D domain?}. These questions represent key areas for further investigation in the realm of 3D perception and robotics.

Given the challenges associated with acquiring 3D data and annotations, a portion of this thesis delves into the exploration of synthetic 3D data to establish categorical priors. However, the real world presents a highly dynamic and versatile environment, making it exceedingly difficult to capture the data distribution of a diverse range of objects within simulated settings. In contrast, 2D data is readily accessible on the internet in the form of images or captions. One potential avenue for future research involves leveraging pre-trained 2D vision foundation models (VFM) or vision-language models (VLM) to facilitate zero-shot 3D scene comprehension. This leads to fundamental questions: Are there specific 2D priors that can enhance 3D scene understanding, and what are the optimal methods for transposing 2D priors into the 3D domain? Subsequent work could address these critical questions and explore efficient approaches for translating high-dimensional features, such as semantics or visual features, from vision-foundation models into the 3D domain.

Furthermore, an additional dimension of research could entail proposing a unified model capable of supporting various querying and prompting methods to attain a singular 3D representation suitable for different use cases, such as language-guided 3D segmentation, point-guided 3D segmentation, or language-guided object selection and editing. The efficient implementation of such a unified model would eliminate the need for distributed 2D querying, which tends to be slow and necessitates multiple specialized models for distinct prompting styles.


Another promising avenue for future research entails exploring the extraction of the appropriate output formulation from pre-trained 2D diffusion~\cite{rombach2021highresolution} or foundation models. These foundation models, like Stable Diffusion, have undergone pre-training on extensive datasets. A potential direction involves fine-tuning or transferring these models to cater to specific 3D applications, possibly by imbuing them with 3D awareness through the inclusion of camera viewpoint annotations during fine-tuning. Consequently, these adapted models become valuable assets for tasks such as novel-view synthesis, particularly in scenarios characterized by an extreme scarcity of data, such as when dealing with a single image.

Furthermore, future research should emphasize the acquisition of high-quality 3D data for training these models. To provide context, object-centric datasets have expanded to a substantial scale of around 10 million assets~\cite{objaverse}. The size of these datasets has facilitated remarkable advancements in zero-shot novel view synthesis, reconstruction, and generation. Nevertheless, existing datasets often lack the essential compositional and scene-centric attributes necessary for a comprehensive 3D understanding. Hence, dedicated efforts in curating high-quality datasets, each focusing on compositional or scene-centric aspects, can significantly enhance the process of refining 2D foundational models to achieve effective 3D comprehension.


A way to generate high-quality datasets involves the real2sim approach. This approach entails capturing scenes in high quality and subsequently integrating them into simulators, allowing agents to learn over vast quantities of steps, ranging from millions to billions. While rendering times in these simulators have seen improvements, the question of fidelity still lingers. An intriguing path for future research lies in the exploration of novel-view synthesis techniques tailored for real-time simulation of authentic scenes. One such method, Gaussian splitting~\cite{kerbl20233d}, has emerged as a real-time explicit representation of 3D scenes derived from posed 2D images. A promising avenue for future research could revolve around Gaussian splatting or similar fast 3D scene representations as foundations for real-world simulators in the context of embodied AI research. Training agents within "real-world simulations" holds the potential to eliminate the need for specialized techniques designed to bridge domain gaps and facilitate the transfer of knowledge from simulation to reality. The central focus could be on constructing a robust infrastructure around classical reinforcement learning observation-action loops, encompassing elements like collision avoidance, path planning, and agent dynamic modeling within these realistic simulations. Such an approach would pave the way for efficient and large-scale learning from 2D captures in the real world, unlocking a multitude of possibilities for addressing edge cases and enabling agents to reason about their 3D environment despite limited real-world data.


Within this thesis, the construction of categorical models demonstrated exceptional simulation-to-real transfer capabilities, particularly when fine-tuned using limited real-world data. A promising direction for future research involves the expansion of these efficient categorical models to scenarios where 3D data is not a prerequisite for 3D-aware learning~\cite{lunayach2023fsd}. A crucial aspect of this approach could involve the utilization of multi-stage transfer learning strategies, harnessing self-supervised cues derived from the visual world. These cues might encompass elements such as chamfer distance losses, geometric correspondences, or optical flow. Furthermore, other avenues for future work may entail the application of the successful categorical models explored in this thesis to various other use cases. These potential applications encompass articulated object reconstruction~\cite{heppert2023carto}, efficient real-world grasping, the detection of openable and articulated parts, and a range of other scenarios. These endeavors could lead to broader and more impactful contributions to the field of 3D-aware learning and its applications.


Another exciting avenue for future research involves the extension of the versatile neural field representations explored in this thesis to diverse use cases. For instance, these representations could be applied to scenarios where language-guided querying of a neural field is required, built from representations generated using sparse viewpoints. Within a similar domain, investigating the concept of compositionality in the context of locally editing 3D objects, guided by image or text conditioning using 2D priors, could offer a convenient way to achieve the desired output without relying on time-consuming manual editing efforts. Furthermore, exploring the applicability of these adaptable neural field approaches to dynamic scenes, such as scenarios involving robots manipulating or grasping objects, holds the potential to imbue these agents with dynamic awareness. This, in turn, could unlock a range of applications, including object tracking and 4D editing, paving the way for more sophisticated and efficient solutions in these areas.


To summarize, the outlook for future work can be segmented into four key areas. 1.~\textit{Real-worlds as simulators for embodied AI research using some of the recent advances in real-time neural rendering of 3D scenes}. 2.~\textit{Self-supervised 2D prior distillation into the 3D domain using efficient 3D representations} 3.~\textit{Self-supervised efficient categorical models for a variety of use cases such as articulated object understanding and reconstruction} and 4.~\textit{Dynamic 3D-aware language guided generalizable neural fields}. All of these future avenues hold significant promise for advancing the field of robotics and 3D perception, ultimately enabling robots to better adapt to and navigate their 3D environments with increased efficiency.

    \makeBibliography
\end{thesisbody}

\end{document}